\documentclass[12pt,a4paper,twoside]{book}    
\usepackage[top=2.5cm,bottom=2.5cm,left=3.5cm,right=1.5cm]{geometry}
\renewcommand{\baselinestretch}{1.5}

\usepackage[centertags]{amsmath}
\usepackage{times} 
\usepackage{textcomp}
\usepackage{amsfonts}
\usepackage{amssymb}
\usepackage{cite}
\usepackage{amsthm}
\usepackage{csquotes}
\usepackage{algpseudocode}
\usepackage{rotating} 
\usepackage{tablefootnote}
\usepackage{algorithmicx}
\usepackage[ruled]{algorithm}
\usepackage{graphicx} 
\usepackage{caption,subfig}    
\usepackage{multirow}
\usepackage[T1]{fontenc} 
\usepackage{titlesec}
\titlespacing*{\section}
{0pt}{.2cm}{0.2cm}
\titlespacing*{\subsection}
{0pt}{0.1cm}{0.1cm}
\usepackage{enumitem}
\setlist{nolistsep}
\usepackage{longtable}
\usepackage{bm} 
\usepackage{adjustbox}
\usepackage{blindtext}
\usepackage{url,hyperref}
\usepackage{cite}

\newtheorem{assumption}{Assumption}

\newtheorem{definition}{Definition}
\newcommand{\argmin}[1]{\underset{#1}{\operatorname{argmin}} }
\graphicspath{{ch1_figures/}{jiis_images/}{aspgd_images/}{dscil_images/}{cilsd_images/}{techsym_images/}}

\makeatletter
\newcommand{\removelatexerror} 
 {\let\@latex@error\@gobble}
\makeatother

\begin{document}    

\begin{titlepage}

\newgeometry{left=2.1cm,right=2cm,top=2cm,bottom=2cm}
\renewcommand{\baselinestretch}{1.5}
 \setlength{\parindent}{0pt}

\textheight = 630pt \topmargin=0pt \voffset=1cm \headheight = 0pt
\marginparwidth= 0pt \headsep = 0pt

\thispagestyle{empty}
\begin{center}
\vspace{-10pt}
\begingroup
    \fontsize{16pt}{12pt}\fontfamily{ptm} \selectfont
    \textbf{ ANOMALY  DETECTION IN BIG DATA}

\endgroup
\begingroup
    \fontsize{12pt}{12pt}\fontfamily{ptm} \selectfont
    \bfseries
\vspace{2.5cm}
\textbf{Ph.D. THESIS} \\
\vspace{1.5cm}
{\normalsize \textbf {by}}\\  \vspace{1.5cm}
{\normalsize \textbf {CHANDRESH KUMAR MAURYA}}
\endgroup
\vspace{3.5cm}
%
\renewcommand{\baselinestretch}{1.2}
\vspace{3.5cm}


\begingroup
    \fontsize{14pt}{12pt}\fontfamily{ptm} \selectfont
    \bfseries
DEPARTMENT OF COMPUTER SCIENCE AND ENGINEERING\\
INDIAN INSTITUTE OF TECHNOLOGY ROORKEE\\
ROORKEE - 247 667 (INDIA)\\
\vspace{-01ex} AUGUST, 2016  
\endgroup

\end{center}

\end{titlepage}

%
%
\newgeometry{left=2.1cm,right=2cm,top=2cm,bottom=2cm}
\renewcommand{\baselinestretch}{1.5}
 \setlength{\parindent}{0pt}

\textheight = 630pt \topmargin=0pt \voffset=1cm \headheight = 0pt
\marginparwidth= 0pt \headsep = 0pt

\thispagestyle{empty}
\begin{center}

\renewcommand{\baselinestretch}{1.2}
\thispagestyle{empty}

\begingroup
    \fontsize{16pt}{12pt}\fontfamily{ptm} \selectfont
    \bfseries
\vspace{-10pt}
 ANOMALY  DETECTION IN BIG DATA\\
\endgroup

\vspace{0.8cm}
\begingroup
    \fontsize{12pt}{12pt}\fontfamily{ptm} \selectfont
    \bfseries
\textbf{A THESIS} \\
\vspace{0.8cm}
{
{\emph{\emph{\textbf{\textit{Submitted in partial fulfilment of
the \\ requirements for
the award of the degree \\\vspace{0.8cm}
of}}}}\\\vspace{0.8cm}

 \textbf{DOCTOR OF PHILOSOPHY}} \\\vspace{0.8cm}
\emph{\emph{\textbf{\textit {in}}}} \\\vspace{0.8cm}
 {\textbf{COMPUTER SCIENCE AND ENGINEERING}}}\\

\vspace{1.0cm}
{\emph {\emph{\textbf{\textit{by}}}}} \\\vspace{0.8cm}
\textbf{CHANDRESH KUMAR MAURYA}\\
\endgroup

\vspace{1.5cm}
%
\renewcommand{\baselinestretch}{1.2}
\vspace{1.5cm}
\begingroup
    \fontsize{14pt}{12pt}\fontfamily{ptm} \selectfont
    \bfseries
DEPARTMENT OF COMPUTER SCIENCE AND ENGINEERING\\
INDIAN INSTITUTE OF TECHNOLOGY ROORKEE\\
ROORKEE - 247 667 (INDIA)\\
\vspace{-01ex} AUGUST, 2016  
\endgroup

\end{center}

\newgeometry{top=2.5cm,bottom=2.5cm,left=3.5cm,right=1.5cm}


\thispagestyle{empty}
\tableofcontents
\cleardoublepage 
\addcontentsline{toc}{chapter}{\listfigurename}
\listoffigures
\clearpage 
\addcontentsline{toc}{chapter}{\listtablename}
\thispagestyle{empty}
\listoftables
\clearpage
\thispagestyle{empty}
\cleardoublepage 

\clearpage
\pagenumbering{arabic}
\pagenumbering{arabic}
\chapter{Introduction}\label{intro}
\setcounter{secnumdepth}{4}
\setcounter{tocdepth}{4}
\renewcommand{\baselinestretch}{1.5}
\setlength{\parskip}{0.5cm} 
Data mining is the process of discovering hidden patterns in the data through computational techniques \cite{Vipin2005,SKG:2005,SKG:2001}. Anomaly detection is one of the sub-fields of data mining. Anomaly is defined as a state of the system that does not conform to the normal behavior  of the system/object \cite{Varun2009,Varun:2012,Chandola:2013}. For example, emission of neutrons in a nuclear reactor channel above the specified threshold is an anomaly. Similarly, a suspicious activity of a person over a metro station is an anomaly. As a third example, abnormal usage of a credit card refers to the anomalous event.  Above examples indicate that anomaly detection is an important data mining/machine learning task. The main focus in anomaly detection is to discover the unusual pattern in the data. The term has also been referred to as outlier mining, event detection, exception, contaminant  mining, intrusion detection system (IDS) \cite{Hanan:2014,NCD:2014,Gonsa:2006,Gonsa:2010}, fraud-detection \cite{SKG:2006}, fault-detection \cite{Gonsa:2000} etc. depending on the application domain. We emphasize here that outlier mining (anomaly detection) is not a new task. It has its root dated back to $19^{th}$ century \cite{Edgeworth1887,grubbs:1969}. Since then lots of work has been done till date by different researchers and has resulted in different automated systems for anomaly detection, e.g, Aircraft monitoring sensor, patient health monitoring, credit-card-fraud detection system, video surveillance system etc.

The significance of anomaly detection is owing to the nature of anomaly that is often critical and needs immediate action. For example, modern aircraft record gigabytes of highly complex data from its propulsion system, navigation, control system, and pilot inputs to the system giving rise to so-called {\bfseries "Big data"}. Analyzing such a complex and heterogeneous data indeed requires automated and sophisticated system \cite{Das2010,NCD:2015,AT:2013}. As an another example, social media analytic looks for potential anomalous nodes in the graph created from its users. Before we proceed further, let us define anomaly formally \cite{Vipin2005}.

\begin{definition}
Anomalies are patterns in the data that deviate from the normal behavior or working of the system.
\end{definition}

\begin{figure}[t]
\centering
\includegraphics[width=6in,height=3.5in]{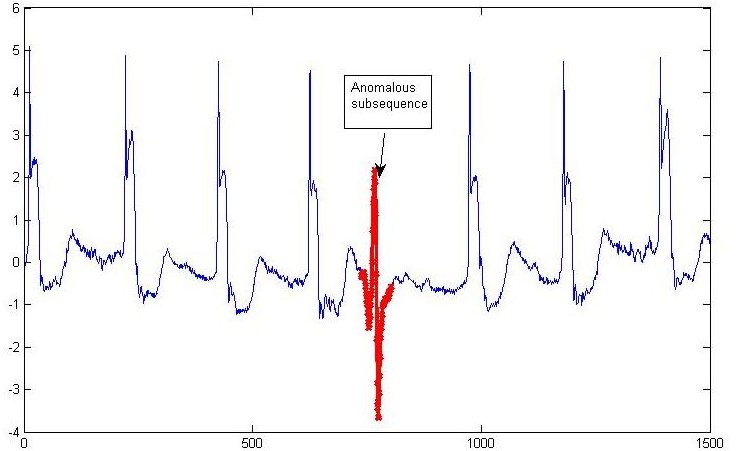}
\caption{EEG data (heart rate shown with respect to time)}
\label{eeg}
\end{figure}

\begin{figure}[t]
\centering
\includegraphics[width=5in,height=2in]{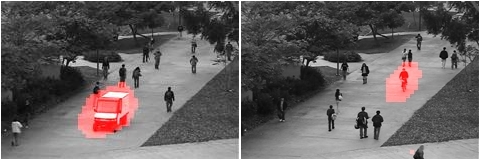}
\caption{Anomaly detection in crowd scene  \cite{Mahadevan2010}}
\label{crowd}
\end{figure}
\vspace{-.7cm}
As it can be seen clearly in Fig. \ref{eeg} around time $t=700$, there is sudden drop in heart beat which might be indicative of potential anomaly occurred in the patient at that time. As an another example of anomaly in crowd scene, see Fig. \ref{crowd}.  In the Fig. \ref{crowd}, automobile in the park, where only humans are allowed, is an example of anomaly. Similarly, bicyclist in the park is also an anomaly. Since, our aim is to detect the aforementioned anomaly in big data, we define the big data as:
\begin{definition}
Big data refers to a data which is complex in nature and requires lots of computing resources for its processing.
\end{definition}
\vspace{-.7cm}
It should be  clear that  big data may be small in sample size but huge number of dimensions or large number of samples with small number of dimensions (2 to 4, say). The data having large number of dimensions and samples is trivially big data. We define  the term big data formally later in this chapter. Examples of big data can be patient monitoring sensor data that consists of hundreds of attributes each of various types. 
Next we describe the various types of anomalies as defined in \cite{Varun2009}.
\begin{itemize}
\item {\bf  Categorization based on the nature of the anomaly}
\begin{itemize}
\item {\bfseries Point Anomaly}
 is a point of usually
high or low value with respect to other data instances.. For example, in Fig.\ref{pointAno}, point O denotes a point anomaly in the dataset. Example of  point anomalies is a small number of malicious transactions in a huge transaction database.  In the present work, we tackle  the point anomalies only.
\item {\bfseries Contextual Anomaly}  is an anomaly that is considered as an anomaly in a specific context and not an anomaly in another context. For example, as shown in Fig.\ref{contxano} monthly temperature at $t_1$ and $t_2$ are the same but temperature at $t_2$ is anomalous because temperature is usually high in the month of June.  
\item {\bfseries Subsequence Anomaly}
 is a collection of continuous records that are abnormal with respect to the entire sequence \cite{DBLP:conf/icdm/KeoghLF05} as shown in Fig.\ref{eeg} in red color.
\begin{figure}
\centering
\includegraphics[width=4in, height=3in ]{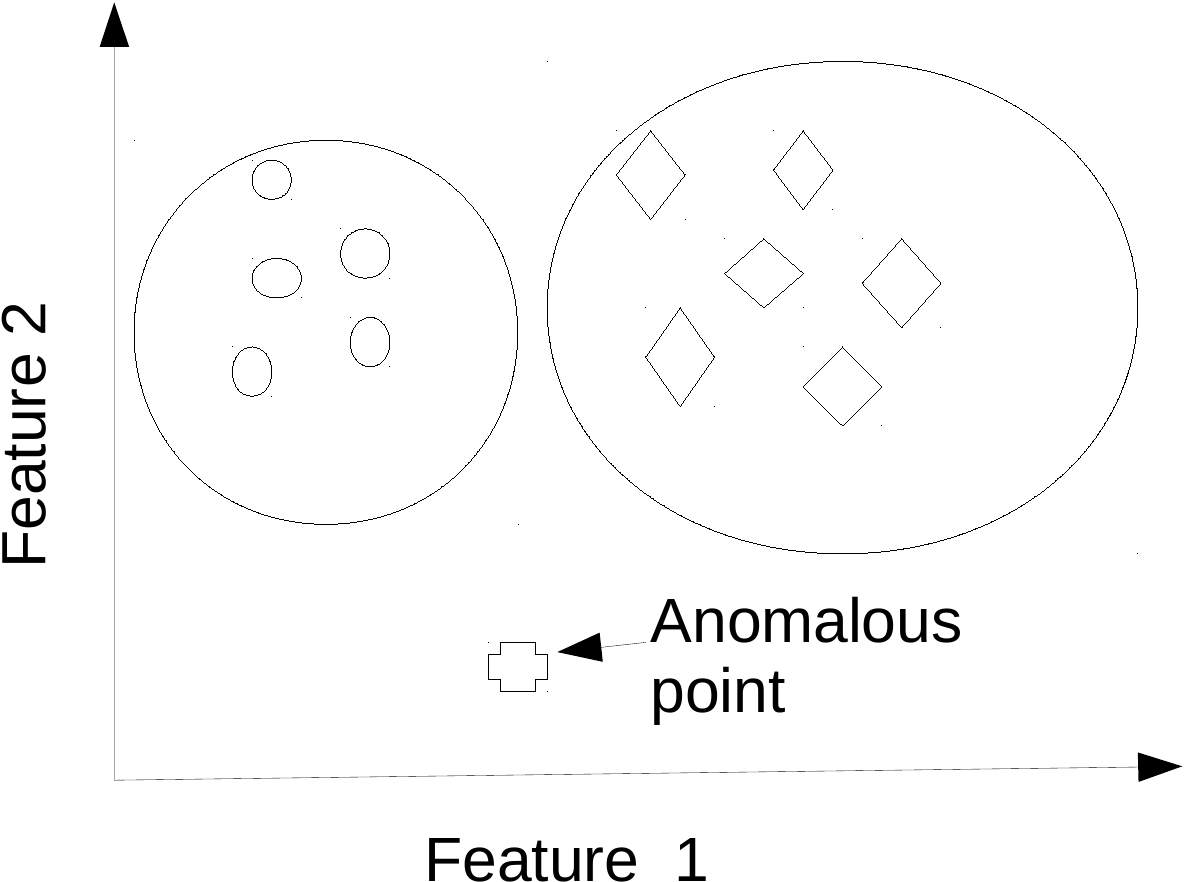}
\caption{An example of point anomaly}
\label{pointAno}
\end{figure}
\begin{figure}
\centering
\includegraphics[width=4.5in, height=2.5in ]{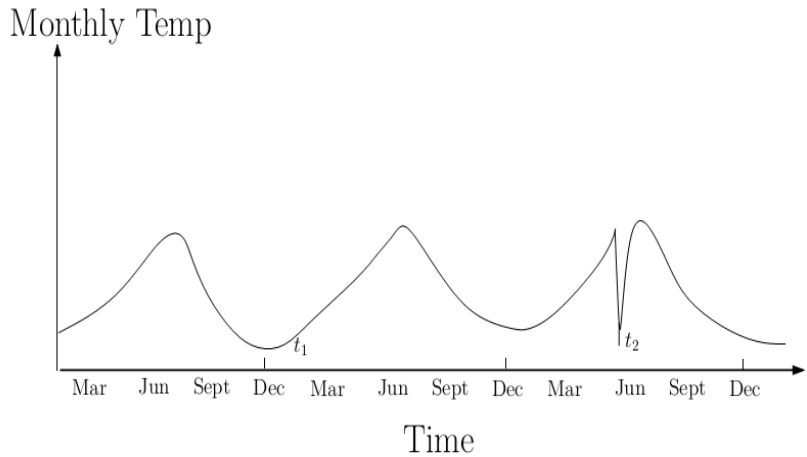}
\caption{Contextual Anomaly at $t_2$  in monthly temperature time series \cite{Varun2009}}
\label{contxano}
\end{figure}
\end{itemize}

\item {\bf  Categorization based on the neighborhood}

\begin{itemize}
\item {\bf Local Anomaly}  is an anomaly if it is quite dissimilar with respect to it values in its neighborhood.
\item {\bf Global Anomaly}  is distinct with respect to  the entire data set.
\end{itemize}
\end{itemize}
We emphasize that there have been several research works in the direction of finding global anomaly in large databases but finding local anomaly is limited. In the present work, we do not distinguish between the local and global anomaly. As we discuss later in the thesis that we take a different approach to handle anomaly.

\section{Motivation}
Our motivation comes from data laden domain in the industries what is being termed as \emph{big data}. The natural question arises is: why anomaly detection in big data?  There are numerous  reasons we can account for.  Let us look at them one by one.
Big data is often classified to emerge along five dimensions: \emph{Volume, Velocity, Variety, Veracity, Variability} (5 V's). Below,  we make the meaning  of  the 5 V's clear.

\begin{itemize}
\item {\bf Volume: } Massive volume of data (Petabytes even Zettabytes) is overwhelming enterprises. This leads to gleaning insights from high dimension data where scalability of algorithms is challenged\cite{Charu2001}.  In fact, only a limited set of work has been done to detect anomalies in high dimension data\cite{Charu2001, Vries2012}. The reason is that \emph{curse of dimensionality} prohibits traditional anomaly detection techniques perform effectively. Curse of dimensionality refers to the problem where the meaning of nearest-neighbors become vacuous \cite{Beyer1999}.
 
\item {\bf Velocity: } Massive data is gathering at a very high rate. A survey by IBM says that about 90\% of the whole data in the world today has  piled up in the last two year only.
For a time-critical process such as fraud detection, data must be scrutinized as it enters into the enterprise.
\item {\bf Variety: }Big data can be structured and unstructured. For example, sensor data, online transaction data, audio, video, click stream data etc. For knowledge discovery and new insights, these datasets must be analyzed together - Monitoring hundreds of live video  from CCTV cameras to target points of interest is a daunting task.
\item {\bf Veracity: } Veracity refers to the truthfulness  of the data. That means that the quality of the data can change severely which hampers accurate  anomaly detection and analysis .

\item {\bf Value: } Having storage and access  to big data is well and good but unless we can turn it into value it is useless. In terms of anomaly detection, it refers to efficiently and accurately finding anomalies.
\end{itemize}

Thus, above points indicate that anomaly detection in big data can be quite tedious and cumbersome. Till now, there exist only a few approaches that address curse of dimensionality, noise, sparsity, streaming, heterogeneity issues while targeting anomalies efficiently.

\section{Applications to Big data}
Anomaly detection  in big data finds several usage emerging from data laden-domains.  Some of them have been described below.
\begin{itemize}
\item {\bf  Business:} Anomaly detection in business has tremendous applications. For example, millions and billions of  banking transactions happen on a daily basis around the world. This has given rise to what is being called as big data. Anomaly detection arises in the form  of very tiny fraction of fraudulent transactions among the large set of normal transactions. The main challenge is that the data is continuously flowing from one point to the other, is high dimension, distributed, and often secure (secure here means that not all data is exposed to the analyst for confidentiality reasons).  The task is to detect the fraudulent transaction on the fly in the big data and possibly prevent it from happening. 

 \item {\bf  Healthcare:} In health care domain, patient's health is continuously monitored through various sensors giving real-time condition of the patient. Due to unavailability of the sufficient number of staff, any abnormal situation must be brought to the notice of the doctors instantly to prevent loss of life. Such a situation can be modeled as the anomaly detection problem in health care data such as electroencephalogram (EEG), electrocardiogram (EEG) etc. The main idea is to detect different abnormal conditions such as cardiac arrest, low blood pressure, low glucose level etc. in real-time. 

\item {\bf  Computer Network and Data Centers:}  Anomaly detection has several usage in computer network and data centers. For example, intrusion detection system (IDS) is a form of anomaly detection where the task  is to find potential denial-of-service attack (DoS attack), unauthorized access to computing resources, servers etc. Similarly, anomaly detection in data center is related to the problem of finding abnormal system conditions from  log files.

\item {\bf  Plant Monitoring:} Power plants, Nuclear plants are monitored through wireless sensors, which are distributed in space. The  task is to detect abnormal conditions of the plant, and perhaps before they occur since when anomaly enters into the system, it is impossible to prevent it.   The anomaly detectors are installed to continuously monitor the proper working conditions of the machines.

\item {\bf  Surveillance:} Remote surveillance is another area of anomaly detection. Now-a-days, CCTV can be seen installed at metro stations, shopping malls, pedestrian crossings etc. in order to monitor the possible malicious activities  For example,  CCTV footage are used to find potential terrorist activity. Data set from CCTV is in the form of videos that requires state-of-the-art video processing technology in order to track the anomalous activity. 

\item {\bf Satellite Imagery:} Satellite images are hyper spectral images and huge in size.  They are used to find water bodies, rare metals etc. on far distant planets and Galaxies.   Anomaly detection is concerned with finding these rare events from hyper spectral images.

\section{The Problem Statement and Research Scope}
Big data brings with it great opportunity and challenge together. Opportunity comes in the form of plenty of data to glean insightful information. The challenge is that data is too huge to mine efficiently using conventional knowledge discovery methods. Our primary aim is therefore to look at the problems posed by big data. We formulate our objective as follows:

\emph{``To efficiently detect anomalies in big data which is sparse, high-dimensional, streaming and distributed''}.

 Since tackling all of the characteristics of big data together is a complicated task, we  make some assumptions and consider different scenarios of big data characteristics. Our assumptions are as follows:
\begin{itemize}
\item Our proposed algorithms work with numeric data only.
\item
 Concept Drift \cite{Mirza2015,VB:2009} (sudden change in data distribution or concepts) in not handled in the thesis.
\item We only tackle \emph{point} anomaly.
\end{itemize}
Along the line of solution methodologies, we adopt the class-imbalance learning approach to solve the \emph{point} anomaly detection problem. We argue that the class-imbalance learning problem is similar to the point anomaly detection problem and therefore, can be used to solve the point anomaly detection. Some works that have followed this approach include \cite{Dayong:2015}\cite{Jialei2014}\cite{ Wang:2013}\cite{ Shou:2015}\cite{Ajith:2016b} and discussed in detail in \cite{Charu:2013}.

 In order to detect anomalies from big data , the following scenarios have been considered in the present work:
\begin{itemize}
\item {\bf Scenario 1:} To detect anomalies when the data is \emph{streaming}.
\item {\bf Scenario 2:} To detect anomalies in the data when it is \emph{streaming, sparse, high dimension}.
\item {\bf Scenario 3:} To detect anomalies in the data when it is \emph{sparse, high dimension, distributed} data.
\end{itemize}
By efficiently solving the problem, we mean the technique is able to detect \emph{anomaly} in a timely manner, is scalable (in terms of numbers of instances as well as dimensions), and incur small false positive and small false negative. In other words, we aim to achieve higher \emph{Gmean} and lower \emph{Mistake rate} (defined in the next Chapter) than the existing methods in the literature.

\section{Specific Research Contributions }
Our contributions are as follows:
\begin{itemize}
\item To solve the problem of anomaly detection in big data in Scenario 1, we propose an algorithm based on \emph{online learning} referred to as {\bf P}assive-{\bf A}ggressive GMEAN (PAGMEAN). PAGMEAN is an improved version of a popular online learning algorithm called Passive-Aggressive (PA) of \cite{Crammer2006}. In other words, PA algorithm is sensitive to outliers as shown in Chapter \ref{Chapter3}. To alleviate this problem, PAGMEAN algorithm utilizes the modified hinge loss function that is a convex surrogate loss for the $0-1$ loss function. In turn, the $0-1$ loss function is obtained from directly optimizing the \emph{Gmean} performance metric. The challenge here is that \emph{Gmean} metric is a non-decomposable metric, i.e., it can not be written as the sum of losses over individual data points, and hence traditional classification models based on statistical learning can not be used.  We show the efficiency  and effectiveness of the proposed algorithms on several popular real and benchmark datasets and demonstrate the competitiveness with respect to the state-of-the-art algorithms in the literature.

\item  PAGMEAN algorithm discussed previously is not able to handle \emph{high dimension} and \emph{sparse} data.  Therefore, we tackle the Scenario 2 using  {\bf A}ccelerated-{\bf S}tochastic-{\bf P}roximal {\bf G}radient {\bf D}ecent (ASPGD) algorithm. Specifically, we use a smooth version of the modified hinge loss used within the PAGMEAN algorithm.  Smooth loss function gives us the freedom to employ any gradient based algorithm. For that purpose, we use the stochastic proximal learning framework algorithm with Nesterov's acceleration. $L_1$-regularization is used to handle the sparsity. In the experiment section, we show encouraging results  on several benchmark and real data sets and compare with the state-of-the-art techniques in the literature.

\item In order to solve the \emph{sparse, high dimension} and \emph{distributed} problem of big data in Scenario 3, we propose two novel \emph{distributed} algorithms called {\bf D}istributed {\bf S}parse {\bf C}lass-{\bf I}mbalance  {\bf L}earning (DSCIL) and {\bf C}lass-{\bf I}mbalance  {\bf L}earning on {\bf S}parse data in a {\bf D}istributed environment (CILSD).   DSCIL is based on the distributed alternating direction method of multiplier (DADMM) framework. The loss function used in DADMM is a  cost-sensitive,  smooth and strongly convex hinge loss.  Due to the linear convergence of DSCIL, we propose CILSD algorithm. CILSD uses the same loss function as DSCIL but is based on FISTA-like update rule in a distributed environment.  As it is known that FISTA algorithm converges quadratically \cite{Beck:2009}, we get a faster algorithm (CILSD) than DSCIL. We demonstrate the efficiency and effectiveness of these algorithms in terms of various metrics like \emph{Gmean, F-measure, Speedup, Training time} etc. and compare with the state-of-the-art techniques in the literature. We also show real-world application of the proposed algorithms on KDD Cup 2008 anomaly detection challenge data set.

\item Our algorithms based on online learning and distributed learning are \emph{supervised} machine learning algorithms, i.e., they require labeled data for normal as well as anomalous classes. However, in a real world, data is often noisy and is unlabeled. Therefore, aforementioned techniques can not be applied in a real-world setting. We seek a solution for a real-world problem where our data is coming from nuclear reactor channel (obtained from Bhabha Atomic Research Center, Mumbai, India). The data set contains the count of neutron emission from the nuclear reactor and the task is to find at what point of time a particular channel was behaving maliciously? Since this is a \emph{unsupervised} learning task, we utilize support vector data description (SVDD) algorithm \cite{Tax2004,ckm2014} for finding anomalies.

\end{itemize}
\end{itemize}
\vspace{-0.5cm}
\section{Organization of the Thesis}
The thesis is divided into 7 seven chapters. Each chapter can be read independently without requiring going back and forth.  The content of each chapter is described below.

{\bf Chapter 1:} This Chapter gives the introduction of the proposed work. In particular, we talk about what is an anomaly?  Why is anomaly detection important? Then, we talk about various kinds of anomalies and how to report the result of anomaly detection algorithm?  We also establish the connection between anomaly detection and related problems like outlier detection, class-imbalance problem etc. This chapter also covers  the  motivation and contribution of the work.

{\bf Chapter 2:} 
In this chapter, we exhaustively review related works on anomaly detection. In particular,  statistical based, clustering based, density based, nearest neighbor based, information theoretic based anomaly detection techniques are discussed in great depth. We argue why traditional techniques for anomaly detection fail on large data sets? Then, we take a digression and talk about some nonconventional techniques for anomaly detection. Specifically, multiple kernel learning based, non-negative  matrix factorization based, the random projection based, ensemble based anomaly detection techniques  and their key limitations are discussed. From these discussions, we  find research gaps which we fill up with our contribution.

{\bf Chapter 3:} This chapter begins with a literature survey of algorithms targeted for anomaly detection in \emph{streaming} data followed by their limitations. Then, we propose our first online algorithm for class-imbalance learning and anomaly detection. In particular, passive-aggressive algorithms (PA) \cite{Crammer2006}, which have been successfully applied in online classification setting, are sensitive to outliers because of their dependence on the norm of the data points and as such can not be applied for class-imbalance learning task. In the proposed work, we make it insensitive to outliers by utilizing a modified hinge loss that arises out of the maximization of Gmean metric and derives some new algorithms called Passive-Aggressive GMEAN (PAGMEAN). Second, It is found that the derived algorithms either outperform or perform equally good as compared to some of the state-of-the-art algorithms in terms of the Gmean and mistake rate over various benchmark data sets in most of the cases. Application to online anomaly detection on real world data sets is also presented which shows the potential application of PAGMEAN algorithms for real-world online anomaly detection task.

{\bf Chapter 4:} The work presented in Chapter 3, although scalable to a large number of samples,  does not exploit sparsity in the data, which is not uncommon these days. In this chapter, we propose a novel $L_1$ regularized smooth hinge loss minimization problem and an algorithm based on accelerated-stochastic-proximal learning framework (called ASPGD) to solve the above problem. We demonstrate the application of proximal algorithms to solve real world problems (class imbalance, anomaly detection), scalability to big data and competitiveness  with recently proposed algorithms in terms of \emph{Gmean, F-measure} and \emph{Mistake rate} on several benchmark data sets.

{\bf Chapter 5:} This chapter begins with the survey of techniques developed for handling anomaly detection in \emph{large, sparse, high dimension, distributed} data and their limitations. Then, we describe our proposed framework.   DSCIL and CILSD algorithms are described in detail followed by their distributed implementation in MPI framework.  Finally, we show the efficacy of the proposed approaches on benchmark and real-world data sets and compare the performance with the state-of-the-art techniques in the literature.

{\bf Chapter 6:}  Chapter 6 elucidates the application of support vector data description algorithm for finding anomalies in a real-world data. We illustrate the working mechanism of the algorithm and show the experimental results on nuclear power plant data.

 {\bf Chapter 7:} This chapter summarizes our main finding and ends with some open problems that we plan to explore in future.
 \thispagestyle{empty}
\chapter{Literature Survey}
In this chapter, we present relevant works that have addressed the problem of anomaly detection in general. In particular, we discuss the research works in three flavors. The first is based on the traditional approach to anomaly detection, the second is based on the modern approach to anomaly detection, and the third one is based on the online learning. The reason for such categorization is that anomaly detection has been tackled in statistics as \emph{outlier detection} since 1969 \cite{grubbs:1969} and recently, there has been an advancement in research which utilizes the more state-of-the-art machine learning approach to solving the problem. We present the discussion that assumes anomaly detection, outlier detection, novelty detection as the similar problem. Further, we also present the literature work that has used class-imbalance learning to address the \emph{point} anomaly detection problem.

\section{Traditional Approaches to Anomaly Detection }
As we mentioned in chapter \ref{intro} that anomaly detection is not a new task. Lots of work has been done in statistics community. However, the ultimate goal of anomaly detection is to find any outlying pattern in the data. For example, given a dataset from health care, the key task is to find any anomalous point, subsequence (in the case of gene expression data) anomaly. Traditional approaches to anomaly detection are known by the umbrella term \enquote{outlier detection} \cite{Varun2009}. There exist some literature which distinguishes the anomaly detection and outlier detection such as \cite{Zimek:2012}. However, we make no difference between the two terms while presenting the survey below due to the possible overlap. 

Now, let us  look at traditional approaches to anomaly detection  and study some of the algorithms developed and what kind of issues they aim at to solve. Broadly speaking, they can be categorized based on the use of available data labels: Supervised,  semi-supervised, and unsupervised.
\begin{itemize}
\item{\bf Supervised: } In the supervised setting of anomaly detection, the common assumption is that the data is available for both normal as well as anomalous class. First, a model is built using data from the normal class. Any unseen data instance is then given to the model which predicts the class label. The main challenges associated  with  supervised mode of anomaly detection is that (i) data from anomalous class is often rare (imbalanced) (ii) getting  representative samples from the anomalous class is difficult \cite{Varun2009}. In the present work, we  adopt the supervised mode of anomaly detection under the \emph{class-imbalance} setting. 
 Later in this chapter, we discuss the classification-based approaches that come under \emph{supervised} setting of anomaly detection.  

\item{\bf Semi-supervised: }Semi-supervised setting assumes that small amount of labeled data from normal class and a large amount of unlabeled data is available. Because of the availability of small amount of data from the normal class, semi-supervised techniques often perform better than supervised counterpart \cite{Blanchard2010}. Another advantage  of small amount of  normal class instances is that  semi-supervised mode is far wider applicable than the supervised mode. The typical approach adopted by semi-supervised mode is that they build the model to capture the normal behavior of the system and any deviation from the normal behavior raises an alarm. On the other  line of  work, there exist literature which builds the model using the data from the anomalous operation of the system \cite{dasgupta2000comparison,Dasgupta:2002}.  However, it is difficult to characterize all the abnormal behavior of the system and therefore, approach mentioned thereof are rarely employed in practice.

\item{\bf Unsupervised: } Unsupervised setting assumes that normal instances are available in abundant whereas anomalous instances are rare.  A false alarm is raised if the assumption made turns out to be wrong \cite{Varun2009}. Because of the less strict assumption on the availability of the abnormal instances, the unsupervised mode is more generic than the supervised and semi-supervised mode. Later in this chapter, we will discuss clustering, density, and nearest-neighbor based approaches which work in the unsupervised mode.
\end{itemize}

After anomalies have been found, we need a way to report the result. An anomaly detection algorithm typically outputs its result in one of the following two ways:-
\begin{itemize}
\item{\bf Scores: } 
Score reports the degree of outlierness of a data instance. Some techniques assume high score to be a high degree of outlierness and some assume vice-versa.
\item{\bf Labels :} Label is used when the category of anomalies is small and the algorithm reports whether a data instance is anomalous or normal. 
\end{itemize}

Below we describe various approaches that have been developed in statistics community as well as data mining community for anomaly detection. These are:
\begin{itemize}
\item Classification-Based Approaches
\item Statistical Approaches
\item Clustering-Based  Approaches
\item Information-Theoretic Approaches
\item Spectral-Theoretic Approaches
\end{itemize}


\subsection{Classification-Based Approaches}
Classification based anomaly detection techniques are built on the assumption that the labeled instances (for both the normal as well as anomalous classes) are available for learning the model (typically a classifier). They work in two phases: (i) a model is trained using data from both the normal class and anomalous class (ii) trained model is presented with unseen data to predict its class label. Techniques falling under \emph{classification-based} approaches are classified  either as one-class classification or multi-class classification.

\begin{figure}
\centering
\includegraphics[width=4in, height=3in ]{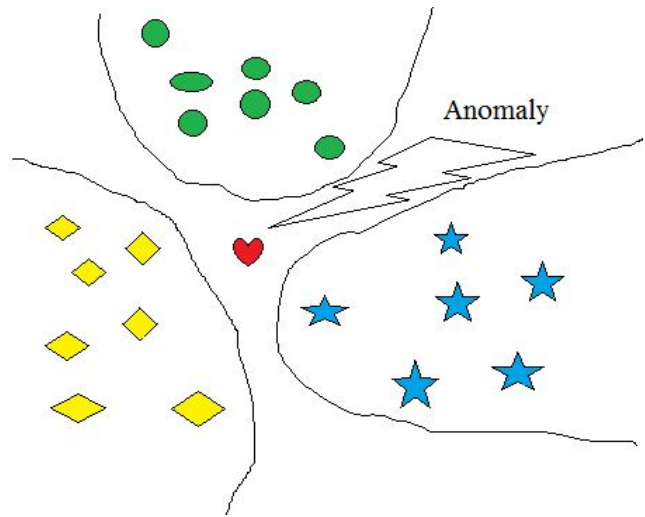}
\caption{Anomaly detected by multi-class classifier}
\label{multiclass}
\end{figure}

In {\bf Multi-Class Classification} setting, learner (classifier) is trained on labeled data comprising of various labels corresponding to normal classes as shown in Fig. \ref{multiclass}.
For example, we can model the anomaly detection as training several binary classifiers  where one class is normal and the rest of the classes as anomalous. During testing, the unseen example is presented to each of the binary classifiers. An unseen example is declared as an anomaly if (i) none of the classifiers predicts it as normal  (ii)  take the majority vote to decide for abnormality. In some cases, classifiers assign a confidence score  to the unseen data and declare it as an anomaly if the confidence score is below some threshold. Popular classifiers used for multi-class classification based anomaly detection include Neural network \cite{Stefano2000}, Bayesian Network \cite{Siaterlis2004,Das2007}, Support Vector Machine (SVMs) \cite{Vapnik1995, Davy2002}, Rule-based \cite{Agrawal1995, Fan2001} classifiers etc. 

\begin{figure}
\centering
\includegraphics[width=4in, height=3in ]{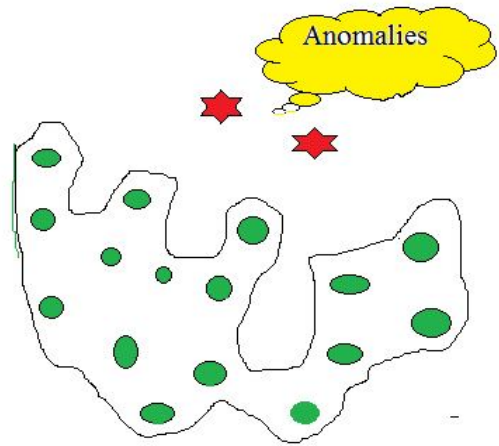}
\caption{Anomaly detected by one-class classifier}
\label{oneclass}
\end{figure}

{\bf One-Class Classification}  builds discriminatory model using only labeled data of normal instances as shown in Fig. \ref{oneclass}. In this setting, learner draws a boundary around normal instances and leaves abnormal instances untouched. Indeed, complex boundaries can be drawn using non-linear models such as kernel-methods, Neural Networks etc. 
Popular techniques for anomaly detection using one-class classification are one-class SVM and its various extensions \cite{Ratsch2002, Amer2013}, one-class Kernel Fisher Discriminants \cite{Roth2005}, support vector data description (SVDD) \cite{Tax2004} etc.

\noindent \textbf{Pros and Cons of Classification-Based Approach:}
\noindent The performance of classification-based technique on anomaly detection task depends on the generalization ability of classifier on unseen data. There are many robust linear and nonlinear classifier with a provable guarantee that they will find decision boundary between normal and abnormal instances if any. 

The major disadvantage in using classification based techniques is the availability of the training data for normal instances and the training time. If there are not sufficient examples from abnormal classes, it is very hard to build a meaningful decision boundary. Secondly, in real-time anomaly detection application, the classifier is expected to produce anomaly score within reasonable time limits. Thirdly, they assign class labels to each test data point that may become a disadvantage in case a meaningful anomaly score is required.
\subsection{Statistical Approaches}
Statistical approaches to anomaly detection are model based. That is, it is assumed that the data is coming from some distribution (but unknown).  Model is built  by estimating the parameters of the probability distribution from the data.  An anomalous object is such that it does not fit the model very well.  Statistical anomaly detection techniques are based on the following assumption:
\begin{assumption} Normal data instances reside in the high probability region of the stochastic model while anomalous data instances reside in the low probability region of the stochastic model. 
\end{assumption}
\vspace{-.7cm}
In clustering problem, anomalous objects are such that do not fall in some particular cluster and lie in a low-density region (we shall see in the clustering-based approach it is not always the case and the problem becomes tricky). Likewise, in regression problem, anomalies fall apart from the regression line.

Briefly, we describe some of the techniques under this category.

\begin{itemize}
\item {\textbf{Box Plot Rule}:} 
Box plot has been used to detect anomalies in univariate and multivariate data and is, perhaps the simplest anomaly detection technique. A box plot shows the various statistical estimator on a graph such as largest non-anomaly, upper quartile (Q3), median, lower quartile (Q1), and the smallest non-anomaly as shown in Fig. \ref{boxplot}. The difference $Q3-Q1$ is called \emph{Inter Quartile Range} (IQR) and shows the range of the most normal data (typically 99.3\%). A data point that lies $1.5IQR$ above the $Q3$ or below the $Q1$ is declared as an anomaly. Some works that have used box plot rule to identify anomalies are \cite{laurikkala2000informal,solberg2005detection,horn2001effect}.
\begin{figure}
\centering
\includegraphics[width = 4cm, height=6cm]{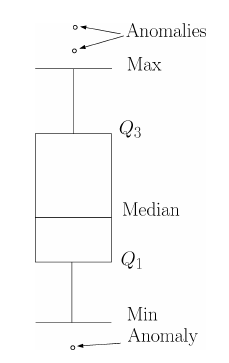}
\caption{Box plot showing anomaly \cite{Varun2009}.}
\label{boxplot}
\end{figure}
\item {\bfseries Univariate Gaussian Distribution:}
As we described previously that most of the real data can be modeled using some of the distributions. When the data set is large (social media data, aircraft navigation data etc.), it is assumed that the data follows the normal distribution.
Model parameters mean $\mu$ and standard deviation $\sigma$ are computed from the data using \emph{Maximum Likelihood Principle}. A point \emph{P} with attribute value $x$ and confidence level $\alpha$ is predicted as outlier with probability $p(|x|>=c) =\alpha$  using this model. The main difficulty encountered in univariate Gaussian distribution assumptions is choosing parameters of the model using sampling theory. As a result, the accuracy of the prediction is reduced. 

\item {\bfseries Multivariate Gaussian Distribution:} Univariate Gaussian assumption is applicable to the univariate data.  In order to handle the case of multivariate data, multivariate Gaussian assumption is used. To model the problem, a point is classified as normal or anomalous depending upon its probability from the distribution of the data above or below a certain threshold. Since multivariate data tends to have a high correlation among its different attributes, asymmetry is invariably present in the model. To cope up with this problem, we need a metric that takes into account the shape of the distribution. The Mahalanobis distance is such a metric given by \eqref{maha}.
\begin{equation}
\label{maha}
Mahalanobsisdist(\bf x, \bar{\bf x})= (x - \bar{x}){\bf S}^{-1}(\bf x - \bar{\bf x})^T ,
\end{equation}
Where $\mathbf{x}$ is the data point and $\mathbf{\bar{x}}$ is the mean data  point and $\mathbf{S}$ is the covariance matrix.
\item {\bfseries Mixture Model Approach:}
The mixture model is a widely used technique in modeling problems that assume that the data is generated  from several distributions. For example, one sample can be generated from many distributions with certain probabilities given by \eqref{mix}
\begin{equation}
\label{mix}
p({\bf x};\Theta) = \sum_{i=1}^m w_i p_i({\bf x}|\Theta) ,
\end{equation}
where $\Theta$ is the parameter vector, $w_i$ is the weight given to $i^{th}$ component in the mixture. The basic idea of a mixture model for anomaly detection task is the following.  Two sets of objects are created; one for the normal object N and the other for anomalous objects A. Initially, set  N contains all the objects and set A is empty. An iterative procedure is applied to move anomalous objects from set N to set A. The algorithm stops as soon as there is no change in the likelihood of the data. Eventually, the set A will have all anomalous objects and the set N will have normal objects.

%
%

\end{itemize}

\noindent \textbf{Pros and Cons of Statistical-Based Approach:}
Below, we describe the  pros and cons of using the statistical approach for anomaly detection task. 
\begin{itemize}
\item If the distribution underlying the data can be estimated accurately, statistical techniques provide a reasonable solution for anomaly detection.
\item Statistical techniques can be used in an unsupervised mode provided the distribution estimation is robust to anomalies.
\item The anomaly score produced by statistical techniques is often equipped with a confidence interval. The confidence interval may be used to gain better insight into the fate of a test instance.

\end{itemize}

Some of the cons associated with statistical techniques for anomaly detection are:
\begin{itemize}
\item The major challenge in using statistical techniques is that they assume that the data is distributed according to a particular distribution. However, this assumption is rarely followed by real-world data and the problem becomes severe in big data.
\item Final decision about a test instance, whether to declare an anomaly or not,  depends on the test statistics used; the choice of which is non-trivial \cite{motulsky1995intuitive}
\item They are unable to detect the anomaly in \emph{streaming, sparse, heterogeneous}  data efficiently.
\end{itemize}

\subsection{Clustering-Based Approaches}
Clustering is the process of grouping data into different clusters based on some \emph{similarity} criteria. Clustering-based approaches for anomaly detection are not new.  In fact, outliers are found as a by-product during clustering provided outliers do not form coherent, compact group of their own. Clustering-based approaches can have two subcategory namely proximity-based and density-based approaches which are described in subsections \ref{proximity} and \ref{density}
Clustering based techniques can be put into three categories depending on the assumption made by different researchers on anomalies \cite{Varun2009}. 

\begin{assumption}
 It says that normal instances form a coherent cluster while anomalies do not.
\end{assumption}
\vspace{-.5cm}
Techniques built on the above assumption use clustering algorithm on the given data set and report points as anomalous that can not be put into any cluster by the algorithm. Notable work based on the above assumption are of \cite{Ester1996, Guha1999}. Clustering algorithms that do not require data instances to necessarily belong to some cluster can be used under the above assumption such as ROCK \cite{guha1999rock}, DBSCAN \cite{ester1996density}, and SNN \cite{ertoz2004finding}.

\begin{assumption} It says that normal records lie close to the center of gravity  of the cluster while anomalous records reside for away from the closest center of gravity of the cluster. 
\end{assumption}
\vspace{-.5cm}
Above  definition works in two step. In the first step, clustering algorithm runs over the entire data so as to form a natural cluster. In the second step, anomaly score is calculated by computing distance of each data point from their closest centroid. A noteworthy point is that clustering algorithm based on the second assumption can be executed in either unsupervised or semi-supervised setting. In \cite{Hanan:2005}, the author propose rough set and fuzzy clustering based approach for intrusion detection. One limitation  of the techniques based on the  second assumption is that they will fail to find anomalies if they form a homogeneous group of their own.

\begin{assumption} It says that normal records form a dense and huge cluster while abnormal records form tiny and sparse cluster.
\end{assumption}
\vspace{-.5cm}
Techniques in this subcategory fix some threshold or size on the cluster that will enable them to accumulate outliers in a different group. One notable work in this category is Cluster-based local outlier factor (CBLOF) \cite{He2003}. The CBLOF computes two things: (i) the size of the cluster (ii) distance of the data instance to its cluster centroid. They declare a data instance as anomalous when the size and/or density of the cluster, in which it falls, is below a certain threshold.

\noindent \textbf{Pros and Cons of Clustering-Based Approaches:}
The performance of clustering based outlier detection algorithm depends on the training time. Some clustering algorithm run in quadratic time and hence several optimizations have been proposed to reduce it to linear time $O(Nd)$ but they are approximation algorithm.

Some salient features of clustering based anomaly detection approaches are the following:
\begin{itemize}
\item They perform well in unsupervised and semi-supervised settings.
\item They can be utilized to complex data type just by changing the baseline clustering algorithm.
\end{itemize}
Downside, however, encompasses the following:
\begin{itemize}
\item Performance of clustering algorithm depends on  how well the underlying algorithm capture the intrinsic structure of the data?
\item Many techniques are not optimized for anomalies.
\item Their clustering performance hinges on the assumption that anomalies do not form significant clusters.
\end{itemize}
\vspace{-.9cm}
\subsubsection{Proximity-Based} \label{proximity}
\begin{figure}
\centering
\includegraphics[width=5in, height=4in ]{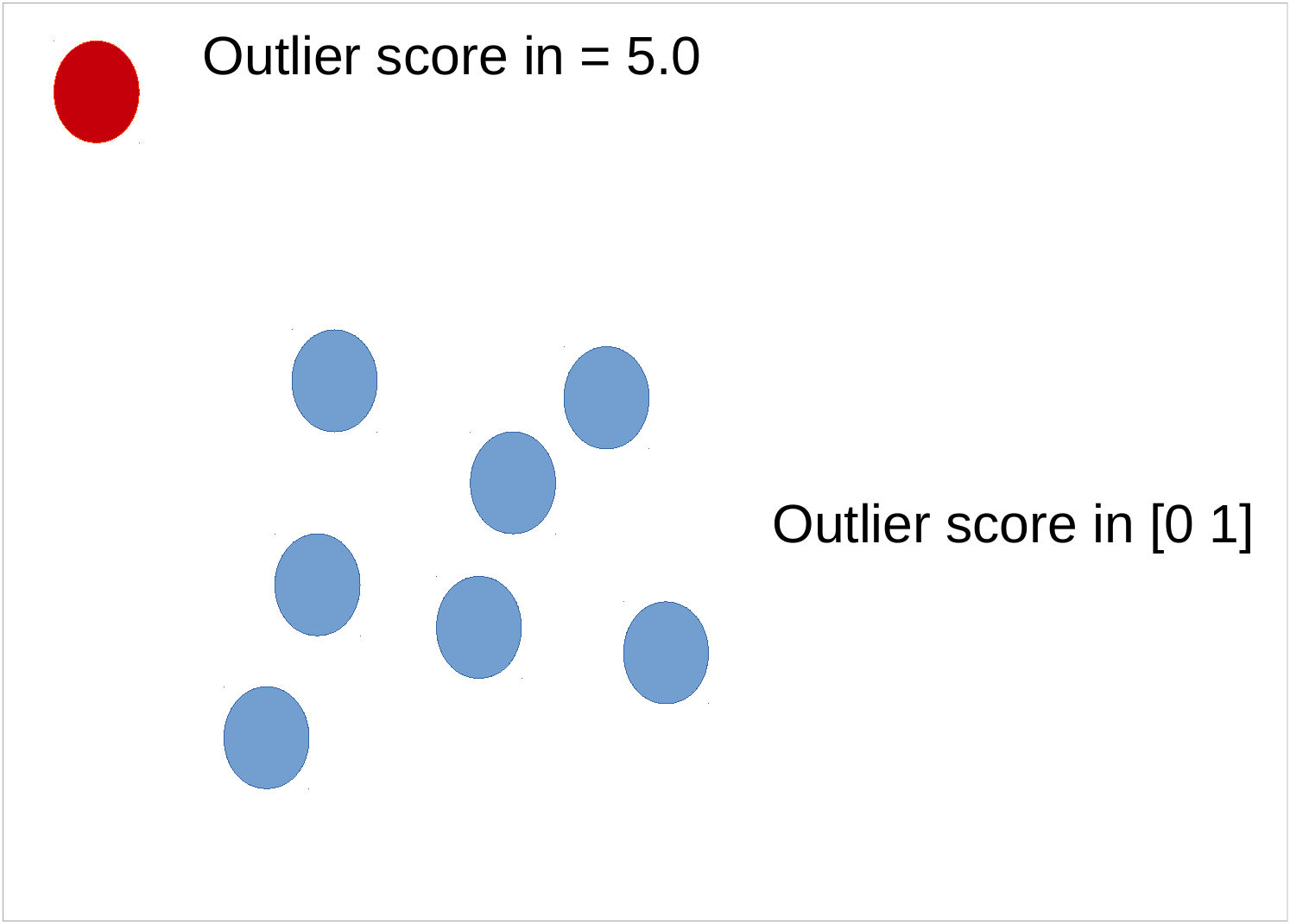}
\caption{Proximity based outlier score with $k=5$}
\label{knnk5}
\end{figure}

\vspace{-.3cm}
Proximity-based approaches (also known as nearest-neighbor based approaches), as the name suggests, are approaches that rely on some metric for computing the proximity (distance) between data points. Clearly, the performance of these techniques depend on how good our metric is? The idea of the proximity-based approach is simple. Anomalous objects are points that are far away from most of the points. The general strategy to compute the distance is to use k-nearest neighbor based techniques. The outlier score of an object is given by its distance to its k-nearest neighbors \cite{Vipin2005}. For example, in Fig. \ref{knnk5}, the red marked point has a very large outlier score as compared to green points.

\begin{figure}
\centering
\includegraphics[width=4in, height=3.5in ]{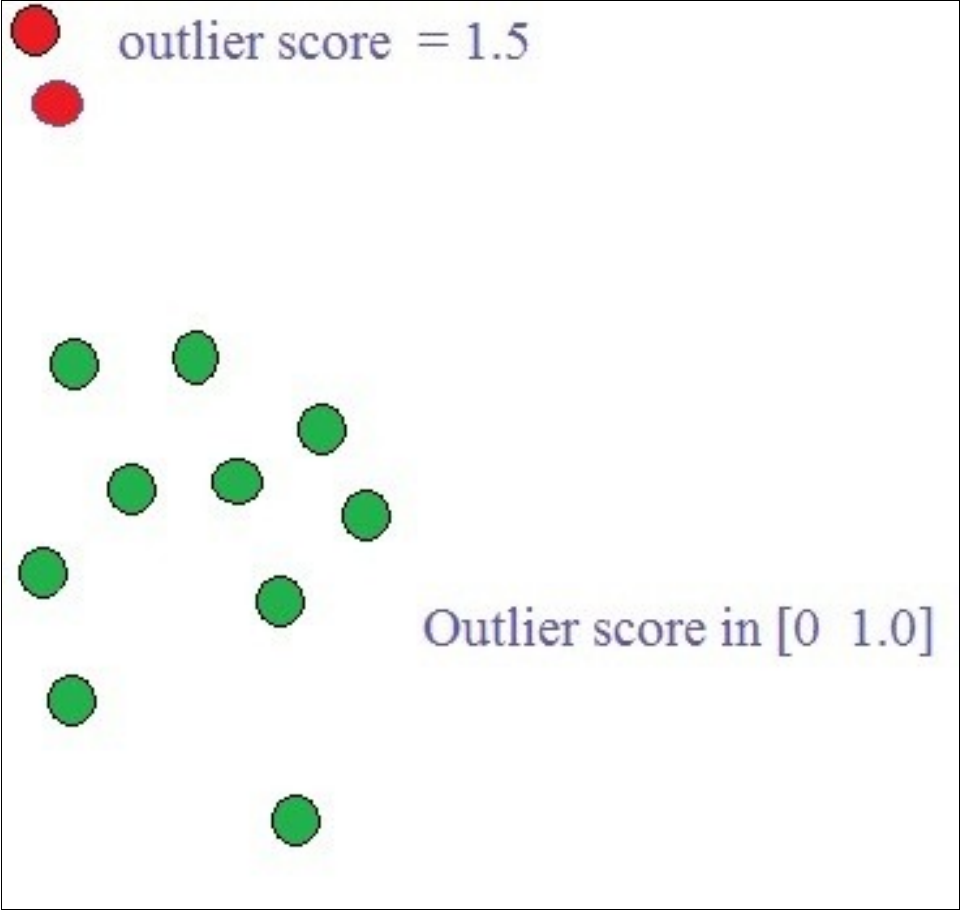}
\caption{Proximity based outlier score with $k=1$}
\label{knn}
\end{figure}

\begin{figure}
\centering
\includegraphics[width=4in, height=3.5in ]{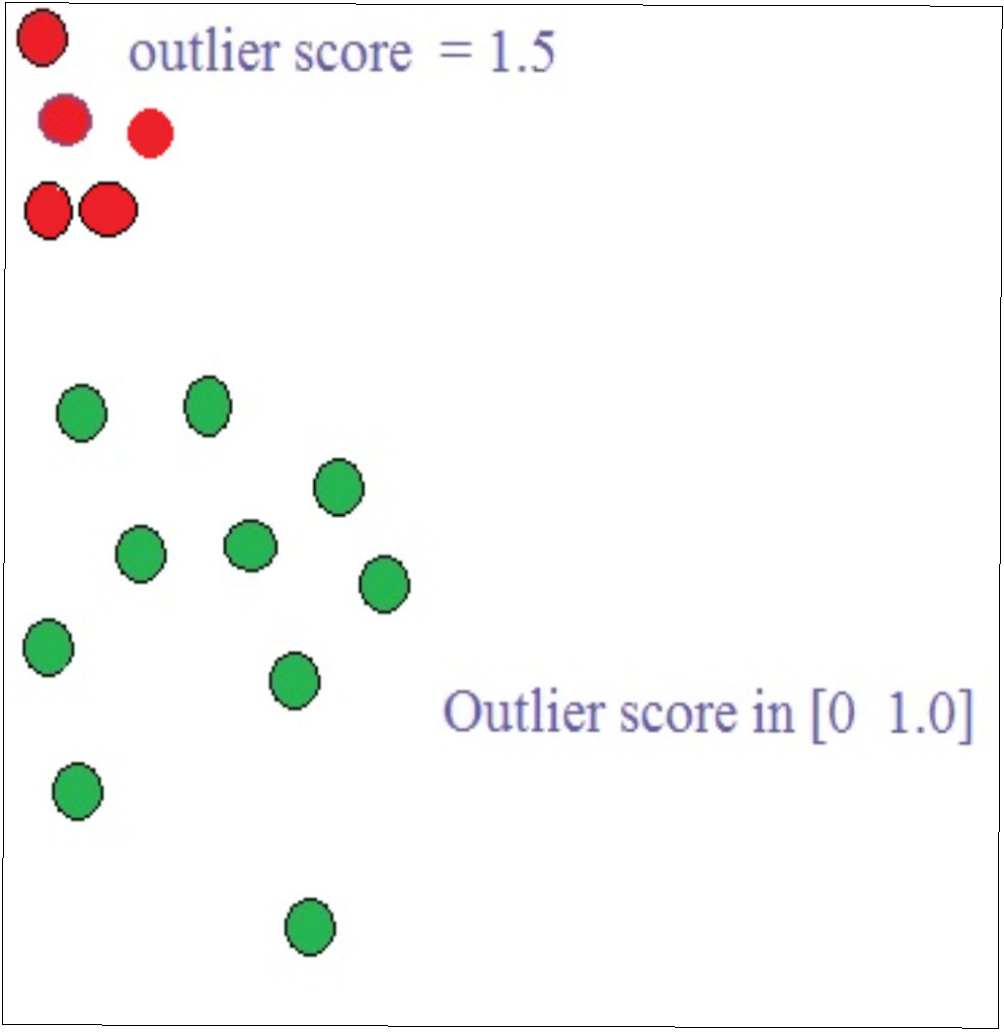}
\caption{Proximity based outlier score with $k=5$}
\label{knn5}
\end{figure}

We note that the outlier score of a data point depends on the value of $k$, the number of nearest neighbors. If the value of $k$ is too small, say 1, then a small number of neighboring outliers  will contribute a small outlier score and thus hamper the performance of the algorithm (see Fig. \ref{knn}). On the other hand, a large value of $k$ will make a group of points having nearest neighbors less than $k$ to become outlier Fig. \ref{knn5}. 

\begin{figure}
\centering
\includegraphics[width=4in, height=3in ]{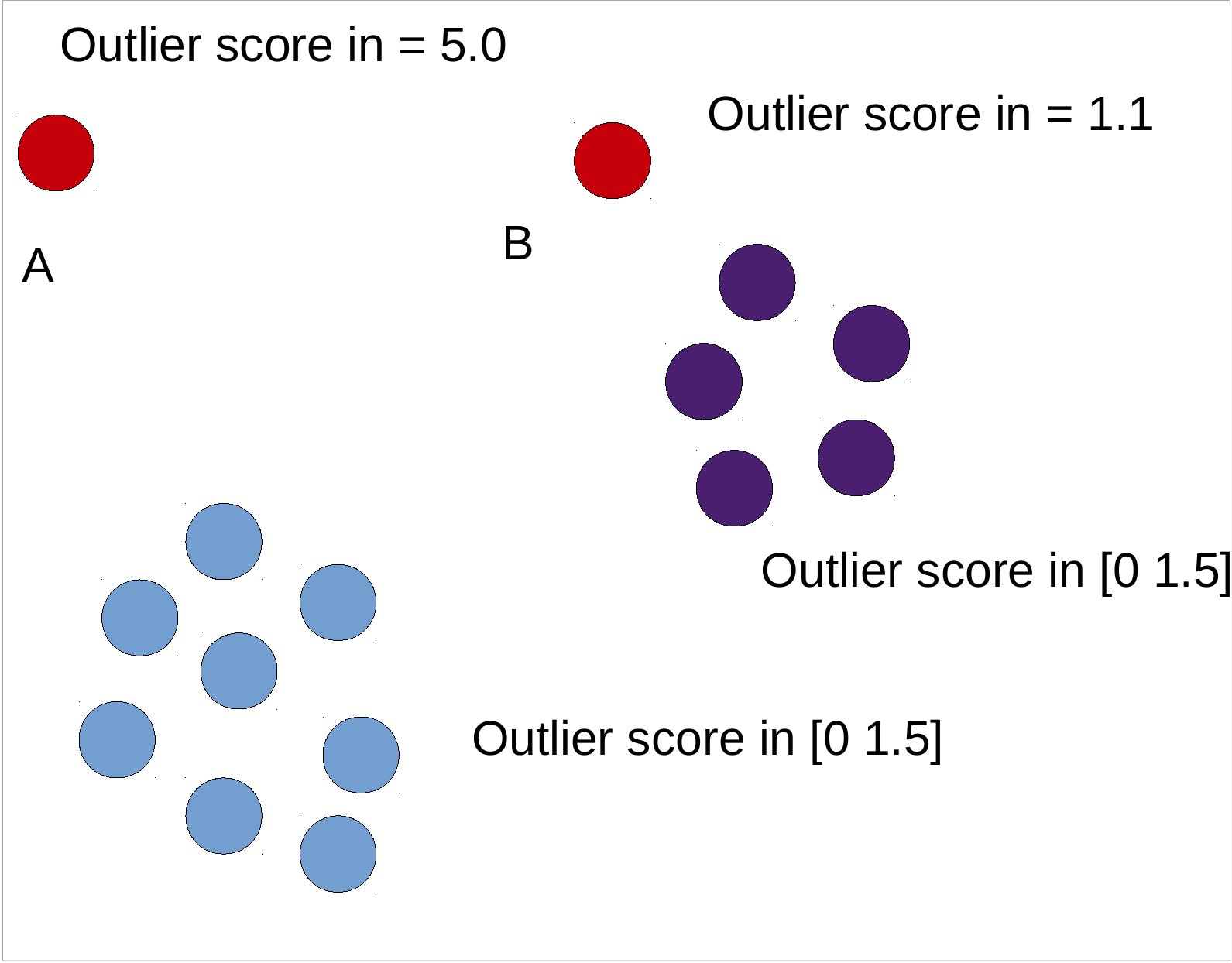}
\caption{Proximity based outlier score with $k=5$}
\label{knn5diffden}
\end{figure}

Proximity-based anomaly detection using the sum of distances of the given data point from its $k$-nearest neighbors as anomaly score has been used in \cite{eskin2002geometric,angiulli2002fast,zhang2006detecting}.  Nearest-neighbor based anomaly detection similar to the aforementioned technique has been used to detect fraudulent credit card transactions in \cite{bolton2001unsupervised}.

\noindent \textbf{Pros and Cons of Proximity-Based Approach: }
Proximity-based approaches are simple and easy to apply in comparison to statistical techniques.
However, their running time is in the order of $O(n^2)$ and this makes them less efficient over high dimensional data comprising millions and billions of points. Secondly, outlier score is sensitive to the value of $k$, which is NP-hard to determine in practice. In addition, they perform poorly over clusters of varying densities. To see this, consider outlier score of points $A$ and $B$ in Fig. \ref{knn5diffden}. Clearly, point $A$ is correctly identified as an outlier but point $B$ has outlier score even less than points in the green cluster. The reason is due to  different sparsity and density of clusters.

\subsubsection{Density-Based  } \label{density}
\vspace{-.5cm}
Density-based outlier detection scheme states that outliers are found in sparse region. In fact, density-based approach is similar to nearest-neighbor based approach in the sense that density can be calculated as the inverse of the distance to $k$-nearest neightbors. For example, $k$-nearest neighbor of a data instance is the number of points enclosed within the hypersphere centered at the given data instance. Taking reciprocal of this distance will give us the density of the so called point. Despite this, density-based technique can  not solve the issue of varying densities similar to the reasons described for proximity-based approach. Hence, the concept of relative density is introduced as given by \eqref{relden}.
\begin{equation}\label{relden}
avg relative density (x,k)= \frac{density(x,k)}{\sum_{y \in N(x, k)}density(y,k)/|N(x, k)|} 
\end{equation}
where $density(x,y)$ is the density of the point $x$. 
LOF (local outlier factor) proposed by Breunig et. al.\cite{Breunig2000} uses the concept of relative density. LOF score for a point is equal to the ratio of the average relative density of the $k$-nearest neighbor of the point and local density of the data point. The local density of a data point is found by dividing $k$ (the number of nearest neighbors) to the volume of the hypersphere containing $k$-data points centered at the data instance. Clearly, the local density of normal points lying in the dense region will be high while the local density of anomalous points in the sparse region will be low.

In literature, several researchers have worked upon the variants of LOF. Some of them have attempted to reduce the original time-complexity of LOF lower than $O(N^2)$. Some compute local density in a different way and others have proposed a variant of LOF suitable for different kinds of data. A very recently, Kai Ming Ting et al. \cite{Ting2013} have proposed a novel density estimation technique that has the average case sublinear time complexity and constant space complexity in the number of instances. This order of magnitude improvement in performance can deal with anomaly detection in big data. They have also proposed DEMass-LOF algorithm that does not require distance calculation and runs in sublinear time  without any indexing scheme.

 \noindent \textbf{Pros and Cons of Density-Based Approach: }
Density-based approaches suffer from the same malady as their counterpart (proximity-based). That is, they have the computational complexity of $O(N^2)$. In order to minimize it, efficient data structures like \emph{ k-d tree} and \emph{R-tress} have been proposed \cite{Bentley1975}. Despite this, modified techniques do not scale over multiple attributes nor do they provide anomaly score for each test instance, if required.

\subsection{Information-theoretic Approaches}
Besides approaches to anomaly detection mentioned above, there are several others approaches to solving the anomaly detection. Below, we describe one such approach based on  information theory. It is a stream of Applied mathematics, Computer science etc. that deals with quantitative  information that can be gleaned from the data. It uses several such measures like    \emph{    Kolmogorov Complexity, entropy, relative entropy}. etc.
\begin{assumption} 
Anomalies in the data infuses erratic information content in the data set.
\end{assumption}
\vspace{-.7cm}
Basic anomaly detection algorithm in this category works as follows. Assume $\Theta(N)$ is the complexity (Kolmogorov) of the given data set D. The objective is to find the minimal subset of instances \emph{I} such that $\Theta (N) -\Theta (N-I)$ is maximum. All the instances thus obtained will be anomalies. 

One notable work under this assumption is \cite{Keogh2004}. In \cite{Keogh2004}, the author uses the size of the compressed data file as a measure of the dataset's \emph{Kolomogorov Complexity}. \cite{arning1996linear} utilizes the size of the regular expression to estimate the \emph{Kolomogorov Complexity}. Besides the above, information theoretic measures such as \emph{entropy, relative uncertainty} etc. has been used in \cite{ando2007clustering,he2005optimization,lee2001information}.
\\
\noindent \textbf{Pros and Cons of Information Theoretic Approach: }
A major drawback of Information theoretic approach is that they involve dual optimization. First, minimize the subset size and second, maximize the decrease in the complexity of the data set. Hence their running time is exponential in the number of data points. Some approximation search techniques have been proposed. For example, \emph{Local Search algorithm} to approximately find such a subset in $O(n)$ time.
The advantage of the information-theoretic approach is that they can be employed in an unsupervised setting and does not make any assumption pertaining to the distribution of the data.

\subsection{Spectral-Theory Based Approaches}\label{spectral}
Spectral theory deals with the problems in high dimensions. They assume that the data set can be embedded into much lower dimensions while still preserving the intrinsic structure. In fact, they are derived from Johnson-Lindenstrauss lemma \cite{Johnson1984} (see Appendix \ref{jl} for the definition).

\begin{assumption} Dataset can be projected into lower dimensional subspace such that normal and anomalous instances appear significantly different.
\end{assumption}
\vspace{-.7cm}
A consequence of the projection in lower dimension manifold is that not only dataset size is reduced but also we can search outliers in the latent space because of correlation among several attributes. The fundamental challenge encountered by such techniques is to determine such lower embeddings which can sufficiently distinguish anomalies from the normal instances. This problem is nontrivial because there are an exponential number of dimensions on which data can be projected. Some notable work in this domain  use Principal Component Analysis (PCA) for anomaly detection \cite{Shyu2003} in network intrusion, Compact Matrix Decomposition (CMD) for anomaly detection in a sequence of graph etc \cite{Sun2007}.

\noindent \textbf{Pros and Cons of Spectral-Theoretic Approach:}
Dimensionality reduction techniques like PCA work linearly in data size but quadratic in the number of dimensions. On the other hand, nonlinear techniques run linearly in the number of dimensions but polynomial in the number of principal components \cite{Gunter2007}. Techniques performing SVD on the data have $O(N^2)$ time complexity.

The advantage of spectral methods is that they are suitable for anomaly detection in high dimension data. Also, they can work in an unsupervised setting as well as the semi-supervised setting. The disadvantage of spectral techniques is that they will separate the anomaly from the normal instances provided there exist a lower dimension embedding. Another disadvantage is that they suffer from high computation time.

\section{Modern Approaches to Anomaly Detection}
Traditional approaches to anomaly detection in big data suffers from miscellaneous issues. For example, statistical techniques require underlying distribution to be known a priori.  Proximity-based and density-based approaches require appropriate metric to be defined for calculating anomaly score and run in quadratic time with the number of data instances. Clustering based techniques need some kind of optimization for reducing quadratic time complexity and do not generalize for heterogeneous data. Similarly, Information-theoretic and spectral techniques require an appropriate measure of information in case of the former and embedding for the latter. 

The point is that they can not handle the case of high dimension, heterogeneous, noisy, streaming, and distributed data that is found ubiquitously everywhere. For example, aircraft navigation data is highly complex, heterogeneous, and noisy in nature that requires sophisticated tools and techniques for online processing so as to thwart any likely accident. Similarly, mobile phone call record demands batch processing of millions and billions of calls every day for a potential terrorist attack.

Above points indicate that we need to have some kind of mechanism that not only reveals potential anomalous record in the complex data but also gives us insight about  it. This makes sense because manual knowledge discovery in big data  is a nontrivial task.
Therefore, we will look at techniques that meet some of the aforementioned goals in the forthcoming sections. In particular, we will discuss recent approaches to anomaly detection,  their results, and drawbacks. Further, techniques going to be covered in this chapter are suitable for applying anomaly detection in unsupervised as well as semi-supervised mode. This is also an important point since real world data sets are mostly unlabeled. 

\subsection{Non-Parametric Techniques}
 Non-parametric technique refers to a technique in which the number of parameters grows with the size of the data set or that does not assume that the structure of the model is fixed. Some examples of non-parametric models are histograms, kernel density estimator, non-parametric regression, models based on Dirichlet process, Gaussian Process etc. A key point about non-parametric techniques is that they do not assume that data come from some fixed but unknown distribution. Rather, they make fewer assumptions about the data and hence are more widely applicable. Some notable works in anomaly detection using non-parametric models are described below.

 Liang et al. \cite{Liang2011} propose Generalized latent Dirichlet allocation (LDA) and a mixture of Gaussian mixture model (MGMM) for unimodal and multimodal  anomaly detection on galaxy data. They assume that sometimes data besides being anomalous at an individual level is also anomalous at the group level and hence techniques developed for point anomaly detection fails to identify anomalies at the group level. However, their model is highly complex and learns a lot of parameters using variational inference method. 
 
 In the same line of work, Rose et al. \cite{Rose2014} have proposed  group latent anomaly detection(GLAD) algorithm for mining abnormal community in social media. Their model also suffers from the same problem as that of Liang et al. above, i.e., it is complex and involves learning of a large number of parameters. In \cite{Rodner2011}, the author uses Gaussian process (GP) for one-class classification similar to one class SVM approach but in a non-parametric way for identifying anomalies in wire ropes. However, their approach generates falls alarm and does not incorporate prior knowledge about the structure of the rope. The same group also combined GP with kernel functions for one class classification \cite{Rodner2013}. They show that GP combined with kernel functions can outperform support vector data description \cite{Tax2004} over various data sets.

The potential advantage of the non-parametric techniques is that they do not assume that the data is coming from some fixed but unknown distribution.  Also as the  number of parameters grows linearly with the size of the input, these techniques can handle dynamic nature of the data. However, on the flip side, there is a lack of suitable methods for  estimating hyperparameters  such as kernel bandwidth when GP prior is combined with the kernel function. Unless one has the right kernel bandwidth, performance of GP methods for anomaly detection is poor. Secondly, doing cross-validation for finding parameters is infeasible since non-parametric techniques involve parameters which can easily go beyond hundreds and thousands in number.

\subsection{Multiple Kernel Learning}
\begin{figure}
\centering
\includegraphics[width =5in, height=3in]{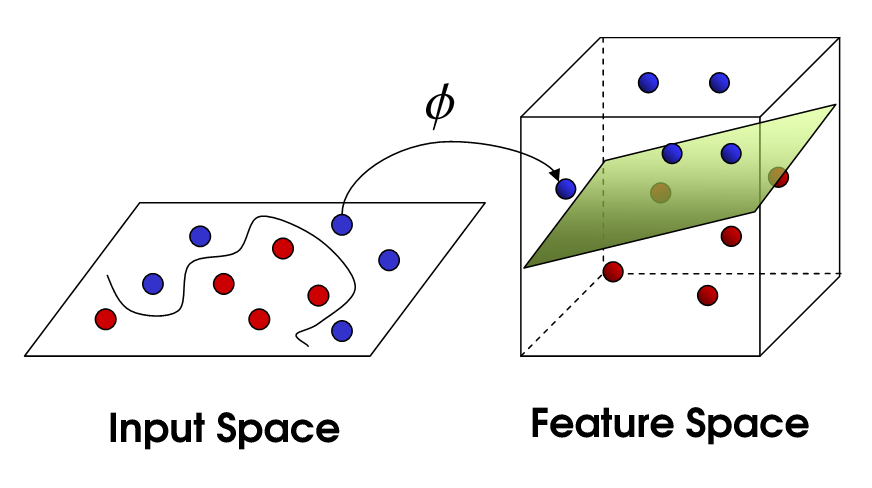}
\caption{An Example of working of the kernel function. Any non-linearly separable data set can be mapped to higher dimension feature space through the kernel function where data set can be separated via a linear decision boundary.}
\label{kernel}
\end{figure}

Kernel methods \cite{Taylor2004} provide a powerful framework for analyzing data in high dimension \ref{kernel}. They have been successfully applied in ranking, classification, regression over multitude of data. 

Let us define some term before delving into depth.
Kernel is a function $\kappa$ such that for all ${\bf x}, {\bf z} \in \cal{X} $ satisfies 	
\begin{equation}
\kappa(\mathbf{x},{\bf z}) = \langle \phi ({\bf x}), \phi({\bf z}) \rangle
\end{equation}
where $\phi$ is a mapping from some Hilbert space $\cal{X} $ to an (inner product) feature space $\cal{F}$
\begin{equation}
\phi :{\bf x} \in \cal{X} \longmapsto \phi ({\bf x}) \in \cal{F}
\end{equation}
Intuitively, it says that kernel implicitly computes the inner product between two feature vectors in high dimensions feature space, i.e., without actually computing the features.

Some examples of kernel function are:
\begin{itemize}
\item Gaussian Kernel(RBF):   $ \kappa({\bf x},{\bf z}) = exp \left( -\frac{||{\bf x} - {\bf z}||^2_2}{2\sigma^2}\right) $

\item Polynomial Kernel :  $\kappa({\bf x},{\bf z}) = \left(1 +  {\bf x}^T{\bf z}\right)^n $
\end{itemize}
\noindent Multiple Kernel Learning (MKL) \cite{Bach2004} has recently gained a lot of attention among data mining/machine learning community. The growing popularity of MKL is due  to the fact that it combines the power of several kernel functions into one framework. This arouses the curiosity: can we apply MKL to anomaly detection  in heterogeneous data coming from multiple sources?  The answer has been recently given by empirical result on Flights Operation Quality Assurance (FOQA) archive data of S. Das et al.\cite{Das2010}
at NASA. 
MKL theory is essentially based on the theory of kernel methods \cite{Smola2001}. The idea is to use a kernel function satisfying \emph{Mercer's condition} (see Appendix \ref{mercer} for the definition) that finds similarity between pair of objects in a high dimension feature space. The salient feature is that a valid kernel can compute the similarity between objects of any kind. This leads to the idea of combining multiple kernels into one kernel and use it for classification, regression, anomaly detection   task etc.

 MKL learns kernel from the training data. More specifically, how the kernel $\kappa$ can be learned as a linear(convex) combination of the given base kernels $\kappa_i$ as shown in \eqref{mkleq}? 
\begin{equation}\label{mkleq}
\kappa({\bf x}_i,{\bf z}_j) =\sum_{k}{\theta_k \kappa_k({\bf x}_i,{\bf z}_j)}
\end{equation}
where $\theta_k \geq 0$, $k=1,2...,K$. The goal is to learn parameters $\theta_k $ such that resultant kernel $\kappa$ is positive semi-definite(PSD) (see the appendix \ref{psd} for the definition of positive-semi definiteness). 

Recently, Verma et al. \cite{Varma2009} have proposed Generalized MKL that can learn millions of kernels over half a billion of training points. After learning multiple kernels in a joint fashion, any anomaly detection technique capable of using kernels as a similarity measure can distinguish between  normal and anomalous instances such as one-class SVM. In \cite{Das2010}, the author uses two base kernels; one for discrete sequences and one for continuous data. The kernel $\kappa_d$ corresponding to discrete sequences is computed using LCS (longest common sub-sequence) while kernel $\kappa_c$ corresponding to continuous data is inversely proportional to the distance between the Symbolic Aggregate Approximation (SAX) \cite{DBLP:conf/icdm/KeoghLF05} representation of the points ${\bf x}_i$ and ${\bf z}_j$. The combined kernel is fed to one-class SVM and its performance is compared with two baseline algorithms namely Orca and SequenceMiner. The results over various simulated and real data demonstrate that  multiple kernel anomaly detection algorithm (MKAD) outperforms the baseline algorithms in terms of detecting different kinds of faults (discrete and continuous).

In \cite{Song2010}, the author uses MKL approach to anomaly detection in network traffic data which is  heavy-flow, high-dimension and non-linear. Essentially, they use sparse and non-sparse kernel mixture based on $L_p$ norm MKL proposed by Kloft et al. \cite{Kloft2011}. Another  work proposed by Tax et al. \cite{Tax2004} is based on support vector data description (SVDD). It builds hypersphere around normal data leaving outliers either at the boundary or outside of it. Their work  uses support vector classifier that gives an indication that MKL approach can be exploited to build hypersphere in  a high dimensions space. This forms the line of the motivation of Liu et al. \cite{Liu2013}, Gornitz et al. \cite{Goernitz2013} in a semi-supervised as well as unsupervised setting to use MKL for anomaly detection.

Thus we see that MKL can tackle high dimension and heterogeneous nature of big data very nicely. However, further work needs to be done to explore the possibility of using MKL in the streaming and distributed anomaly detection scenario.


\subsection{Non-negative Matrix factorization}
Non-negative matrix factorization (NNMF) as a technique for anomaly detection in image data was studied by Lee and Seung in 1999 \cite{Lee1999}. The non-negative matrix factorization problem is  posed as follows:
 
Let ${\bf A}$ be $m\times n$ matrix whose components $a_{ij}$ are non-negative i.e. $a_{ij}\geq 0$. Our goal is to find non-negative matrices ${\bf W}$ and ${\bf H}$ of size $m \times k$ and $ k \times n$ such that \eqref{nnf} is minimized.
\begin{equation}\label{nnf}
F({\bf W},{\bf H}) = \frac{||{\bf A} - {\bf W}{\bf H}||^2_F}{2}.
\end{equation}
where $k \leq min\{m,n\}$ and depends upon the specific problem to be solved. In practice, \emph{k} is much smaller than  \emph{rank({\bf A})}. The product \emph{{\bf W}{\bf H}} is called non-negative matrix factorization for the matrix \emph{{\bf A}}. It should be noted that the above problem is non convex in \emph{{\bf W}} and \emph{H} jointly. Therefore, algorithms proposed so far in the literature seek to approximate matrix \emph{{\bf A}} via product \emph{{\bf W}{\bf H}} i.e. ${\bf A} \approx {\bf W}{\bf H} $. Thus, it is obvious that \emph{{\bf W}{\bf H}} represents \emph{{\bf A}} in a very compressed form.

%
After Lee and Seung initial NNMF algorithm based on multiplicative update rule, several variants have been proposed in order to solve \eqref{nnf}. For example,  modified multiplicative update \cite{Lin2007}, projected gradient descent \cite{Lin07projectedgradient}, alternating least (ALS) \cite{Berry2006}, alternating non-negative least square (ANLS)\cite{Lin07projectedgradient} Quasi-Newton \cite{Kim2007}  etc. have been proposed. 

\begin{figure}

\centering
\includegraphics[width=6in, height=2.5in ]{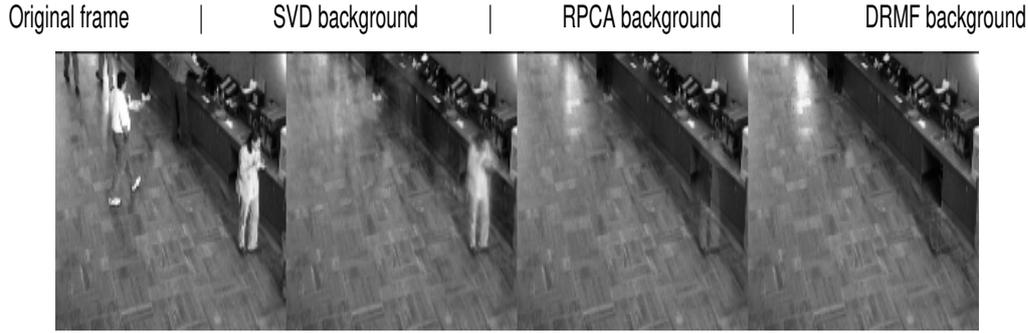}
\caption{Video activity detection  from left to right are: the original frame, background and foreground  \cite{Liang2011}}
\label{drmfv}
\end{figure}

Recently, Liang et al. \cite{Liang2011} propose direct robust matrix factorization (DRMF) for anomaly detection. The basic idea used by them is to exclude some outliers from the initial data and then ask the following question: What is the optimal low rank you can obtain if you ignore some data? They formulate the problem as an optimization problem with constraints on the cardinality of the outlier set and the rank of the matrix. Essentially, they solve the problem shown in \eqref{drmf}.
\begin{equation}\label{drmf}
\begin{aligned}
& \underset{{\bf L},{\bf S}}{\text{minimize}}
& &  ||({\bf X} -{\bf S}) - {\bf L}||_F \\
& \text{subject to}
& & rank({\bf L}) \leq K\\
& &  &||{\bf S}||_0 \leq e
\end{aligned}
\end{equation}
where ${\bf S}$ is the outlier set and ${\bf L}$, low rank approximation to ${\bf A}$. $K$ is the rank desired and $e$ is the maximal number of nonzero entries in ${\bf S}$. 
$\|\cdot\|_F$ denotes the frobenious norm of the matrix (square root of the sum of squares of each element).
In the matrix factorization paradigm, solution to optimization problems involving rank or set cardinality is nontrivial. The author in the aforementioned work uses the trick that the problem is decomposable in nature and hence solvable by block-coordinate descent algorithm \cite{Richt:2012}. They use the DRMF algorithm to separate background from foreground(noise) which has application in video surveillance \ref{drmfv}.\\
In an another work by Allan et al. \cite{Allan2008}, they use NMF to generate feature vectors that can be used to cluster text documents. More specifically, they exploit the concept, derived from NMF, called sum-of-parts representation that shows term usage pattern in the given document. The coefficient matrix factors and features thus obtained are used to cluster documents.  This procedure maps the anomalies of training documents to feature vectors.
In addition to locating outliers in latent subspace, NMF has been used to interpret outliers in that subspace as well. In this domain, work of Fei et al. \cite{Wang2013} is significant. More specifically, they combine NNMF with subspace analysis so that not only outliers are found but also they can be interpreted.

From the above discussion, we see that non-negative matrix factorization can be employed for anomaly detection in large and sparse data. However, suitability of NMF for anomaly detection in streaming, heterogeneous and distributed setting is still unexplored. 

\subsection{Random Projection}
\begin{figure}

\centering
\includegraphics[width=4.5in, height=3.5in ]{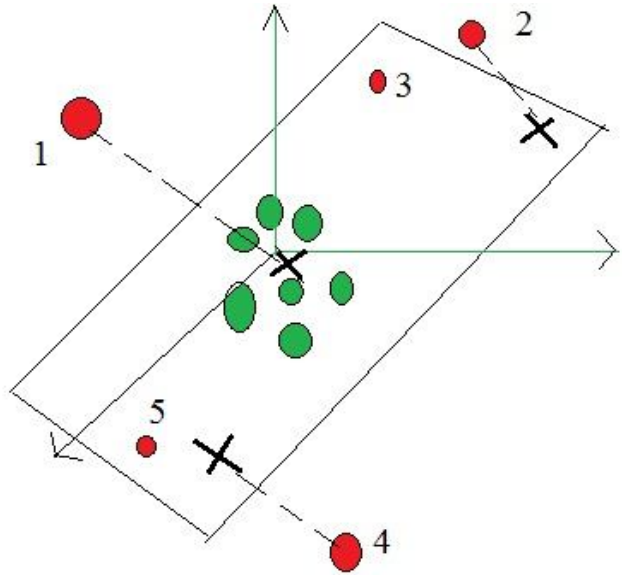}
\caption{Outliers  marked in red color, orientation after projection from 3D to 2D \cite{Hubert2005}.}
\label{proj}
\end{figure}

In section \ref{spectral}, we discussed spectral-techniques to detect outliers. Specifically, spectral-techniques are based on old idea of PCA, CMD etc. Random projection is also based on spectral-theory. However, the main motivation to present random projection here is that there has been recent surge in the theory and algorithms for random projection based techniques (see for example \cite{Durrant:2013,DBLP:conf/icml/DurrantK13}).
Random Projection pursuit is a spectral technique that looks for anomalies in a latent subspace. This technique is particularly suitable for high dimensional data with noise and redundancy. The assumption is that effective dimensionality in which outliers and normal data reside is very small. The technique works on the principle of multiple subspace view \cite{Muller2012}. The basic idea is to project high dimension data into lower dimension subspace such that outliers stand out even after projection (see Fig. \ref{proj}). In Fig. \ref{proj}, after projection, it turns out that point 1 loose its identity as outlier. Point 2 and 4 continue to remain outlier while points 3 and 5 have zero effect of projection.

Note that projection is done to reduce the dimensionality of the data set. Subsequently, we can apply any anomaly detection approach provided certain conditions are met as described below.
\vspace{-.7cm}
\begin{itemize}
\item {\bf Condition 1: } Projection should preserve the pairwise distance metric between data points with very high probability (informal statement of Johnson-Lindenstrauss lemma).
\item {\bf Condition 2: }	Projection should preserve the distance of points to their $k$-nearest neighbor (result from Vries et al. work \cite{Vries2012}).
\end{itemize}

Note that the $1^{st}$ condition is applicable to approaches that use some metric (e.g. distance) to calculate outlier. On the other hand, Condition 2 applies to proximity-based approaches discussed in section \ref{proximity}.

In \cite{Vries2012}, Vries et al. introduce projection-indexed nearest neighbor (PINN) approach that is essentially based on projection pursuit. They first apply random projection (RP) to reduce the dimension of the dataset so as to satisfy the condition 1. Thereafter, they use local outlier factor (LOF) \cite{Breunig2000} to find the local outlier in a data set of size 300,000 and 102,000 dimensions.  In \cite{Huan2010},  the author uses convex-optimization approach to outlier (anomaly) pursuit for \emph{matrix recovery} problem. Their approach recovers the optimal low-dimensional subspace and marks the distorted points ( the anomaly in image data). In \cite{Aouf2012}, Mazin et al. apply a variant of random projection which they call \emph{Random Spectral Projection}. The idea is to use Fourier/Cosine spectral projection for dimension reduction. They show that random samples of Fourier spectrum performs better than random projection in terms of accuracy and storage over text document. In \cite{Muller2012}, Muller et al. propose \emph{OutRank}, a novel approach to rank outliers. OutRank essentially uses subspace view of the data and compares clustered regions in arbitrary subspaces.  It produces the degree outlierness score for each object.

 \noindent \emph{\bf Challenges in using Random Projection pursuit:} Major challenge in using  anomaly detection techniques employing random projection is that they should be able to work in reduced dimension with intrinsic structure of the data. Second issue is how to efficiently choose the number of dimensions to project the data?

\subsection{Ensemble Techniques}
Ensemble technique works on the principle of "Unity is strength". That is, they combine the power of individual techniques of outlier detection and produce astounding results provided certain criteria is met. Although, ensemble techniques have been miraculously applied for classification\cite{Minz:2015}, clustering task long  ago, they have recently been used in anomaly detection scenario. Increasing stardom of ensemble techniques is their ability to locate outliers effectively in high dimension and noisy data \cite{Charu2001}.  
A typical outlier ensemble contains a number of components that aids to its power.  These are:
\begin{itemize}
\item {\bf Model Creation}: An individual model/algorithm is required to create ensemble in the first place. In some cases, the methodology can be simply random subspace sampling.
\item{\bf Normalization}:  Different techniques produce outlier scores that assign different meaning to outlierness of a point. For example, some models assume low outlier score meaning high degree of outlierness. Whereas others assume vice versa. So it is important to consider different  ways for merging anomaly scores  together to create meaningful outlier score.
\item{\bf Model Combination}: This refers to the final combination function which is used to create the final outlier score.
\end{itemize}
In literature, ensembles have been categorized on the basis of component independence and constituent component. 
The first categorization assumes whether the components are developed independently or they depend on each other. These are of two types: In \emph{sequential} ensemble, algorithms are applied in tandem so that the output of one algorithm affects the other; producing either better quality data or specific choice on the algorithm. The final output is either weighted combination or the result of finally applied algorithm. In \emph{Independent} ensemble, completely different algorithm or the same algorithm with different instantiations is applied on the whole or part of the data under analysis.

In categorization by constituent component, \emph{data centric} ensemble picks a subset of data or data dimension (e.g. bagging/boosting in classification) in turn and apply anomaly detection algorithm. \emph{Model} centric approach attempts to combine outlier scores from different models built on the same data. The challenge encountered is that how to combine scores if they have been produced on different scales or format? Different combination functions like \emph{min, max, avg.} etc have been proposed so as to make ensemble techniques work in practice. Further, noteworthy point is that ensemble approach will prove effective when classification based anomaly detection techniques are employed.

\begin{figure}
\centering
\includegraphics[width=4in, height=3in ]{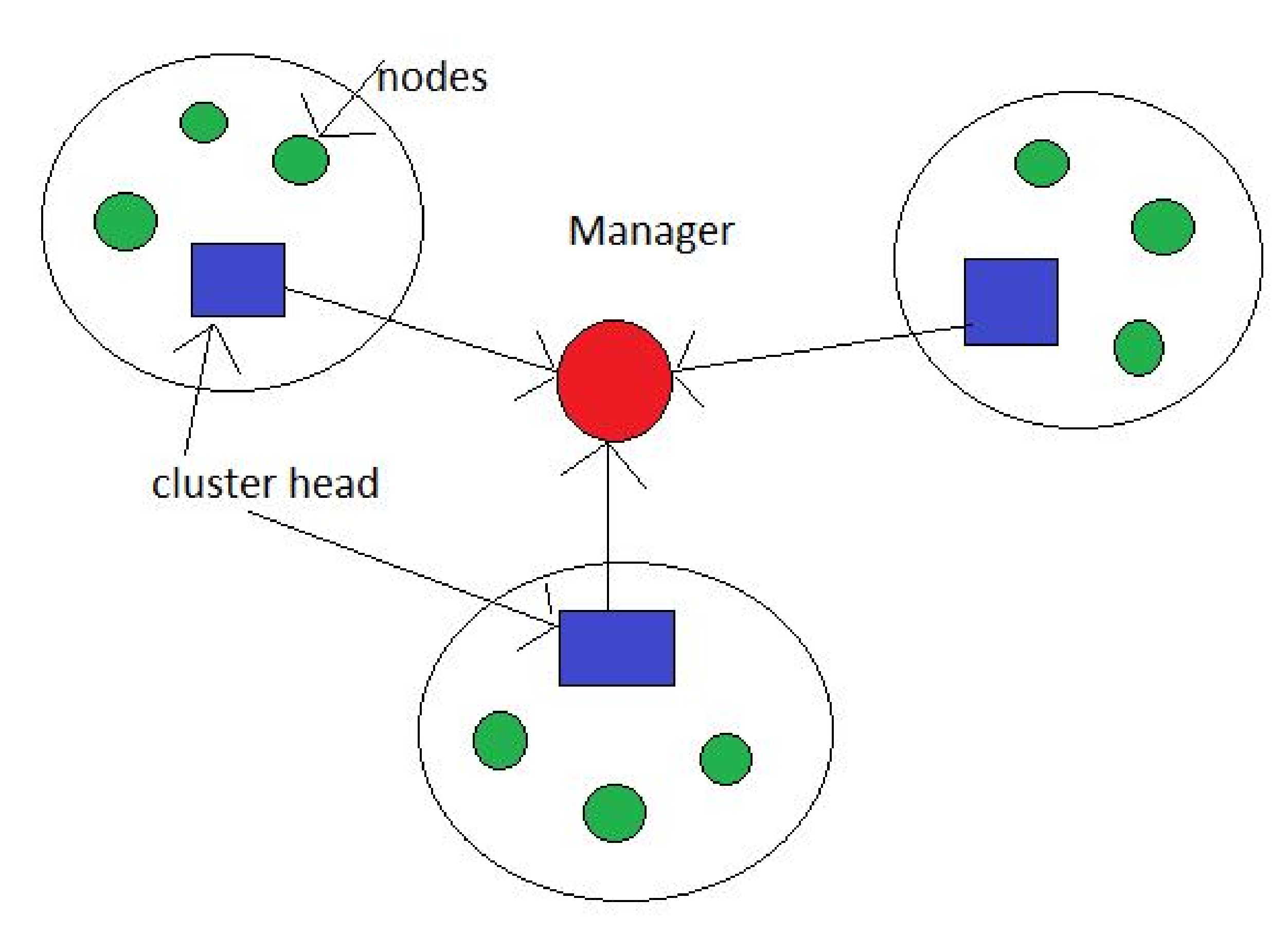}
\caption{Overall infrastructure to support the fusion of anomaly detectors \cite{Cabrera2008}}
\label{manet}
\end{figure}

Although ensemble techniques have deep foundations in classification/clustering, there true power in outlier detection is revealed by Lazarevic et al. \cite{Lazarevic2005} as feature bagging recently. They show that the proposed feature bagging approach can combine outlier scores from the different execution of LOF algorithm in a \emph{breadth-first} manner. That is, first combine the highest scores from all algorithms; second, largest scores are combined next and so on. In \cite{Noto2010}, Noto et al.  use feature prediction using the combination of three classifiers and predictor (FRaC). They combine the anomaly score of FraC using \emph{surprisal} anomaly score given by \eqref{sur} which is an information-theoretic measure of prediction. FraC is a semi-supervised approach that learns conserved relationship among features and characterizes the distribution of ``normal'' examples.
\begin{equation}\label{sur}
surprisal (p)= -log(p)
\end{equation}
In the same line of work, Cabrera et al.\cite{Cabrera2008} proposed anomaly detection in a distributed setting of Mobile Ad-Hoc Networks(MANET). They combine the  anomaly scores from local IDS (intrusion detection systems) attached to each node through averaging operation and this score is sent to the cluster head. All cluster head send cluster-level anomaly index to a manager which averages them (see Fig. \ref{manet}). Dynamic Trust Management scheme is proposed in \cite{NCD:2008} for detecting anomalies in wireless sensor network (WSN).  Hybrid ensemble approach for class-imbalance and anomaly detection is recently proposed in \cite{Ajith:2016a}. The author uses the mixture of oversampling and undersampling with bagging and Adaboost and  show improved performance.  The ensemble of SVMs for imbalanced data set is proposed in \cite{VB:2010,VB:2015}.

\noindent \emph{\bf  Challenges in using ensemble approach to combat Anomalies:} Ensemble techniques, although produce high accuracy, brings several challenges with them such as (1) Unsupervised Nature (2) Small Sample Space Problem (3)  Normalization Issues. This first problem refers to the real-world scenario where data is often unlabeled. In such cases, ensemble techniques, which are mostly based on classification, can not be applied. The second issue alludes to the case that anomalies are present in a tiny amount among the huge bundle of normal instances. Normalization issue is related with  different output formats of  different classifiers, specially when heterogeneous models are being trained on. 

\section{Relevant Algorithms for Anomaly Detection}\label{relevant}
Since our work focuses on anomaly detection in big data, we discuss related research works that have been tailored to the data mentioned thereof. Below, we present related work that  includes techniques based on (1) Online learning (2) Class-Imbalance learning  (3) Anomaly detection in a streaming environment and discuss the differences with our work.

\subsection{Online Learning}
The online learning  refers to a learning mechanism where the learner is given one example at a time as shown in Algorithm \ref{online}.
\begin{algorithm}
  \caption{Online Learning Algorithm
    \label{online}}
  \begin{algorithmic}[1]
		\Repeat
    		\State  receive instance: $\mbox{\bf x}_t$
 		\State predict:  $\hat{y}_t$ 
 		\State receive correct label: $y_t $
 		\State  \emph{\bf suffer loss}: $\ell(y_t, \hat{y}_t)$
		\State \emph{\bf update model}	
		\Until{All examples processed}
 \end{algorithmic}
\end{algorithm}
In algorithm \ref{online}, the learner is presented with an example $\textbf{x}_t$ in line no 2. It makes its prediction  $\hat{y}_t$ in line no. 3 and receives correct label  $y_t $ in line no. 4. In line no. 5, it computes its loss due to mistakes made in the prediction and subsequently updates the model in line no. 6. Clearly,  the online learning algorithm is out of \emph{memory issues}  when processing  massive datasets as it looks at one example at a time. Secondly, it has optimal running time of $O(nd)$ provided line 5 and 6 take time $O(d)$, where $n$ is the number of samples processed so far and $d$ is dimensionality of the data.   Thirdly, it is easy to implement. In the next paragraph, we discuss relevant literature work based on online learning and their limitations in tackling outliers. 

 Online learning has its origin from classic work of Rosenblatt on perceptron  algorithm \cite{Rosenblatt1958}. Perceptron algorithm is based on the idea of a single neuron. It simply takes an input instance  $\mbox{\bf x}_t$ and learn a linear predictor of the form $f_t(\mathbf{x}_t)=\mathbf{w}^T_t\mbox{\bf x}_t$, where $\mathbf{w}_t$ is weight vector.  If it makes a wrong prediction, it updates its parameter vector as follows:
\begin{equation}
\mathbf{w}_{t+1}=\mathbf{w}_t +y_t\mathbf{x}_t
\end{equation}
where $\mathbf{w}_{t+1}$ is weight vector at time $t+1$.

\cite{Kivinen2010} propose online learning with kernels. Their algorithm, called $NORMA_\lambda$,  is based on regularized empirical risk minimization which they solve via regularized stochastic gradient descent. They also show empirically how this can be used in anomaly detection scenario. However, their algorithm requires tuning of many parameters which is costly for time critical applications. Passive-Aggressive (PA) learning \cite{Crammer2006} is another online learning algorithm based on the idea of maximizing ``margin'' in online learning framework. PA algorithm updates the weight vector whenever ``margin'' is below a certain threshold on the current example. Further, the author introduce the idea of a slack variable to handle non-linearly separable data. Nonetheless, PA algorithms are sensitive to outliers. The reason is as follows: PA algorithm applies the update rule $\mathbf{w}_{t+1}-\mathbf{w}_t =\tau_t y_t\mathbf{x}_t$, where $\tau_t$ is a learning rate. In the presence of outliers, the minimum of $\frac{1}{2} \|\mbox{\bf w} - \textbf{w}_t \|_2$ could be high since $\|\mbox{\bf x}_t\|_2$ is high for outliers. Other online learning algorithms in the literature are MIRA \cite{Crammer:2003}, ALMA \cite{Gentile:2002}, SOP \cite{Cesa-Bianchi:2005}, ARROW \cite{Crammer:2013}, NAROW \cite{Orabona:2010}, CW \cite{NIPS2008_3554}, SCW \cite{Jialei:2012} etc. Many of these algorithms are variant of the basic PA algorithm and perceptron algorithms and hence sensitive to outliers. A thorough survey of these algorithms is not feasible here. For an exhaustive survey on online learning see \cite{Nico2006}.

 Online lerning based algorithms presented above, although scale to the number of data points, do not scale with the number of data dimensionality. For example, Passive-Aggressive (PA) learning \cite{Crammer2006},  MIRA \cite{Crammer:2003}, ALMA \cite{Gentile:2002}, SOP \cite{Cesa-Bianchi:2005}, ARROW \cite{Crammer:2013}, NAROW \cite{Orabona:2010}, CW \cite{NIPS2008_3554}, SCW \cite{Jialei:2012} etc. have running time complexity of $O(nd)$.  Noteworthy points about algorithms mentioned above are: (i) they are sensitive to outliers and cannot handle class-imbalance problem without modification (ii) they do not consider sparsity present in the data except CW and SCW. Though CW and SCW exploit sparsity structure in the data, they are not designed for the class-imbalance problem. In the present work, we attempt to address these issues through the lens of online and stochastic learning in Chapter \ref{Chapter3}.

\subsection{Class-Imbalance Learning}
Class-imbalance learning aims to correctly classify minority examples  in a binary classification setting. In the literature, there exist solutions that are either based on the idea of sampling or weighting scheme. In the former case, either majority examples are undersampled or minority examples are oversampled.  In the latter case, each example is weighted differently and the idea is to learn these weights optimally. Some examples of sampling based technique are $SMOTE$ \cite{Chawla2002}, $SMOTEBoost$ \cite{Chawla2003}, $AdaBoost.NC$ \cite{Shou2010} and so on. Works that use weighting scheme include cost-sensitive learning \cite{Elkan2001}, Adacost \cite{Fan1999}, \cite{Ling2006}, \cite{Liu2006}, \cite{Lo2011} and so on. It is worthwhile to mention here that only a few work exist that jointly solve the class-imbalance learning and online learning. Below, we mention some work which closely match our work.

In \cite{Wang2013}, the author proposed sampling with online bagging (SOB) for class-imbalance detection. Their idea essentially is based on resampling, that is, oversample minority class and undersample the majority class from Poisson distribution with average arrival rate of $N/P$ and $R_p$ respectively, where $P$ is the total number of positive examples, N is the total number of negative examples and $R_p$ is the recall on positive examples. Essentially, \cite{Wang2013} propose an online ensemble of classifiers where they may achieve high accuracy, but the training of ensemble of classifiers is a time-consuming process. In addition, \cite{Wang2013} does not use the concept of surrogate loss function to maximize \emph{Gmean}.

\cite{Jialei2014} proposes an online cost-sensitive classification for imbalanced data. One of their problem formulation is based on the maximization of the weighted sum of sensitivity and specificity and the other is the minimization of the weighted cost. Their solution is based on minimizing convex surrogate loss function (modified hinge loss) instead of the non-convex 0-1 loss. Their work closely matches our work. But, in section \ref{algo} we show that the problem formulation of \cite{Jialei2014} is different from our formulation and the solution technique they adopt is based on the online gradient descent while ours is based on the online passive-aggressive framework. Specifically, the problem formulation of \cite{Jialei2014} is a special case of our problem formulation. The further difference will become clear in section \ref{exp4}. In \cite{Zheng201527}, the author proposes a  methodology to detect spammers in a social network. Essentially, \cite{Zheng201527} introduces which features might be useful for detecting spammers on online forums such as facebook, twitter etc. One of the major drawbacks of their proposed method is that it is an offline solution  and, therefore, can not handle big data. Secondly, they apply vanilla SVM for spammer detection which could be not  effective due to the use of hinge loss within SVM.
 The present work also attempts to solve the open problem in \cite{Zheng201527} by online spammer detection with low training time. 
 
 Recently Gao et al. \cite{Gao:2016}  propose a class-imbalance learning method based on two-stage extreme learning machine (ELM). Although, they are able  to handle class-imbalance, scalability of two-stage ELM in high dimensions is not shown. Besides, two-stage ELM solves the class-imbalance learning problem in the \emph{offline} setting. ESOS-ELM proposed in \cite{Mirza2015} uses an ensemble of a subset of sequential ELM to detect concept drifts in the class-imbalance scenario. Although ESOS-ELM is an online method, they demonstrate the performance on low dimensional data sets only (largest dimension of  the data set tested is  $<$ 500) and do not exploit sparsity structure in the data  explicitly.  However, none of the techniques mentioned above tackles the problem of class imbalance in huge dimension (quantity in millions and above), nor do they handle the problem structure present in the data such as sparsity. In our present work, we propose an algorithm that is scalable to high dimensions and can exploit sparsity present in the data. 
 
 Cost-sensitive learning can be further categorized into the offline cost-sensitive learning (OffCSL) and the online cost-sensitive learning(OnCSL).  OffCSL incorporates costs of misclassification into the offline learning algorithms such as cost-sensitive decision tress \cite{Drummond2005}, \cite{Ling:2004}, \cite{Ling2006}, cost-sensitive multi-label learning \cite{ Lo2011}, cost-sensitive naive Bayes \cite{Chai:2004} etc. On the other hand, OnCSL-based algorithms use cost-sensitive learning within the \emph{Online} learning framework. Notable work in this direction includes the work of Jialei et al. \cite{Jialei2014}, Adacost \cite{Fan1999}, SOC \cite{Dayong:2015}. It is to be noted that the cost-sensitive learning methods have been proved to outperform sampling-based methods over Big data \cite{Gary:2007}. On the other hand, OnCSL-based methods are more scalable over high dimensions when compared to their counterpart OffCSL-based methods due to the processing of one sample at a time in  case of the former.

\subsection{Anomaly Detection in a Streaming environment}
Online outlier detection in sensor data is proposed in \cite{Subramaniam:2006}. Their method uses \emph{kernel density estimation} (KDE) to approximate the data distribution in an online way and employ distance-based algorithms for detecting outliers. However, their work suffers from several limitations. Firstly, although the approach used in \cite{Subramaniam:2006} is scalable to multi-dimensions, their method does not take into account \emph{evolving} data stream. Secondly, using KDE to estimate data distribution in the case of \emph{streaming} data is a non-trivial task. Thirdly, their algorithm is based on the concept of sliding-window. Determining the optimal width of the sliding window is again non-trivial. Abnormal event detection using online SVM is presented in \cite{Davy:2006}. In \cite{Otey:2006}, the author presents a \emph{link-based} algorithm (called LOADED) for outlier detection in mixed-attribute data. However, LOADED does not perform well with continuous features and experiments were conducted on data sets with dimensions at most 50.

Fast anomaly detection using Half-Space Trees was proposed in \cite{Tan:2011}. Their Streaming HS-Tree algorithm has constant amortized complexity of $O(1)$ and constant space complexity of $O(1)$. Essentially, they build an ensemble of HS-tress and store \emph{mass}\footnote{  Data mass is defined as the number of points in a region, and two groups of data can have the same mass regardless of the characteristics of the regions  \cite{Ting:2010}.} of the data in the nodes. Their work is different from our work in the sense that we use \emph{Online} learning to build our model instead of an ensemble of HS-tress. Numenta \cite{Ahmad:2015} is a recently proposed anomaly detection benchmark. It  includes algorithms for tackling anomaly detection in a streaming setting. However, the working and scoring mechanism of Numenta is different from our work. Specifically, their Hierarchial Temporal Memory (HTM) algorithm is a window based algorithm and uses \emph{NAB} scores (please see \cite{Ahmad:2015}) to report anomaly detection results. Whereas, we use \emph{Gmean} and \emph{Mistake rate} to report the experimental results.
 
\subsection{Anomaly Detection in Nuclear Power Plant}
In this Section, we describe research works that closely matches our work in Chapter \ref{Chapter6}. There exist some work that have tackled anomaly detection in nuclear power plant. In \cite{Asok:2011}, the author study health monitoring of nuclear power plant. They propose an algorithm based on symbolic dynamic filtering (SDF) for feature extraction for time series data followed by optimization of partitioning of sensor time series. The key limitation of their work is that their algorithm is supervised anomaly detection  and they  tested their model on small number of features and data set only (training  and test set each has 150 samples), hence, can not be applied as such on big data. \cite{Marklund:2014} proposed a spectral method for feature extraction  and passive acoustic anomaly detection in nuclear power plants. \cite{Seong:1995} developed an online fuzzy logic based expert system for providing clean alarm pictures to the system operators for nuclear power plant monitoring. The key limitation of their method is that they model is depends on hand-crafted rule which may be not very accurate, given the many possibilities of anomaly occurrence. Model-based nuclear power plant monitoring is proposed in \cite{Nabeshima:2002}. Their model consists of neural network that takes input signals from the plant. Next component is the expert system that takes input from neural network and human operator for making informed decision about system's health. 
The shortcomings of the proposed approach in \cite{Nabeshima:2002} is that neural network requires lots of data for training and is supervised. The approach that we take  in this chapter builds on \emph{unsupevised} learning paradigm and hence differs from the previous studies on nuclear power plant condition monitoring.
\section{Datasets Used}
In this section, we  discuss  the datasets used in our experiments.  The datasets with their train/test size, feature size, ratio of positive to negative samples, and sparsity are shown in Tables \ref{data_pagmean},\ref{data_aspgd} and \ref{data}.  Note that the imbalance ratio shows the ratio of the positive to negative class in the training set. The test set can have different imbalance ratio.  These are the  benchmark datasets which can be freely downloaded from LIBSVM website\cite{CC01a}  pageblock from \cite{KEEL} also at \cite{Lichman:2013}.  The datasets in Table \ref{data_pagmean} have  a small number of features and with little to no sparsity while datasets in Tables \ref{data_aspgd} and \ref{data} are high dimension data with sparse features.  For our purpose, in each dataset, the positive class will be treated as an anomaly that we wish to detect efficiently. We briefly describe the various datasets used in our experiments.

\begin{itemize}
\item {\bf Kddcup 2008} dataset  is a breast cancer detection dataset that consists of 4 X-ray images; two images of each breast. Each image is represented by several candidates. After much pre-processing, kddcup 2008 dataset overall contains information of 102294 suspicious regions, each region described by 117 features.  
Each region is either ``benign'' or ``malignant'' and the ratio of malignant to benign regions is 1:163.19. Due to this huge imbalance ratio, the task of identifying malignant  (anomaly) is challenging.

\item {\bf Breast Cancer } Wisconsin Diagnostic data contains the digitized image of a fine needle aspirate of a breast mass. They delineate the characteristics of the nucleus of the cell present in the image. Some of the features include the radius, texture, area, perimeter etc. of the nucleus. The key task is to classify images into benign and malignant.
\item {\bf Page blocks}  dataset consists of blocks of the page layout of a document. The block can be one out of the five block types: (1) text (2) horizontal line (3) pictures (4) vertical line (5) graphic.  Each block is produced by a segmentation process.  Some of the features are height, length, area, blackpix etc.  We converted the multi-class  classification problem into binary classification problem by changing the labels of horizontal lines by the positive class (+1) and rest of the labels to  the negative class (-1). The task is to detect the positive class (anomaly) efficiently. Note that positive class is only a small fraction (6\%) of the total data.
\item {\bf W8a} is a dataset of keywords extracted from a web page and  each feature is a sparse binary feature. The task is to classify whether a web page falls into a category or not. W8a and a9a (described below) dataset were originally used by J.C. Platt \cite{Platt:1999}.
\item {\bf A9a} is a census data that contains features such as age, workclass, education, sex, martial-status etc. The task is to predict whether the income exceeds \$50K/yr. The challenge is that the number of individual having income more than \$50K/yr is very less. 

\item {\bf German} dataset contains credit assessment of customers in terms of good or bad credit risk. Some of the features in the dataset are credit history, the status of existing checking account, purpose, credit amount etc.  The challenge comes in the form of identifying  a small fraction of fraudulent customers from a huge number of loyal customers.

\item {\bf Covtype}  dataset contains information about the type of forest  and associated attributes. The task is to predict the type of forest cover from cartographic variables (no remotely sensed images). The cartographic variables were derived from data obtained from US Geological Survey and USFS data. Some of the features include Elevation, Aspect, Slope, Soli\_type etc. Covtype is multi-class classification dataset. To convert the multi-class dataset to binary class dataset, we follow the procedure given in \cite{Collobert:2002}. In short, we treat class 2 as the positive class and other 6 classes as negative class.

\item {\bf ijcnn1} dataset consists of time-series samples produced by 10-cylinder internal combustion engine.  Some of the features include crankshaft speed in RPM, load, acceleration etc. The task is to detect misfires (anomalies) in certain regions on the load-speed map.

\item {\bf Magic04}  dataset comprises of simulation of  high energy gamma particles  in a ground-based gamma telescope. The idea is to discriminate the action of primary gamma (called signal)  from the images of hadronic showers \cite{Lichman:2013} caused by cosmic rays (called background). The actual dataset is generated by the Monte Carlo Sampling.

\item {\bf Cod-rna} dataset comes from bioinformatics domain. It consists of a long sequence of coding and non-coding RNAs (ncRNA). Non-coding RNAs play a vital role in the cell, several of which remain hidden until now. The task is to detect the novel non-coding RNA (anomalies) to better understand their functionality.  
\end{itemize}
\vspace{-.5cm}
Next, we describe the large-scale datasets used in chapter 4 and chapter 5.
\vspace{-.5cm}
\begin{itemize}
\item {\bf News20} dataset is a collection of 20,000 newsgroup posts on 20 topic.  Some of the topics include comp.graphics, sci.crypt,sci.med,talk.religion etc. Original news20 dataset is a multi-class classification dataset. However, Chih-Jen Lin et al. \cite{CC01a} have converted the multi-class dataset into the binary class dataset and we use that dataset directly.
\item {\bf Rcv1}  dataset is a benchmark  collection of newswire stories that is made available by Reuters, Ltd. Data is organized  into four major topics:  ECAT (Economics), CCAT (Corporate/industrial), MCAT (Markets), and GCAT (Government/Social).  Chih-Jen Lin et al. have preprocessed the dataset and assume that ECAT and CCAT denote the positive category whereas MCAT and GCAT designate the negative category.

\item {\bf Url} dataset \cite{Ma:2009} is a collection of URLs. The task is to detect malicious URLs (spam, exploits, phishing, DoS  etc.) from the normal URLs. The author  represents the URLs  based on host-based features and lexical features.  Some of the lexical feature types  are hostname, primary domain, path tokens etc. and host-based features are WHOIS info, IP prefix, Connection speed etc. 

\item {\bf Realsim} dataset is a collection of UseNet articles \cite{Andrew:Mac} from 4 discussion groups: real autos, real aviation, simulated auto racing, simulated aviation. The data is often used in binary classification separating real from simulated and hence the name.

\item {\bf Gisette} dataset was constructed from MINIST dataset \cite{Yann}. It is a handwritten digit recognition problem and the task is to classify confusing digits. The dataset also appeared in NIPS 2003 feature selection challenge \cite{Isabelle}.

\item {\bf Pcmac} dataset is a modified form of the news20 dataset.

\item {\bf Webspam} dataset contains information about web pages. There exists the category of web pages whose primary goal is to manipulate the search engines and web users. For example, phishing site is created to duplicate the e-commerce sites  so that the creates of the phishing site can divert the credit card transaction to his/her account. To combat this issue, web spam corpus was created in 2011.  The corpus consists of approximately 0.35 million web pages; each web page represented by bag-of-words model. The dataset also appeared in Pascal Large-Scale Learning Challenge in 2008 \cite{Soeren}. The task is to classify each web page as spam or ham. The challenge comes from the high dimensionality and sparse features of the dataset.

\end{itemize}

\begin{table}[t]
\centering
\caption{Summary of datasets used in the experiments in Chapter 3}
\label{data_pagmean}
\begin{tabular}{|l|l|l|l|}
\hline
Dataset & \#Test set size(validation set size) & \#Features &\#Pos:Neg \\ \hline \hline
page blocks & 3472(2000) & 10 & 1:8.7889 \\\hline
w8a &14951(49749)  &300 &1:31.9317 \\\hline
a9a &16281(32561)  &122 &1:3.2332 \\\hline
german & 334(666) & 24 & 1:2.3\\\hline
covtype & 207711(100000) & 54 & 1:29 \\\hline
ijcnn1 & 91691(50000)&22 & 1:10.44 \\\hline
breast cancer & 228(455) & 10 & 1:1.86 \\\hline
kddcup2008 &68196(34098)  &117 &1:180\\\hline
magic04 & 9020(10000) & 10 & 1:1.8  \\\hline
cod-rna & 231152(100000) & 8 & 1:2 \\\hline
\end{tabular}
\end{table}

\begin{table*}
\centering
\caption{Summary of sparse data sets used in the experiment in Chapter 4 }
\label{data_aspgd}
\begin{tabular}{|l|r|r|r|r|r|r|}
\hline
Dataset & Balance & \# Train &\# Test  & \# Features  &\#Pos:Neg \\ \hline \hline
news20 & False & 3,000&7,000 &1,355,191&1:9.99 \\\hline
rcv1& false &20,370 &677,399 &47,236& 1 : 0.9064 \\\hline
url &False &2,000&8,000&3,231,961&1:10 \\\hline
realsim &False&3,000&4,000&20,958&1:2.2515 \\\hline
gisette &False&1,000&2,800&5,000&1:11\\\hline
news2& True &10,000&9,000&1,355,191&1:1\\\hline
pcmac & True &1,000 &900&3,289&1: 1.0219 \\\hline
webspam &False&1,000&1,000&16,609,143&1:19\\\hline
\end{tabular}
\end{table*}

\begin{table*}[htbp]
\centering
\caption{Summary of sparse data sets used in the experiment in Chapter 5}
\label{data}
\begin{tabular}{|l|l|l|l|l|l|l|}
\hline
Dataset & Balance & \#Train &\# Test  & \#Features  & Sparsity (\%) &\#Pos:Neg \\ \hline \hline
news20 & False & 6,598&4,399 &1,355,191& 99.9682&1:6.0042 \\\hline
rcv1& True &10,000 &10,000&47,236&99.839& 1:1.0068 \\\hline
url &False &8,000&2,000&3,231,961& 99.9964&1:10.0041 \\\hline
realsim &False&56,000&14,000&20,958& 99.749&1:1.7207 \\\hline
gisette &False&3,800&1,000&5,000& 0 .857734&1:11.667\\\hline
webspam &False&8,000&2,000&16,609,143& 99.9714 &1:15\\\hline
w8a & false & 40,000&14,951&300& 95.8204&1:26.0453\\\hline
ijcnn1 &false &48,000&91,701&22&39.1304&1:9.2696\\\hline
covtype&false  &2,40,000&60,000&54  &0 & 1:29.5344\\\hline
pageblocks & false&3,280&2,189&10&0&1::11.1933\\\hline
\end{tabular}
\end{table*}
\clearpage
\newpage
\section{Research Gaps Identified}
In this chapter, we presented the detailed summary of the relevant work in data mining  and machine learning domain to combat the anomaly detection problem. For our literature survey, we find that:
\vspace{-.5cm}
\begin{itemize}
\item Traditional approaches for anomaly detection in big data have a number of limitations. Some important ones are as follows: statistical techniques require underlying data distribution to be known a priori, proximity-based and density-based techniques require appropriate metrics to be defined for calculating anomaly score and have high time complexity. Clustering based techniques are also computationally intensive. Most of the traditional anomaly detection techniques assume a static candidate anomaly set. They are not able to handle evolving anomalies. 
\item Non-Parametric techniques are useful for anomaly detection in real world data for which class labels and data distribution are not known in advance. Further, the non-parametric techniques are also able to handle high dimensional data with varying data distribution. But, mostly the research work based on non-parametric anomaly detection techniques have assumed data to be homogeneous and static in nature. Therefore, there is scope of extending the existing non-parametric techniques for heterogeneous, distributed data streams. 

\item
 Multiple kernel learning (MKL) method and its variants  have the advantage of addressing the issue curse of dimensionality.  But, such techniques have been applied mostly on homogeneous and static data.  Further research work needs to be done to explore the possibility of using MKL in streaming and distributed anomaly detection scenarios.  In addition, the hyper parameters of the kernel function are also being set using some pre-defined constant values. Automatic learning of hyper parameters is also an open problem.

\item
Non-negative matrix factorization based methods have the advantage of being able to handle  anomaly detection in high dimensional and sparse data scenarios. But, they have been mostly applied to centralized  data although many real world data is distributed in nature. Therefore, there is scope for further research in this direction. 
\item

Random projection based techniques are useful once the intrinsic structure of the data  and number of dimensions to be used for projection is known. Therefore, there is need for devising methods that help us to choose the correct number of dimension for projection without their prior knowledge. 
\item

Finally, ensemble techniques have been able to address the issue of  anomaly detection in heterogeneous setting. But, most of the ensemble based techniques are not able to handle sparse data sets. To account for sparsity and high dimension case, further work using ensemble approach is required. Combining anomaly scores produced from different ensemble models also needs to be looked into.

\end{itemize}

In conclusion, we find that neither traditional approaches nor modern approaches are able to detect anomalies in big data efficiently, i.e., solving major big data issues such as \emph{streaming, sparse, distributed} and \emph{high dimensions}. In the next and subsequent chapters, we propose our work to tackle the aforementioned issues in an \emph{incremental} fashion.

 \thispagestyle{empty}
\chapter{Proposed Algorithm : PAGMEAN}\label{Chapter3}
In this chapter, we  tackle the anomaly detection problem in a \emph{streaming}  environment using \emph{online} learning. The reason to conduct such a study is that most of the real-world data is streaming in nature. For example,  the measurement from sensors forms a stream. Because of the dynamic nature of streaming data and the inability to store it, there is an urgent need to develop efficient algorithms to solve  the anomaly detection problem in a \emph{streaming} environment. We propose  an algorithm called Passive-Aggressive GMEAN (PAGMEAN). This algorithm is based on the classic online algorithm called Passive-Aggressive (PA) \cite{Crammer2006}. In PA algorithm, we show that it is sensitive to outliers and can not be directly applied for anomaly detection. Therefore, we introduce a modified hinge loss that is a convex surrogate for the indicator function (defined later in this chapter). The indicator function is obtained from maximizing the \emph{Gmean} metric directly. The major challenge is that \emph{Gmean} metric is non-decomposable, that means, it can not be written as the sum of losses over individual data points. We exploit the modified hinge loss within the PA framework and come up with PAGMEAN algorithm. We empirically show the effectiveness  and efficiency of PAGMEAN over various benchmark data sets and compare with the state-of-the-art techniques in the literature. 

\section{Introduction}
In this chapter, we aim at to tackle the \emph{streaming} problem of big data while detecting anomalies. Throughout this chapter and subsequent chapters, we make the following assumption:
\begin{assumption}
The terms outlier and anomaly are used interchangeably. 
\end{assumption}
\vspace{-0.4cm} \noindent
The reason for using the above assumption is that outliers/anomalies are present in a tiny amount compared to the normal samples and the outlier/anomaly detection problem can be addressed via class-imbalance learning problem. Our focus will be the detection of \emph{point} anomalies through the use of class-imbalance learning.

\section{Proposed  Algorithm - PAGMEAN} \label{algo}

First we introduce some notation for the ease of exposition. Examples in our data  come in a streaming fashion. At time $t$, instance-label pair is denoted as $(\mbox{\bf x}_t, y_t) $ where $\mbox{\bf x}_t \in \mathcal{R}^n $ and $y_t \in \{-1, +1\}$.  We consider linear classifier of the form $f_t(\mathbf{x}_t)=\mathbf{w}^T_t\mbox{\bf x}_t$, where $\mbox{\bf w}_t$ is the weight vector.  Let $\hat{y}_t$ be the prediction for the $t^{th}$ instance, i.e., $\hat{y}_t = sign(f_{t}(\mbox{\bf x}_t))$,  whereas the value $|f_{t}(\mbox{\bf x}_t)|$, known as ``margin'', is used as the confidence of the learner on the $t^{th}$ prediction step.
\subsection{Problem Formulation}
In a binary classification, there are two classes. We assume that minority class is positive class and labeled as $+1$.  Let $P$ and $N$ denote the total number of positive  and negative examples received so far, respectively. When there are two classes, four cases can happen during prediction. True positive $(T_p)$, true negative $(T_n$), false positive $(F_p)$ and false negative $(F_n)$. They are defined as follows: 
\begin{definition}
 True positive $T_p = \{ y =\hat{y}=+1  \}$,  True negative $T_n =\{y = \hat{y} = -1 \}$,   False positive $F_p= \{y= -1, \hat{y} = +1\}$, False negative $ F_n = \{y = +1, \hat{y} = -1\} $.
\end{definition}
\vspace{-0.4cm} \noindent Note that time $t$ is implicit in the above notations, that is, $T_p$ can also denote the total number of examples classified as positive up to time $t=1, 2,...,T$. Meaning will become clear from the context. Our objective is to maximize $Gmean$ metric for class-imbalanced problem. 

\begin{definition}Gmean is defined as:
\begin{equation}
\label{gm}
Gmean = \sqrt{sensitivity \times specificity}
\end{equation}
where $sensitivity$ and $specificity$ are defined as:
\begin{equation}
sensitivity = \frac{T_p}{T_p + F_n},   specificity =\frac{T_n}{T_n + F_p}
\end{equation}
\end{definition}

\section{Experiments} \label{exp4}
For comparative evaluation  of our proposed algorithms, we use the dataset presented in Table \ref{data_pagmean}. Note that the PAGMEAN algorithms are tested on only small-scale datasets. The reason is that PAGMEAN algorithms, though being online, may not handle high dimension data (the number of features going into millions and above) in a \emph{timely} manner. That means they will run slow. Secondly, PAGMEAN algorithms do not exploit \emph{sparsity} structure present in the data. 

\subsection{Experimental Testbed and Setup}
We compare our algorithms with the  parent algorithm PA and its variants along with recently proposed cost-sensitive algorithm CSOC of \cite{Jialei2014}. In \cite{Jialei2014}, it is claimed that CSOC outperforms many state-of-the-art algorithms such as PAUM, ROMMA, agg-ROMMA, CPA-PB. Hence, we only compare with CSOC. We further emphasize that PA, PAGMEAN and CSOC all are first-order methods that only use the gradient of the loss function. On the other hand, comparison with  ARROW \cite{Crammer:2013}, NAROW \cite{Orabona:2010}, CW \cite{NIPS2008_3554} etc. is not presented since they use second-order information (Hessian) and their scalability over large data sets is poor. Comparison with SVM$^{perf}$\cite{Joachims:2005} is also not fair due to (i) It does not optimize \emph{Gmean} using \emph{surrogate} loss function (ii) It is an offline solution.

To make a fair comparison, we first split all the dataset into validation and test set randomly (online algorithms do not require separate train and test set). The validation set is used to find the optimal value of the parameters, if any, in the algorithm. Aggressiveness parameter $C$ in PA algorithm is searched over $2^{[-6:1:6]}$ and parameter $\lambda$ in PAGMEAN, its variants as well as CSOC is searched over $2^{[-10:1:10]}$.  We report our results on test data over 10 random permutations.  Note that we do not perform feature scaling as it is against the ethos of online learning where we can access one example at  a time and as such, doing \emph{z-score} normalization is not feasible.

Secondly, most implementations (in Matlab) of online learning available today, e.g., Libol \cite{hoi2014libol}, DOGMA \cite{Orabona09},UOSLIB  \cite{UOSLib2013} are not \emph{out-of-core} algorithm. They process examples by fetching entire data into main memory and thus violates the principle of online learning. Our implementation is based on the idea that  data set is too large to fit into RAM and we can see one example at a time. 
\subsection{Performance Evaluation  Metrics}
As described  in Section \ref{intro} that \emph{Gmean} is a robust metric for the class-imbalance problems. Hence, we use \emph{Gmean} to measure the performance of various algorithms.  Results on 6 benchmark datasets ( ``pageblock'', ``w8a'', ``a9a'', ``german'', ``ijcnn1'', and ``covtype'' ) and 4 real world data sets ( ``breast-cancer'', ``kddcup2008'', ``magic04'', and ``cod-rna'') are shown in (Fig. \ref{benchmark1}, Table \ref{tab:gmean_benchmark} ) and (Fig. \ref{real1}, Table \ref{tab:gmean_real})  respectively. We also show the \emph{mistake rate} of all the compared algorithms on benchmark and real data sets  in (Fig. \ref{benchmark2}, Table \ref{tab:gmean_benchmark} ) and (Fig. \ref{real2} ,Table \ref{tab:gmean_real})  respectively. \emph{Mistake rate} of an online algorithm is defined as the number of mistakes made by the algorithm over time.  It is noted that we do not show the   running time of \emph{out-of-core} implementation since it depends upon the speed of device where data is stored (hard disk, tape or network storage). We assume that data is stored in files and read sequentially in \emph{mini-batches}, where batch-size is arbitrary or can depend on the available RAM.
\vspace{-0.5cm}
\subsection{Comparative Study on Benchmark Data sets}
\begin{enumerate}
\item {\bf Evaluation of Gmean}\\
We first evaluate \emph{Gmean} on the various benchmark data sets as given in Table \ref{data_pagmean}. Online average of \emph{Gmean} with respect to the sample size is reported in Fig. \ref{benchmark1}. From the Figure, we can observe several things. Firstly,  PAGMEAN2 algorithm outperforms their parent algorithms PA on all 6 data sets. Secondly, PAGMEAN2 beats CSOC on all data sets but ijcnn1. Thirdly, among the PAGMEAN algorithms, PAGMEAN2 outperforms PAGMEAN and PAGMEAN1 on pageblock, w8a, german, and a9a data sets. On the other side, Table \ref{tab:gmean_benchmark} reports \emph{Gmean} averaged over 5 runs. From the Table, we can see that PAGMEAN algorithms outperform their parent algorithms PA on all data sets. These results indicate the potential applicability of PAGMEAN algorithms for real-world class-imbalance detection task.

Another observation that can be drawn from the Fig.  \ref{benchmark1} is that initially there is performance drop in all the algorithms due to small number of samples available for learning. This phenomena is noticeable in german and pageblocks data set as these are small data sets. Online performance of algorithms is not smooth on some data sets, e.g., pageblock, german, ijcnn1. This could be due to sudden change in class-distribution (also known as concept drift).  This means  that  the presence of cluster of examples from the positive class or the negative class in the data has severe effect over the model. 
 \item {\bf Evaluation of Mistake Rate}\\
In this Section, we discuss the \emph{mistake rate} of PAGMEAN algorithms. The online average of \emph{mistake rate }  of various algorithms with respect to the  sample size is shown in Fig. \ref{benchmark2}. We can draw several conclusions out of it. Firstly, as more samples are received by the online algorithm, \emph{mistake rate} is decreasing on all data sets except pageblock and german where it seems to increase due to the small sample problem. Secondly, PAGMEAN algorithms suffer higher \emph{mistake rate} as compared to their parent algorithms PA. This is in contrast  to common intuition where the convex surrogate loss is supposed to be more sensitive to class-imbalance and hence lesser \emph{mistake rate} compared to the \emph{mistake rate} due to the hinge loss employed by PA algorithms. Thirdly, among PAGMEAN algorithms, PAGMEAN1 suffers smaller \emph{mistake rate} compared to PAGMEAN2 on all data sets. 

\emph{Mistake rate } averaged over 5 runs of all the compared algorithms is shown in Table \ref{tab:gmean_benchmark}. 
\begin{figure*}
\centering
\subfloat[pageblock]{\includegraphics[width=7cm, height=5cm]{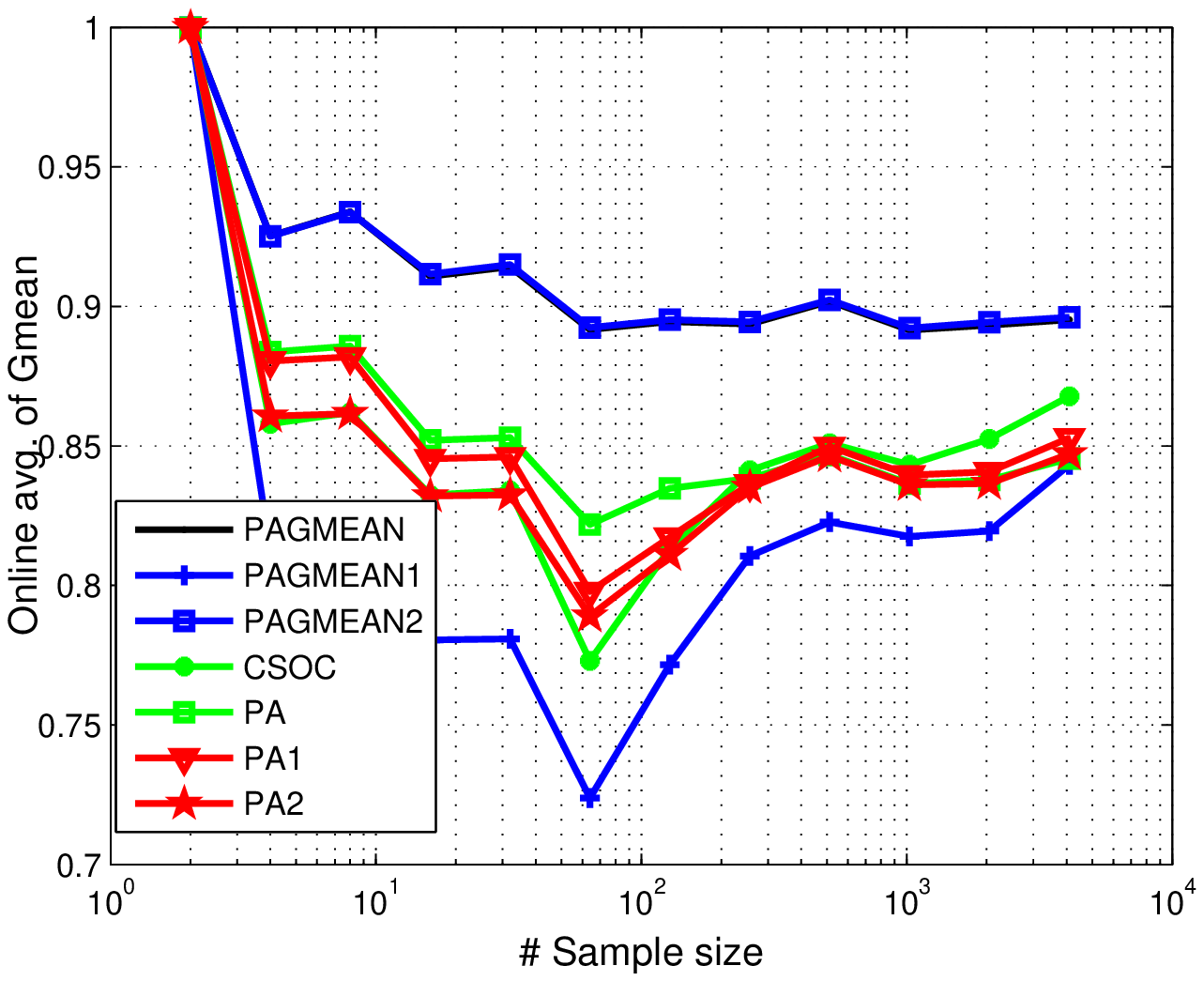}}  
\subfloat[w8a]{\includegraphics[width=7cm, height=5cm]{w8a_gmean}} \\
\subfloat[german]{\includegraphics[width=7cm, height=5cm]{german_gmean}} 
\subfloat[a9a]{\includegraphics[width=7cm, height=5cm]{a9a_gmean}}\\
\subfloat[covtype]{\includegraphics[width=7cm, height=5cm]{covtype_gmean}}
\subfloat[ijcnn1]{\includegraphics[width=7cm, height=5cm]{ijcnn1_gmean}} 

\caption{Evaluation of \emph{Gmean} over various benchmark data sets. (a) pageblock (b) w8a (c) german (d) a9a (e) covtype (f) ijcnn1. In all the figures, PAGMEAN algorithms either outperform or are equally good with respect to its parent algorithms PA and CSOC algorithm. } .
\label{benchmark1}
\end{figure*}
\begin{figure*}
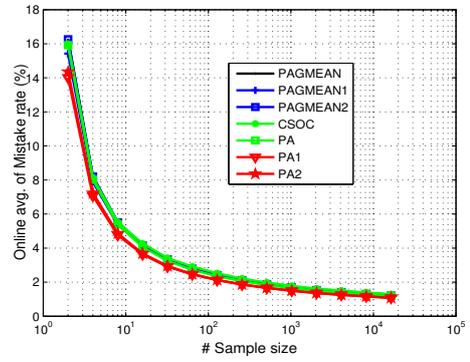
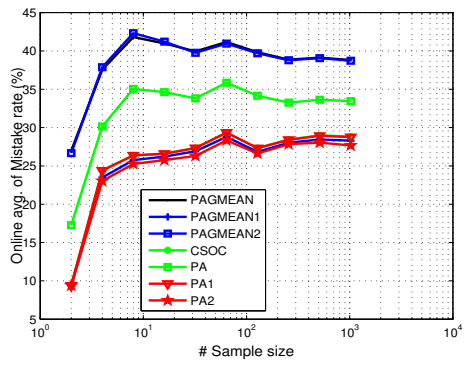
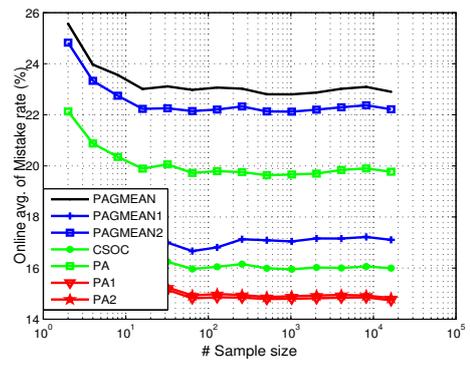
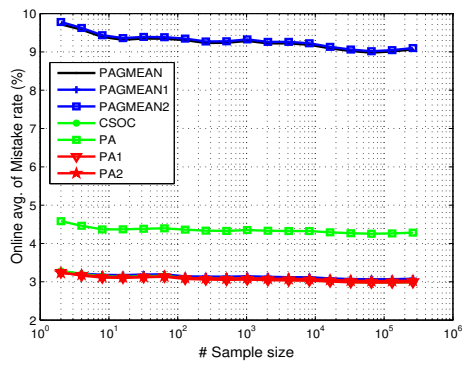
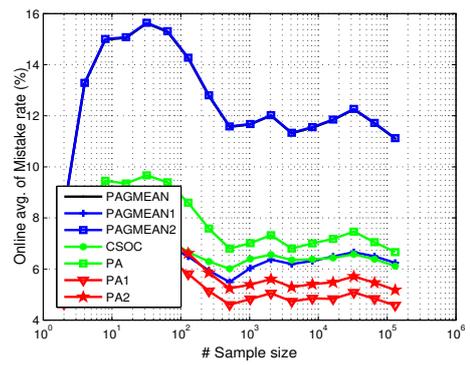

\centering
\begin{tabular}{cc}
\subfloat[pageblock]{\includegraphics[width=7cm, height=5cm]{pageblocks_mistake.eps}} & 
\subfloat[w8a]{\includegraphics[width=7cm, height=5cm]{w8a_mistake.eps}} \\
\subfloat[german]{\includegraphics[width=7cm, height=5cm]{german_mistake.eps}} &
\subfloat[a9a]{\includegraphics[width=7cm, height=5cm]{a9a_mistake.eps}}\\
\subfloat[covtype]{\includegraphics[width=7cm, height=5cm]{covtype_mistake.eps}}&
\subfloat[ijcnn1]{\includegraphics[width=7cm, height=5cm]{ijcnn1_mistake.eps}} 
\end{tabular}
\caption{Evaluation of \emph{Mistake rate} over various benchmark data sets.}
\label{benchmark2}
\end{figure*}
\newpage
It can be observed that PAGMEAN algorithms suffer \emph{mistake rate} that is not \emph{statistically significantly} (on wilcoxon rank sum test )  higher than that of their counterpart PA algorithms on 3 out of 6 data sets (w8a, german, covtype). All of these observations indicate that  further work in theory and practice is  to be investigated. 
\subsection{Comparative Study on  Real and Benchmark Data sets} \label{application}
The proposed PAGMEAN algorithms can be potentially applied to real world data for anomaly detection task. To illustrate this, we performed experiments on 4 real data sets (``breast cancer'', ``kddcup 2008'', ``magic04'', and ``cod-rna''). Performance is measured with respect to \emph{Gmean} and \emph{mistake rate}. The Results are reported in Fig. \ref{real1}, \ref{real2} and Table \ref{tab:gmean_real}.  In Fig. \ref{real1}, empirical evaluation of \emph{Gmean} with respect to sample size is shown. It is clear from the figure that PAGMEAN algorithms outperform PA algorithms on breast-cancer and kddcup 2008 data sets. Their performance is equally good  as compared to PA and CSOC algorithms on magic04 and cod-rna data sets. Interestingly, all algorithms achieve \emph{Gmean} of 0.97 approximately on cod-rna data set. On the other hand, \emph{Gmean} on kddcup 2008, surprisingly, is falling as more sample are received by all algorithms. This may be due to high class-imbalance ratio (1:180) of positive and negative examples present in kddcup 2008 data set and algorithms receive very less positive examples at the end. However, drop in the \emph{Gmean} value of PAGMEAN and PAGMEAN2 algorithms is significantly lower than other algorithms compared  beyond $10^4$ samples. Similar conclusions are drawn on \emph{Gmean} from the results in Table \ref{tab:gmean_real}. These results show the potential applicability of PAGMEAN algorithms for real-world online anomaly detection task.

\emph{Mistake rate} with respect to sample size of all the algorithms is shown in Fig. \ref{real2} and Table \ref{tab:gmean_real}. Similar to conclusions drawn on benchmark data sets, we observe that \emph{mistake rate} of all the algorithms is dropping with increasing sample size. At the same time, it is clear that among PAGMEAN algorithms, PAGMEAN1  suffers the smallest number of  \emph{mistake rate} on all 4 data sets.   Whereas on kddcup 2008 and magic04 data sets, PAGMEAN1 outperforms all other algorithms except PA1 and CSOC.  Lastly, PAGMEAN1 outperforms CSOC and PA on cod-rna data sets.

\end{enumerate}


\begin{table*}[!p]
\footnotesize
\centering
\caption{Evaluation of \emph{Gmean} and \emph{Mistake rate (\%)} on benchmark data sets. Entries marked by * are not statistically significant at 95\% confidence level than the entries marked by ** on wilcoxon rank sum test. }
\label{tab:gmean_benchmark}
\begin{tabular}{|l|c|c|c|c|c|}
\hline
\multirow{2}{*}{Algorithm}  &   Gmean  & Mistake rate(\%) & &Gmean & Mistake rate(\%)\\
\cline{2-3} \cline{5-6}
 &\multicolumn{2}{c|}{w8a} &  & \multicolumn{2}{c|}{german}   \\ \hline \hline
PA & 0.878 $\pm$ 0.025 & 1.194 $\pm$ 0.341   && 0.536$\pm$ 0.031 & 35.255 $\pm$ 2.336 \\ \hline
PA1 & 0.873 $\pm$ 0.031 & 1.101$\pm$ $0.352^{**}$  && 0.102 $\pm$ 0.000 & 28.859$\pm$ 0.095  \\ \hline
PA2 & 0.872$\pm$ 0.032 & 1.116$\pm$ 0.344  &&0.339$\pm$ 0.043 & 27.057$\pm$ $1.383^{**}$  \\ \hline
CSOC & 0.849 $\pm$ 0.055 & 1.346$\pm$ 0.411  && 0.102 $\pm$ 0.000 & 28.859$\pm$ 0.095 \\ \hline
Proposed PAGMEAN & 0.894 $\pm$ 0.014& 1.243$\pm$ 0.379  &&  0.602 $\pm$ 0.028& 39.820$\pm$ 2.933 \\ \hline
Proposed PAGMEAN1 & 0.822$\pm$ 0.059 & 1.725$\pm$ 0.411 &&0.396$\pm$ 0.014 & 27.628$\pm$ $1.464^*$  \\ \hline
Proposed PAGMEAN2 & \bf{ 0.894$\pm$ 0.014} & 1.241$\pm$ $0.372^*$ && \bf{ 0.610$\pm$ 0.019} & 41.502$\pm$ 1.976 \\  \hline
\hline
Algorithm  &  \multicolumn{2}{c|}{covtype} & &   \multicolumn{2}{c|}{ijcnn1} 
  \\ \hline \hline
PA & 0.537$\pm$ 0.003 & 4.288 $\pm$ 0.024  &&0.776$\pm$ 0.009 & 7.017 $\pm$ 0.220  \\ \hline
PA1 & 0.463 $\pm$ 0.006 & 2.992$\pm$ 0.017 && 0.802 $\pm$ 0.024 & 4.747$\pm$ 0.484  \\ \hline
PA2 & 0.397$\pm$ 0.004 & 3.029$\pm$ 0.008 &&0.754$\pm$ 0.025 & 5.391$\pm$ 0.478 \\ \hline
CSOC & 0.449 $\pm$ 0.008 & 2.988$\pm$ $0.018^{**}$ &&0.871 $\pm$ 0.017 & 6.271$\pm$ 0.526 \\ \hline
Proposed PAGMEAN & 0.748 $\pm$ 0.002& 9.056$\pm$ 0.058&& 0.846 $\pm$ 0.005& 11.761$\pm$ 0.446 \\ \hline
Proposed PAGMEAN1 & 0.329$\pm$ 0.003 & 3.074$\pm$ $0.009^*$&& \bf{ 0.877$\pm$ 0.012} & 6.610$\pm$ 0.587\\ \hline
Proposed PAGMEAN2 &\bf{  0.749$\pm$ 0.002} & 9.096$\pm$ 0.061 &&0.847$\pm$ 0.005 & 11.782$\pm$ 0.454 \\ \hline
\hline
Algorithm  &  \multicolumn{2}{c|}{a9a} & &   \multicolumn{2}{c|}{pageblocks} 
  \\ \hline \hline
PA & 0.708$\pm$ 0.007 & 20.105 $\pm$ 0.571 && 0.857$\pm$ 0.024 & 4.749 $\pm$ 0.795 \\ \hline
PA1 & 0.748 $\pm$ 0.016 & 14.875$\pm$ 0.670 &&0.832 $\pm$ 0.059 & 4.083$\pm$ 1.051 \\ \hline
PA2 & 0.744$\pm$ 0.013 & 14.931$\pm$ 0.511&&0.850$\pm$ 0.039 & 3.884$\pm$ 0.958 \\ \hline
 CSOC & 0.798 $\pm$ 0.014 & 16.106$\pm$ 0.568&& 0.875 $\pm$ 0.036 & 6.393$\pm$ 1.053 \\ \hline
Proposed PAGMEAN &0.755 $\pm$ 0.009& 23.294$\pm$ 0.758&&\bf{ 0.900 $\pm$ 0.021}& 7.223$\pm$ 1.083 \\ \hline
Proposed PAGMEAN1 &0.785$\pm$ 0.008 & 17.295$\pm$ 0.388&&0.852$\pm$ 0.029 & 6.831$\pm$ 1.047 \\ \hline
Proposed PAGMEAN2 &\bf{ 0.806$\pm$ 0.007} & 22.617$\pm$ 0.652&& \bf{ 0.900$\pm$ 0.021} & 7.252$\pm$ 1.080 \\ \hline

\end{tabular}
\end{table*}

\begin{table*}[htbp]
\footnotesize
\centering
\caption{Evaluation of \emph{Gmean} and \emph{Mistake rate (\%)} on real data sets. Entries marked by * are not statistically significant at 95\% confidence level than the entries marked by ** on wilcoxon rank sum test. }
\label{tab:gmean_real}
\begin{tabular}{|l|c|c|c|c|c|}
\hline
\multirow{2}{*}{Algorithm}  &  \multicolumn{2}{c|}{breast cancer} & \multirow{2}{*}{} &   \multicolumn{2}{c|}{kddcup2008} \\
\cline{2-3} \cline{5-6} 
&  Gmean  & Mistake rate(\%) & &Gmean & Mistake rate(\%)  \\ \hline \hline
PA & 0.990$\pm$ 0.009 & 0.580 $\pm$ 0.597 &&0.563$\pm$ 0.022 & 25.436 $\pm$ 0.748 \\ \hline
PA1 & 0.980 $\pm$ 0.001 & 0.938$\pm$ 0.141 &&0.542 $\pm$ 0.020 & 9.846$\pm$ 0.826 \\ \hline
PA2 & 0.987$\pm$ 0.005 & 0.625$\pm$ 0.312 &&0.572$\pm$ 0.014 & 17.893$\pm$ 1.208 \\ \hline
CSOC & 0.987 $\pm$ 0.006 & 0.670$\pm$ 0.482&&0.561 $\pm$ 0.017 & 9.840$\pm$ 0.808 \\ \hline
Proposed PAGMEAN & \bf{ 0.991 $\pm$ 0.011}& 0.804$\pm$ 0.960 &&0.720 $\pm$ 0.019& 28.596$\pm$ 1.017 \\ \hline
Proposed PAGMEAN1 & 0.984$\pm$ 0.006 & 1.027$\pm$ 0.368&& 0.555$\pm$ 0.014 & 10.585$\pm$ 0.813 \\ \hline
Proposed PAGMEAN2 &\bf{  0.991$\pm$ 0.007} & 0.759$\pm$ 0.597 &&\bf{ 0.767$\pm$ 0.025} & 27.931$\pm$ 1.016 \\ 
\hline
\hline
\multirow{2}{*}{Algorithm}  &  \multicolumn{2}{c|}{magic04} & \multirow{2}{*}{} &   \multicolumn{2}{c|}{cod-rna} \\
\cline{2-3} \cline{5-6} 
&  Gmean  & Mistake rate(\%) & &Gmean & Mistake rate(\%)  \\ \hline \hline
PA & 0.934$\pm$ 0.002 & 5.282 $\pm$ 0.137 &&0.981$\pm$ 0.001 & 1.780 $\pm$ 0.101 \\ \hline
PA1 & 0.946 $\pm$ 0.003 & 4.286$\pm$ 0.193  &&0.986 $\pm$ 0.002 & 1.293$\pm$ 0.108 \\ \hline
PA2 & \bf{ 0.946$\pm$ 0.002} & 4.281$\pm$ $0.152^{**}$  && 0.986$\pm$ 0.002$^{**}$ & 1.299$\pm$ 0.159 \\ \hline
CSOC & 0.942 $\pm$ 0.002 & 4.118$\pm$ 0.178  &&0.985 $\pm$ 0.001 & 1.614$\pm$ 0.062 \\ \hline
Proposed PAGMEAN & 0.926 $\pm$ 0.002& 5.064$\pm$ 0.133  &&0.982 $\pm$ 0.000& 1.774$\pm$ 0.041 \\ \hline
Proposed PAGMEAN1 & \bf{ 0.946$\pm$ 0.002} & 4.181$\pm$ $0.476^*$  &&0.985$\pm$ $0.003^*$ & 1.389$\pm$ 0.192 \\ \hline
Proposed PAGMEAN2 & 0.926$\pm$ 0.002 & 5.059$\pm$ 0.135  && 0.982$\pm$ 0.000 & 1.768$\pm$ 0.041 \\ \hline

\end{tabular}
\end{table*}

\begin{figure*}
\centering
\begin{tabular}{cc}
\subfloat[breast cancer]{\includegraphics[width=7cm, height=5cm]{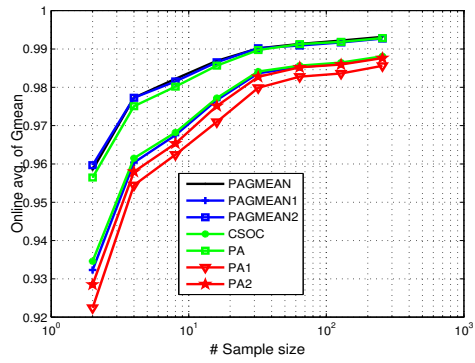}} & 
\subfloat[kddcup 2008]{\includegraphics[width=7cm, height=5cm]{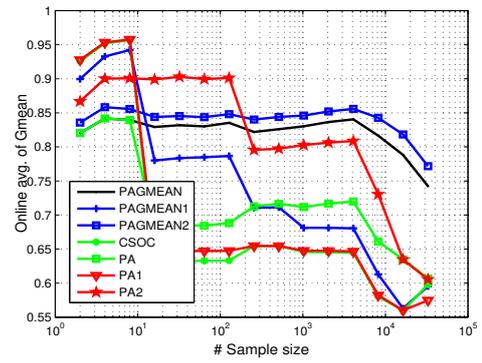}} \\
\subfloat[magic04]{\includegraphics[width=7cm, height=5cm]{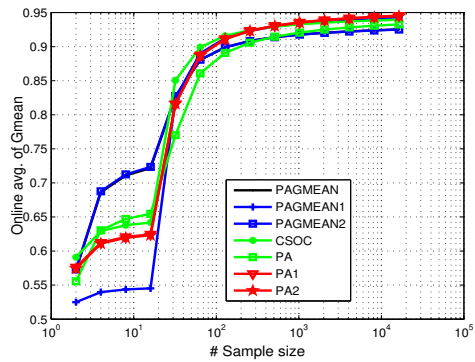}} &
\subfloat[cod-rna]{\includegraphics[width=7cm, height=5cm]{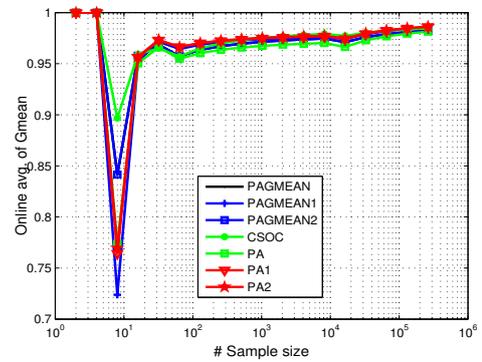}}
\end{tabular}
\caption{Evaluation of \emph{Gmean} over various real data sets.}
 \label{real1}
\end{figure*}

\begin{figure*}
\centering
\begin{tabular}{cc} 
\subfloat[breast cancer]{\includegraphics[width=7cm, height=5cm]{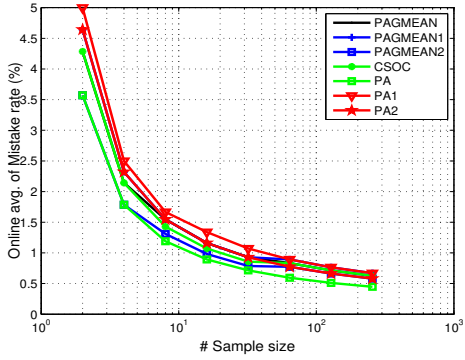}} & 
\subfloat[kddcup 2008]{\includegraphics[width=7cm, height=5cm]{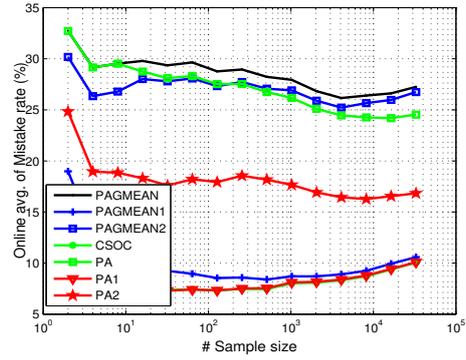}} \\
\subfloat[magic04]{\includegraphics[width=7cm, height=5cm]{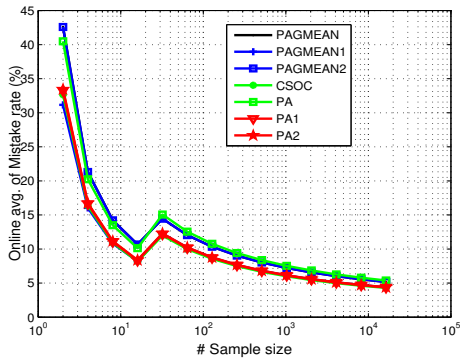}} &
\subfloat[cod-rna]{\includegraphics[width=7cm, height=5cm]{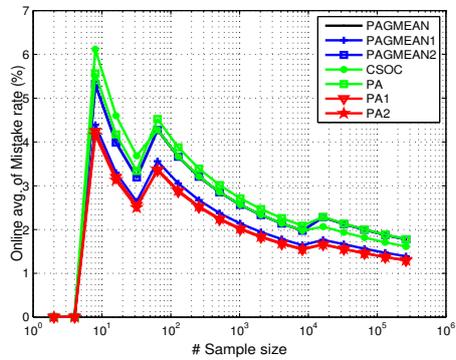}}
\end{tabular}
\caption{Evaluation of \emph{Mistake rate} over various real data sets.}
\label{real2}
\end{figure*}

\newpage
\section{Discussion}\label{conc}
In the proposed work, we attempt to make the classical passive-aggressive algorithms insensitive to outliers and apply it to the class-imbalance learning and anomaly detection problems. To solve the aforementioned problem, we maximize the \emph{Gmean} metric directly.  Since direct maximization of \emph{Gmean} is NP-hard, we resort to convex surrogate loss function and minimize a modified hinge loss instead. The modified hinge loss is utilized within the PA framework to make it insensitive to outliers and new algorithms are derived called PAGMEAN.  Empirical performance of all the derived algorithms is tested on various benchmark and real  data sets. From the discussion above, we conclude that our derived algorithms  perform equally good  as compared to other algorithms (PA and CSOC) in terms of  \emph{Gmean}. This indicates the potential applicability of PAGMEAN algorithms for real world class-imbalance and anomaly detection problems.  However,  the \emph{mistake rate}  of the proposed algorithms are surprisingly higher than the compared algorithm on \emph{some} datasets. Therefore, further work is required to identify the exact reasons for higher mistake rate.

Finally, we would like to highlight is that results on high dimensional data (high dimension means the number of features is in millions) could not be included. The reason is that even online algorithms run slowly when working on full feature space. In the next chapter, we propose an algorithm that exploits the \emph{sparsity} present in the  data and scales over millions of dimensions.

 \thispagestyle{empty}
\chapter{Proposed Algorithm : ASPGD}\label{Chapter4}

In the previous Chapter, we proposed an \emph{online} algorithm that tackles \emph{streaming} anomaly detection problem. However,  one of the limitations of the PAGMEAN algorithm is that it is not able to exploit the \emph{sparsity} structure present in the big data. To solve this problem, we propose another \emph{online} algorithm that solves \emph{sparse, streaming, high-dimensional} problem of big data during anomaly detection. As before, we employ the class-imbalance learning mechanism to  handle the \emph{point} anomaly detection problem.   The problem formulation in the present work uses $L_1$ regularized proximal learning framework and is solved via Accelerated-Stochastic-Proximal Gradient Descent (ASPGD) algorithm. Within the ASPGD algorithm, we use a smooth and strongly convex loss function. This loss function is insensitive to the class-imbalance since it is directly derived from the maximization of the \emph{Gmean}. The work presented in the current Chapter demonstrates  (i) the application of proximal algorithms to solve real world problems (class imbalance) (ii) how it scales to big data, and (iii) how it outperforms some recently proposed algorithms in terms of \emph{Gmean, F-measure} and \emph{Mistake rate} on several benchmark data sets.
\section{Introduction}
As discussed in Section \ref{relevant},  there are serious issues in using the classical approaches to anomaly detection. Firstly, sampling-based techniques are neither scalable to  the number of samples nor to the data dimensionality in the case of big data. Secondly, existing works on sampling such as \cite{Wang2013,Shuo:2015} do not exploit the rich structure present in the data such as sparsity. Kernel-based methods suffer from scalability and long training time. For example, if the data dimensionality is of the order of millions, kernel-based methods will require storage of the gram matrix (also known as kernel matrix) of size million $\times$ million which is prohibitive for machines with low memory.  Cost-sensitive learning has recently gained popularity in addressing the class imbalance problem \cite{Jialei2014,Dayong:2015} because of: (i) learning cost to be assigned to different classes in a data dependent way (ii) scalability (iii) ability to exploit sparsity.  The present work builds upon cost-sensitive learning and extends the work present in \cite{Dayong:2015}.

To further address the problem of class imbalance, the choice of  metric  used to evaluate the performance of different methods is crucial. For example, using accuracy as a class imbalance performance measure may be misleading. Consider  99 negative examples and 1 positive example in a dataset, our objective is to classify each positive example correctly. Now,  a classifier which classifies each example as negative will have an accuracy of 99\% that is incorrect because our goal was to detect  the positive example. Therefore,  different researchers have devised alternative performance metrics to assess different methods for class imbalance problem. Among them are the \emph{recall, precision, F-measure, ROC} (receiver operating characteristics) and \emph{Gmean} \cite{Kubat1997}. It is found that \emph{Gmean} is robust to the  class imbalance problem \cite{Kubat1997}. 

In this work, we tackle the class imbalance problem in an online setting exploiting \emph{sparsity} and \emph{high dimensional} characteristics of the big data. Our contributions are as follows:

\begin{enumerate}
  \setlength{\itemsep}{0pt plus 1pt}
\item
We propose a class-imbalance learning algorithm in an online setting within the Accelerated-Stochastic-Proximal Gradient Decent learning framework (called ASPGD). The novelty of ASPGD is that it uses a strongly convex loss function that we come up with direct maximization of the non-decomposable performance metric (\emph{Gmean} in our case). A non-decomposable performance metric  is a metric that can not be written as the sum of losses over data points.

\item We show, through extensive experiments on real and benchmark data sets, the effectiveness and efficiency of the proposed algorithm over various state-of-the-art algorithm in the literature.
\item
In our work, we also show that Nesterov's acceleration \cite{Nesterov:1983} \emph{does not always} helps in achieving higher \emph{Gmean}.
\item The effect of learning rate $\eta$ and sparsity regularization parameter $\lambda$ on \emph{Gmean, F-measure}, and \emph{Mistake rate} is  demonstrated as well.
\end{enumerate}
\vspace{-.5cm}
\section{Proposed Algorithm - ASPGD}
First, we establish some notation for the sake of clarity.   Input data is denoted as instance-label pair $\{{\bf x}_t,y_t \}$ where $t=1,2,...,T$,   ${\bf x}_t \in \mathcal{X} \subseteq R^d $ and $y_t \in \mathcal{Y} \subseteq \{-1, +1\}$. In the offline setting, we are allowed to access entire data and usually, $T$ is finite.  On the other hand, in  the online setting  we see one example at a time and  $T \rightarrow \infty$. At time $t$, an (instance, label) pair is denoted by $({\bf x}_t, y_t) $.  We consider linear functional of the form $f_t(\mathbf{x}_t)=\mathbf{w}_t^T{\bf x}_t$, where ${\bf w}_t$ is the weight vector.  Let $\hat{y}_t$ be the prediction for the $t^{th}$ instance, i.e., $\hat{y}_t = sign(f_{t}({\bf x}_t))$,  whereas the value $|f_{t}({\bf x}_t)|$, known as `margin', is used as the confidence of the learner in the $t^{th}$ prediction step.  
We work under the following assumptions on the model function $f$.\\
Function $f$ is $L$- smooth if its gradient is $L$-Lipschitz, i.e.,
\begin{equation*}
\| \nabla f(x) - \nabla f(y)\| \leq L\| x -y\|.
\end{equation*}
In other words, gradient of the function $f$ is upper bounded by $L>0$. $\|\cdot\|$ denotes the euclidean norm unless otherwise stated.  The function $f$  is $\mu$- strongly convex if
\begin{equation*}
f(y) \geq f(x) + \nabla f(x)^T (y-x) + \frac{\mu}{2} \|y-x\|^2.
\end{equation*}
where,  $\mu > 0$ is the strong convexity parameter. Intuitively, strong convexity is a measure of the curvature of the function. A function having large $\mu$ is has high positive curvature.  Interested readers can refer to \cite{Nesterov:2004} and appendix \ref{smoothstrong}.

\subsection{Problem Formulation and The Loss Function}
Our problem formulation is same as presented in Chapter \ref{Chapter3}. that is, we want to maximize the \emph{Gmean}. To do that, we use the lemma \ref{lemma}.  However, there are certain number of issues in using the loss function  \eqref{loss_pagmean} from Chapter \ref{Chapter3}.
Note that the loss function used in Chapter \ref{Chapter3} is:
\begin{equation}
\label{loss2}
\ell(f;(\mathbf{x},y))=\max \left( {0,\rho  - y f({\bf x}) } \right) 
\end{equation} 
where, 
\begin{equation*}
 \rho = \left( \left( {\frac{N}{P}} \right){I_{(y = 1)}} + \left( {\frac{{P - F_n}}{P}} \right){I_{(y =  - 1)}}\right)  
\end{equation*}
The parameter $\rho$ controls the penalty imposed when the current model mis-classifies the incoming example. The important thing here is that we impose a high penalty (the ratio $N/P$ is high for $N>>P$ in the class imbalance scenario) on misclassifying  a positive example $(y=+1)$. On the other hand, when model mis-classifies a negative sample, the penalty imposed is $(P-F_n)/P$, which is less when we  have a small number of false negatives. 
It is to be noted that the loss function in \eqref{loss2} is non-differentiable at the hinge point. Thus, it is not directly applicable to algorithms that require $L-$smooth and $\mu-$strongly convex loss functions (please refer to the appendix \ref{smoothstrong} for the definition of smooth and strongly convex function). In section \ref{Algorithm}, we review some proximal algorithms that work under the assumption that loss function is smooth and strongly convex. Therefore, we introduce a smooth and strongly convex function that  upper bounds the  indicator function in \eqref{loss2}.
\begin{equation}\label{loss1}
\ell(f, ({\bf x},y)) = \frac{\rho}{2} \max(0, 1-yf({\bf x}))^2
\end{equation}
Above loss function is utilized within the accelerated-stochastic proximal learning framework in section \ref{plf}. Note that the loss function as presented in \eqref{loss1} is strongly convex with strong convexity parameter $\mu=\rho$. Strong convexity parameter depends on the hinge point of the loss function in \eqref{loss1}. Since hinge point is a data dependent term, we estimate it in an online fashion.

\subsection{Stochastic Proximal Learning Framework}\label{plf}
Now, we describe proximal learning framework \cite{Parikh:2014} which aims to solve the composite optimization problem of the following form:
\begin{equation}\label{prox}
\phi(\mathbf{w}) \triangleq \min_{\mathbf{w} \in R^d} f(\mathbf{w}) +r({\bf w})
\end{equation}
\noindent where $f(\cdot)$ is the average  of convex and differentiable functions, i.e., $f(\mathbf{w}) =\frac{1}{T} \sum_{t=1}^T f_t(\mathbf{w})$ and $r: R^d \rightarrow R$ is `simple' convex function that can be non-differentiable. Note that the framework \eqref{prox} is quite general and encompasses many algorithms. For example, if we set $f_t(\mathbf{w}) = \max(0, 1-y\mathbf{w}^T\mathbf{x})$ and $r(\mathbf{w}) =\lambda \|\mathbf{w}\|^2_2$, we get $L_2$ regularized SVM. On the other hand, if we set $f_t(\mathbf{w}) =(y-\mathbf{w}^T\mathbf{x})^2$ and $r(\mathbf{x})=\lambda\|\mathbf{w}\|_2^2$, we get ridge regression. For our problem, $f_t(\mathbf{w})$ is given in \eqref{loss1}. Since we consider to solve \eqref{prox} under sparsity constraint, we utilize sparsity-inducing $L_1$ norm of $\mathbf{w}$, i.e., $r(\mathbf{w})=\lambda\|\mathbf{w}\|
_1$ (which is non-differentiable at $0$).

There are many algorithms to solve the problem \eqref{prox} in the offline setting under different assumptions on the loss function $f(\cdot)$ and regularization parameter $r(\cdot)$ \cite{Beck:2008,Tibshirani:1996,Nesterov:2013,Nesterov:2007}. Since our aim is to solve the problem \eqref{prox} under the online learning framework, techniques mentioned thereof cannot be used. Under our assumptions on the form of $f(\cdot)$ and $r(\cdot)$ (strongly convex and non-smooth, respectively), subgradient methods can be a good candidate. However, these methods have notorious convergence rate of $O(1/\epsilon^2)$(iteration complexity), i.e., to obtain $\epsilon$ accurate solution, we need $O(1/\epsilon^2)$ iteration.  Hence, we resort to proximal learning algorithms because of their simplicity, ability to handle non-smooth regularizer, scalability, and faster  convergence under certain assumptions \cite{Parikh:2014}.

A proximal gradient step at \emph{iteration} $k$ is given by:
\begin{equation}
\mathbf{w_{k+1}} = prox_{\eta r}(\mathbf{w}_k - \eta \nabla f(\mathbf{w_k}))
\end{equation}
where, $prox_{\eta} (\cdot)$ is proximal operator defined as 
\begin{equation}
prox_{\eta r} (\mathbf{u}) \triangleq   \argmin{\mathbf{u}}  \frac{1}{2\eta} \|\mathbf{u}-\mathbf{w}\|_2^2 + r(\mathbf{u})
\end{equation}
where $\eta$ is the step size and $\|\cdot\|_2$ is 2-norm. $\nabla$ denotes the gradient of the loss function $f(\cdot)$.
One of the major drawbacks of using Proximal Gradient Descent (PGD) in an offline setting is that they are not scalable to large-scale data sets as they require entire data set to compute the gradient. In the proposed work, we focus on algorithms that are \emph{memory}-aware. Such algorithms come under online learning framework (also known as stochastic algorithms). A simple Stochastic Proximal Gradient Descent (SPGD) update rule at $t^{th}$ time step is given by:
\begin{equation}
\mathbf{w_{t+1}} = prox_{\eta_t r}(\mathbf{w}_t - \eta_t \nabla f_t(\mathbf{w_t}))
\end{equation}
where $\nabla f_t(\cdot)$ is evaluated at  $t^{th}$ example. In order to achieve acceleration (faster convergence) in  online learning, we follow Nesterov method \cite{Nesterov:1983}. Specifically, Nesterov method achieves acceleration by introducing an auxiliary variable $\mathbf{u}$ such that the weight vector at time $t$ is  the convex combination of weight at time $t-1$ and $\mathbf{u}_t$, i.e.,
\begin{equation}
   \mathbf{w}_{t} = (1-\gamma) \mathbf{w}_{t-1} + \gamma \mathbf{u}_t 
\end{equation}
\noindent With all the tools of  accelerated-stochastic-proximal (ASP)  learning framework in hand, we are ready to present our ASP Gradient Descent (ASPGD) algorithm for sparse class imbalance learning. 

We note that similar ASP algorithms based on Nesterov accelerated method have recently appeared in \cite{Nitanda:2014,Kowk:NIPS2009}. However, \cite{Nitanda:2014}  propose an accelerated algorithm with variance reduction and \cite{Kowk:NIPS2009} propose an  accelerated algorithm that uses two sequences for computing the weight vector $\mathbf{w}$, where one of the sequences utilizes lipschitz parameter of the smooth component of the composite optimization. On the  other hand, we exploit the strong convexity of the smooth component. Besides, in our present work, we aim at  showing how efficient  the  ASP algorithms  are in dealing with the class imbalance problem?

\subsection{ASPGD Algorithm}\label{Algorithm}
In this Section, we present the algorithm to solve \eqref{prox} in an online setting.  Our proposed algorithm is called  ASPGD which is based on SPGD framework  with Nesterov's  acceleration. ASPGD algorithm which is presented in Algorithm \ref{aspgd} is able to handle the class imbalance problems.
  \begin{algorithm}[H]  
   \caption{ASPGD: Accelerated Stochastic Proximal Gradient Descent Algorithm for Sparse Learning}
   \label{aspgd}
 \begin{algorithmic}[1]
   \Require $ \eta > 0, \lambda, \gamma = \frac{1-\sqrt{\mu\eta}}{1+\sqrt{\mu\eta}}$ 
   \Ensure ${\bf w}_{T+1}$
   \For{$t := 1,...,T$}
    \State receive instance  ${\bf x}_t$ 
     \State     $\mathbf{v}_t = \nabla \Phi^*_t({\bm \theta}_t)$
     \State      $\mathbf{u}_t = prox_{\eta r(\mathbf{w})}(\mathbf{v}_t)$
     \State   $\mathbf{w}_{t} = (1-\gamma) \mathbf{w}_{t-1} + \gamma \mathbf{u}_t $
     \State	predict $\hat{y}_t=sign(\mathbf{w}_t^T\mathbf{x}_t)$
     \State	receive true label $y_t \in \{-1,+1\}$
     \State   suffer loss: $\ell_t(y_t, \hat{y}_t)$  as given in \eqref{loss1} 
        \If{ $\ell_t(y_t, \hat{y}_t)>0$}
     \State update:   
     \State  ${\bm \theta}_{t+1} = {\bm  \theta}_{t} -\eta \nabla \ell_t(\mathbf{w}_t)$ 
         \EndIf     
 \EndFor
\end{algorithmic}
  \end{algorithm}

In Algorithm \ref{aspgd}, $\Phi$ is some $\mu-$strongly convex function such as $\| \cdot \|_2^2$ and $*$ above $\phi$ denotes the \emph{dual norm} (see the appendix \ref{norm} for the definition of dual norm); $\bm  \theta$ is some vector in $\mathcal{R}^n$.  At this point, we emphasize that \cite{Dayong:2014} recently proposed algorithms for sparse learning (see Algorithm 1 in \cite{Dayong:2014}). Their algorithm is a special case of our algorithm and can be analyzed under Stochastic Proximal Learning (SPL) framework.  Specifically, setting $\gamma = 1$ and $\ell_t$ to hinge loss in ASPGD, we obtain Algorithm 1 in \cite{Dayong:2014}. The same author's extended paper \cite{Dayong:2015} proposed algorithms for class imbalance sparse learning. Algorithm 6 in \cite{Dayong:2015} is a special case of ASPGD without acceleration and again Algorithm 6 in \cite{Dayong:2015} can be analyzed under SPL framework.

\section{Experiments}\label{exp}
In this Section, we empirically validate the performance of ASPGD algorithm over various benchmark data sets  given in Table \ref{data_aspgd}. Notice that NEWS2  and PCMAC  are balanced data sets. All the algorithms were run in MATLAB 2012a (64-bit version) on 64-bit Windows 8.1 machine.\footnote{Our code is available at https://sites.google.com/site/chandreshiitr/publication}

\subsection{Experimental Testbed and Setup}
For evaluation purpose, we compared ASPGD and it  variants without acceleration (which we call  ASPGDNOACC) with that of a recently proposed algorithm CSFSOL \cite{Dayong:2015}.  In \cite{Dayong:2015}, the author compared their first and second order algorithms with a bunch of cost-sensitive algorithms (CS-OGD, CPA, and PAUM etc.) It is found that CSFSOL and its second order version CSSSOL outperform the aforementioned algorithms in terms of a metric called \emph{balanced accuracy} (which is defined as $0.5\times sensitivity +0.5\times specificity$). For this reason, we compare our ASPGD, ASPGDNOACC with CSFSOL algorithm (notice that CSSSOL is a second order algorithm, hence no comparison with CSSSOL is made). 
For performance evaluation, we use \emph{Gmean} and \emph{Mistake rate} as performance metrics. The time taken by these  algorithms are not shown since each algorithm reads data in \emph{mini-batches} and processes it online. Hence, total time consumed will depend on how fast we can read data from the storage device such as hard disk, network etc.  The most time-consuming operation in Algorithm \ref{aspgd} is evaluating the \emph{prox} operator. In our case,  \emph{prox} operator is soft-thresholding operator that has a closed form solution \cite{Simon}. Other steps in Algorithm \ref{aspgd} take $O(1)$ time. 

For parameter selection (learning rate $\eta$), we fix the sparsity regularization parameter $\lambda=0$ and perform grid search as  in \cite{Dayong:2015}. Note that for ASPGD algorithm, strong convexity parameter $\mu$ is equal to $\rho$ shown in \eqref{loss2} (which is easy to show). The parameter $\rho$ can be calculated in an online fashion. Hence, no parameter tuning is required for ASPGD algorithm compared to CSFSOL and ASPGDNOACC. Further, the parameter $\rho$ (and $\mu$ thereof) can be greater than 1, hence, we have to set $\eta$ such that $\mu \eta <1$ for the parameter $\gamma$ to be a valid convex combination parameter. In our subsequent experiments, we set $\eta=\frac{1}{\mu+1}$. In subsection \ref{learningrate}, we also demonstrate the effect of varying the learning rate. In \cite{Nitanda:2014}, it is stated that diminishing learning rate helps in reducing the variance introduced by random sampling, but it leads to slower convergence rate. Keeping that in mind, we set the aforementioned value of the learning rate.  

The results presented in the next subsection have been averaged over 10 random permutations of the test data and shown on semi log plot.  No feature scaling technique has been used as it is against the ethos of online learning. Online learning dictates that  only a subset of the entire data set is seen at one point of time thus meeting more practical scenario of real world data in our simulation. 

\subsection{Comparative Study on Benchmark Data sets}

\begin{enumerate}
 \item {\bf Evaluation of Gmean}\\
In this Section, we evaluate the \emph{Gmean} over various benchmark data sets as shown in Table \ref{data}. The results are presented in Figure \ref{fig:gmean} and Table \ref{tab:gmean}. From Figure \ref{fig:gmean}, several conclusions can be drawn. First, ASPGD algorithm outperforms ASPGDNOACC on 6 out of 8 data sets (news2, gisette, rcv1, url, pcmac, and webspam ). This indicates that Nesterov's acceleration  helps in achieving higher \emph{Gmean}. Secondly, ASPGD either outperforms or performs equally good compared to CSFSOL on 6 out of 8 data sets (news, gisette, rcv1, url, pcmac, and webspam ). Thirdly, on news2 and realsim data sets, all algorithms suffer performance degradation. This may be due to sudden change in concept or the class distribution. This shows one inherent limitation of ASPGD and CSFSOL algorithms in addressing concept drift. The same observation can be made from the \emph{cumulative  Gmean} results presented in Table \ref{tab:gmean}. For example, on news data set, ASPGD achieves \emph{Cumulative Gmean} which is statistically more significant than the \emph{Cumulative Gmean} achieved by CSFSOL and vice versa on rcv1 data set.

\begin{figure*}
\centering
\subfloat[news]{\includegraphics[width=6.3cm, height=4.5cm]{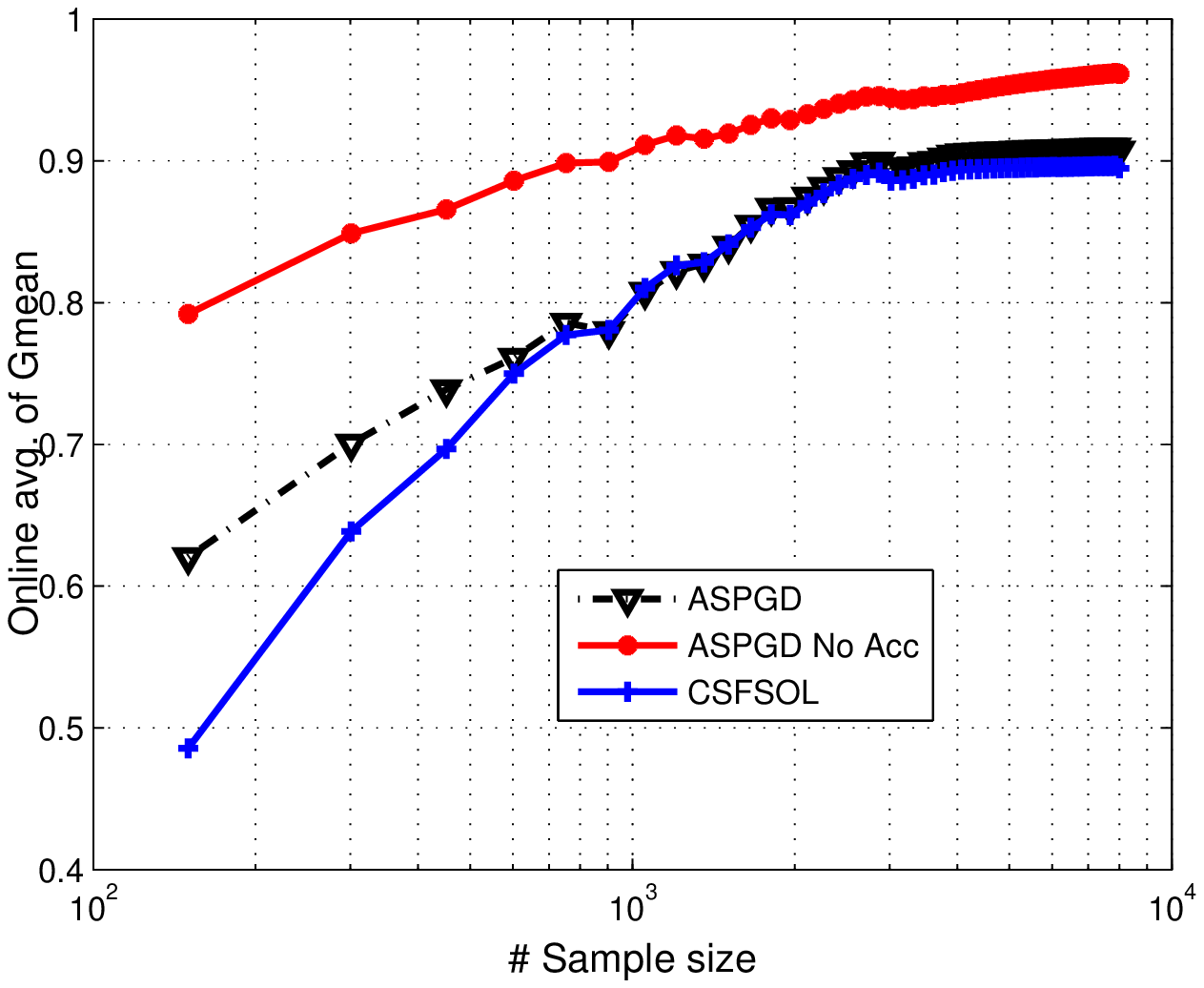}}  
\subfloat[news2]{\includegraphics[width=6.3cm, height=4.5cm]{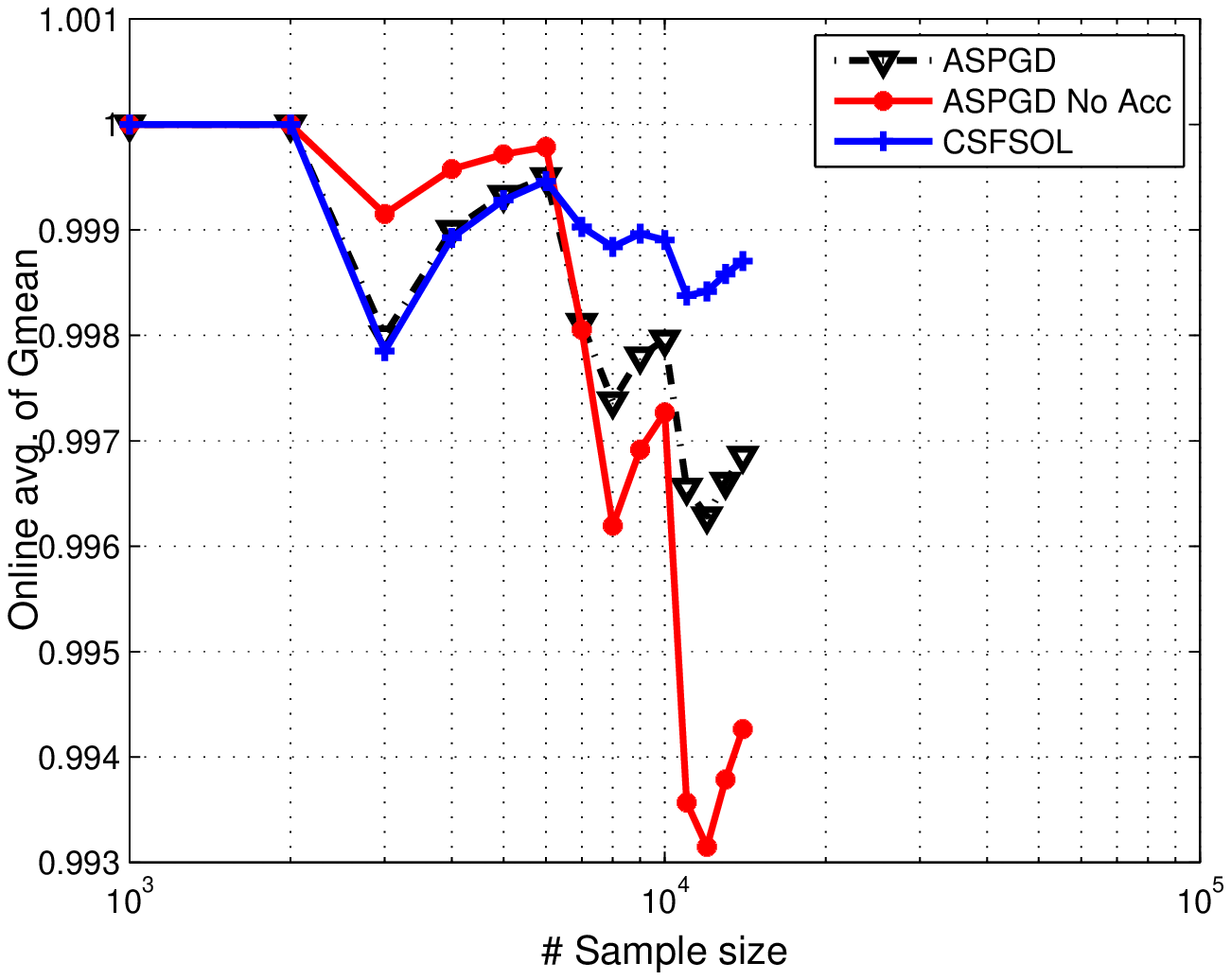}} \\
\subfloat[gisette]{\includegraphics[width=6.3cm, height=4.5cm]{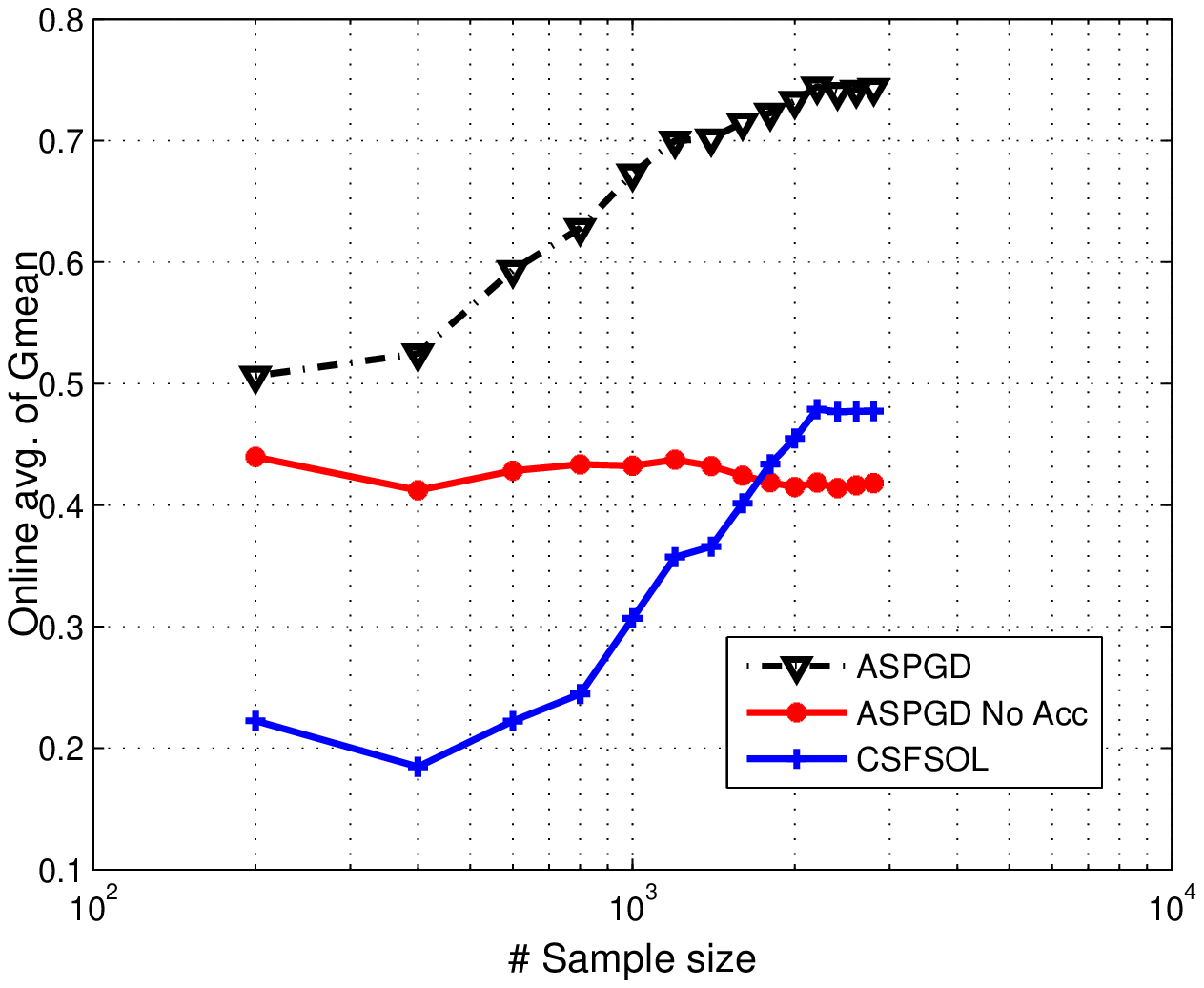}} 
\subfloat[realsim]{\includegraphics[width=6.3cm, height=4.5cm]{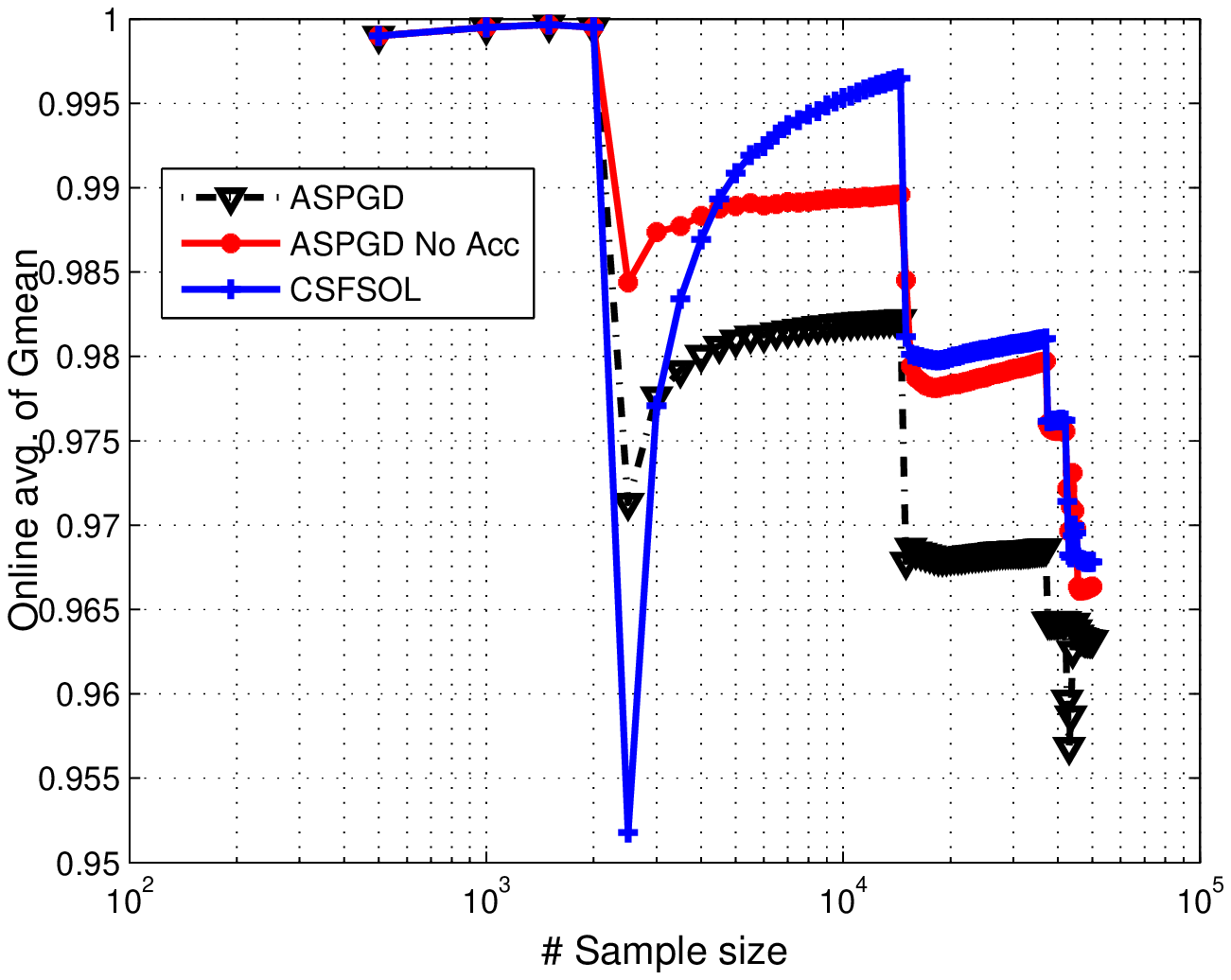}}\\
\subfloat[rcv1]{\includegraphics[width=6.3cm, height=4.5cm]{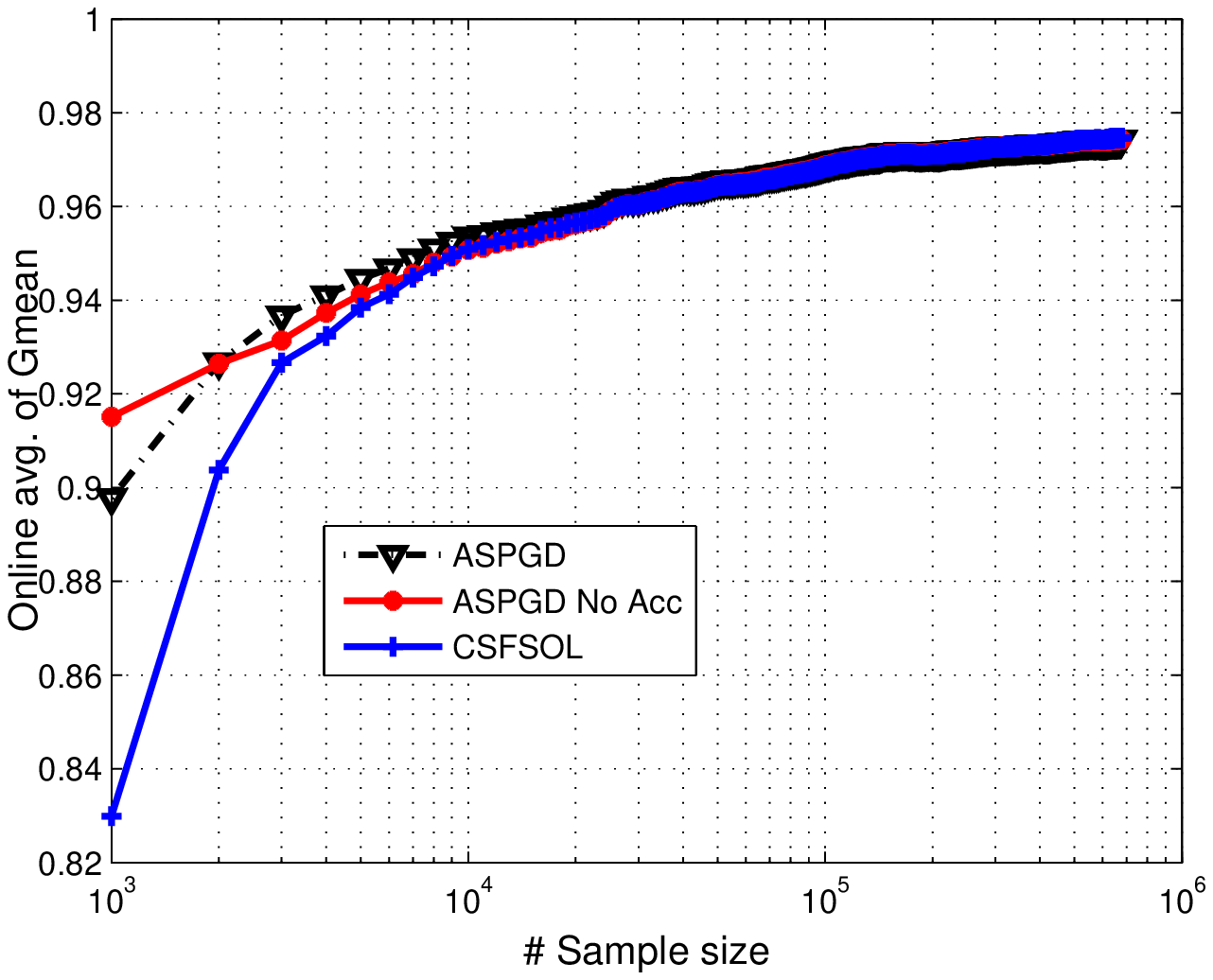}}
\subfloat[url]{\includegraphics[width=6.3cm, height=4.5cm]{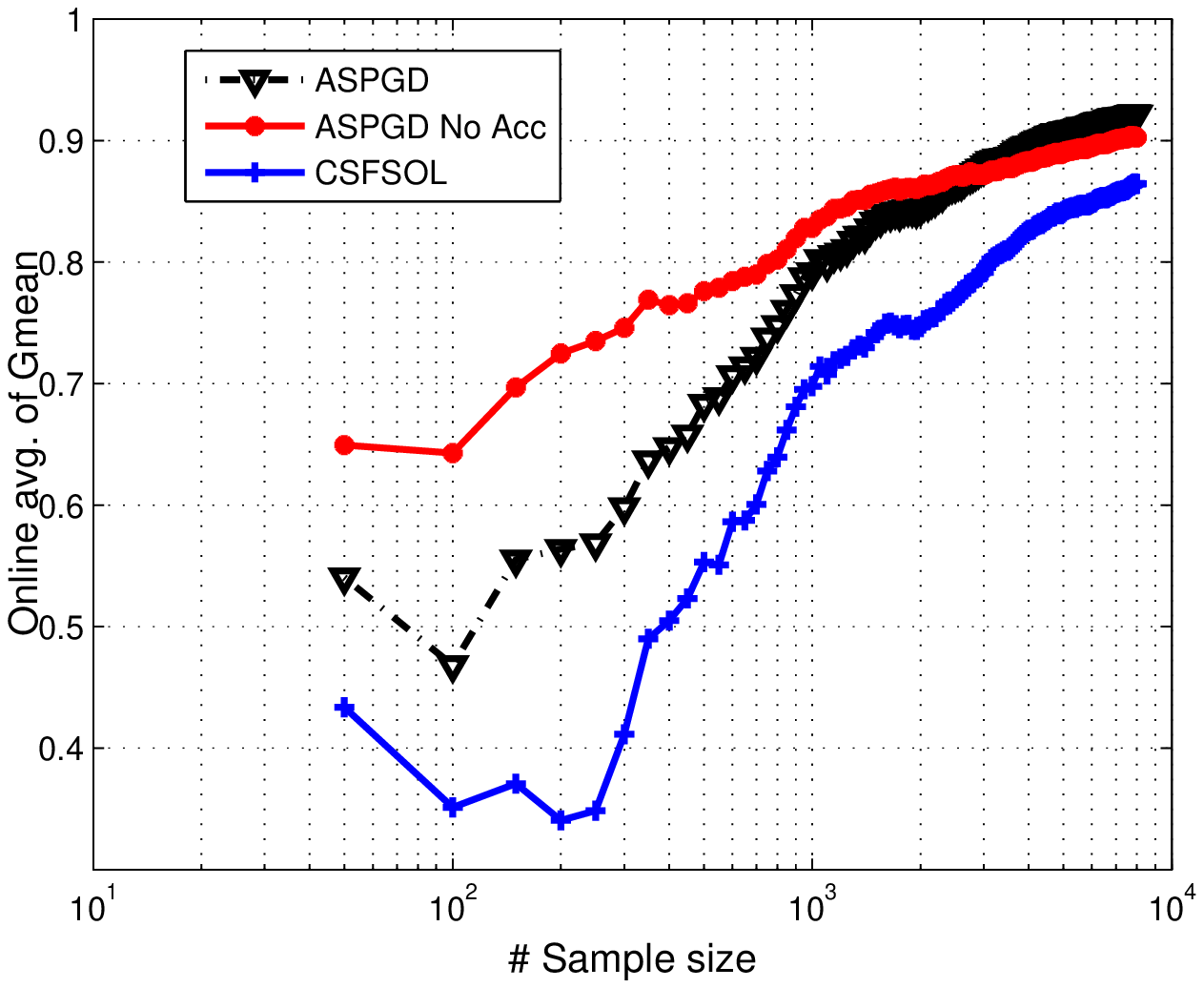}} \\
\subfloat[pcmac]{\includegraphics[width=6.3cm, height=4.5cm]{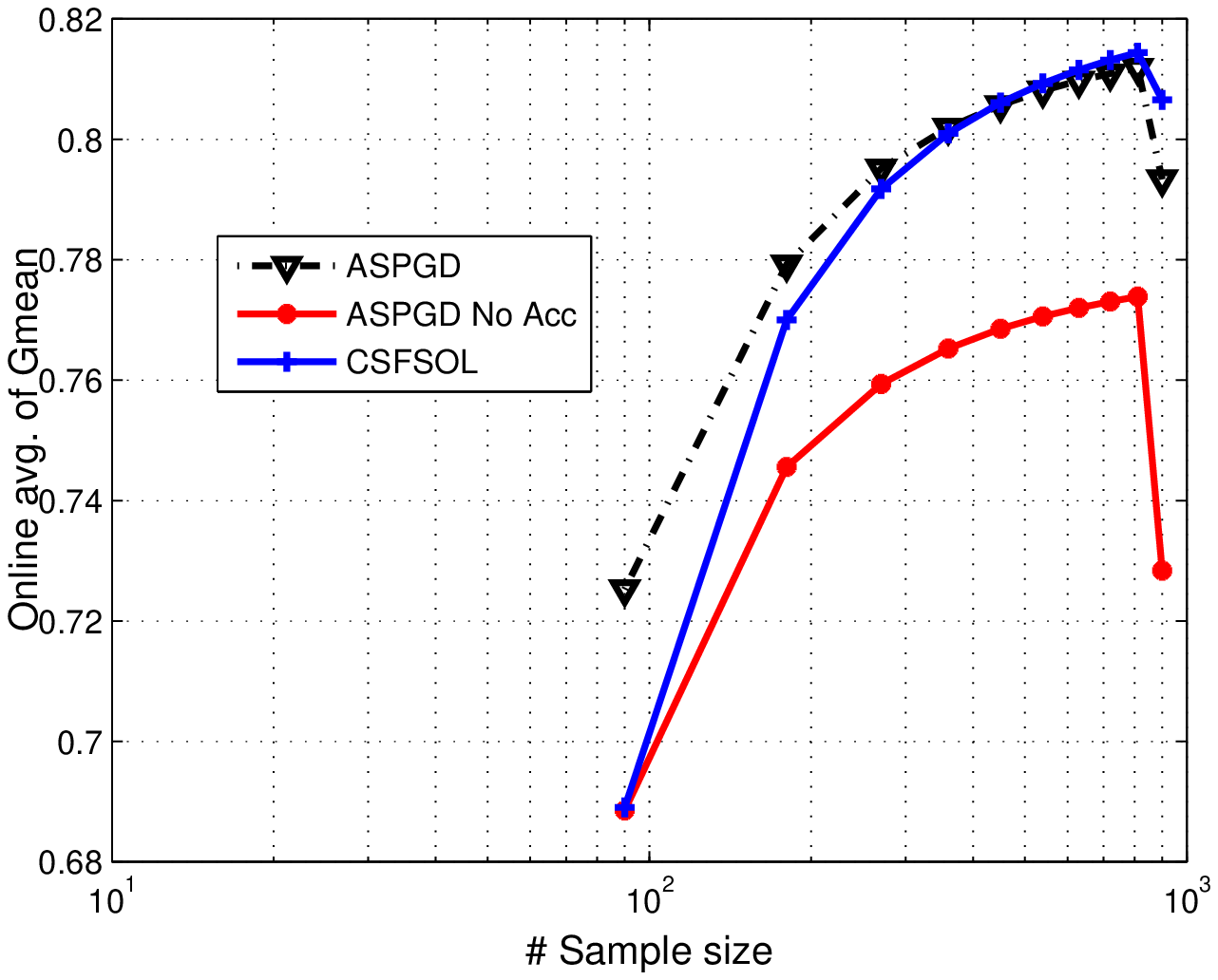}}
\subfloat[webspam]{\includegraphics[width=6.3cm, height=4.5cm]{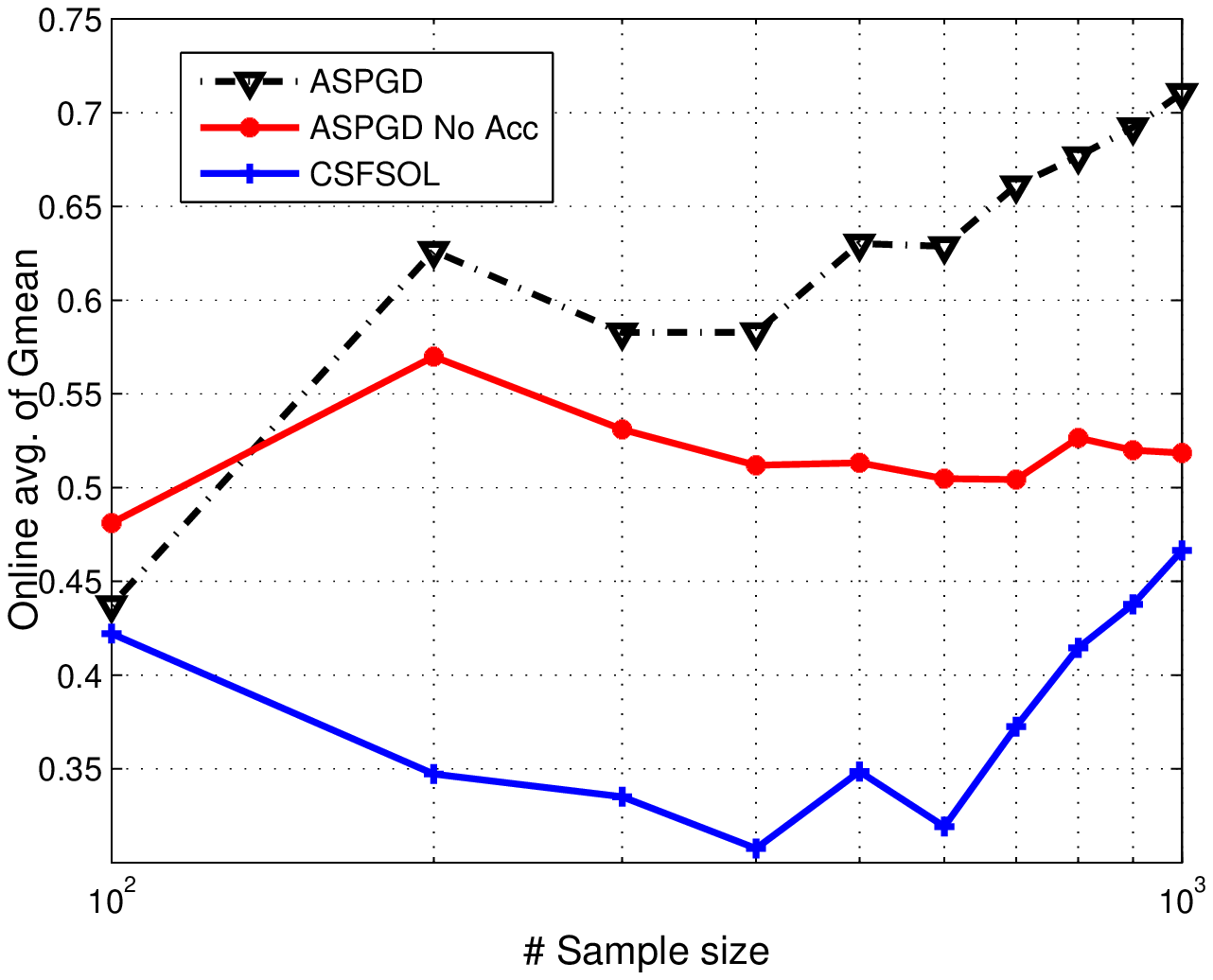}} \\

\caption{Evaluation of online average of \emph{Gmean} over various benchmark data sets. (a) news (b) news2 (c) gisette (d) realsim (e) rcv1 (f) url (g) pcmac (h) webspam.} 
\label{fig:gmean}
\end{figure*}

\item {\bf Evaluation  of Mistake rate}\\
Online average of \emph{Mistake rate} is shown in Figure \ref{fig:mistake} and cumulative \emph{Mistake rate} is shown in Table \ref{tab:gmean}. The following conclusions can be drawn from the results in Figure \ref{fig:mistake}. ASPGD achieves smaller \emph{Mistake rate} than ASPGDNOACC over all the data sets. Secondly, as more and more samples are consumed, \emph{Mistake rate} of ASPGD eventually reaches  the \emph{Mistake rate} of CSFSOL on 5 out of 8 data sets (news, rcv1, url, pcmac, and realsim). Thirdly, all  algorithm's \emph{Mistake rate} is increasing (fluctuating indeed) on news2 and relasim data sets. The same reason as discussed in the previous subsection applies here too. That is, concept drift of classes may result in higher \emph{Mistake rate}. Finally, the \emph{Mistake rate} result presented in Table \ref{tab:gmean} is consistent with the observation drawn from the Figure \ref{fig:mistake}. For example, cumulative  \emph{Mistake rate} obtained by ASPGD is smaller than that of ASPGDNOACC  on all data sets. On the other hand, \emph{Mistake rate} of ASPGD is not statistically significantly higher than the \emph{Mistak rate} suffered by CSFSOL  on news, pcmac, relasim, and rcv1.

\begin{figure*}
\centering
\subfloat[news]{\includegraphics[width=6.3cm, height=4.5cm]{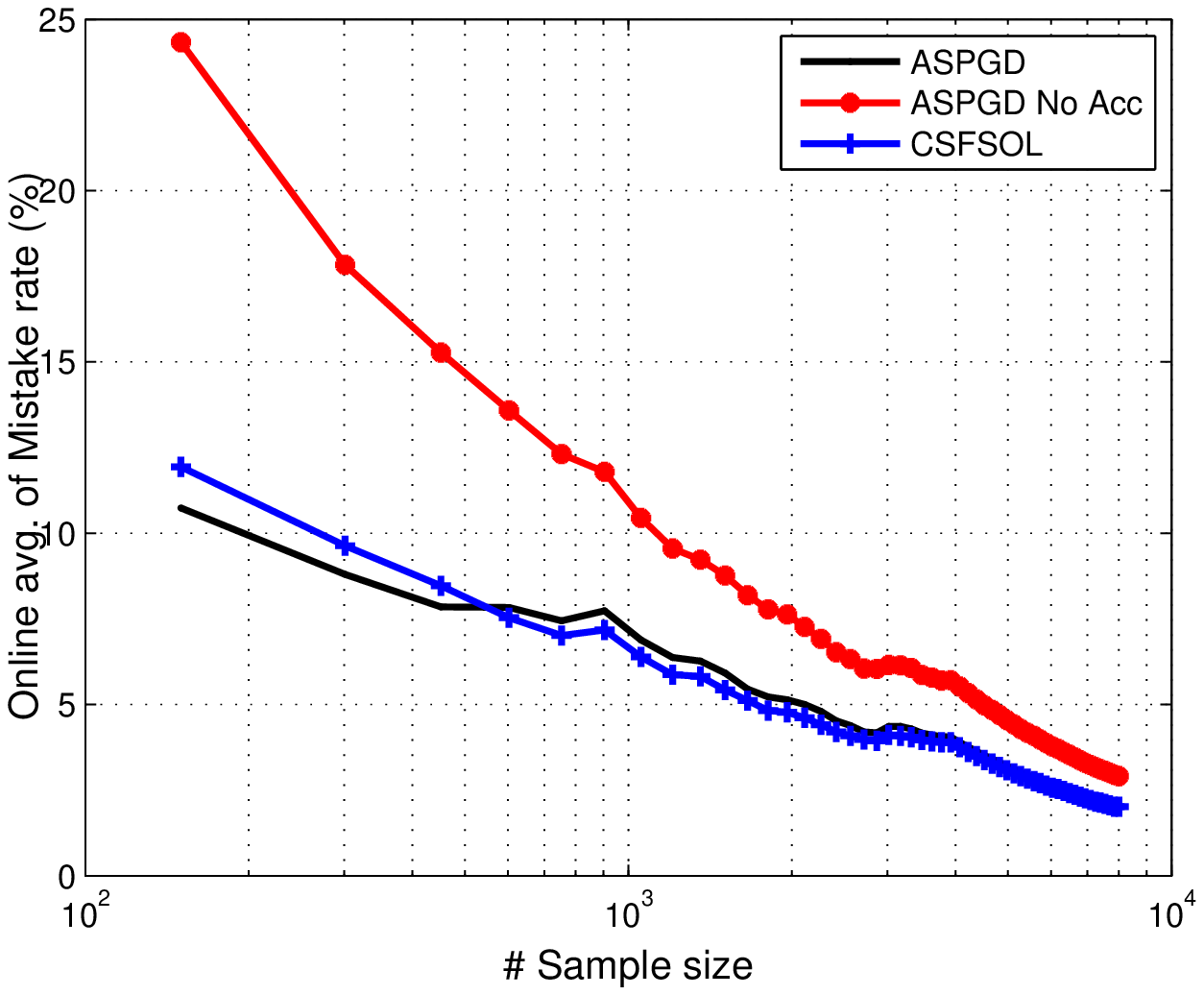}}  
\subfloat[news2]{\includegraphics[width=6.3cm, height=4.5cm]{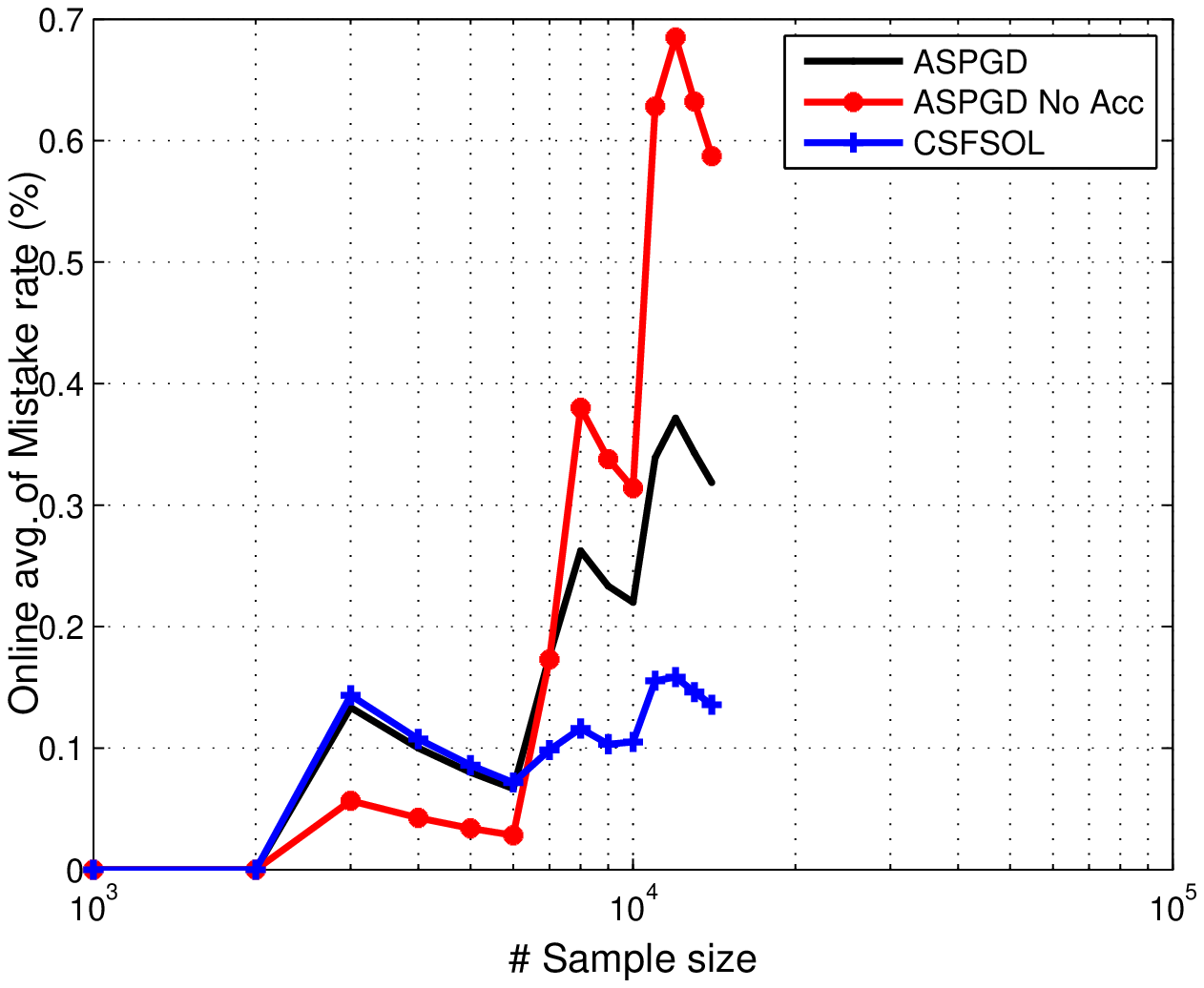}} \\
\subfloat[gisette]{\includegraphics[width=6.3cm, height=4.5cm]{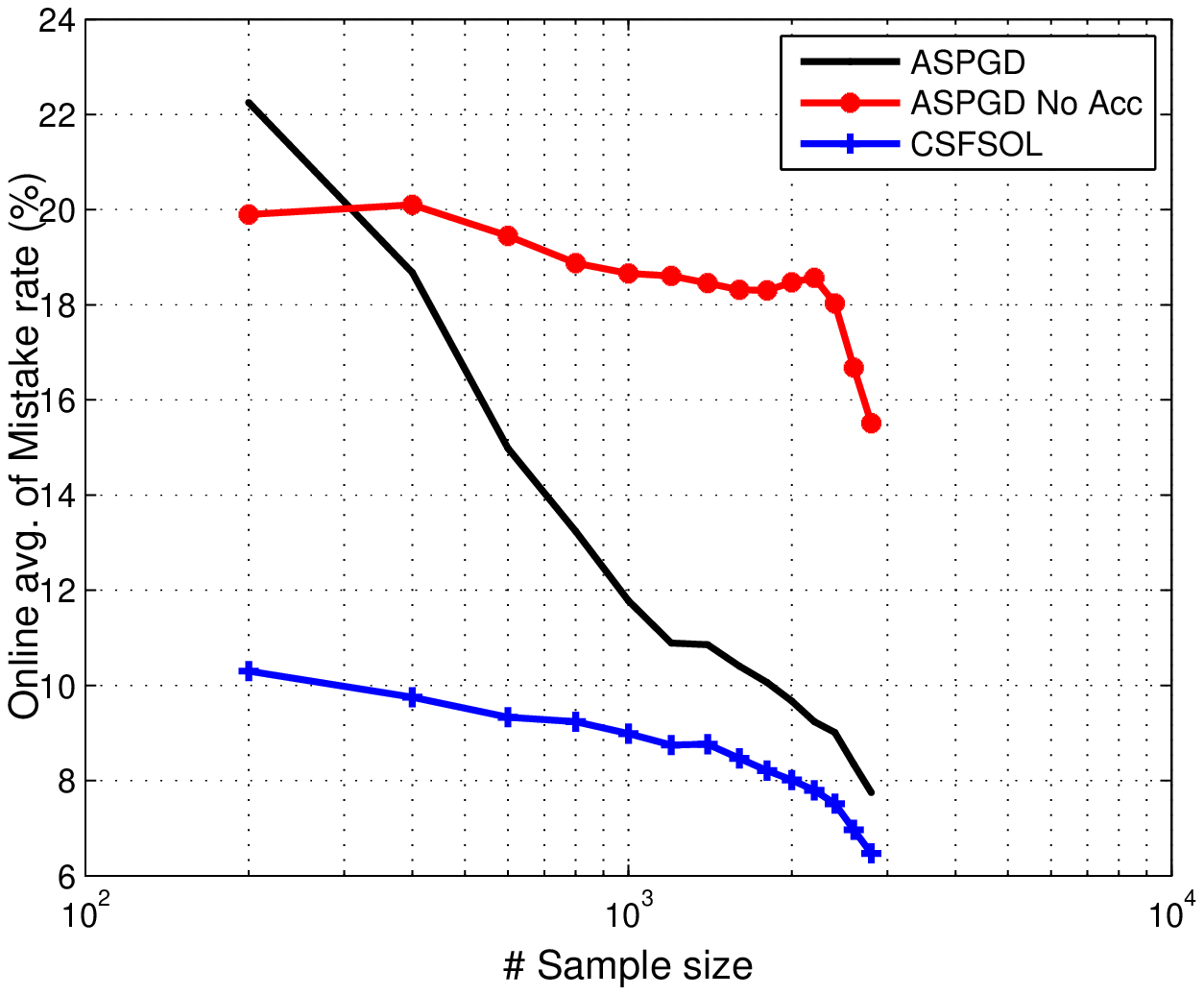}} 
\subfloat[realsim]{\includegraphics[width=6.3cm, height=4.5cm]{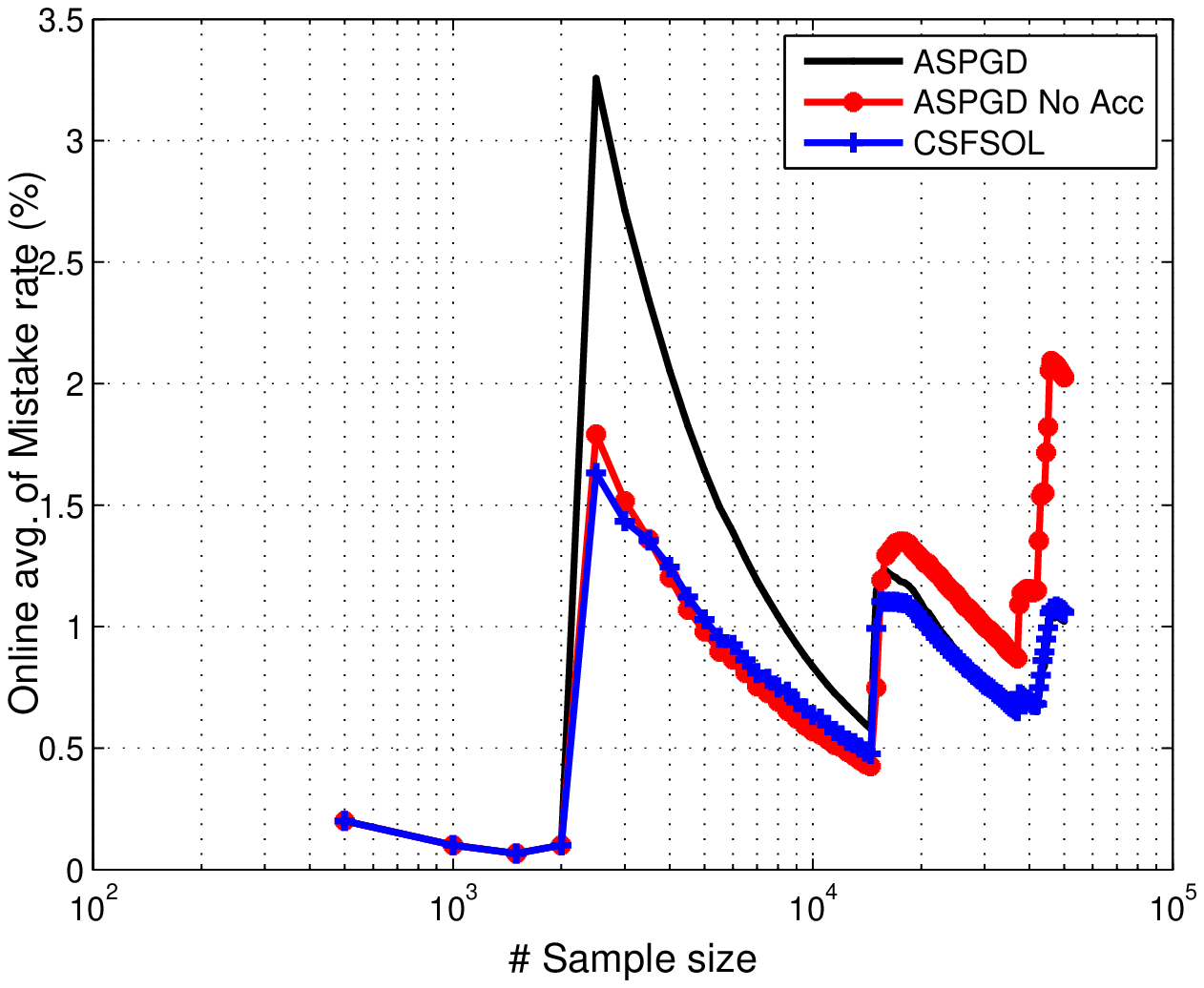}}\\
\subfloat[rcv1]{\includegraphics[width=6.3cm, height=4.5cm]{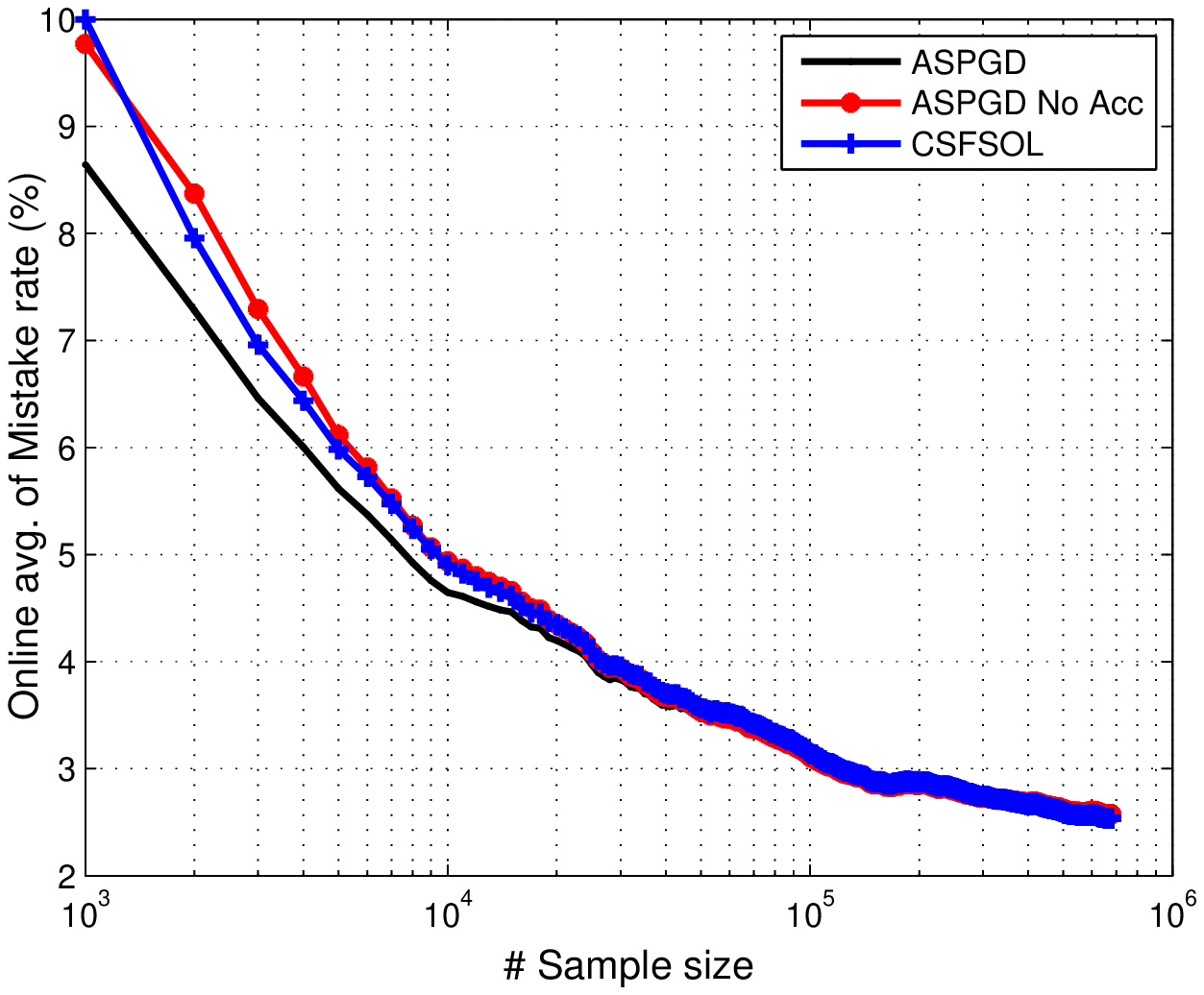}}
\subfloat[url]{\includegraphics[width=6.3cm, height=4.5cm]{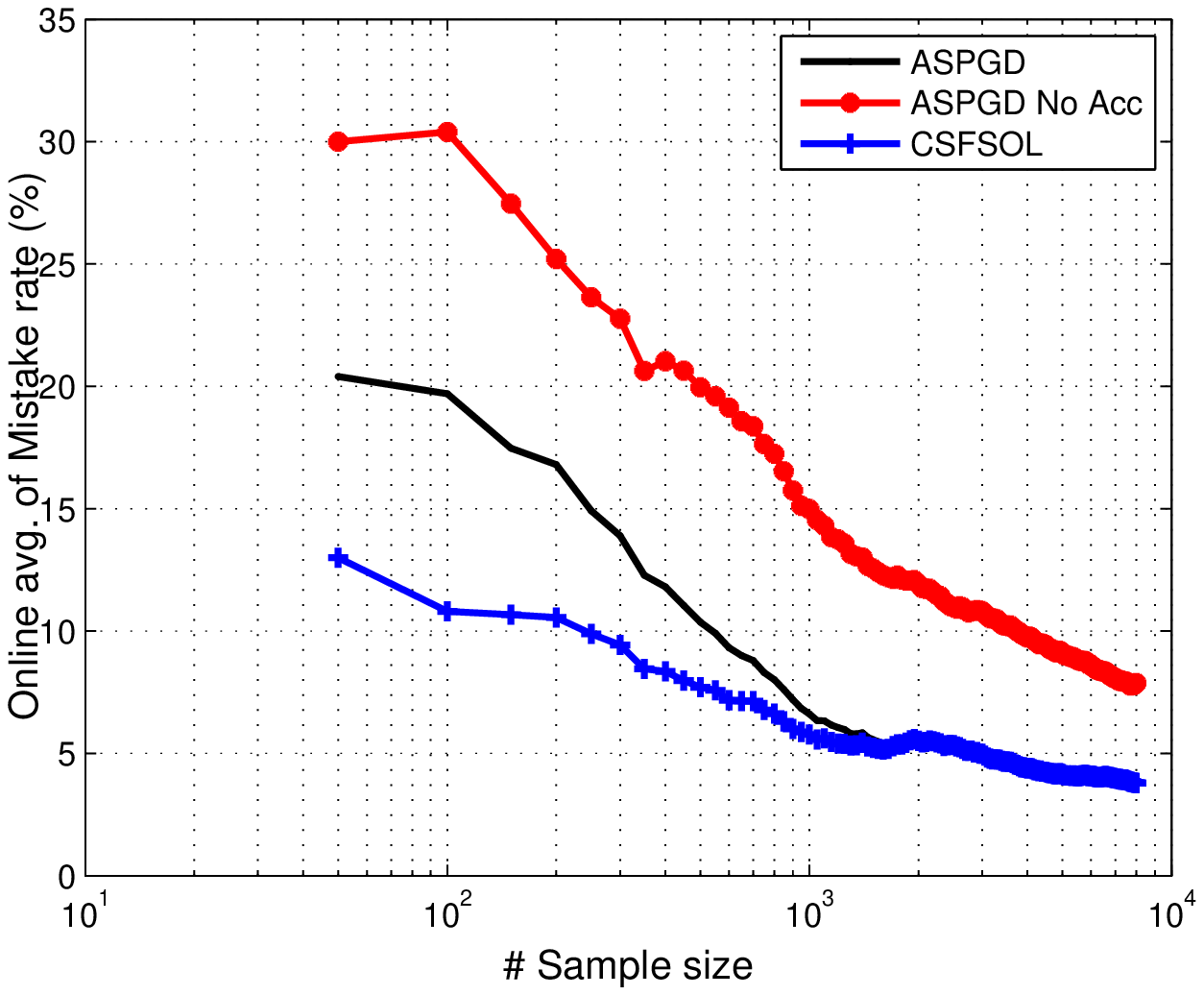}} \\
\subfloat[pcmac]{\includegraphics[width=6.3cm, height=4.5cm]{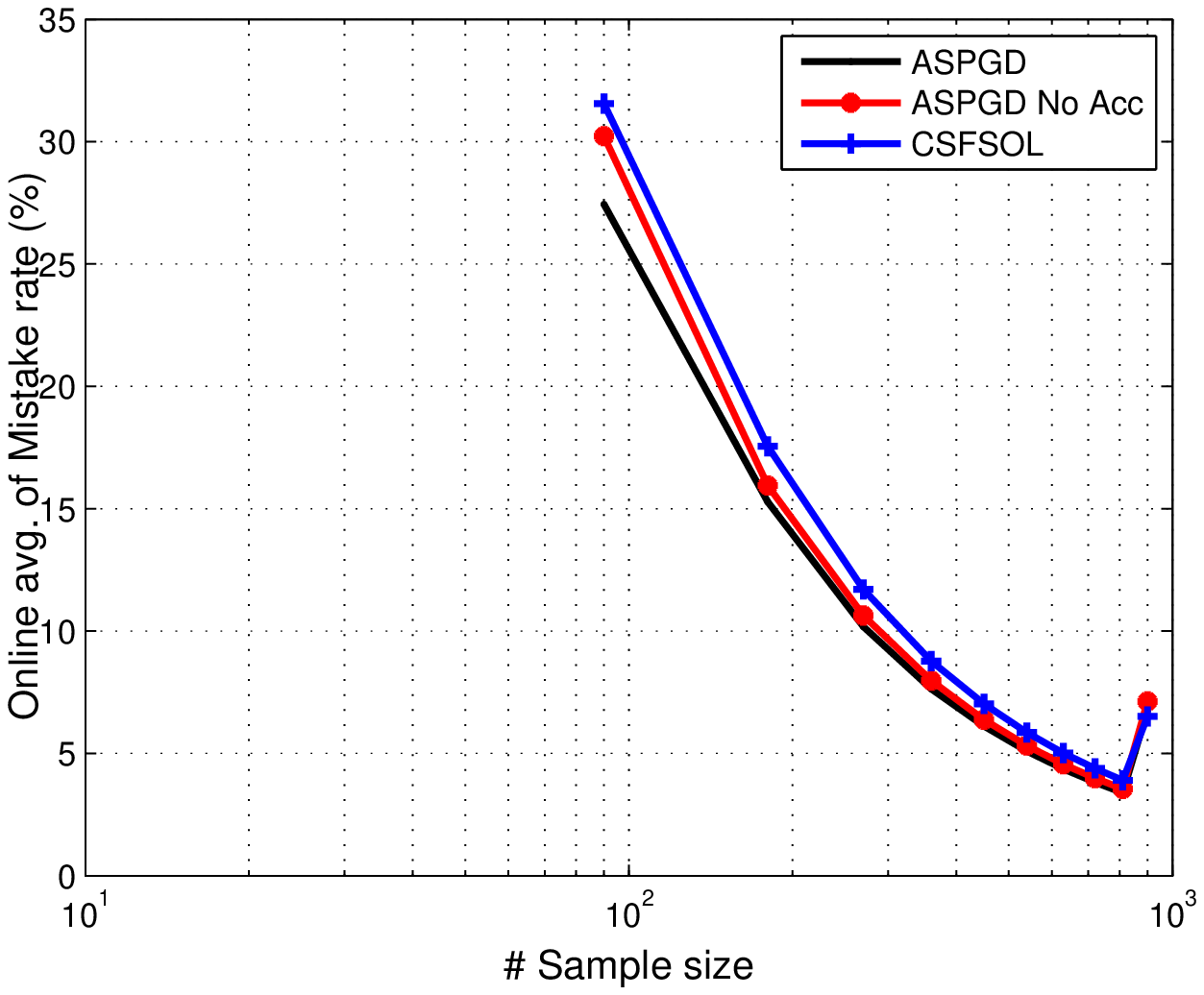}}
\subfloat[webspam]{\includegraphics[width=6.3cm, height=4.5cm]{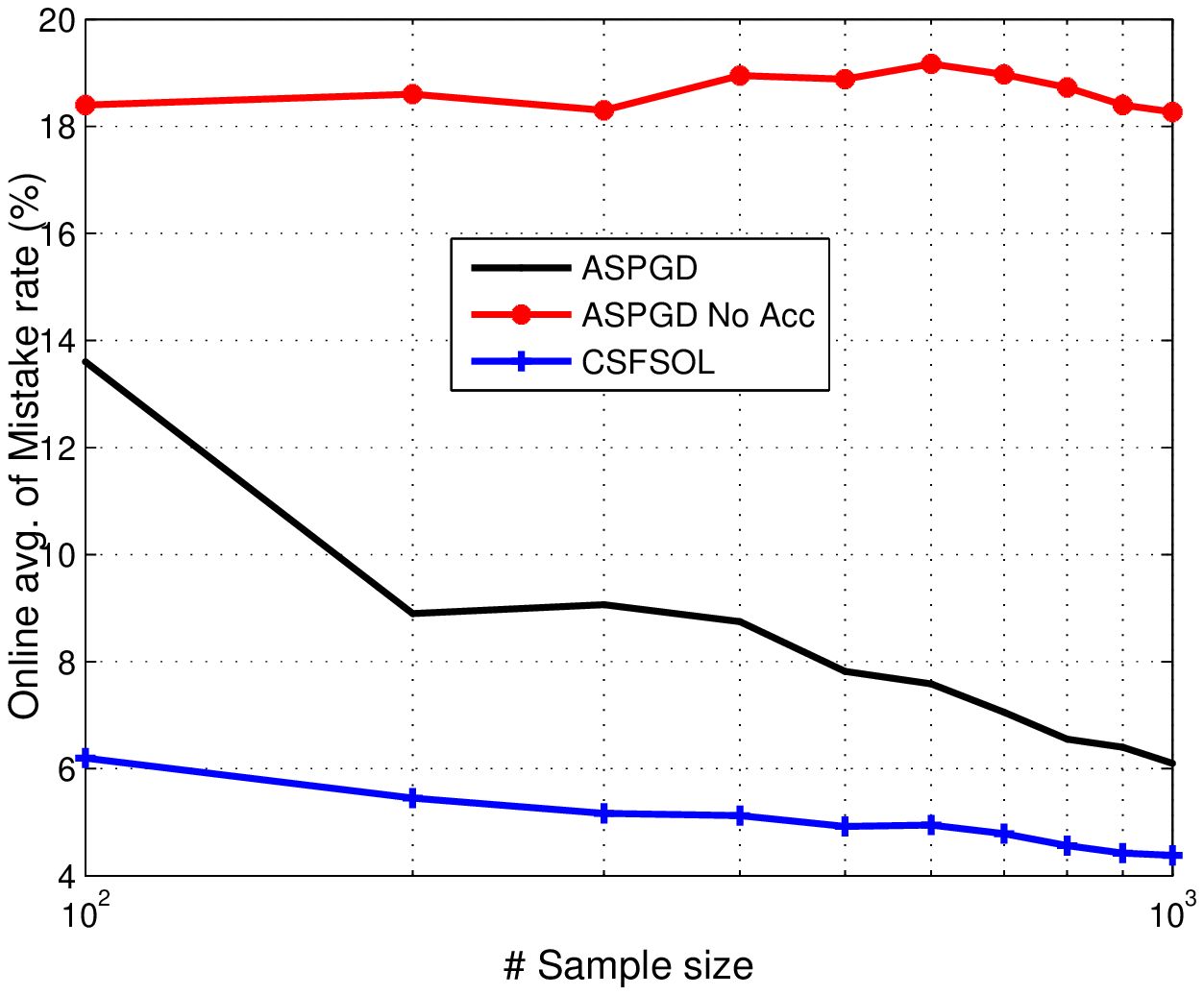}} \\
\caption{Evaluation of \emph{mistake} over various benchmark data sets. (a) news (b) news2 (c) gisette (d) realsim (e) rcv1 (f) url (g) pcmac (h) webspam.}
\label{fig:mistake}
\end{figure*}

\item {\bf Effect of Regularization Parameter on  F-measure}\\
The effect of regularization parameter $\lambda$ on \emph{F-measure} is shown in Figure \ref{fig:f-measure} on various benchmark data sets. From these figures, we can observe that increasing the regularization parameter $\lambda$ decreases the \emph{F-measure}. Important thing to notice here is that decrease in \emph{F-measure} with increasing $\lambda$ is higher in CSFSOL algorithm compared to ASPGD and ASPGDNOACC. In other words, we can obtain higher \emph{F-measure} from ASPGD and ASPGDNOACC compared to CSFSOL for a given $\lambda$.

\begin{figure*}
\centering
\subfloat[news]{\includegraphics[width=7.5cm, height=5.5cm]{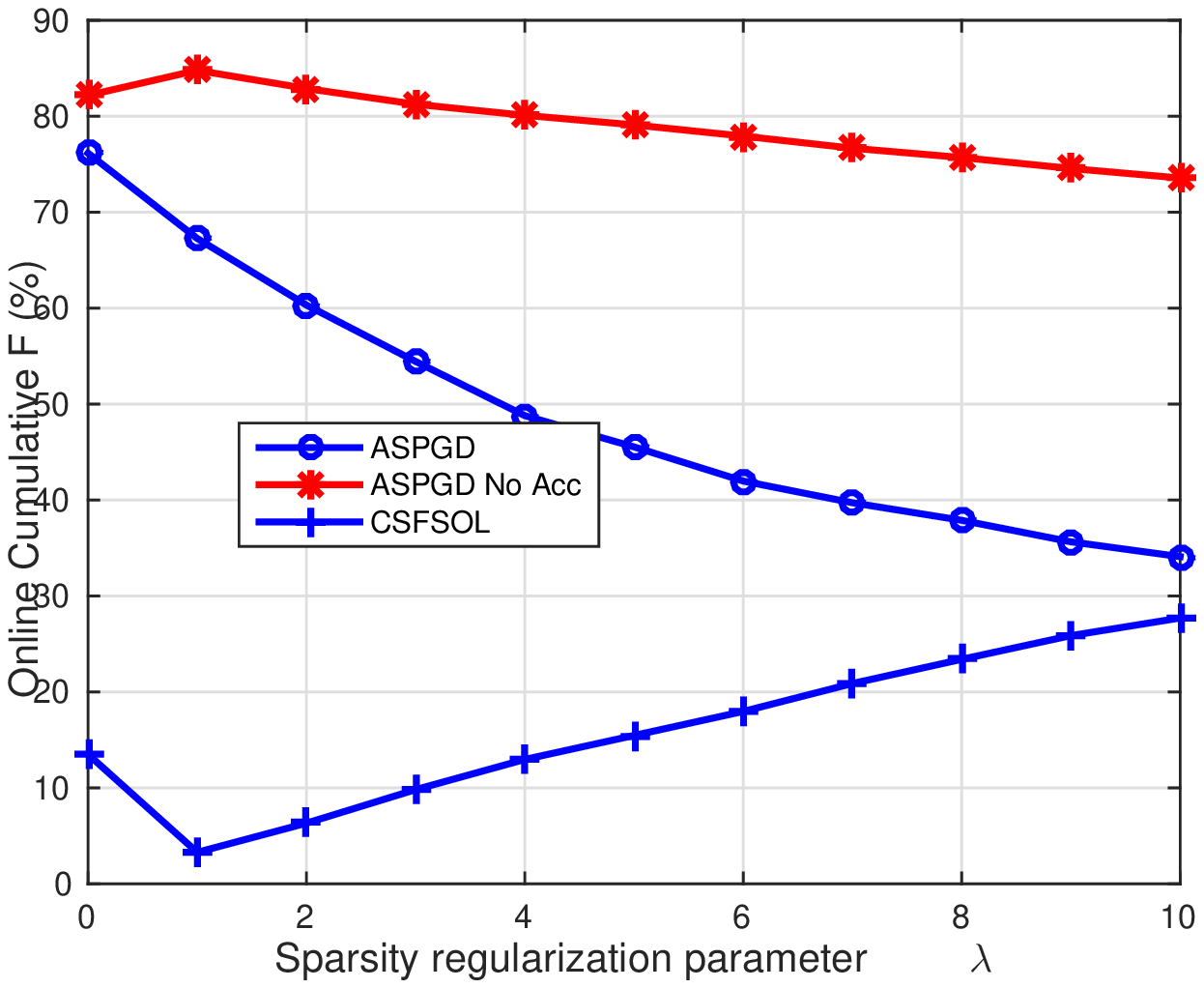}}  
\subfloat[realsim]{\includegraphics[width=7.5cm, height=5.5cm]{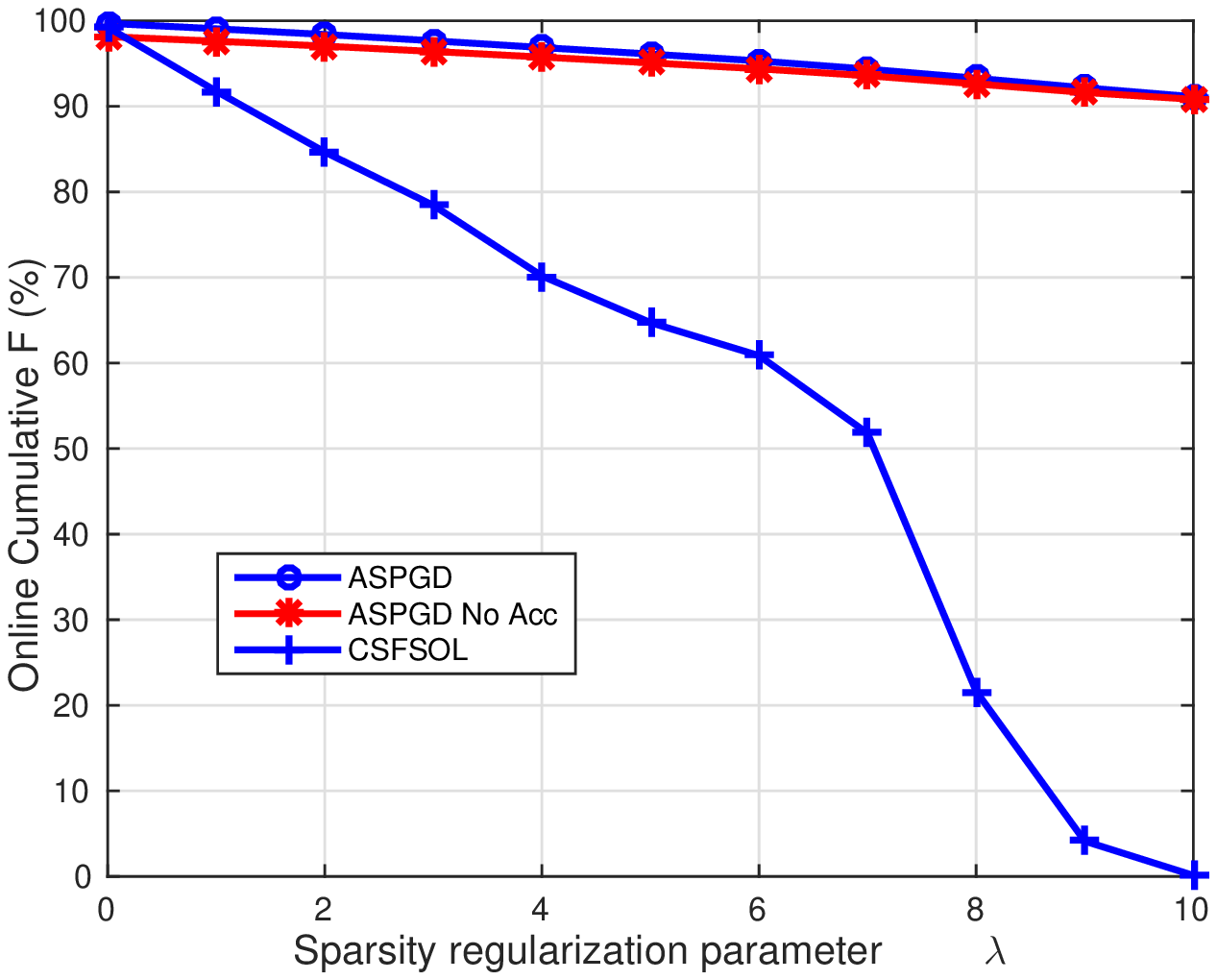}} \\
\subfloat[gisette]{\includegraphics[width=7.5cm, height=5.5cm]{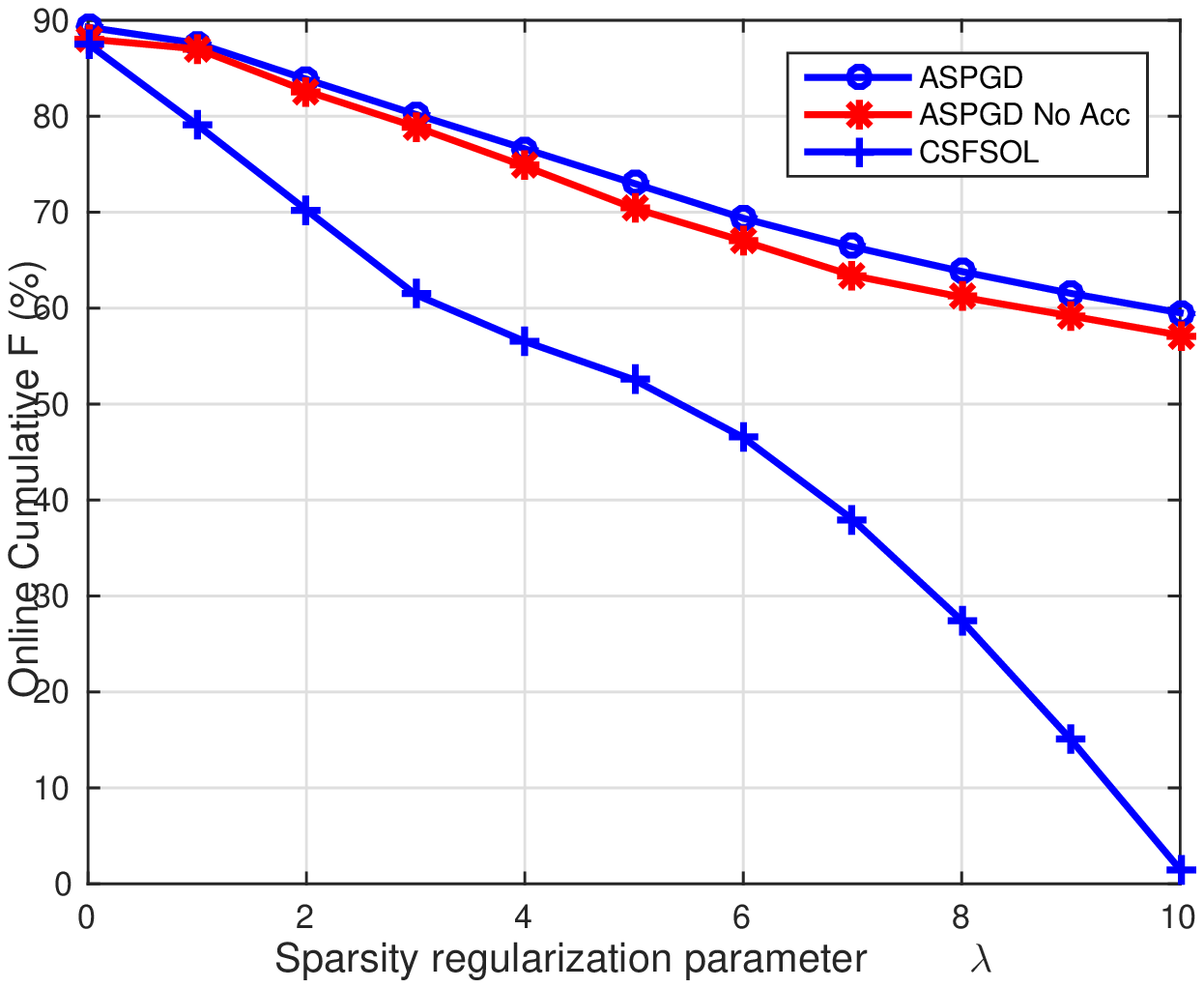}} 
\subfloat[rcv1]{\includegraphics[width=7.5cm, height=5.5cm]{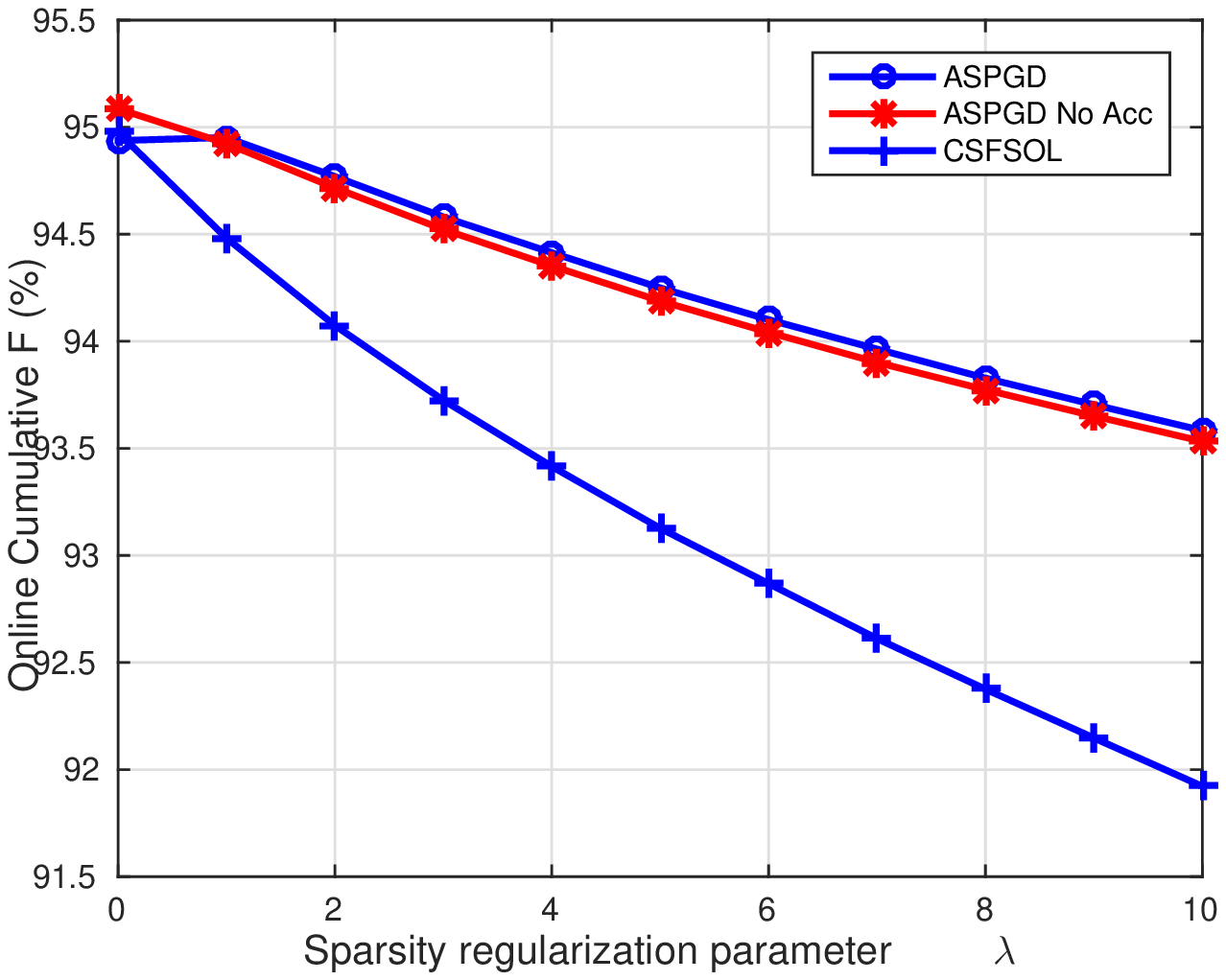}}\\
\subfloat[pcmac]{\includegraphics[width=7.5cm, height=5.5cm]{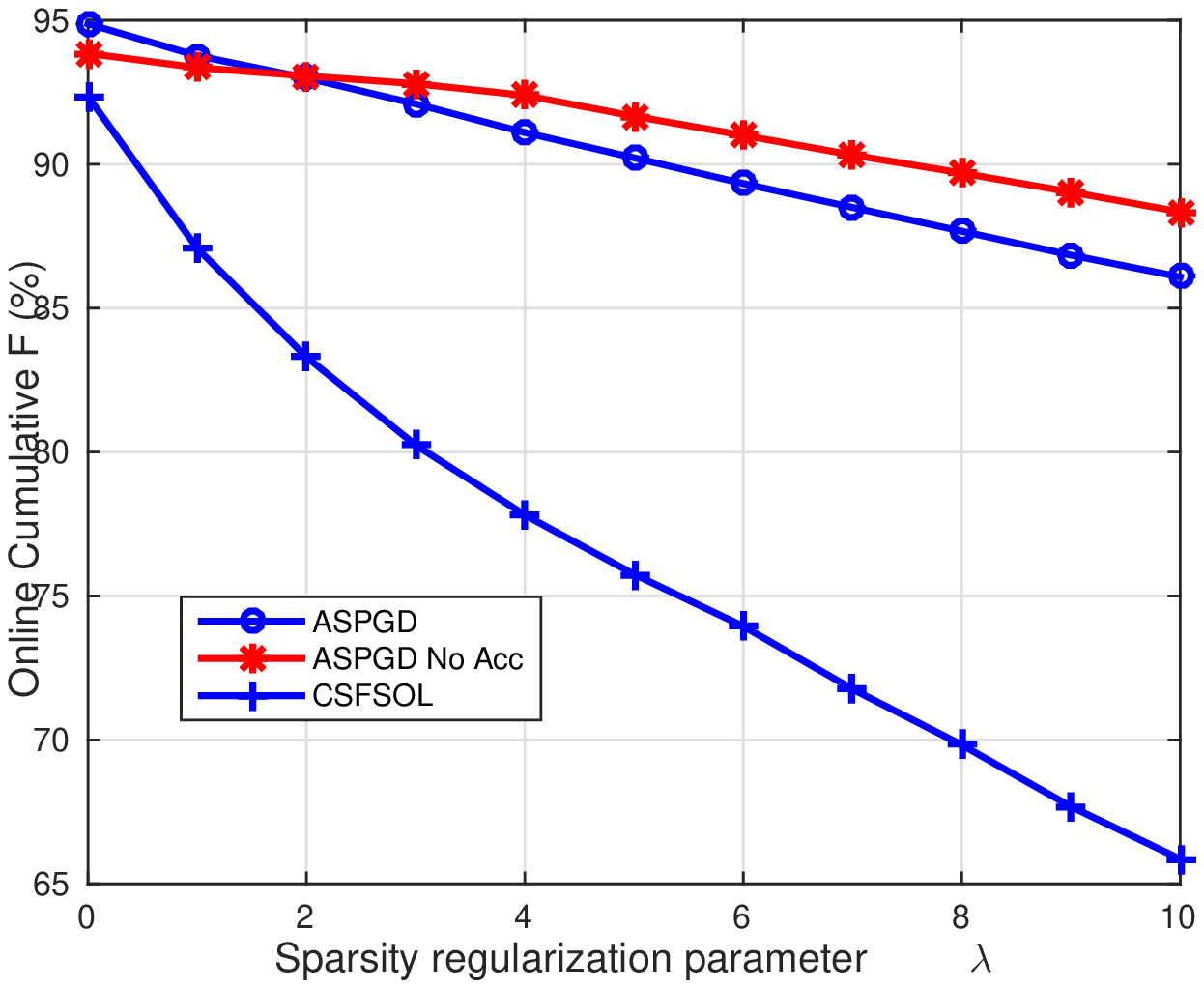}}
\caption{Effect of regularization parameter $\lambda$ on \emph{F-measure}  on (a) news (b) realsim (c) gisette (d)  rcv1 (e) pcmac.}
\label{fig:f-measure}
\end{figure*}

\begin{table*}[]
\footnotesize
\centering
\caption{Evaluation of cumulative \emph{Gmean(\%) } and \emph{Mistake rate (\%)} on benchmark data sets. Entries marked by * are statistically significant than the entries marked by ** and entries marked by $\dagger$ are NOT statistically significant than  the entries marked by $\ddagger$ at 95\% confidence level  on Wilcoxon rank sum test. }
\label{tab:gmean}
\begin{tabular}{|l|c|c|c|c|c|}
\hline
\multirow{2}{*}{Algorithm}  &  \multicolumn{2}{c|}{news} & \multirow{2}{*}{} &   \multicolumn{2}{c|}{news2} \\
\cline{2-3} \cline{5-6} 
&  Gmean(\%)  & Mistake rate(\%) & &Gmean(\%)  & Mistake rate(\%)  \\ \hline \hline
Proposed ASPGD & 90.7$\pm$ 0.003$^{*}$ & 2.087 $\pm$ 0.139$^{\dagger}$   & &  99.686$\pm$ 0.015 & 0.319 $\pm$ 0.014  \\\hline
Proposed ASPGDNOACC & {\bf 96.1$\pm $0.003} & 2.909 $\pm$ 0.117  && 99.426$\pm$ 0.027 & 0.587 $\pm$ 0.029  \\\hline

CSFSOL & 89.5 $\pm$ 0.004$^{**} $ & 2.013$\pm$ 0.069$^{\ddagger}$   &&99.870 $\pm$ 0.014 & 0.136$\pm$ 0.016   
 \\ 
\hline
\hline
\multirow{2}{*}{Algorithm}  &  \multicolumn{2}{c|}{gisette} & \multirow{2}{*}{} &   \multicolumn{2}{c|}{realsim} \\
\cline{2-3} \cline{5-6} 
&  Gmean(\%)   & Mistake rate(\%) & &Gmean(\%)  & Mistake rate(\%)  \\ \hline \hline
Proposed ASPGD  & {\bf 74.333 $\pm$ 1.174} & 7.750 $\pm$ 0.539  && 96.319 $\pm$ 0.077$^{**}$ & 1.022 $\pm$ 0.023$^{**}$ \\ \hline
Proposed ASPGDNOACC  &41.810$\pm$ 1.308 & 15.511 $\pm$ 0.826 && 96.636 $\pm$ 0.116 & 2.025 $\pm$ 0.036  \\ \hline

CSFSOL  &47.727 $\pm$ 2.176 & 6.468$\pm$ 0.184   && 96.783 $\pm$ 0.102$^{*}$ & 1.059 $\pm$ 0.025$^{*}$  \\ \hline

\hline
\multirow{2}{*}{Algorithm}  &  \multicolumn{2}{c|}{rcv1} & \multirow{2}{*}{} &   \multicolumn{2}{c|}{url} \\
\cline{2-3} \cline{5-6} 
&  Gmean(\%)   & Mistake rate(\%) & &Gmean(\%)  & Mistake rate(\%)  \\ \hline \hline
Proposed ASPGD &97.422 $\pm$ 0.003$^{**}$ & 2.576 $\pm$ 0.003$^{**}$  & & {\bf 92.139 $\pm$ 0.285} & 3.684 $\pm$ 0.151$^{*}$ \\ \hline
Proposed ASPGDNOACC &97.426 $\pm$ 0.003 & 2.574 $\pm$ 0.003  & &90.263 $\pm$ 0.380 & 7.880 $\pm$ 0.353  \\ \hline

CSFSOL    & 97.461 $\pm$ 0.004$^{*}$ & 2.535 $\pm$ 0.004$^{*}$ && 86.459 $\pm$ 0.497 & 3.791 $\pm$ 0.086 $^{**}$\\ \hline

\hline
\multirow{2}{*}{Algorithm}  &  \multicolumn{2}{c|}{pcmac} & \multirow{2}{*}{} &   \multicolumn{2}{c|}{webspam} \\
\cline{2-3} \cline{5-6} 
&  Gmean(\%)   & Mistake rate(\%) & &Gmean(\%)  & Mistake rate(\%)  \\ \hline \hline
Proposed ASPGD & 79.339 $\pm$ 3.149$^{\ddagger}$ & 6.778 $\pm$ 0.547$^{\ddagger}$     & &{\bf 71.052 $\pm$ 1.718} & 6.100 $\pm$ 0.727    \\ \hline
Proposed ASPGDNOACC &72.840 $\pm$ 4.024 & 7.122 $\pm$ 0.424   & &51.856 $\pm$ 5.034 & 18.270 $\pm$ 3.196   \\ \hline

CSFSOL    & 80.655 $\pm$ 1.918 $^{\dagger}$& 6.511 $\pm$ 0.360$^{\dagger}$    &&46.655 $\pm$ 5.684 & 4.380 $\pm$ 0.225   \\ \hline
\end{tabular}
\end{table*}
\begin{figure*}
\centering
\subfloat[news]{\includegraphics[width=7.5cm, height=5.5cm]{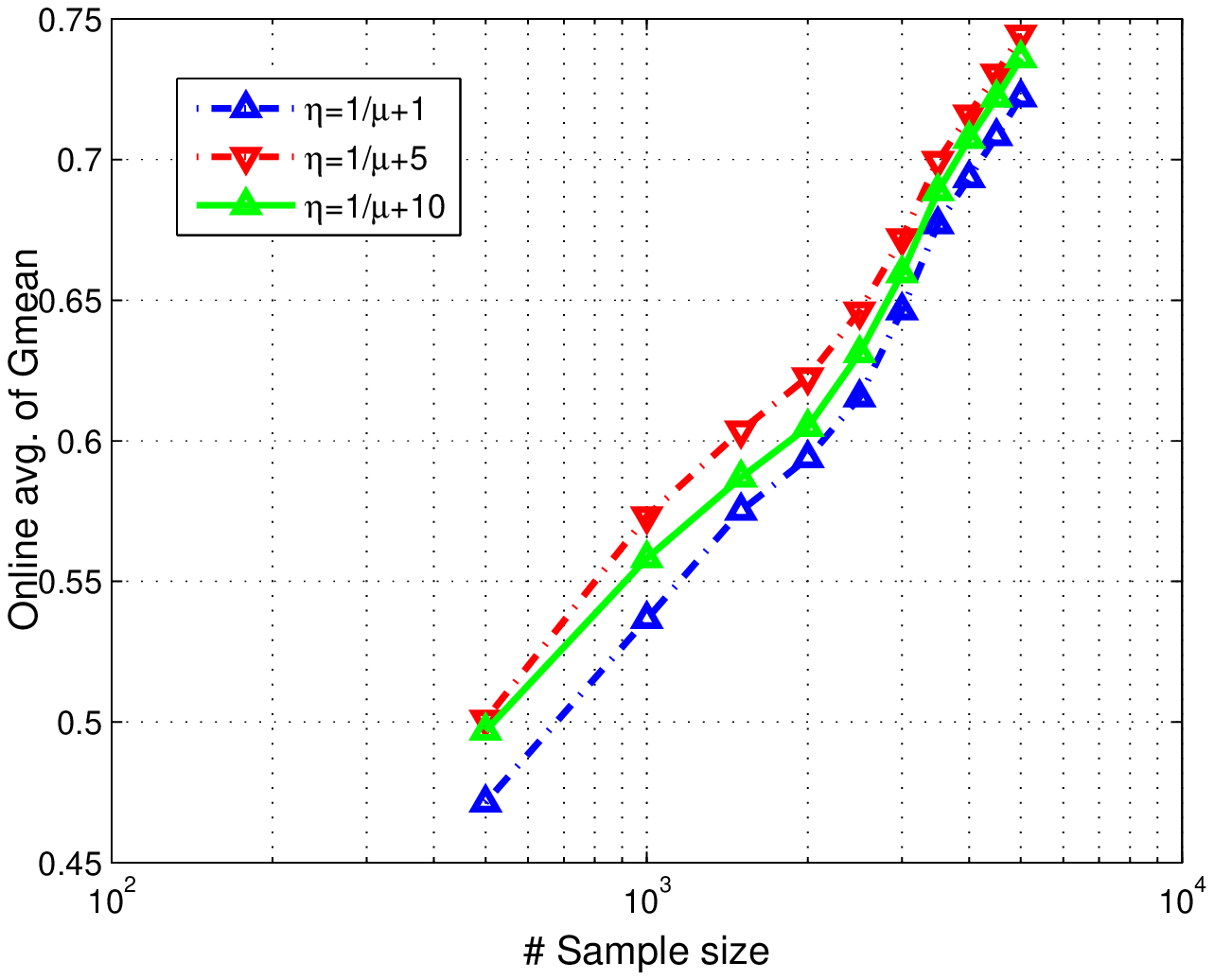}}  
\subfloat[realsim]{\includegraphics[width=7.5cm, height=5.5cm]{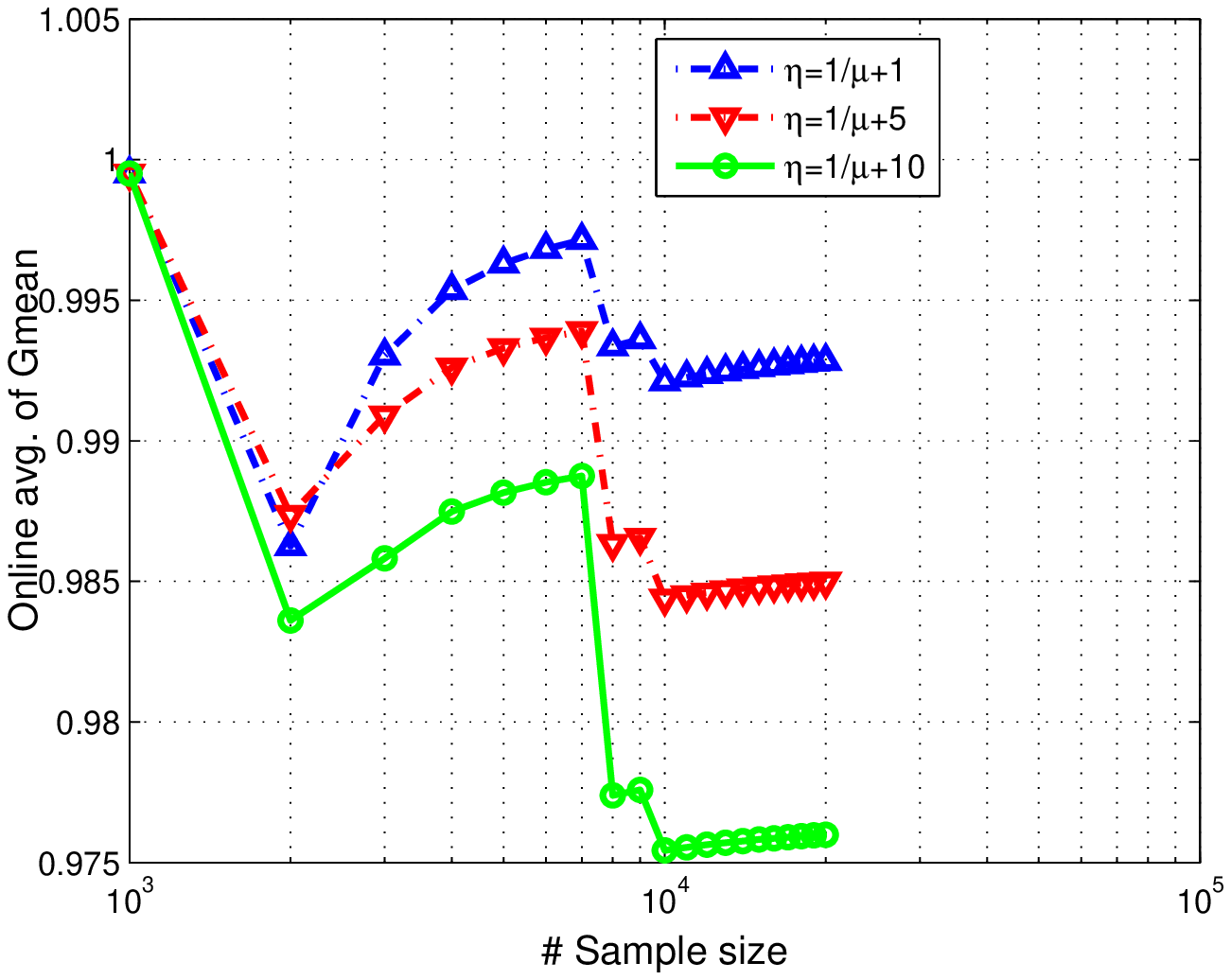}} \\
\subfloat[gisette]{\includegraphics[width=7.5cm, height=5.5cm]{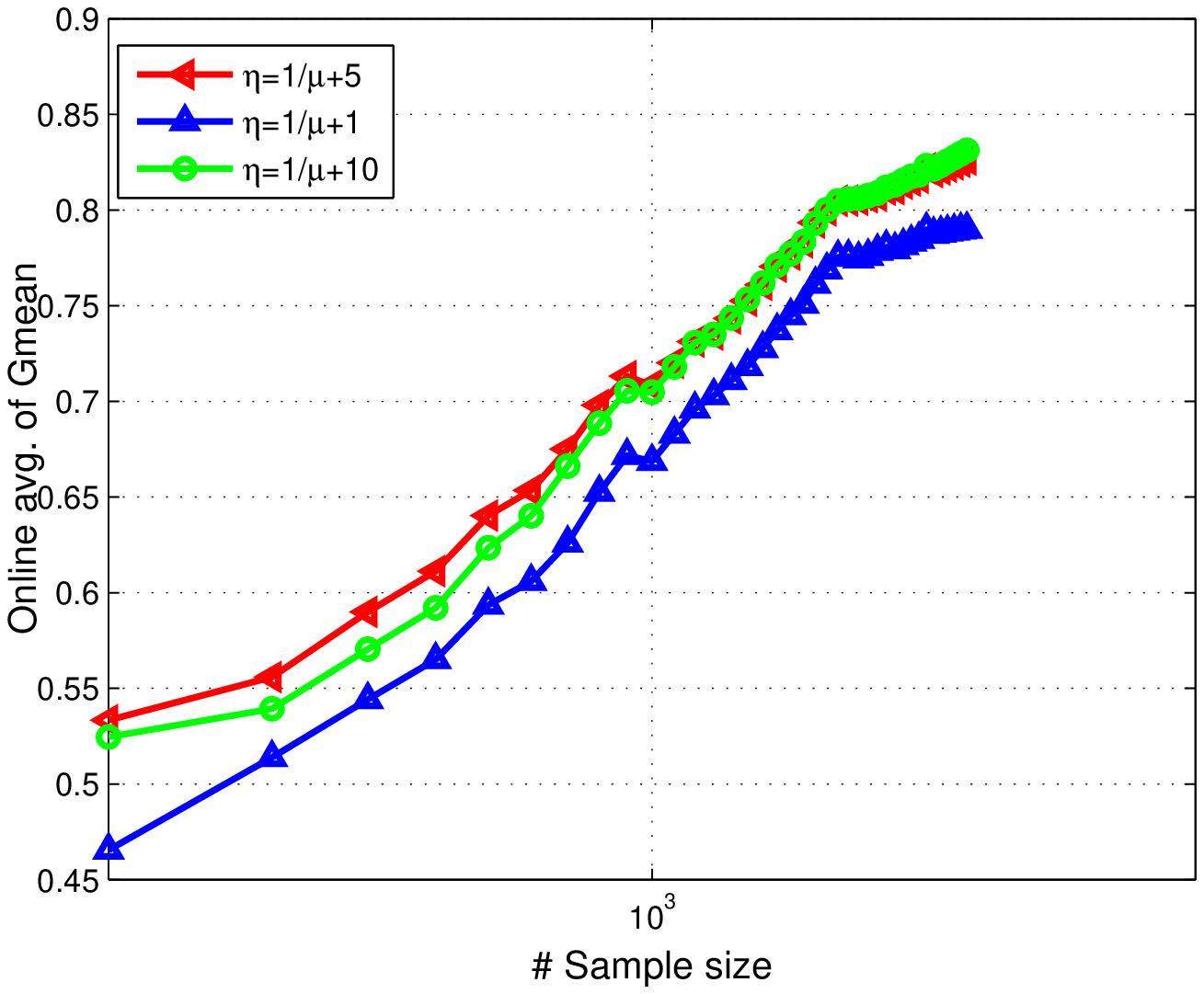}} 
\subfloat[rcv1]{\includegraphics[width=7.5cm, height=5.5cm]{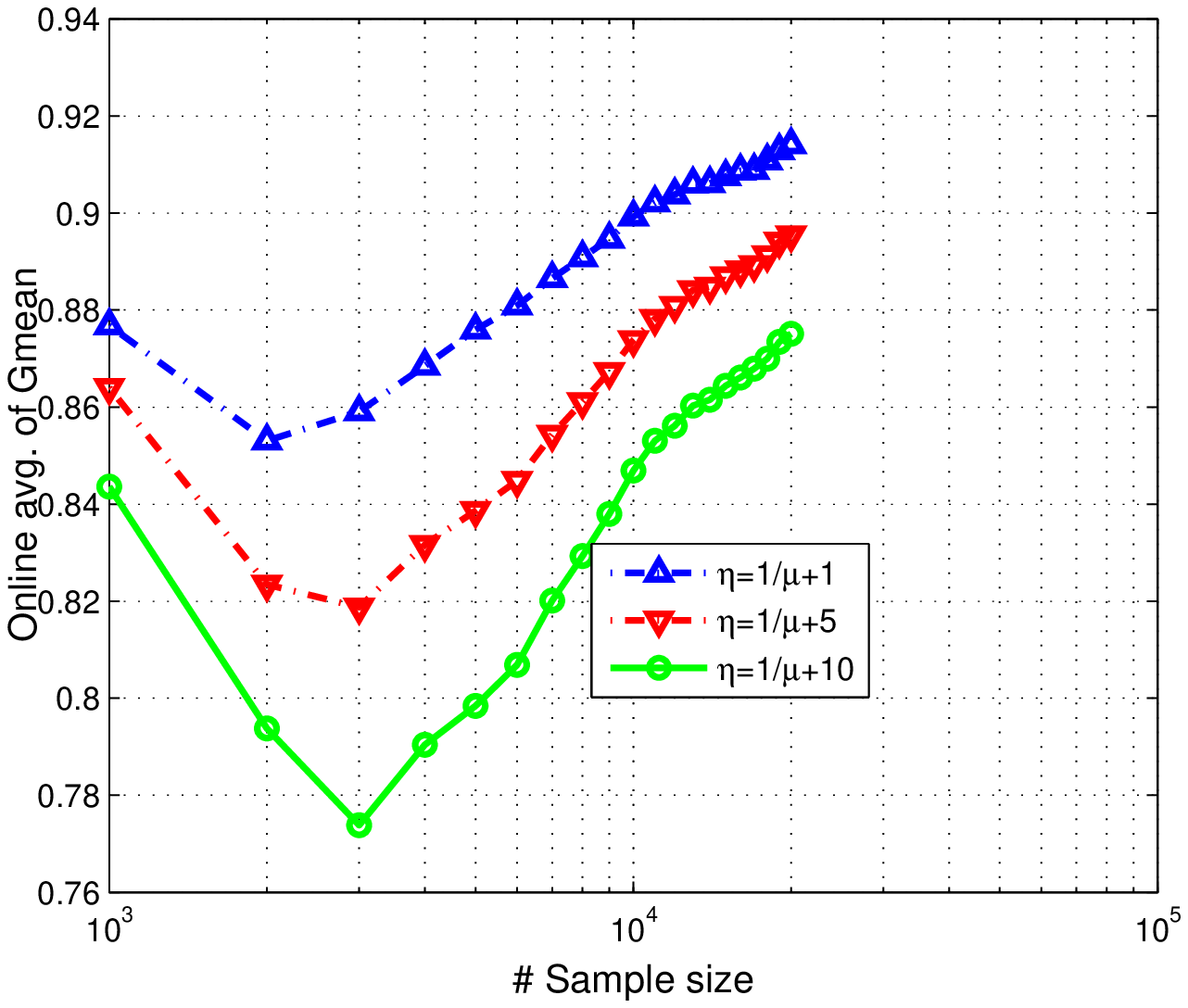}}\\
\caption{Effect of learning rate $\eta$ in ASPGD algorithm for maximizing \emph{Gmean}  on (a) news (b) realsim (c) gisette (d)  rcv1.}
\label{fig:eta}
\end{figure*}

\item {\bf Experiment with Varying Learning Rate}\label{learningrate}\\
In this Section, we demonstrate the effect of using varying learning rate $\eta$. 
 As we can see in Figure \ref{fig:eta} that there is no clear winner for maximizing \emph{Gmean}. On news data set, $\eta=1/(\mu+10)$ obtains highest \emph{Gmean} overall.  On the other hand, $\eta=1/(\mu+1)$ achieves the highest \emph{Gmean} overall on realsim and rcv1. We set the learning rate $\eta=1/(\mu+1)$ in the experiment  of the previous section based on this observation.

\item {\bf Effect of the Regularization Parameter}\\
In this Section, we demonstrate the effect of varying sparsity regularization parameter $\lambda$ on maximizing \emph{Gmean} and minimizing \emph{Mistake rate} in the algorithms compared. The results are shown in Figure \ref{fig:glambda} and \ref{fig:mlambda}. Regularization parameter $\lambda$ is varied in $[0,10]$ in steps of 1. From the Figure \ref{fig:glambda}, we observe that different algorithms achieve highest \emph{Gmean} on different values of $\lambda$ on news and gisette data set. For example, ASPGD achieves highest \emph{Gmean} at $\lambda=0$ whereas ASPGDNOACC  at $\lambda=1$  and CSFSOL at $\lambda=10$ on news data sets. On the other hand, on realsim and rcv1 data sets, all the algorithms achieve highest \emph{Gmean} at $\lambda=0$. Another major observation is that  ASPGD algorithm achieve higher \emph{Gmean} compared to CSFSOL on 3 out of 4 data sets (news, realsim, and rcv1) over the entire range of $\lambda$ values tested.  A higher $\lambda$ value implies the addition of  sparsity and hence more sparse model.  Sparse models are easier to interpret and quicker to evaluate.

In Figure \ref{fig:mlambda}, the effect of regularization parameter on \emph{Mistake rate} is shown. As before, ASPGD algorithm suffers smaller \emph{Mistake rate} compared to CSFSOL on the entire range of $\lambda$ values tested on 3 out of 4 data set (relaism, gisette, and rcv1). Further, smaller values of $\lambda$ lead to smaller \emph{Mistake rate} which is obvious from monotonically increasing \emph{Mistake rate} on all data sets and algorithms except ASPGD on gisette data set.
 
{\bf Remark: } The difference between ASPGD and SOL (generalized version of CSFSOL) is that ASPGD uses (1) smooth and modified hinge loss (2) Nesterov's acceleration. Time complexity of both the algorithms is $O(nd)$, which is linear in $n$ and $d$; where $n$ is the number of data point and $d$ is the dimensionality. Only thing that differs in the time complexity is the hidden constant in Big O notation. In fact, extra step involved in ASPGD is the  summation of two vectors of size d in step 5 of the algorithm that cost O(d). For big data where usually d is sparse, it will take $O(s)$ time where s is the number of nonzeros in ${\bf w}_{t-1}$ and ${\bf u}_t$. The evaluation of the smooth hinge loss  differs by a constant $(O(1))$.  In addition, from implementation point of view, the results reported in the paper assume that  data resides in back-end such as hard disk. We read data in mini batches and process them one by one. Thus, our implementation is more amicable to online learning, where we are not allowed to see all the data in one go. Whereas, implementation provided by authors of SOL load entire data in main memory and process one example at a time. Thus, their implementation violates the principle of  online learning. 

\end{enumerate}

\begin{figure*}
\centering
\subfloat[news]{\includegraphics[width=7.5cm, height=5.5cm]{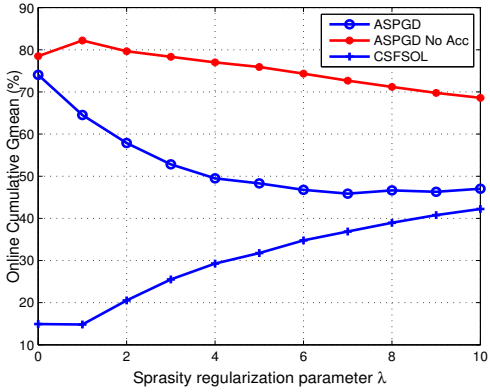}}  
\subfloat[realsim]{\includegraphics[width=7.5cm, height=5.5cm]{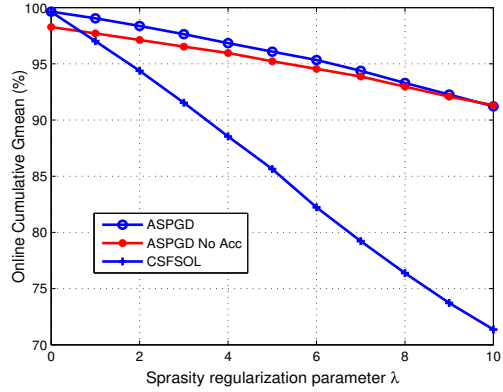}} \\
\subfloat[gisette]{\includegraphics[width=7.5cm, height=5.5cm]{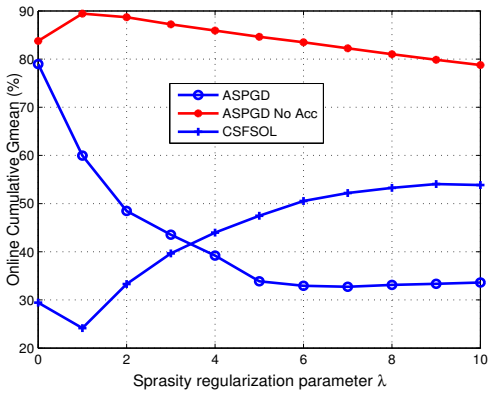}} 
\subfloat[rcv1]{\includegraphics[width=7.5cm, height=5.5cm]{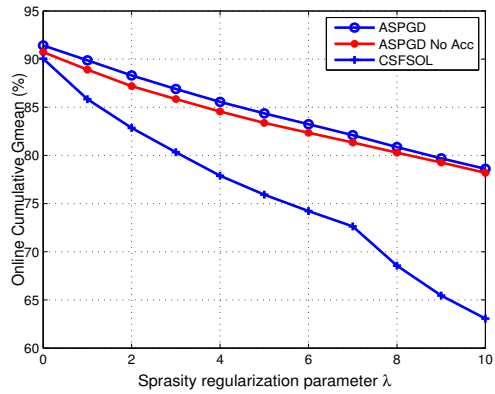}}\\
\caption{Effect of regularization parameter $\lambda$ in ASPGD algorithm  for maximizing \emph{Gmean} on  (a) news (b) realsim (c) gisette (d)  rcv.}
\label{fig:glambda}
\end{figure*}
\begin{figure*}
\centering
\subfloat[news]{\includegraphics[width=7.5cm, height=5.5cm]{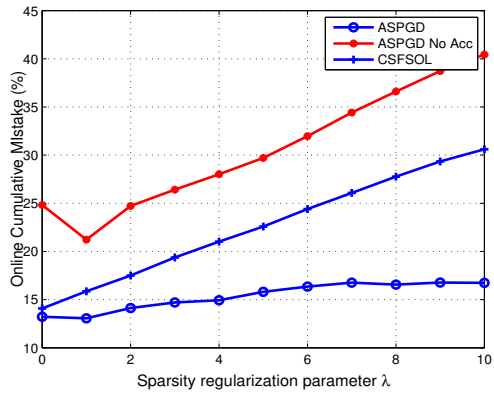}}  
\subfloat[realsim]{\includegraphics[width=7.5cm, height=5.5cm]{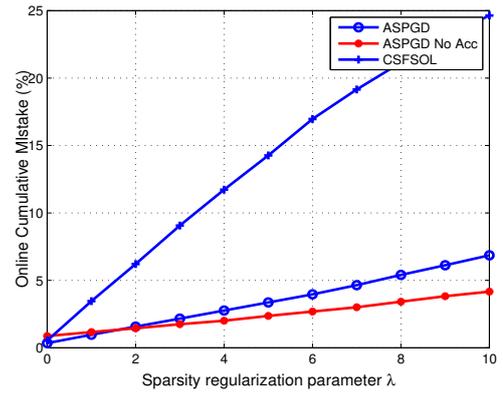}} \\
\subfloat[gisette]{\includegraphics[width=7.5cm, height=5.5cm]{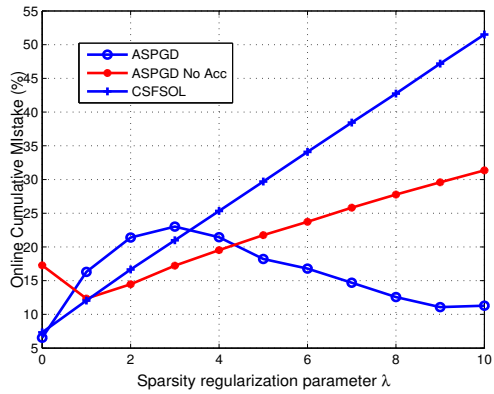}} 
\subfloat[rcv1]{\includegraphics[width=7.5cm, height=5.5cm]{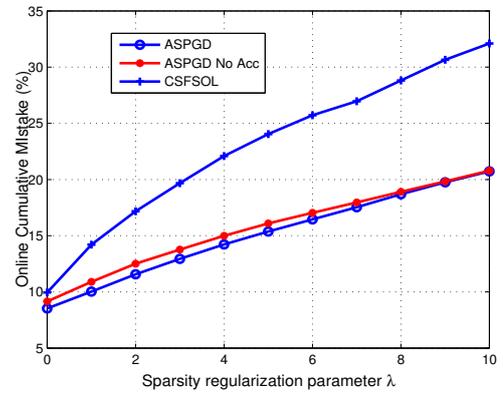}}\\
\caption{Effect of regularization parameter $\lambda$ in ASPGD algorithm  for minimizing \emph{Mistake rate} on  (a) news (b) realsim (c) gisette (d)  rcv1.}
\label{fig:mlambda}
\end{figure*}
\clearpage
\newpage
\section{Discussion}
In the present Chapter, we handle the \emph{streaming, sparse, high dimensional} problem of big data for detecting anomalies efficiently.  As discussed in Chapter \ref{Chapter3}, PAGMEAN algorithm does not scale over high dimensions; nor does it exploit the sparsity present in the big data. We follow the same recipe as in PAGMEAN algorithm to derive the ASPGD algorithm. However, instead of using the loss function employed by PAGMEAN, we use a smooth and strongly convex cost-sensitive loss function that is a convex surrogate for the  $0-1$ loss function.   The relaxed problem is solved via accelerated-stochastic-proximal learning algorithm  called ASPGD. Extensive experiments on several large-scale data sets show that ASPGD algorithm outperforms a recently proposed algorithm (CSFSOL) in terms of \emph{Gmean, F-measure} and \emph{Mistake rate} on many of the  data set tested. Further, we also compared  \emph{non-accelerated} version of ASPGD algorithm called ASPGDNOACC with ASPGD and CSFSOL. From the discussion in Section \ref{exp}, we also conclude that acceleration is not always helpful; neither in terms of \emph{Gmean} nor \emph{Mistake rate}. 

Because of the massive growth in data size and its distributed nature, there is immediate need to tackle the class imbalance in the distributed setting. In the next Chapter, we propose an algorithm for handling the class-imbalance problem in the distributed setting.
\chapter{Proposed Algorithms : DSCIL and CILSD }\label{Chapter5}
Globalization in the $21^{st}$ century has given rise to the distributed work culture.  As a result, data is no longer collected at a single place. Instead, it is gathered at multiple locations in a distributed fashion. Gleaning insightful information from  distributed data is a challenging task. There are several concerns that need to be addressed properly. Firstly, collecting the whole data at a single place for knowledge discovery is costly. Secondly, it involves security risk while transmitting data over the network. For example, credit card transaction data. To save cost and minimize the risk of data transportation, there is an urgent need to develop algorithms that can work in a distributed fashion. This is the main motivation behind the work proposed in the current Chapter. 

We study the  class imbalance problems  in a \emph{distributed} setting exploiting sparsity structure  in the data. We formulate the class-imbalance learning problem  as a cost-sensitive learning problem with $L_1$ regularization. The cost-sensitive loss function is a cost-weighted smooth hinge loss.  The resultant optimization problem is minimized within (i) \emph{Distributed Alternating Direction Method of Multiplier} (DADMM) \cite{Boyd:2010} framework (ii) FISTA \cite{Beck:2008}-like update rule in a distributed environment.  We call the algorithm derived within DADMM framework as \emph{Distributed Sparse Class-Imbalance Learning} (DSCIL) and within the FISTA-like update rule as \emph{Class-Imbalance Learning on Sparse data in a Distributed Environment} (CILSD). The reason for proposing CILSD is that  it improves upon the convergence speed  of DSCIL. 

 In the DSCIL algorithm, we partition the data matrix across samples through DADMM.  This operation splits the original problem into a distributed $L_2$ regularized smooth loss minimization and $L_1$ regularized squared loss minimization. $L_2$ regularized subproblem is solved via L-BFGS and random coordinate descent   method in parallel at multiple processing nodes using Message Passing Interface (MPI, a C++ library)  while $L_1$ regularized problem is just a simple soft-thresholding operation. We show, empirically, that the distributed solution matches  the centralized solution on many  benchmark data sets. The centralized solution is obtained via Cost-Sensitive Stochastic Coordinate Descent (CSSCD). 

In CILSD algorithm, we partition  the data across examples and distribute the subsamples to different processing nodes. Each node runs a local copy of FISTA-like algorithm which is a distributed-implementation of the prox-linear algorithm for cost-sensitive learning. Empirical results on small and large-scale benchmark datasets show some promising avenues to further investigate the real-world application of the proposed algorithms such as anomaly detection, class-imbalance learning etc.  To the best of our knowledge, ours is the first work to study class-imbalance in a \emph{distributed} environment on \emph{large-scale sparse} data.

\section{Introduction}
 In the present work, we have made an attempt to address the \emph{point anomaly detection} through the class-imbalance learning in big data in a distributed setting exploiting sparsity structure. Without exploiting sparsity structure, learning algorithms run slower since they have to work in the full-feature space. We propose to solve the class-imbalance learning problem from cost-sensitive learning perspective due to various reasons. First, Cost-sensitive learning can be directly applied to different classification algorithms. Second,  cost-sensitive learning generalizes well over large data and third, they can take into account user input cost. 


In summary, our contributions are as follows:
\vspace{-0.5cm}
\begin{enumerate}
  \setlength{\itemsep}{0pt plus 1pt}
\item We propose, to the best of our  knowledge, the first regularized cost-sensitive learning problem in \eqref{prob} in a \emph{distributed} setting exploiting \emph{sparsity} in the data.
\item The problem in \eqref{prob} is solved via (i) Distributed Alternating Direction Method of Multiplier (DADMM) (ii) distributed FISTA-like algorithm by example splitting across different processing nodes. The solution obtained by DADMM algorithm is called DSCIL and the solution obtained by distributed FISTA is called CILSD.

\item DADMM  splits the problem \eqref{prob} into two subproblems:  a distributed $L_2$ regularized loss minimization and a $L_1$ regularized squared loss minimization. The first subproblem is solved by L-BFGS as well as random coordinate descent method while the second subproblem is just a soft-thresholding operation obtained in a closed-form. We call our algorithm using L-BFGS method as L- Distributed Sparse Class-Imbalance Learning (L-DSCIL) and using random coordinate descent method as R-DSCIL. 
\item All the algorithms  are tested on various benchmark datasets and results are compared with the start-of-the-art algorithms as well as the centralized solution over various performance measures besides \emph{Gmean} (defined later).
\item
We also show (i) the Speedup (ii)  the effect of varying cost (iii) the effect of the number of cores in the distributed implementation (iv) the effect of the regularization parameter.
\item At the end, we show  the useful real-world application of DSCIL  and CILSD algorithms on KDDCUP 2008 data set.
\end{enumerate}

\section{Experiments}
In this Section, we demonstrate the empirical performance of the proposed algorithm DSCIL over various small and large-scale data sets\cite{CC01a} .  A brief summary of the benchmark data sets and the class-imbalance ratio is given in Table \ref{data}. All the algorithms were implemented in C++ and compiled by g++ on a Linux 64-bit machine containing 48 cores (2.4Ghz CPUs) \footnote{Sample code and data set for R-DSCIL can be downloaded from https://sites.google.com/site/chandreshiitr/publication}. 

\subsection{Experimental Testbed and Setup} \label{testbed}
 We compare our DSCIL algorithm with a recently proposed algorithm called Cost-Sensitive First Order Sparse Online learning (CSFSOL) of \cite{Dayong:2015}. CSFSOL is a cost-sensitive  online algorithm based on mirror descent update rule (see for example \cite{Nico2006}). CSFSOL also optimizes the same objective function as ours but the loss function is not smooth. Secondly, as mentioned in the introduction section, CSFSOL is an online centralized algorithm whereas DSCIL is a distributed algorithm. In the forthcoming subsections, we will show (i) The convergence of DSCIL on benchmark data sets (ii) The performance comparison of DSCIL, CSFSOL, and CSSCD algorithms in terms of \emph{accuracy, sensitivity, specificity, Gmean} and \emph{ balanced\_accuracy} (also called $Sum$ which is =0.5*sensitivity+0.5*specificity).  Note that  within the DSCIL algorithm, $L_2$ minimization  solved via L-BFGS algorithm is referred to as L-DSCIL and via CSRCD algorithm is referred to as R-DSCIL. For our distributed implementation, we use MPICH2 library.  MPICH is a high-performance C++ library based on Message Passing between nodes to realize the distributed implementation.

Note also that the DSCIL algorithm contains  ADMM penalty parameter $\rho$ which we set to 1 and the convergence of DSCIL  does not require tuning of this parameter. Regularization parameter $\lambda$ in DSCIL is set to $0.1\lambda_{max}$ where $\lambda_{max}$ is given by $(1/m)\|X^T\tilde{y}\|_\infty$ (see \cite{Koh:2007}), where $\tilde{y}$ is given by:
\[  \tilde{y}= \begin{cases} 
    m_-/m & y_i =1\\
    -m_+/m & y_i =-1, i=1,...,m\\
   \end{cases}
\]
Setting $\lambda$ as discussed above does not require its tuning  as  compared to $\lambda$ in CSFSOL where no closed form solution is available as such and one needs to do the cross-validation.  DSCIL algorithm is stopped when primal and dual residual fall below the primal and dual residual tolerance (see  chapter 3 of \cite{Boyd:2010} for details). For the CSFSOL algorithm, we see that it contains  another parameter, that  is, the learning rate $\eta$. Both of these parameters were searched in the range $\{3\times10^{-5},9\times 10^{-5},3\times 10^{-4},9\times 10^{-4},3\times  10^{-3},9\times 10^{-2},0.3,1,2,4,8\}$ and $\{ 0.0312, 0.0625, 0.125, 0.25, 0.5, 1, 2, 4, 8,16, 32\}$ respectively as discussed in \cite{Dayong:2015} and the best value  on the performance metric is chosen for testing. As an implementation note, we normalize the columns of data matrix so that each feature value lies in $[-1,1]$. All the results are obtained by running the distributed  algorithms on 4 cores using MPI unless otherwise stated.

\subsection{Convergence of DSCIL}
 In this Section, we discuss the convergence of L-DSCIL and R-DSCIL algorithms . The convergence plot with respect to the DADMM iteration over various benchmark datasets is shown in Figure \ref{conv}. From the Figure \ref{conv}, it is clear that L-DSCIL converges faster than R-DSCIL which is obvious as L-DSCIL is a second order (quasi-Newton) method while R-DSCIL is a first order method.  On another note, we  observe that on w8a  and rcv1 data set,  R-DSCIL  starts increasing the objective function that indicates the CSRCD algorithm, which is a random coordinate descent algorithm, overshoots the minimum after a certain number of iterations. The convergence plot also shows the correctness of our implementation.

\begin{figure*}
\centering
\subfloat[ijcnn1]{\includegraphics[width=8cm, height=6cm]{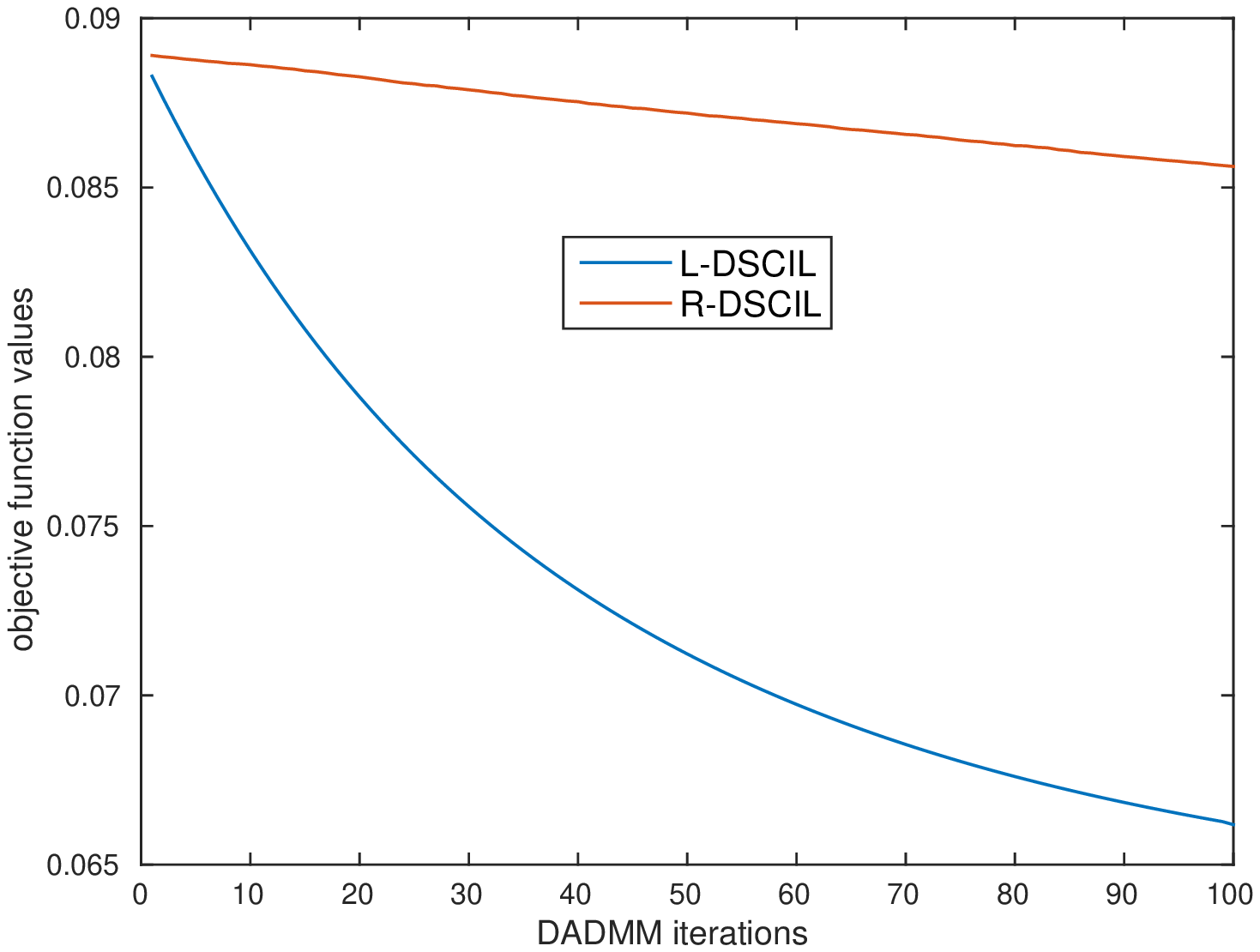}}  
\subfloat[rcv1]{\includegraphics[width=8cm, height=6cm]{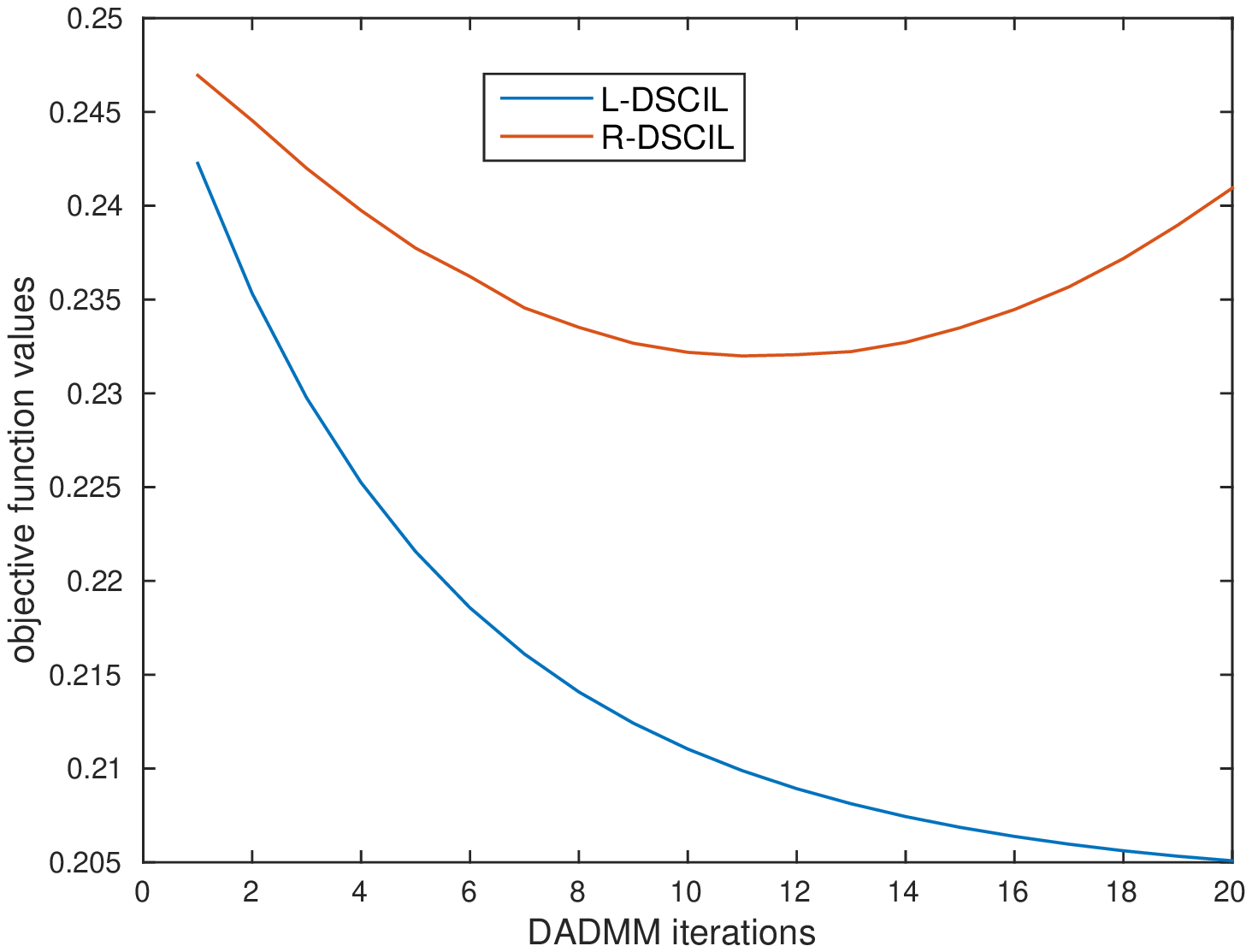}} \\
\subfloat[pageblocks]{\includegraphics[width=8cm, height=6cm]{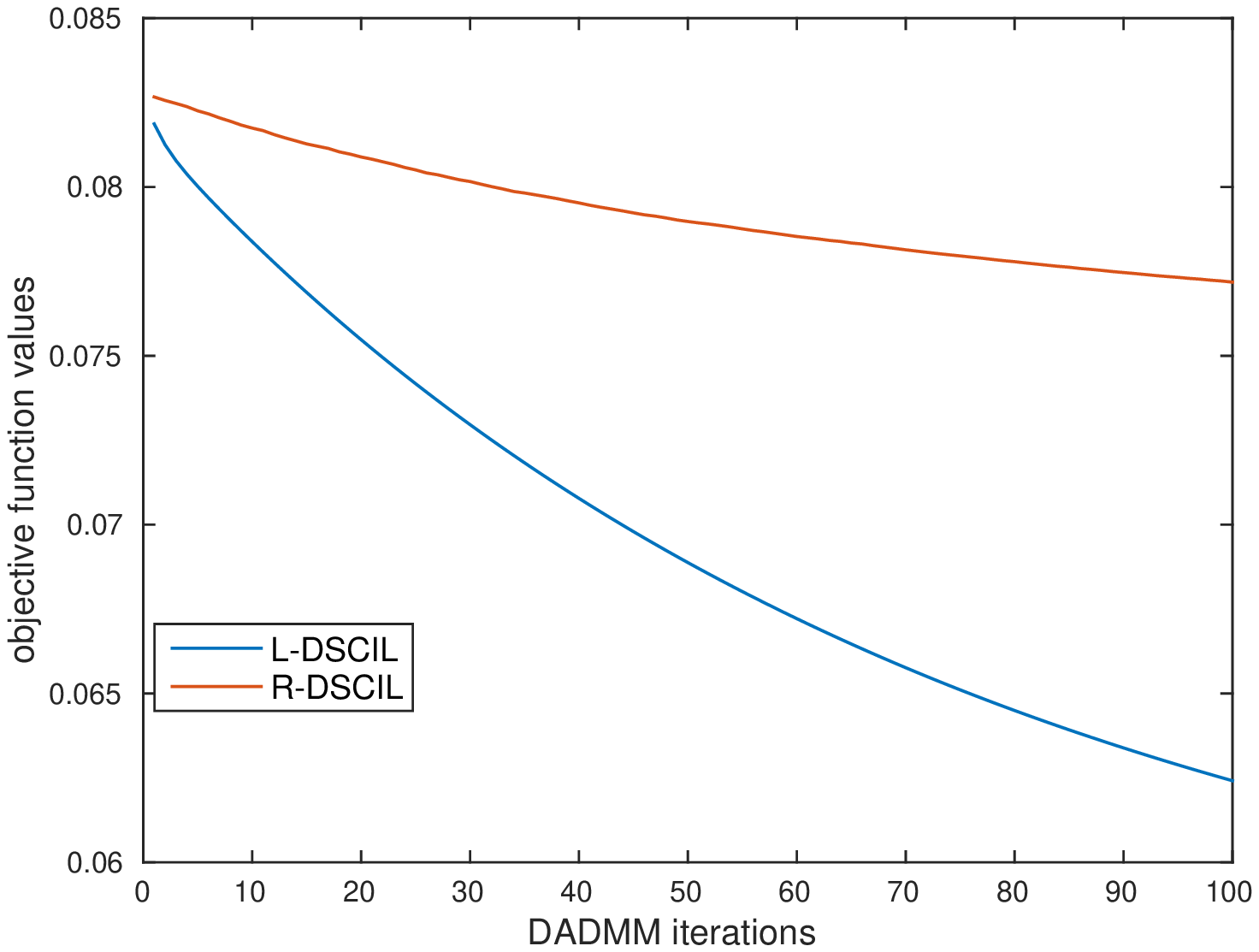}} 
\subfloat[w8a]{\includegraphics[width=8cm, height=6cm]{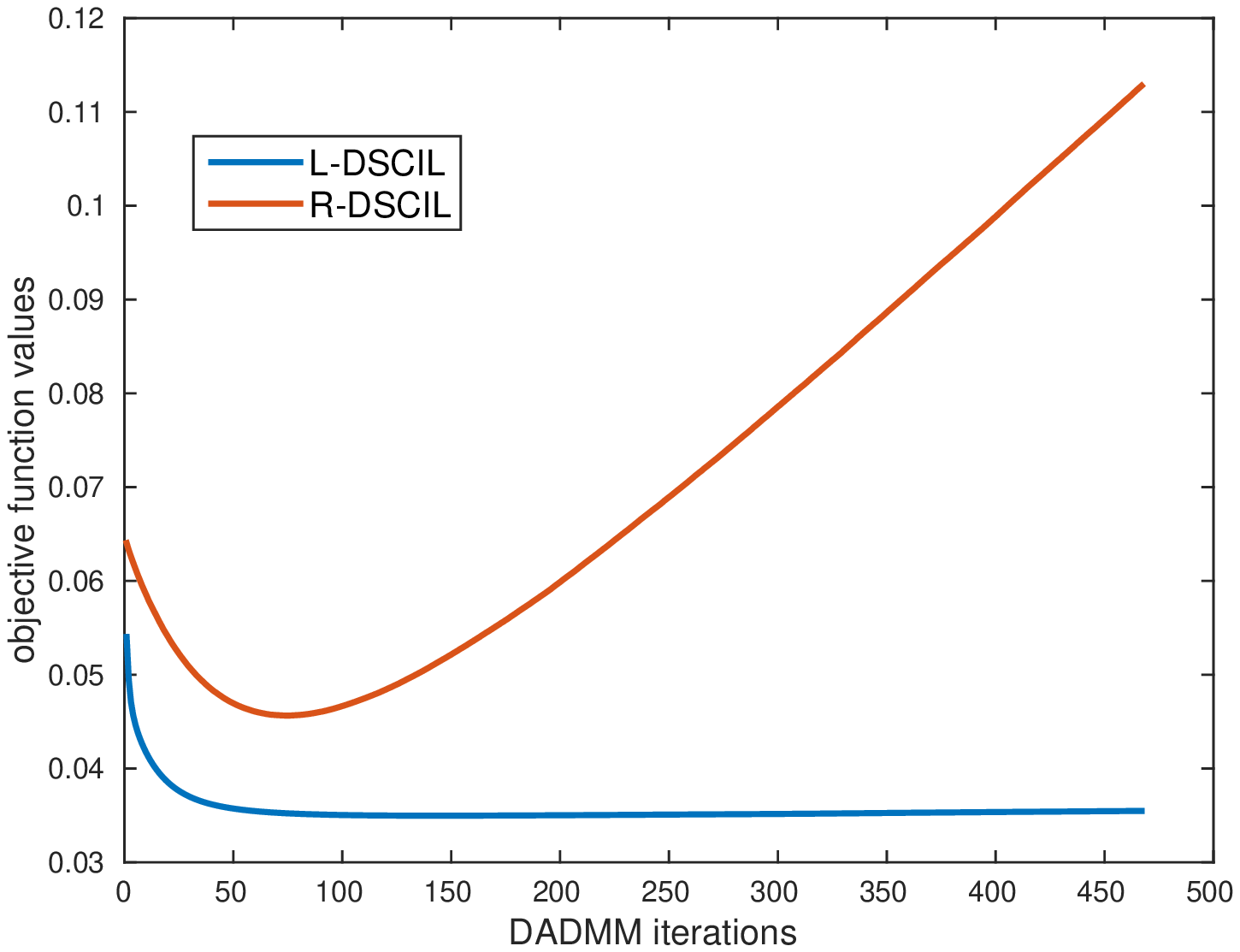}}\\
\caption{Objective Function vs DADMM iterations over  benchmark data sets. (a) ijcnn1 (b) rcv1 (c) pageblocks (d) w8a.} 
\label{conv}
\end{figure*}

\begin{table*}[]
\footnotesize
\centering
\caption{Performance comparison of CSFSOL, L-DSCIL, R-DSCIL and CSSCD over various benchmark data sets.}
\label{tab11}
\begin{tabular}{|l|l|l|l|l|l|}
\hline
\multirow{2}{*}{Algorithm}  &  \multicolumn{5}{c|}{news}  \\
\cline{2-6} 
& Accuracy & Sensitivity& Specificity& Gmean& Sum  \\ \hline \hline
CSFSOL    & 0.966356&	0.982759&	0.966137&	0.974412&	0.974448  \\ \hline
CSSCD  & 0.451466&1	&0.444137&	0.666436&	0.722069  \\ \hline
Proposed L-DSCIL     & 0.992271	&0.568966&	0.997927	&{\bf 0.753516}	&0.783446 \\ \hline
Proposed R-DSCIL       & 0.991134&	0.396552&	0.999079&{\bf 0.629433}	&0.697815 \\ \hline 
\hline
\multirow{2}{*}{Algorithm}  &  \multicolumn{5}{c|}{url}  \\
\cline{2-6} 
& Accuracy & Sensitivity& Specificity& Gmean& Sum  \\ \hline \hline
CSFSOL    &0.9725	&0.994505	&0.970297&	0.982327	&0.982401 \\ \hline
CSSCD  &0.947	&0.967033&	0.944994&	0.95595&	0.956014 \\ \hline
Proposed  L-DSCIL     & 0.9385	&0.824176&	0.949945&{\bf 0.884829} &0.8870605 \\ \hline
Proposed  R-DSCIL       &0.944	&0.791209&	0.959296& 0.871208&0.8752525 \\ \hline
 
\hline
\multirow{2}{*}{Algorithm}  &  \multicolumn{5}{c|}{gisette}  \\
\cline{2-6} 
& Accuracy & Sensitivity& Specificity& Gmean& Sum  \\ \hline \hline
CSFSOL    &0.813&	0.896&	0.73	&0.808752	&0.813\\ \hline
CSSCD  &0.945	&0.918&	0.972&	0.944614&	0.945\\ \hline
Proposed L-DSCIL     & 0.954&	0.926&	0.982	&{\bf 0.953589}&	0.954\\ \hline
Proposed  R-DSCIL       &0.5&	0&	1&	0	&0.5 \\ \hline
 \hline
\multirow{2}{*}{Algorithm}  &  \multicolumn{5}{c|}{ijcnn1}  \\
\cline{2-6}
& Accuracy & Sensitivity& Specificity& Gmean& Sum \\ \hline \hline
CSFSOL    &0.831932	&0.607668	&0.855475&	0.721002	&0.731571 \\ \hline
CSSCD  &0.83456&	0.827824&	0.835267&	0.831537	&0.831546 \\ \hline
Proposed L-DSCIL     &0.868126	&0.576217	&0.89877&	0.719643&	0.737493 \\ \hline
Proposed R-DSCIL       & 0.718258&	0.982438&	0.690525&{\bf 0.823649}&	0.836482 \\ \hline
 \hline
\multirow{2}{*}{Algorithm}  &  \multicolumn{5}{c|}{covtype}  \\
\cline{2-6} 
& Accuracy & Sensitivity& Specificity& Gmean& Sum  \\ \hline \hline
CSFSOL    & 0.908817	&0.86642&	0.910199&	0.88804&	0.888309 \\ \hline
CSSCD  &0.968433	&0	&1&	0&	0.5 \\ \hline
Proposed L-DSCIL     & 0.953617&	0.699578&	0.961897&  0.820318	&0.830737 \\ \hline
Proposed R-DSCIL       & 0.968433&	0&	1&	0	&0.5 \\ \hline

\end{tabular}
\end{table*}

\begin{table*}[]
\footnotesize
\centering
\caption{Performance comparison of CSFSOL, L-DSCIL, R-DSCIL and CSSCD over various benchmark data sets.}
\label{tab12}
\begin{tabular}{|l|l|l|l|l|l|}
\hline
 \multirow{2}{*}{} &   \multicolumn{5}{c|}{rcv1} \\
\cline{2-6} 
&  Accuracy & Sensitivity& Specificity& Gmean& Sum  \\ \hline \hline
CSFSOL    &  0.957&	0.960148&	0.953312&	0.956724&	0.95673 \\ \hline
CSSCD  &  0.8989&	0.871548	&0.930945&	0.900757&	0.901246\\ \hline
Proposed L-DSCIL     &  0.9175	&0.925116	&0.908578	&{\bf 0.916809}&	0.916847\\ \hline
Proposed R-DSCIL      &0.8697	&0.792215	&0.960478	& 0.872299&0.8763465 \\ \hline
 
\hline
 \multirow{2}{*}{} &   \multicolumn{5}{c|}{webspam} \\
\cline{2-6} 
&   Accuracy & Sensitivity& Specificity& Gmean& Sum  \\ \hline \hline
CSFSOL     &0.9925	&0.952&	0.9952	&0.97336&	0.9736\\ \hline
CSSCD  &0.988	&0.808&	1&	0.898888	&0.904\\ \hline
Proposed L-DSCIL      & 0.9705	&0.528&	1&	0.726636&	0.764\\ \hline
Proposed R-DSCIL         &0.885	&0.968	&0.879467	&{\bf 0.922672}&	0.923733 \\ \hline
 
\hline
 \multirow{2}{*}{} &   \multicolumn{5}{c|}{realsim} \\
\cline{2-6} 
&  Accuracy & Sensitivity& Specificity& Gmean& Sum  \\ \hline \hline
CSFSOL   &0.8825&	0&	1	&0	&0.5\\ \hline 
CSSCD   & 0.9105&	0.33617&	0.986969&	0.576012&	0.66157\\ \hline
Proposed L-DSCIL     & 0.812071&	0.900304&	0.800324&{\bf 0.848843}&	0.850314\\ \hline
Proposed R-DSCIL      &0.824286	 &0.90152	&0.814002	&{\bf 0.856644}&	0.857761 \\ \hline
 \hline
\multirow{2}{*}{} &   \multicolumn{5}{c|}{w8a} \\
\cline{2-6}
&   Accuracy & Sensitivity& Specificity& Gmean& Sum  \\ \hline \hline
CSFSOL    & 0.970637&	0.0330396&	1	&0.181768&	0.51652 \\ \hline
CSSCD    & 0.97231	&0.563877&	0.9851&	0.745302&	0.774489\\ \hline
Proposed L-DSCIL     & 0.975587&	0.682819&	0.984755	&{\bf 0.820006}	&0.833787 \\ \hline
Proposed R-DSCIL     &0.978597&	0.348018&	0.998344&	0.589442&	0.673181 \\ \hline
 \hline
 \multirow{2}{*}{} &   \multicolumn{5}{c|}{pageblocks} \\
\cline{2-6} 
&   Accuracy & Sensitivity& Specificity& Gmean& Sum  \\ \hline \hline
CSFSOL    & 0.793056&	0.662069	&0.81306&	0.73369&	0.737564 \\ \hline
CSSCD  & 0.887163	&0.751724	&0.907846&	0.826105&	0.829785 \\ \hline
Proposed L-DSCIL      & 0.919598&	0.706897&	0.95208	&{\bf 0.820379}	&0.829488\\ \hline
Proposed R-DSCIL        &0.90772&	0.344828&	0.993681&	0.585362&	0.669254 \\ \hline 

\end{tabular}
\end{table*}


\subsection{Comparative Study on Benchmark Data sets}
\begin{enumerate}
\item {\bf Performance Comparison with respect to Gmean}\\
We  show the performance comparison of various algorithms compared  with respect to various metrics such as \emph{Accuracy, Sensitivity, Specificity, Gmean, Sum} in Table \ref{tab11} and \ref{tab12}. Here, we want to focus on \emph{Gmean} column as this is the metric of our interest. Now, several conclusions can be drawn from these tables. Firstly, L-DSCIL performance is equally good compared to the centralized solution of CSSCD on many of the data sets. For example, it gives superior performance than CSSCD on \emph{news, gisette, rcv1, realsim} and \emph{ pageblocks} data sets in terms of \emph{Gmean}.  Secondly, the performance of R-DSCIL is not so good compared to CSSCD. It was able to outperform CSSCD on \emph{realsim} and \emph{webspam} data sets only.  Secondly, \emph{Gmean} achieved by CSFSOL on large-scale data sets such as \emph{news, rcv1, url} and \emph{webspam} is higher than any of the other method. This observation can be attributed due to possibly (i) the  use of \emph{strongly} convex objective function used in our present work compared to \emph{convex} but non-smooth objective function in CSFSOL  (ii) we stopped the L-DSCIL algorithm before it reaches optimality (iii) $\lambda_{max}$ value calculated as discussed in subsection \ref{testbed} is not the right choice. We believe that the second and third reason is more likely than the first one as can be seen in the convergence plot of the L-DSCIL algorithm in Figure \ref{conv}. We stopped the L-DSCIL algorithm either primal and dual residual went below the primal and dual feasibility tolerance or maximum iteration is reached (which we set to 20 for large-scale data sets). To verify our point, we ran another experiment with \emph{MaxIter} set to 50. This setting gives the \emph{Gmean} equal to 0.923436 which is clearly larger than previously obtained value 0.916809 for \emph{rcv1} data set. Similarly, by running L-DSCIL and R-DSCIL for a larger number of iterations, we can increase the desired accuracy $\epsilon$.
Finally, we also observe that R-DSCIL fails to capture the class-imbalance on \emph{gisette} and \emph{covtype} whereas CSSCD on \emph{covtype} and CSFSOL on \emph{realsim}  (due to the value of \emph{Gmean} being 0) which indicates the possibility of using L-DSCIL in practical settings.
\item{\bf Study on  Gmean versus Cost}\\
We study the effect of varying the cost on \emph{Gmean}. The results are presented in Figure \ref{lbfgs_cost} and \ref{rcd_cost} for L-DSCIL and R-DSCIL respectively. In each Figure, cost and \emph{Gmean} appear on x and y-axis respectively.  From these Figures, we can draw the following observations. Firstly, \emph{Gmean} is increasing for balanced data such as \emph{rcv1} when the cost for each class equals 0.5. On the other hand, more imbalance the data set is, more the cost given to positive class and higher the \emph{Gmean} (see the hist plot for \emph{news, pageblocks, url}) in Figure \ref{lbfgs_cost}. The same observation can be made from the Figure \ref{rcd_cost}. These observations allude to the fact that right choice for cost is important otherwise classification is affected severely.

\begin{figure*}
\centering
 \subfloat[]{ \includegraphics[width=8cm, height=5cm]{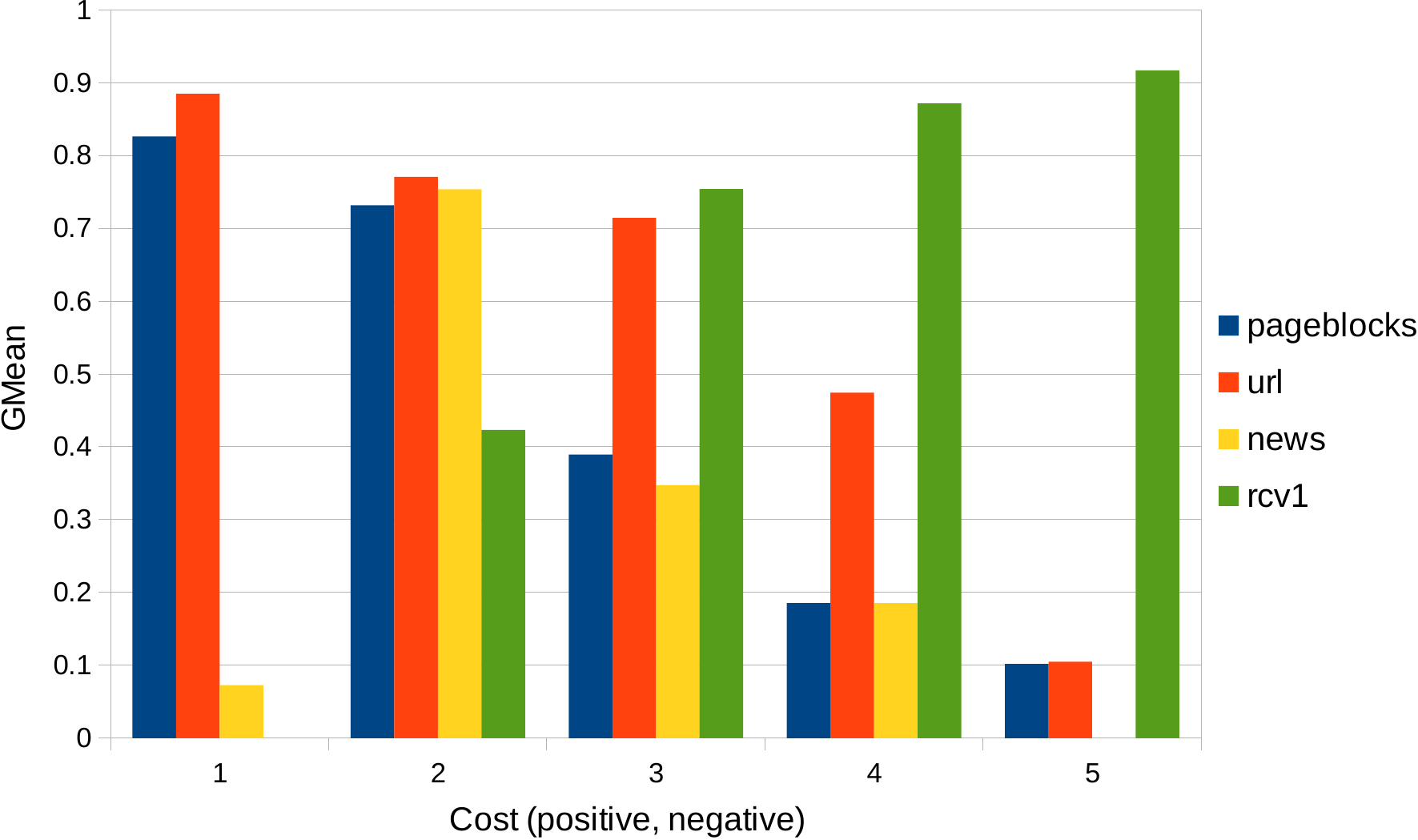}}
  \subfloat[]{ \includegraphics[width=8cm, height=5cm]{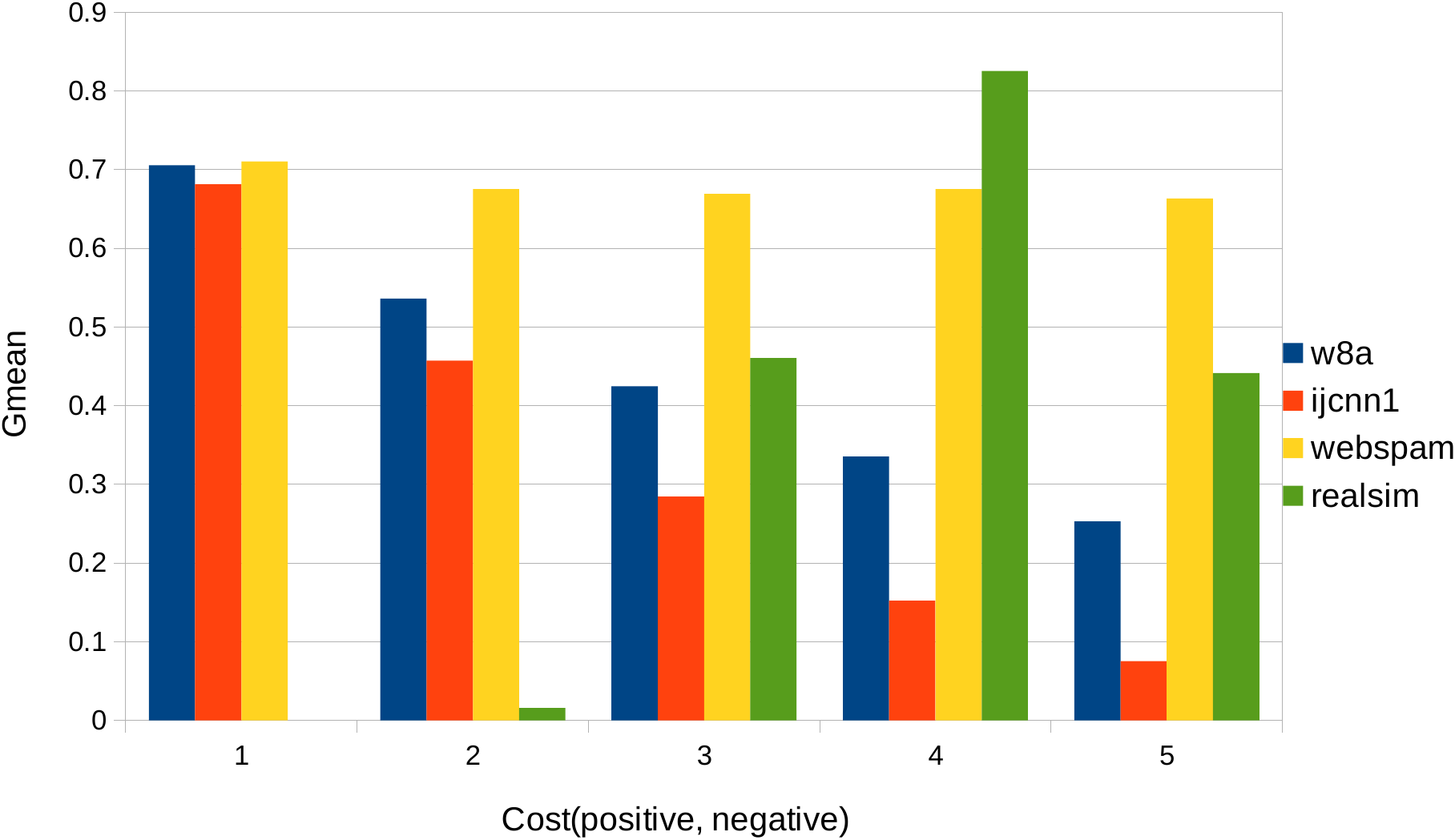}}
\caption{Gmean versus Cost over various data sets for L-DSCIL algorithm. Cost is given on the $x$-axis where each number denotes cost pair such that 1=\{0.1,0.9\}, 2=\{0.2,0.8\}, 3=\{0.3,0.7\}, 4=\{0.4,0.6\}, 5=\{0.5,0.5\}}
\label{lbfgs_cost}
\subfloat[]{ \includegraphics[width=8cm, height=5cm]{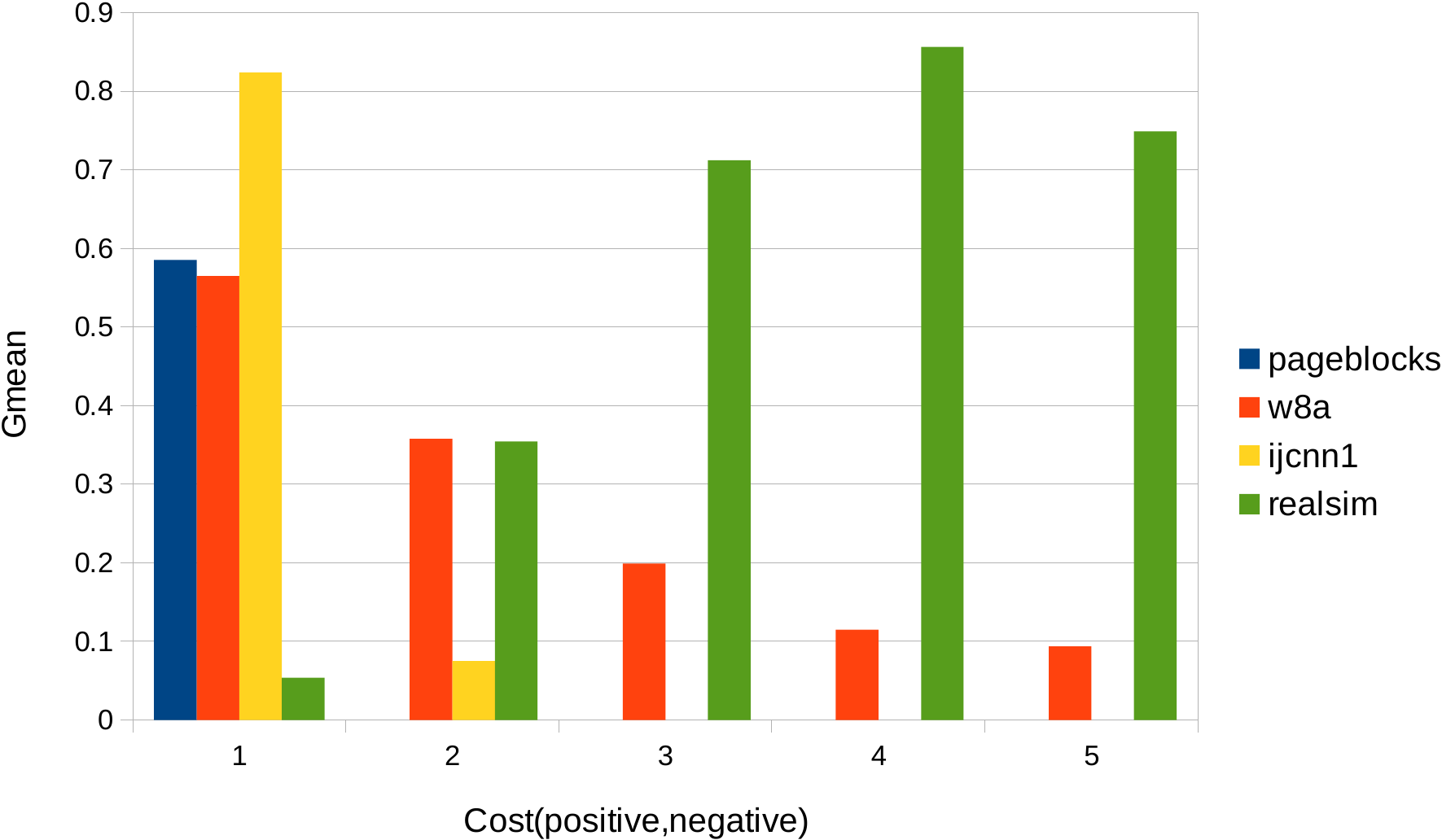}}
  \subfloat[]{ \includegraphics[width=8cm, height=5cm]{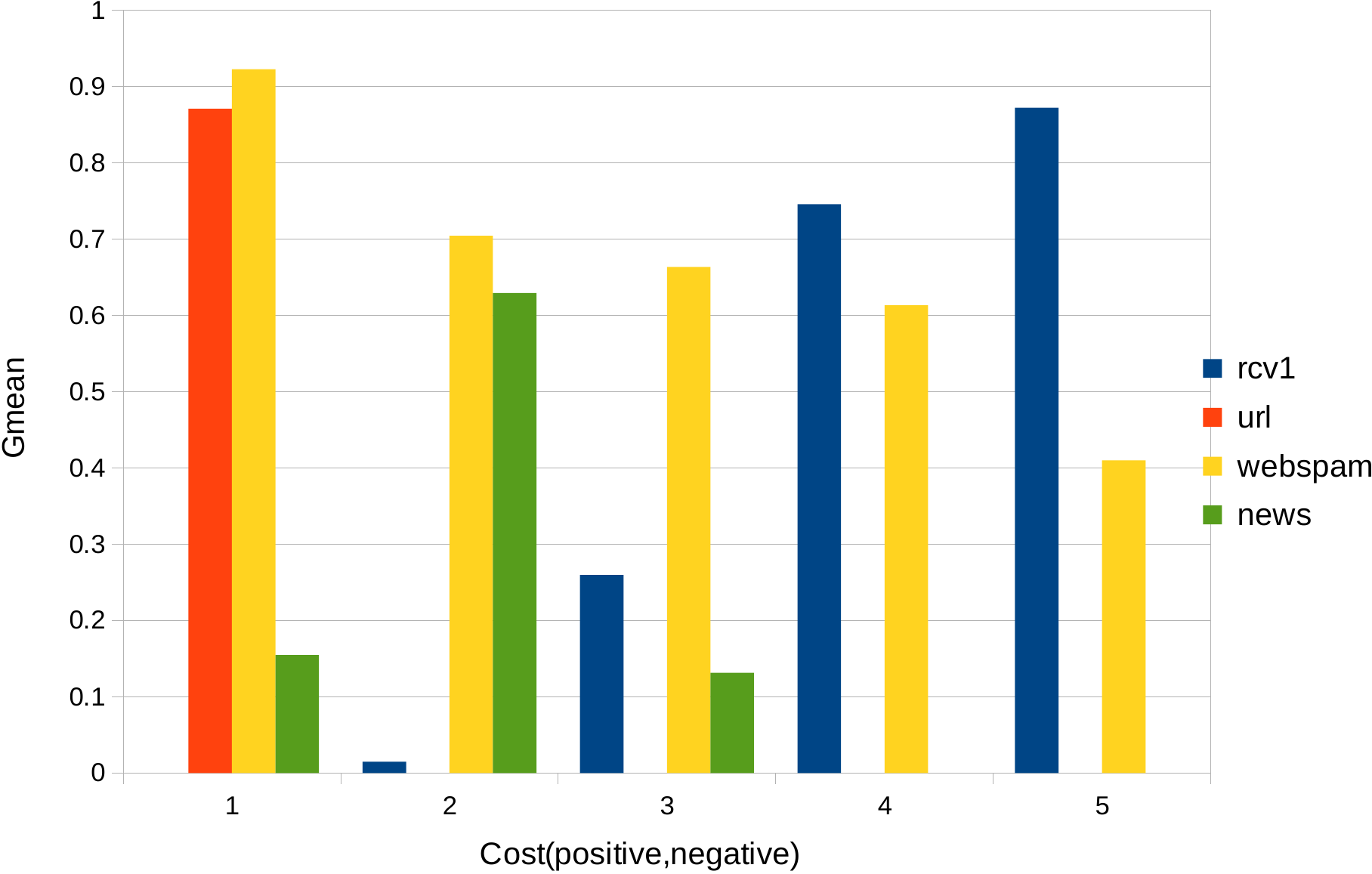}}
\caption{Gmean versus Cost over various data sets for R-DSCIL algorithm. Cost is given on the $x$-axis where each number denotes cost pair such that 1=\{0.1,0.9\}, 2=\{0.2,0.8\}, 3=\{0.3,0.7\}, 4=\{0.4,0.6\}, 5=\{0.5,0.5\}}
\label{rcd_cost}

\end{figure*}

\item {\bf Speedup Measurement}\\
In this Section, we discuss the speedup achieved by R-DSCIL and L-DSCIL algorithms when we run these algorithms on multiple cores.  Speedup results of R-DSCIL algorithm are presented in Figure \ref{rcd_speedup} (a) and (b) and of L-DSCIL in Figure \ref{lbfgs_speedup} (a) and (b).  The number of cores used is shown on the x-axis and the  y-axis shows the training time.  From these figures, we  can draw multiple conclusions. Firstly, from the Figure \ref{rcd_speedup} (a), we observe that as we increase the number of cores, training time is decreasing for all the  data sets except for webspam at 8 cores. The sudden increase in training time of webspam could be explained as follows:  so long as the computation time (RCD running time plus primal and dual variable update) remains above the communication time ($MPI\_Allreduce$ operation), adding more cores reduces the overall training time. On the other hand, in Figure \ref{rcd_speedup} (b), training time is first increasing and then decreasing for all the datasets with increasing number of cores. This could be due to the increasing communication time with the increasing number of cores. After a certain number of cores, computation time starts leading the  communication time and thereby decrease in the training time is observed. Further, in DSCIL algorithm, the communication time is data dependent (it depends on  the data dimensionality and sparsity). Because of these reasons, we observe different speedup patterns on different data sets. Speedup results of L-DSCIL in Figure \ref{lbfgs_speedup} (a) and (b) show the decreasing training time with increasing number of cores for all the data sets. We also observe that training time for L-DSCIL is higher than that of R-DSCIL on all data sets and cores.

\begin{figure*}
\centering
 \subfloat[]{ \includegraphics[width=8cm, height=5cm]{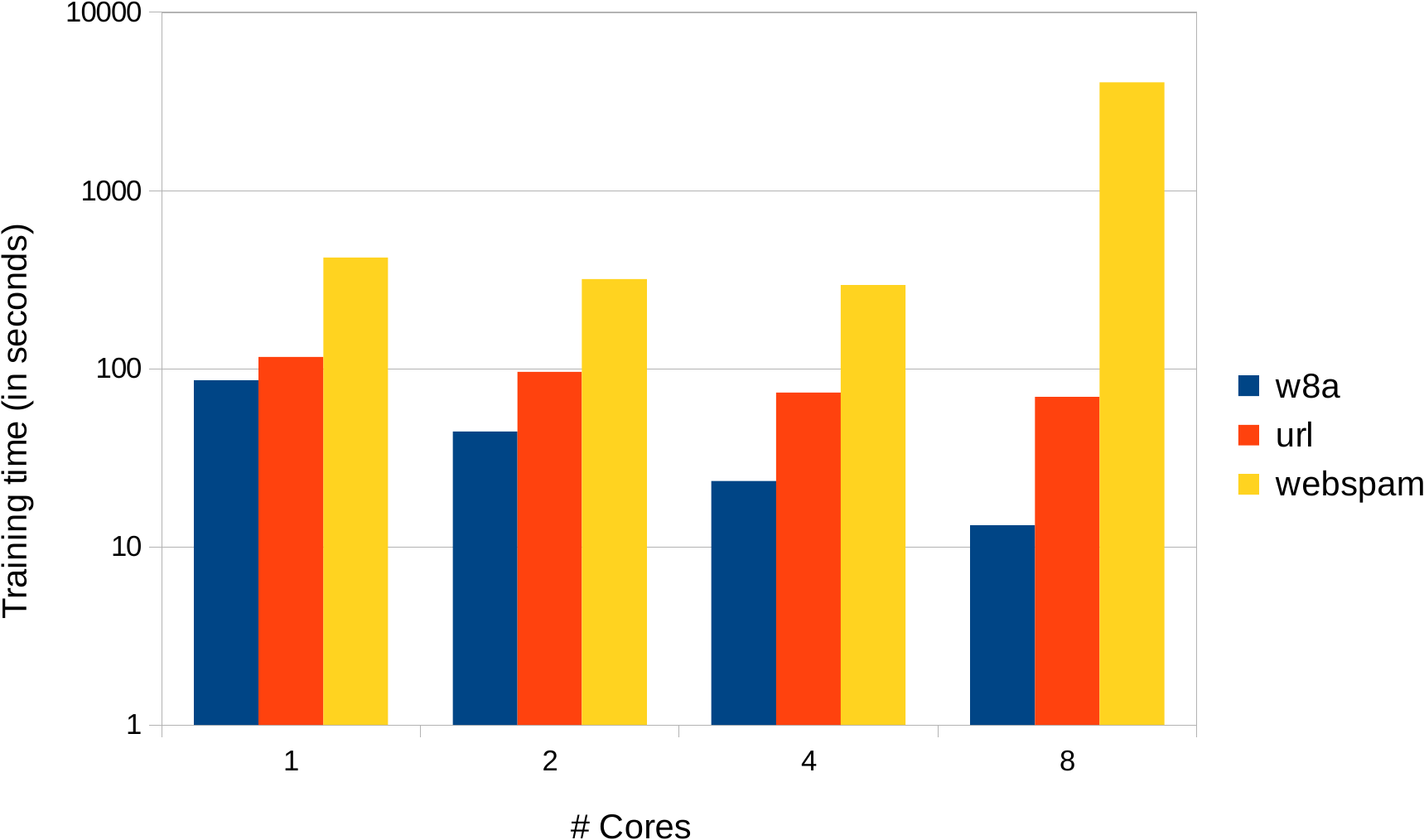}}
  \subfloat[]{ \includegraphics[width=8cm, height=5cm]{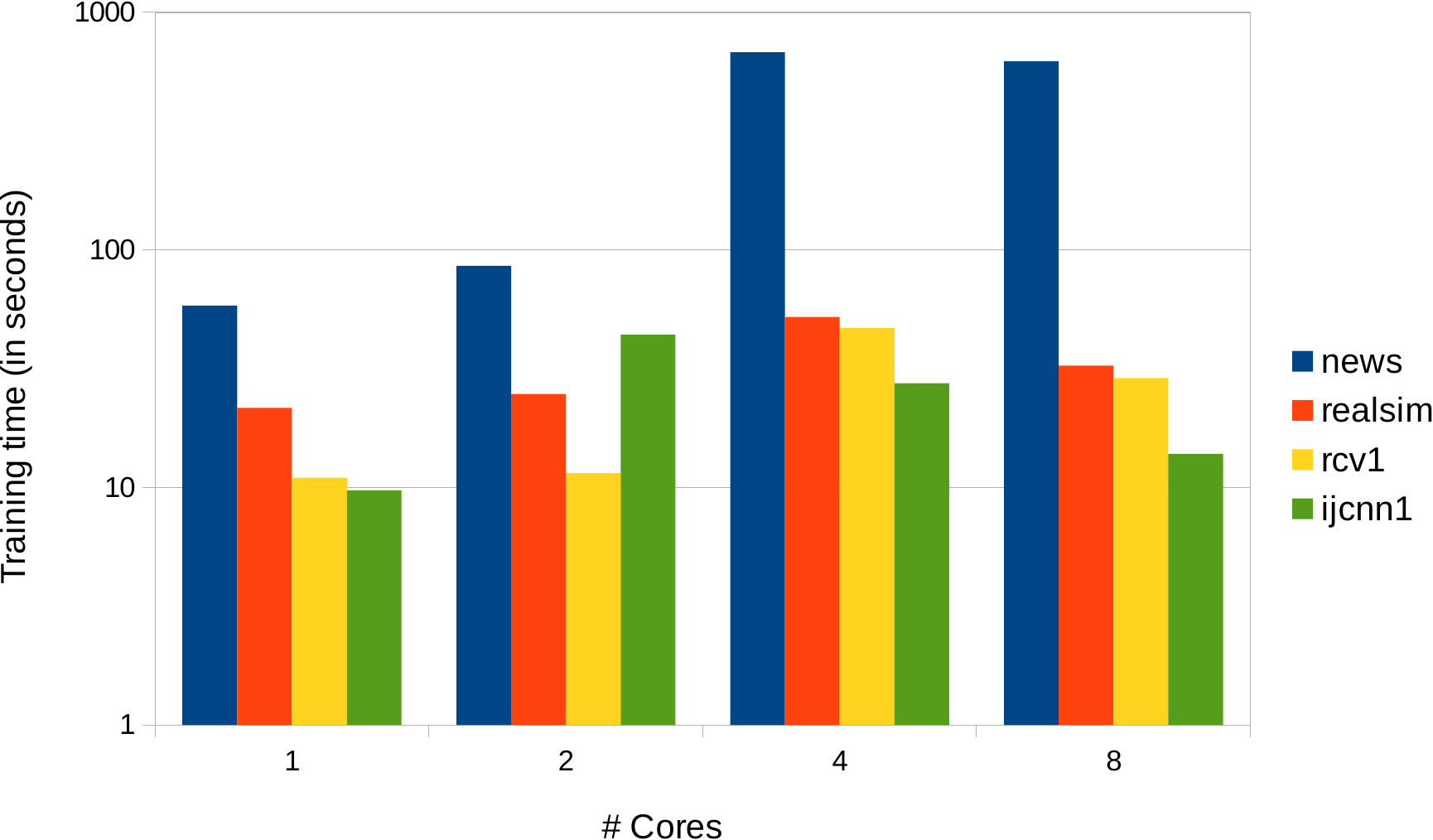}}
\caption{Training time versus number of cores to measure the speedup of R-DSCIL algorithm. Training time in Figure (a) is on the log scale.}
\label{rcd_speedup}
\subfloat[]{ \includegraphics[width=8cm, height=5cm]{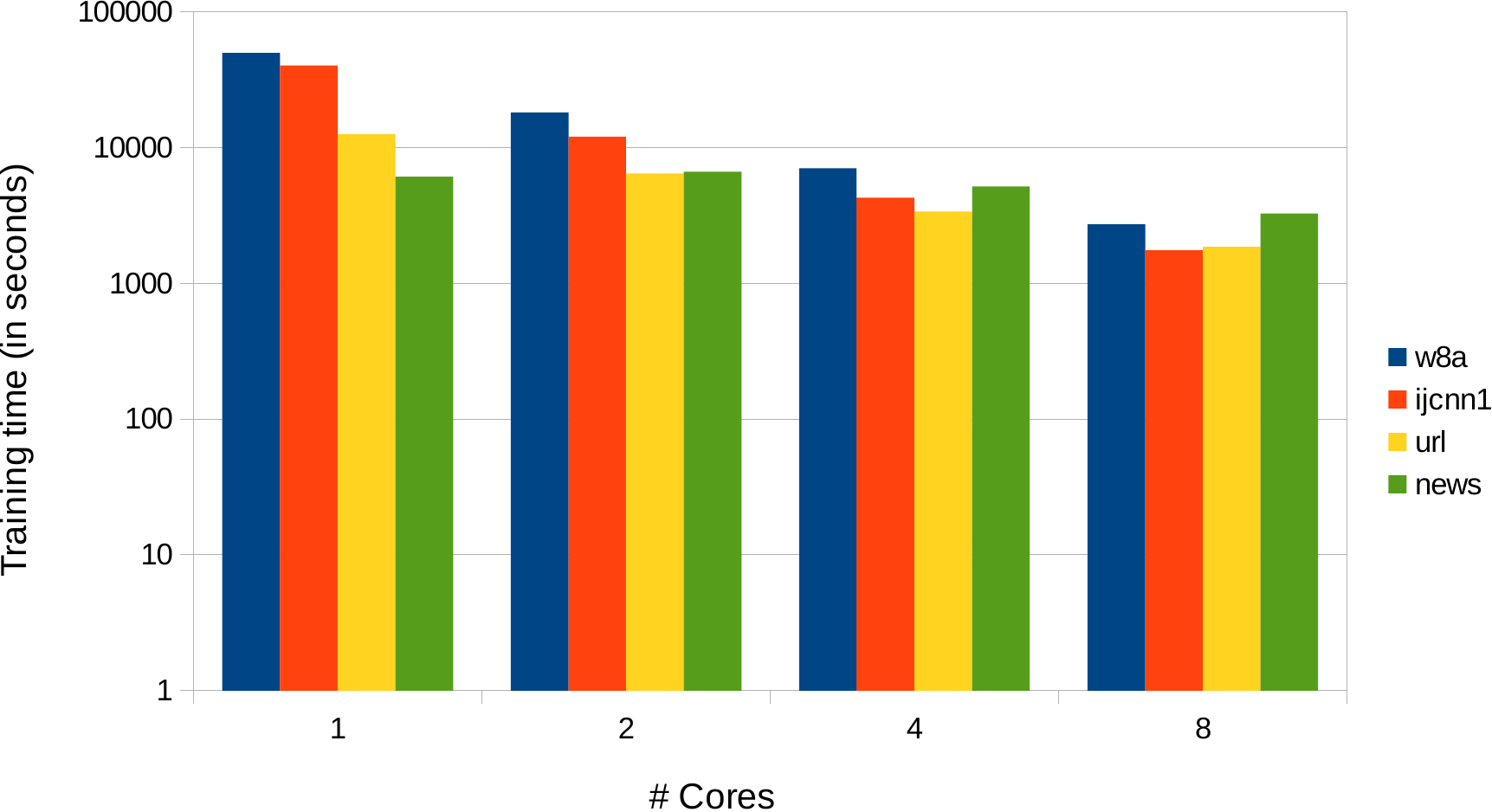}}
  \subfloat[]{ \includegraphics[width=8cm, height=5cm]{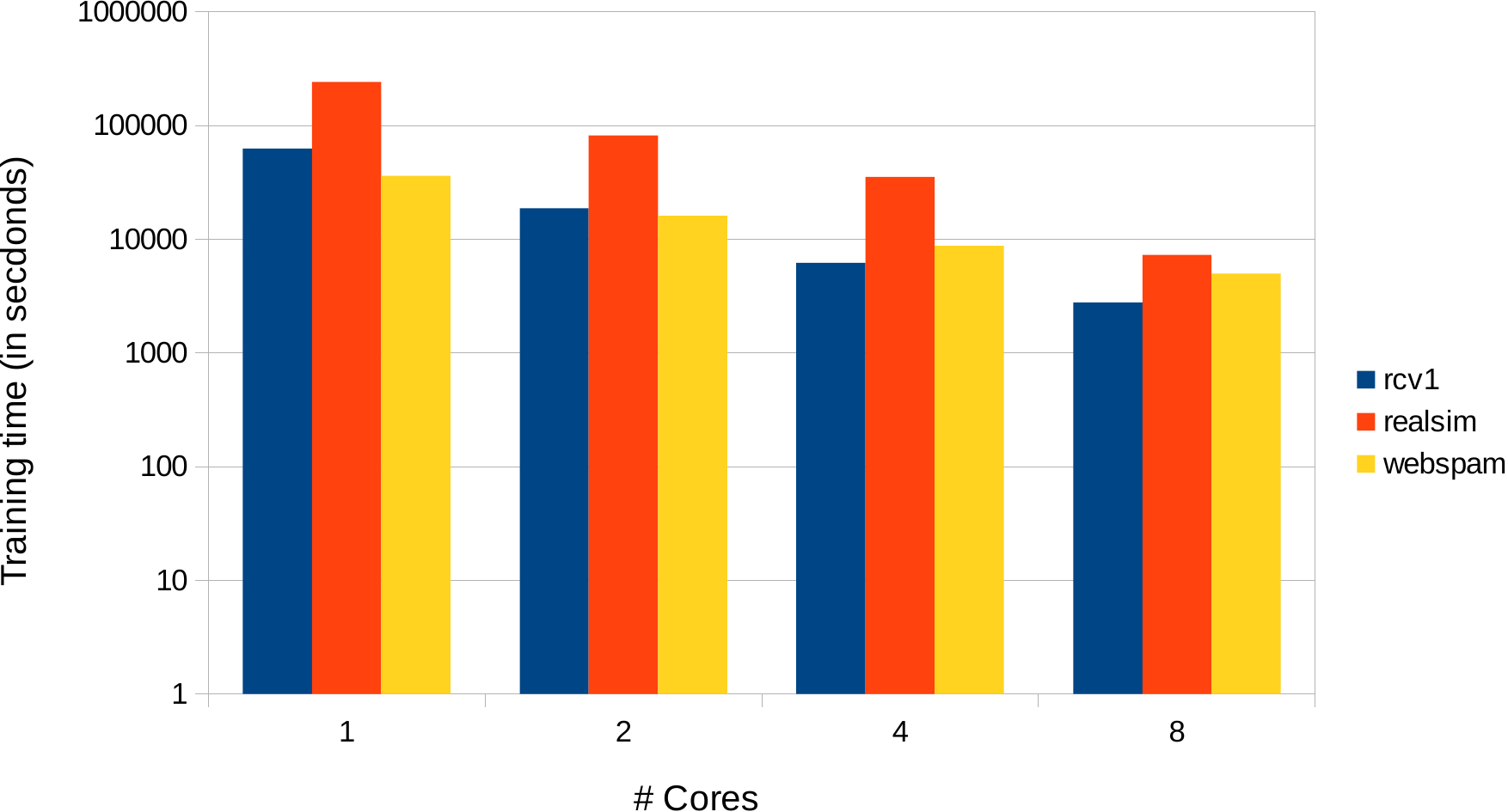}}
\caption{Training time versus number of cores to measure the speedup of L-DSCIL algorithm. Training time in both the figures is on the log scale.}
\label{lbfgs_speedup}
\end{figure*}

\item {\bf Number of Cores versus  Gmean }\\
In this Section, we present the experimental results showing the effect of utilizing a varying number of cores on \emph{Gmean}. This also shows how the different partitioning of the data affects the \emph{Gmean}. We divide the data set into equal chunks and distribute it to various cores. Suppose, we have a data set of size $m$ and want to utilize $n$ cores, we allot samples of size $m/n$ to each core. We choose the data size such that it is divisible by all possible cores utilized. 
 Results are shown in Figures \ref{gmeanvscores1} and \ref{gmeanvscores2} for R-DSCIL and L-DSCIL respectively. 
 In Figure \ref{gmeanvscores1} (a) and (b), we can observe that \emph{Gmean} remains almost constant over various partitioning of the data (various cores). A small deviation is observed over url and w8a data sets in Figure \ref{gmeanvscores1} (b). These observations lead to the conclusion that R-DSCIL algorithm is less sensitive to the data partition.   On the other hand, \emph{Gmean} 
\newpage
\begin{figure*}
\centering
 \subfloat[]{ \includegraphics[width=8cm, height=5cm]{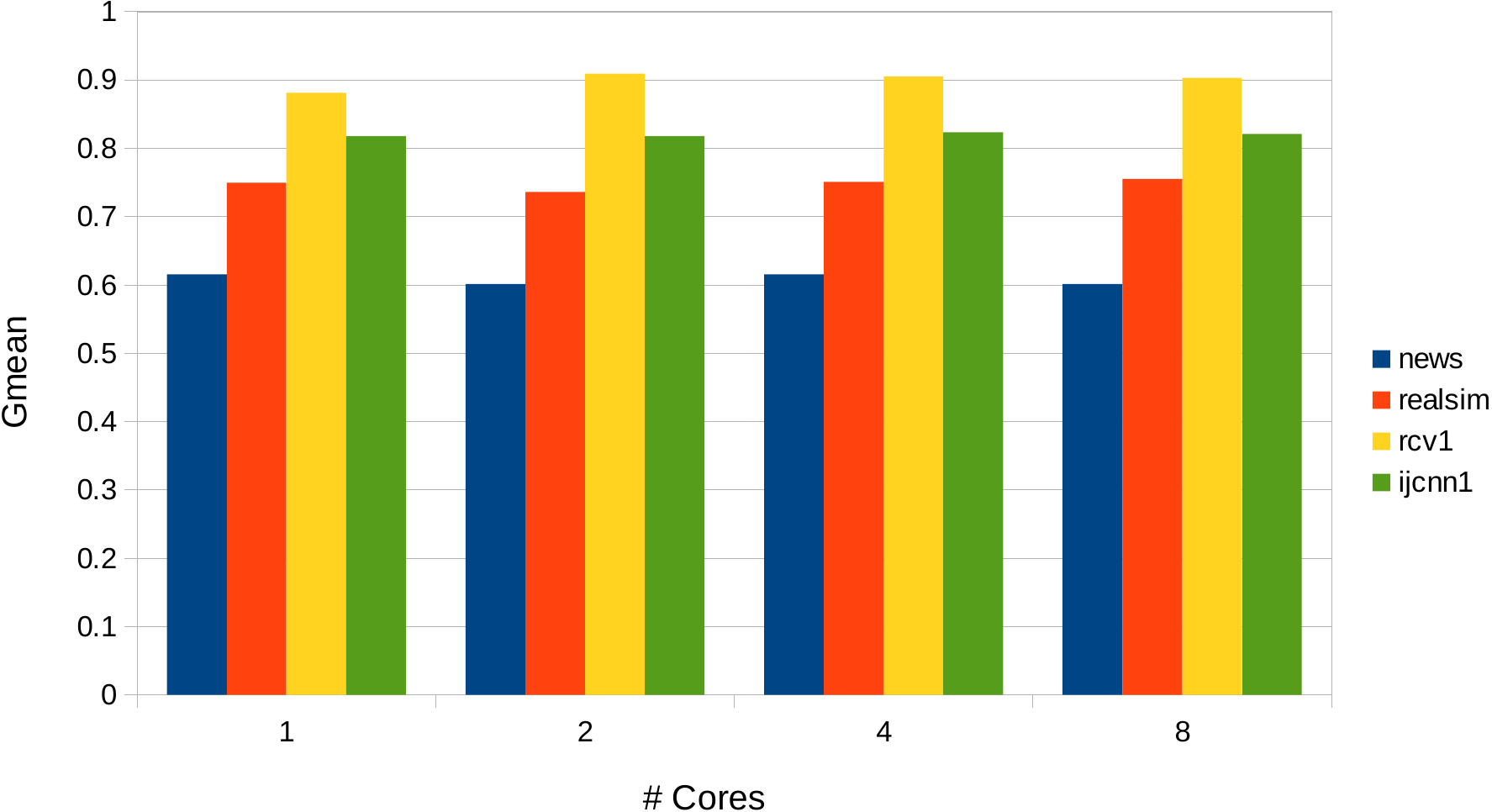}}
  \subfloat[]{ \includegraphics[width=8cm, height=5cm]{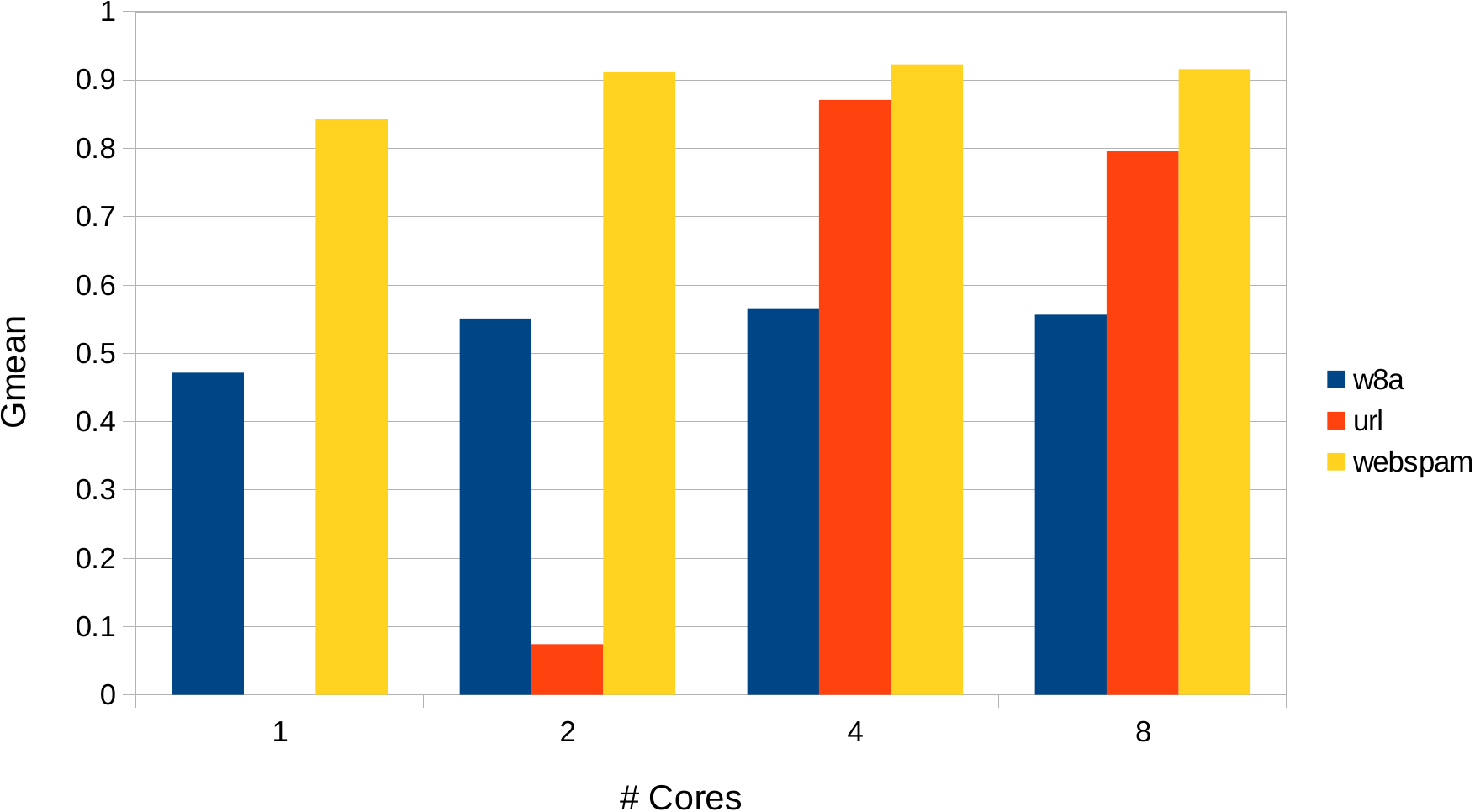}}
\caption{Effect of varying number of cores on \emph{Gmean}  in R-DSCIL algorithm.}
\label{gmeanvscores1}
 \subfloat[]{ \includegraphics[width=8cm, height=5cm]{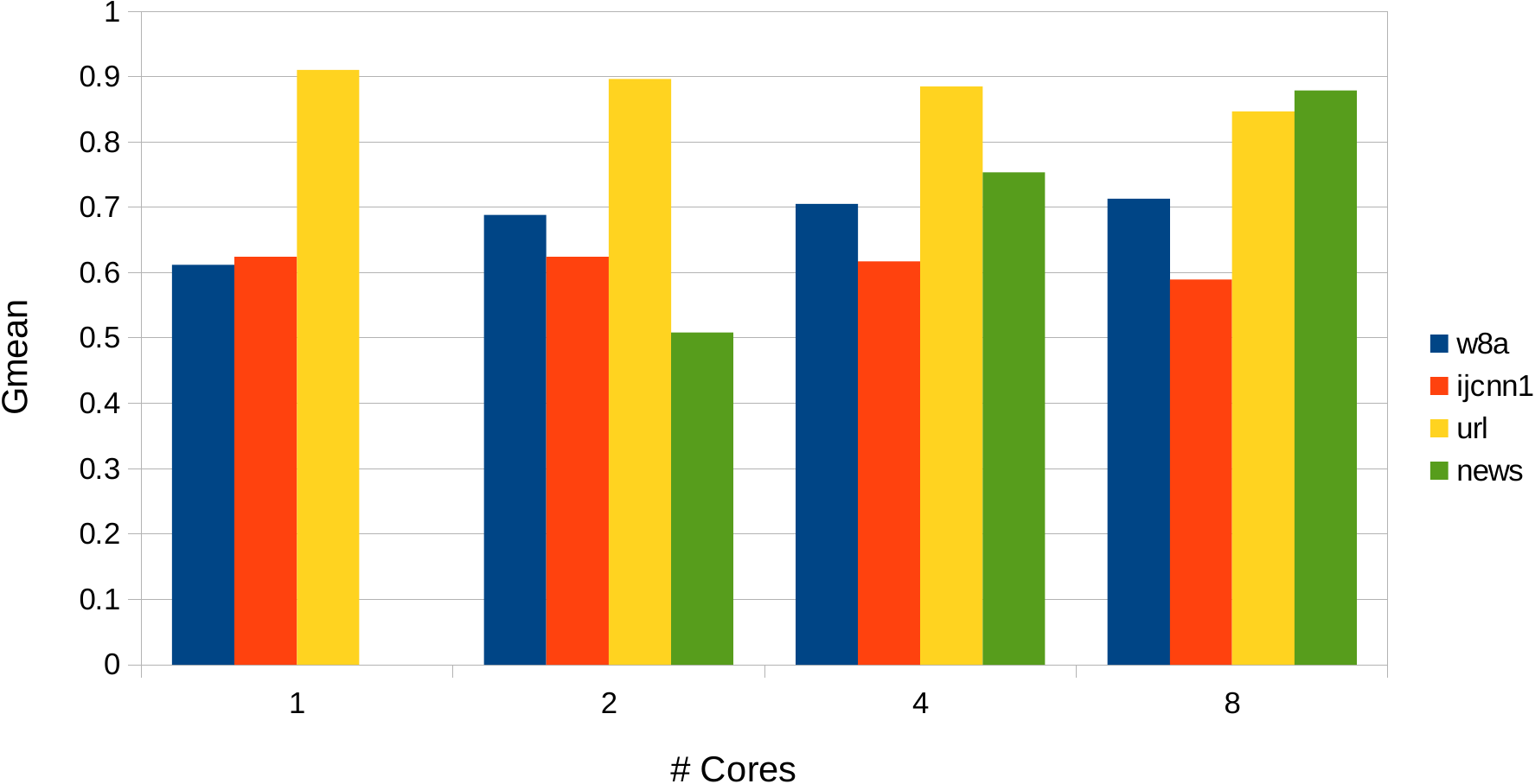}}
  \subfloat[]{ \includegraphics[width=8cm, height=5cm]{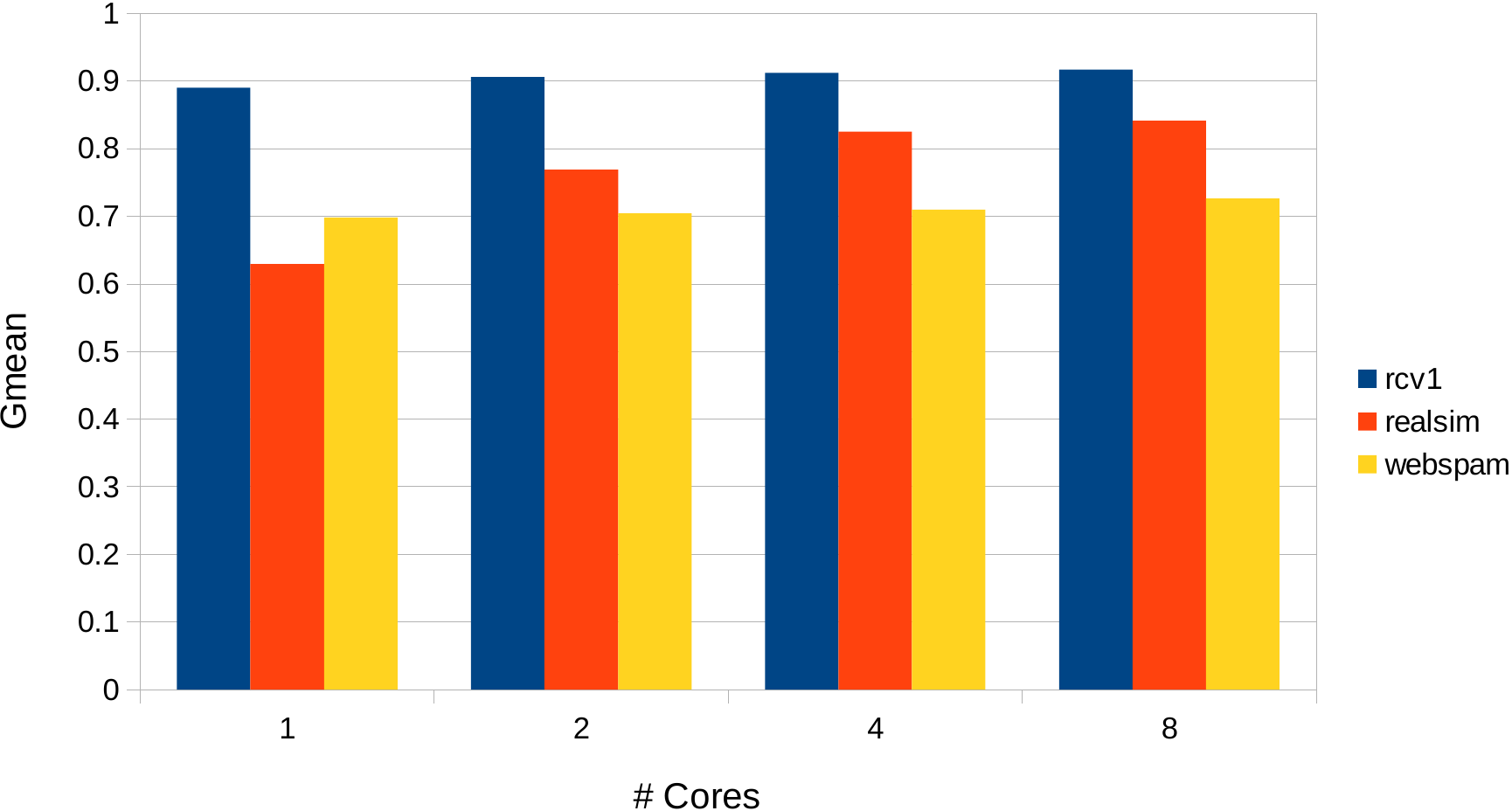}}
\caption{Effect of varying number of cores on \emph{Gmean}  in L-DSCIL algorithm.}
\label{gmeanvscores2}
\end{figure*}
\clearpage
\newpage
results  of the L-DSCIL algorithm for various partitions  in Figures \ref{gmeanvscores2} (a) and (b) show the chaotic behavior. For example,  \emph{Gmean}  changes a lot with the increasing number of cores for news in Figure \ref{gmeanvscores2} (a) and for realsim in  Figure \ref{gmeanvscores2} (b). For other data sets, fluctuations in \emph{Gmean} values are less sensitive.
\item {\bf Effect of Regularization Parameter on Gmean}\\
In this subsection, we discuss the effect of the regularization parameter, $\lambda$, on  \emph{Gmean} produced by the R-DSCIL algorithm as shown in Figure \ref{fig:gmean} on various benchmark data sets. It is clear from the Figure \ref{fig:gmean} that \emph{Gmean} is dropping with increasing regularization parameter on all the data sets tested.  On some data sets such as \emph{rcv1}, \emph{Gmean} drops gradually with increasing $\lambda$. On the other side, some data sets such as \emph{w8a, pageblocks}, \emph{Gmean} falls off to 0 quickly with increasing $\lambda$.   It also shows that higher the sparsity in the data sets such as \emph{rcv1, webspam} etc, higher is the penalty  required to achieve higher \emph{Gmean} and vice versa.

\begin{figure*}
\centering
\subfloat[ijcnn1]{\includegraphics[width=5.5cm, height=5cm]{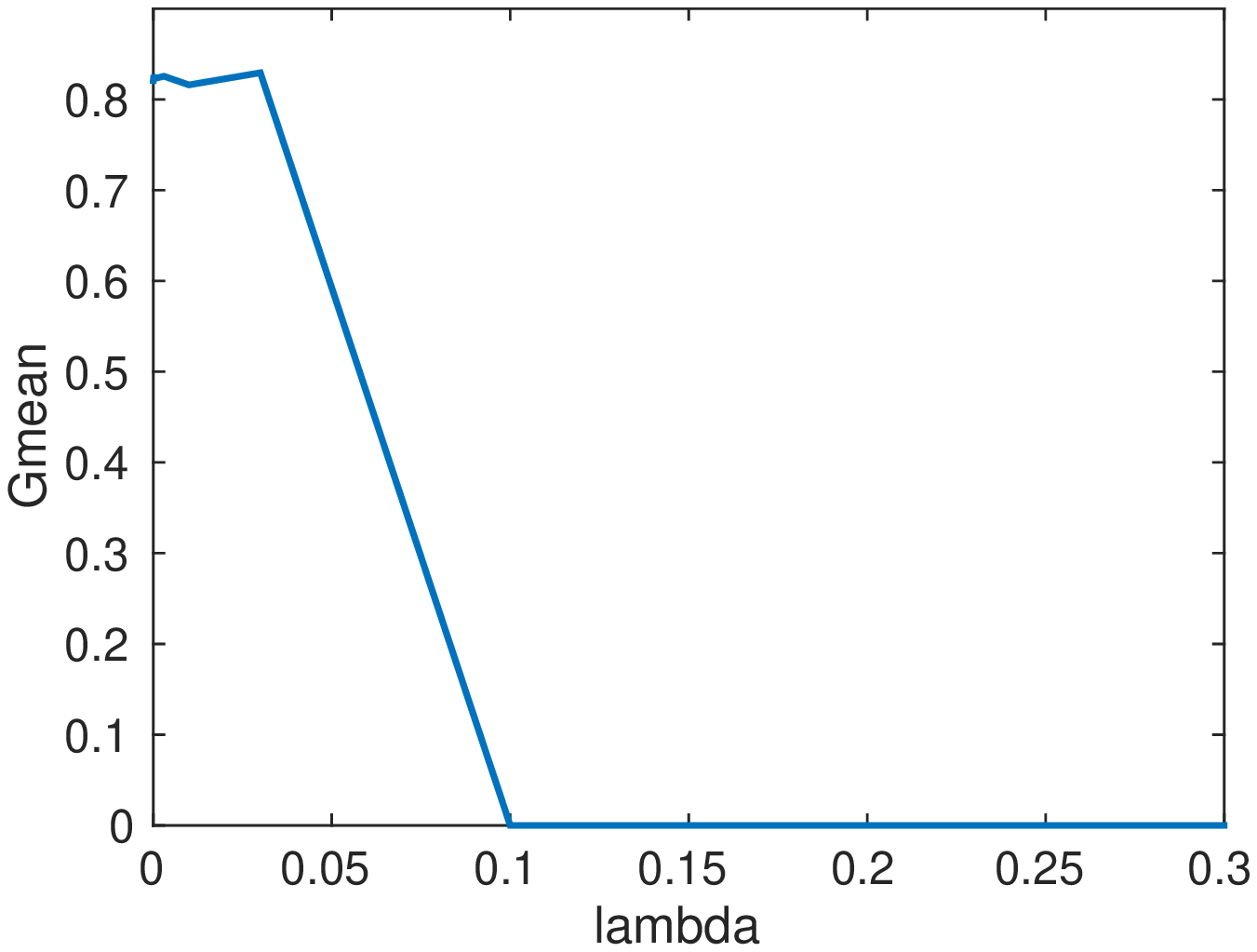}}  
\subfloat[rcv1]{\includegraphics[width=5.5cm, height=5cm]{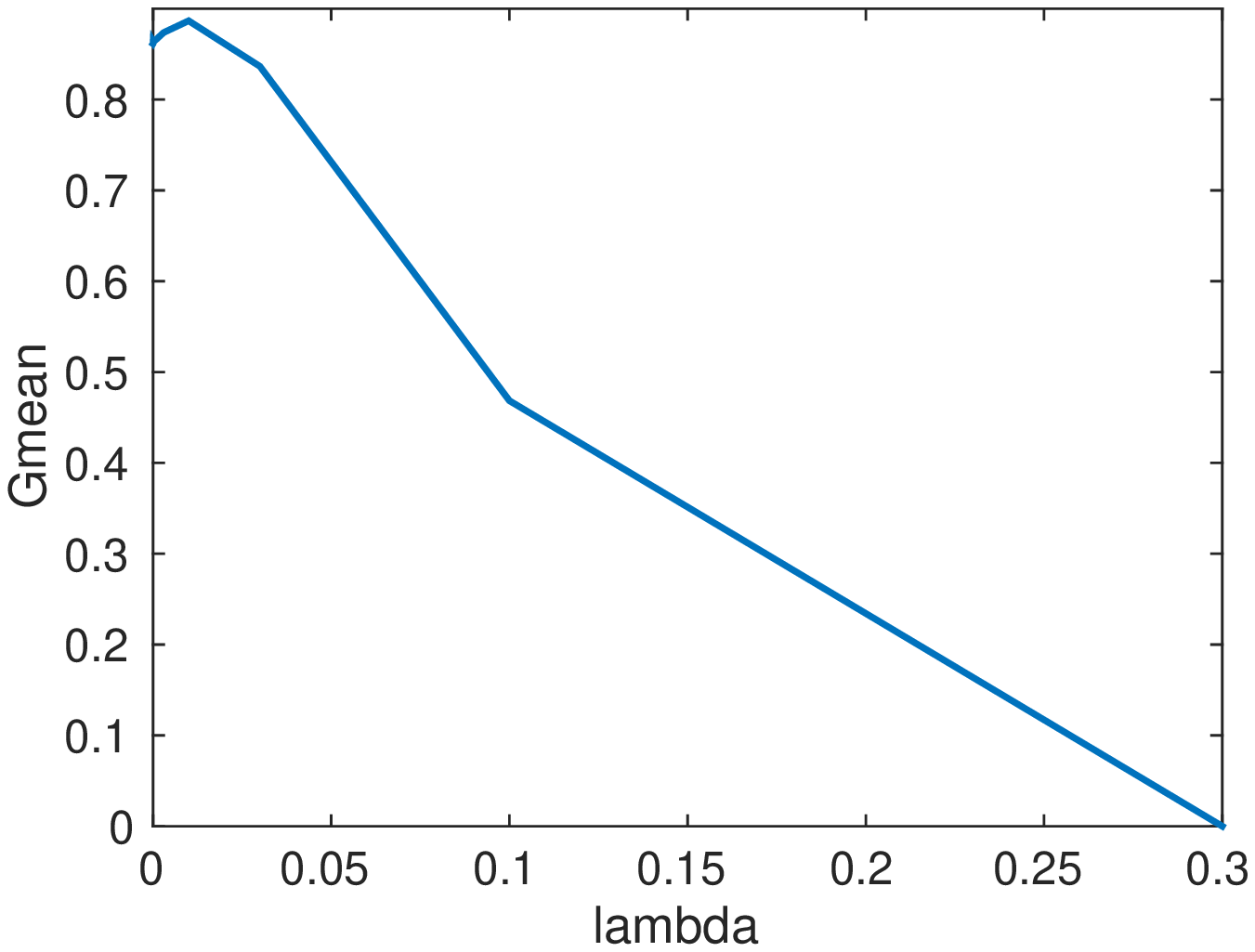}} 
\subfloat[pageblocks]{\includegraphics[width=5.5cm, height=5cm]{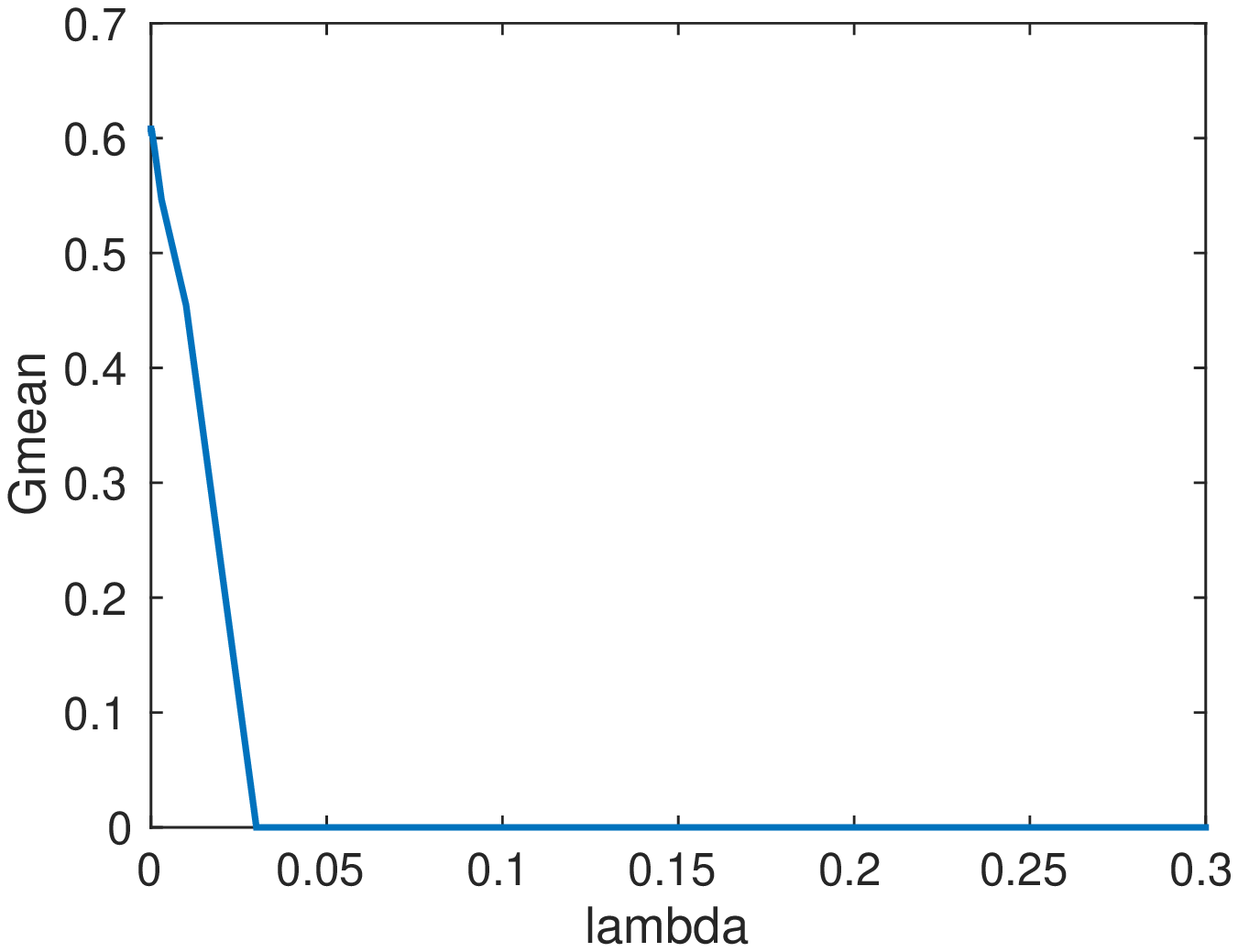}} \\
\subfloat[w8a]{\includegraphics[width=5.5cm, height=5cm]{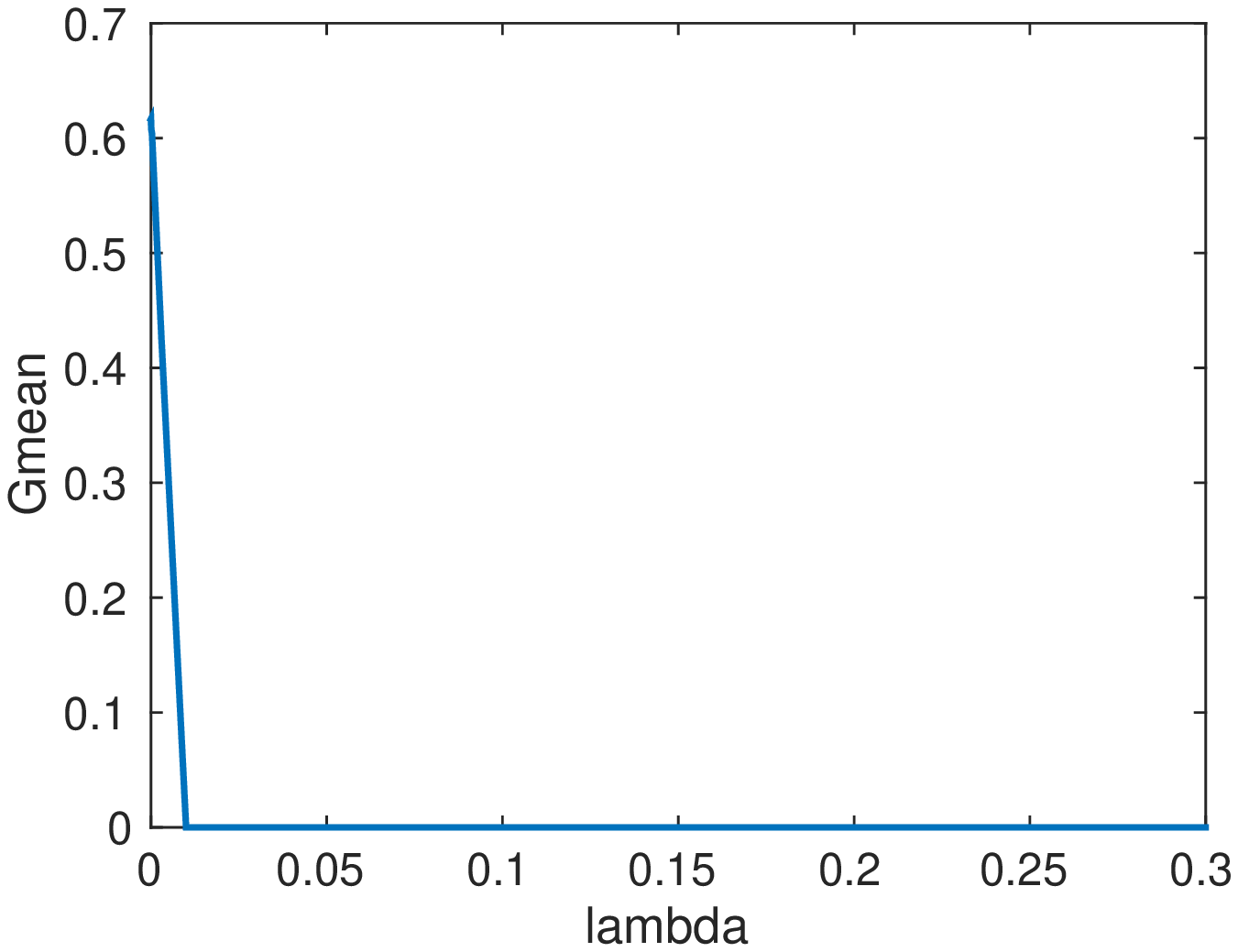}}
\subfloat[news]{\includegraphics[width=5.5cm, height=5cm]{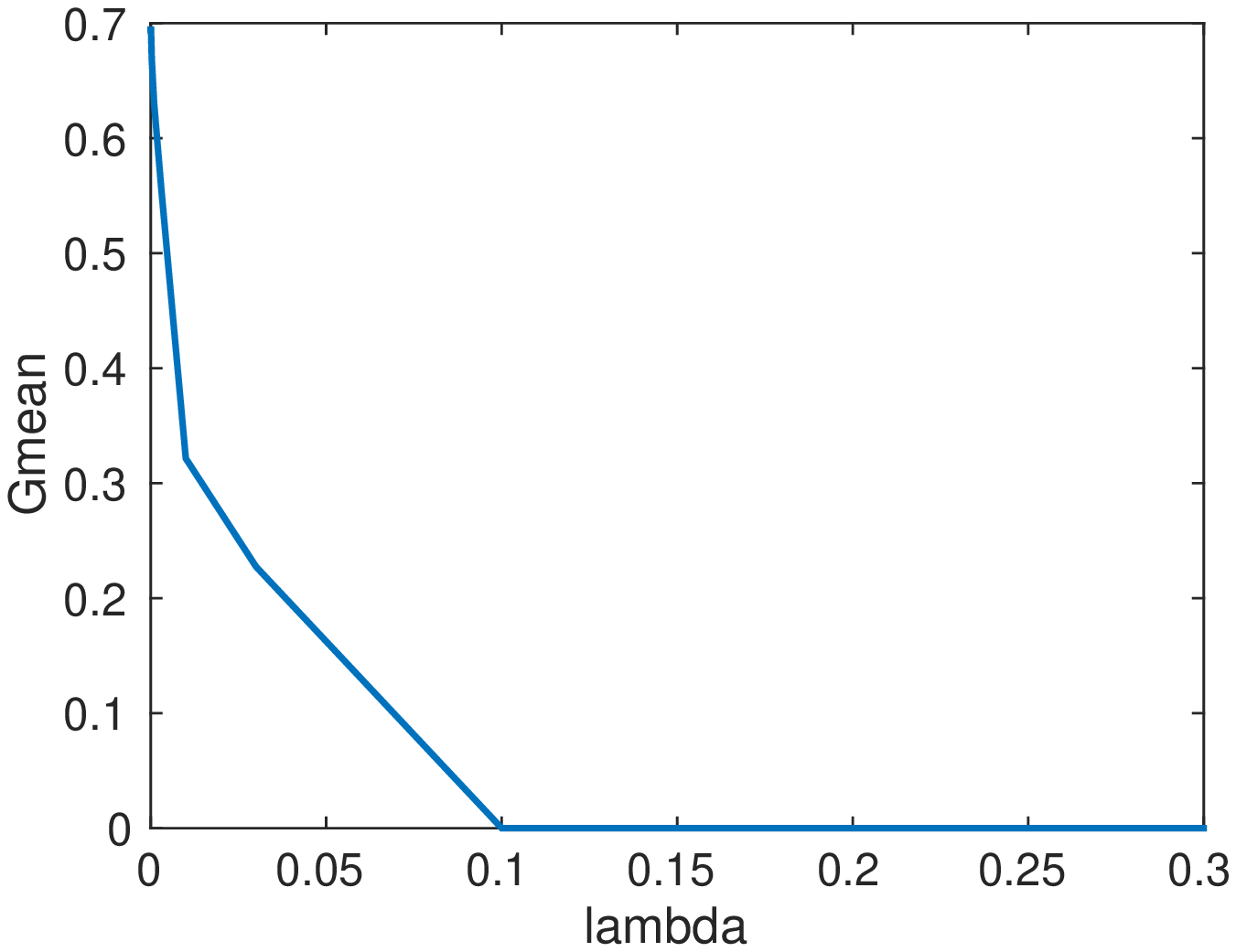}}  
\subfloat[url]{\includegraphics[width=5.5cm, height=5cm]{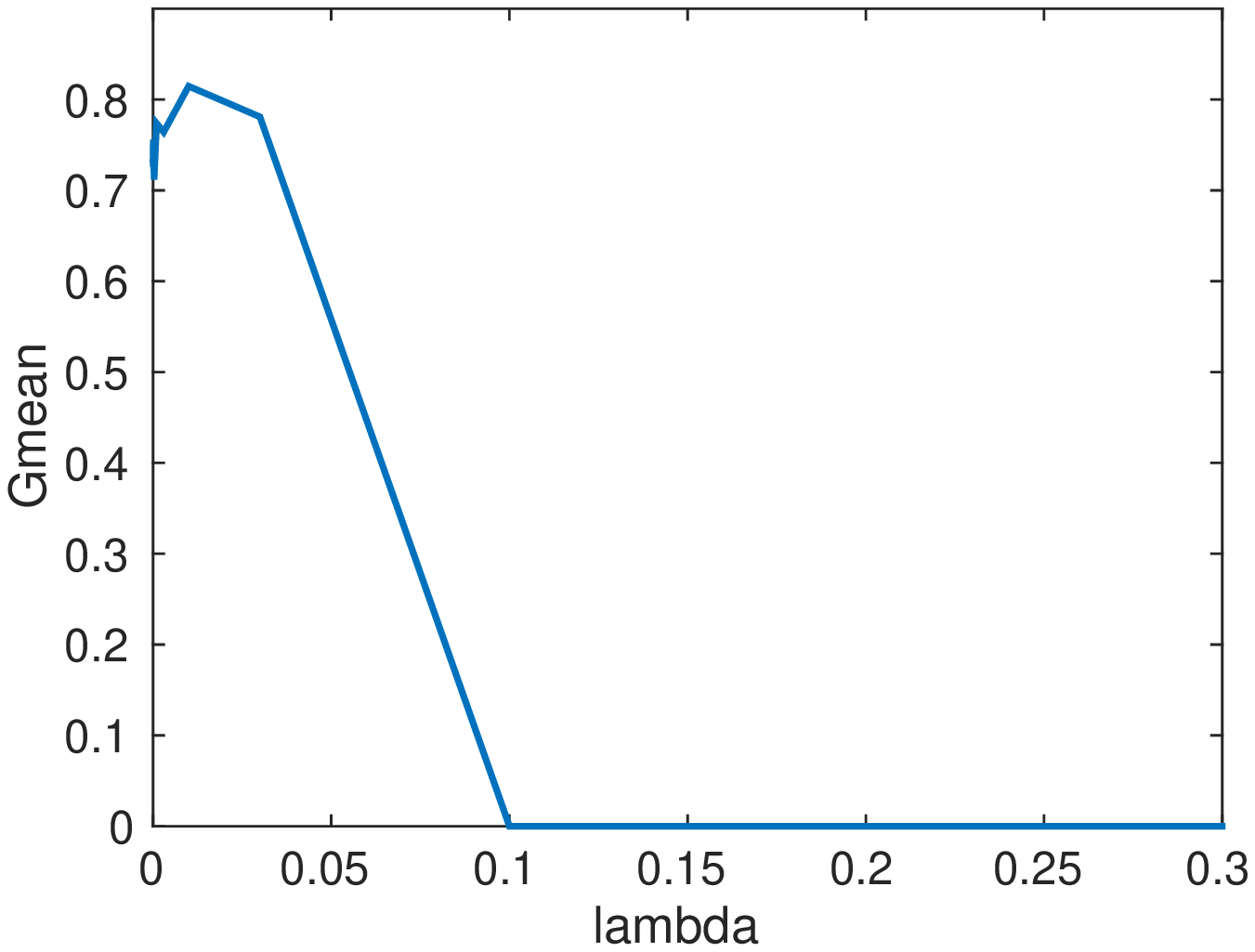}} \\
\subfloat[realsim]{\includegraphics[width=5.5cm, height=5cm]{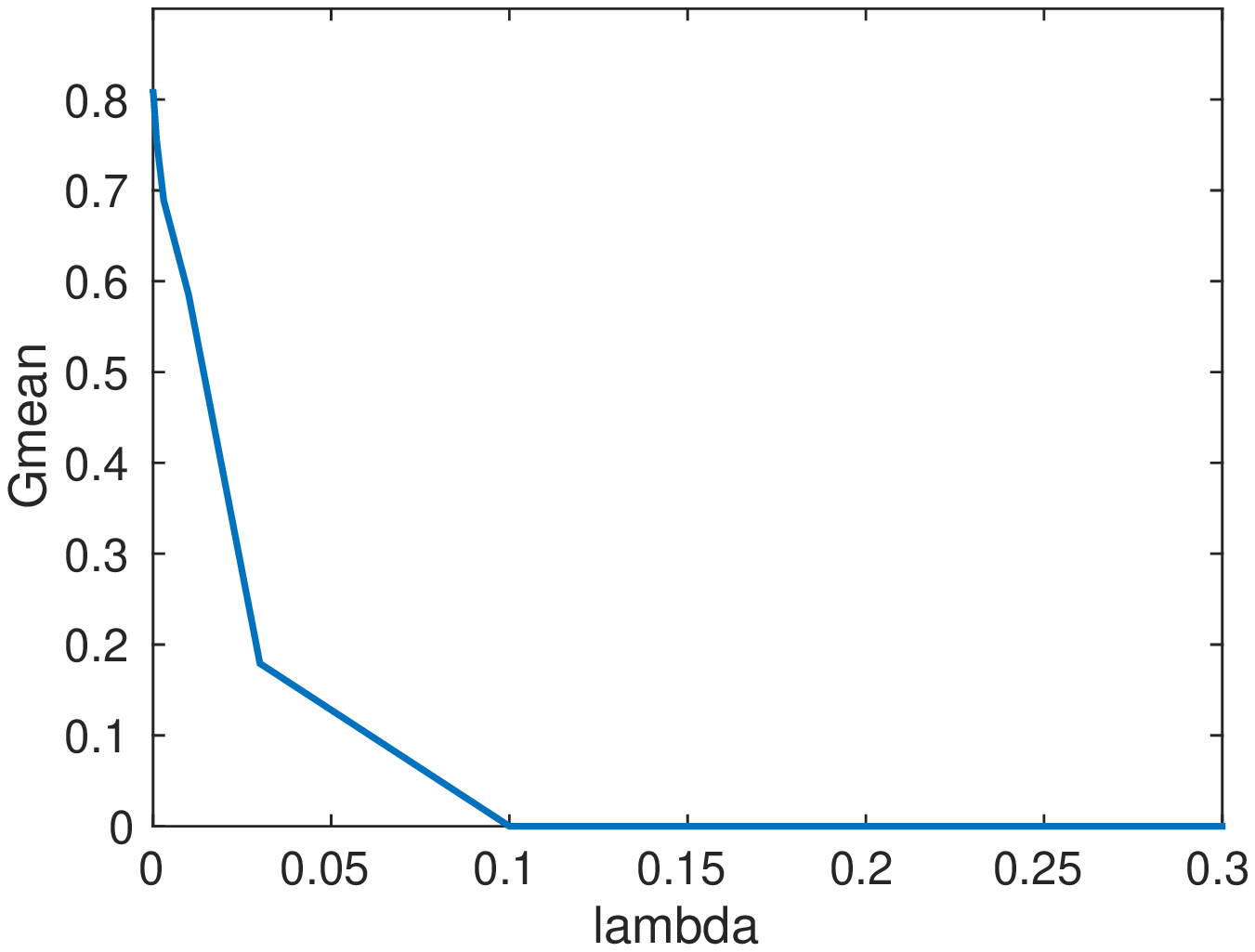}} 
\subfloat[webspam]{\includegraphics[width=5.5cm, height=5cm]{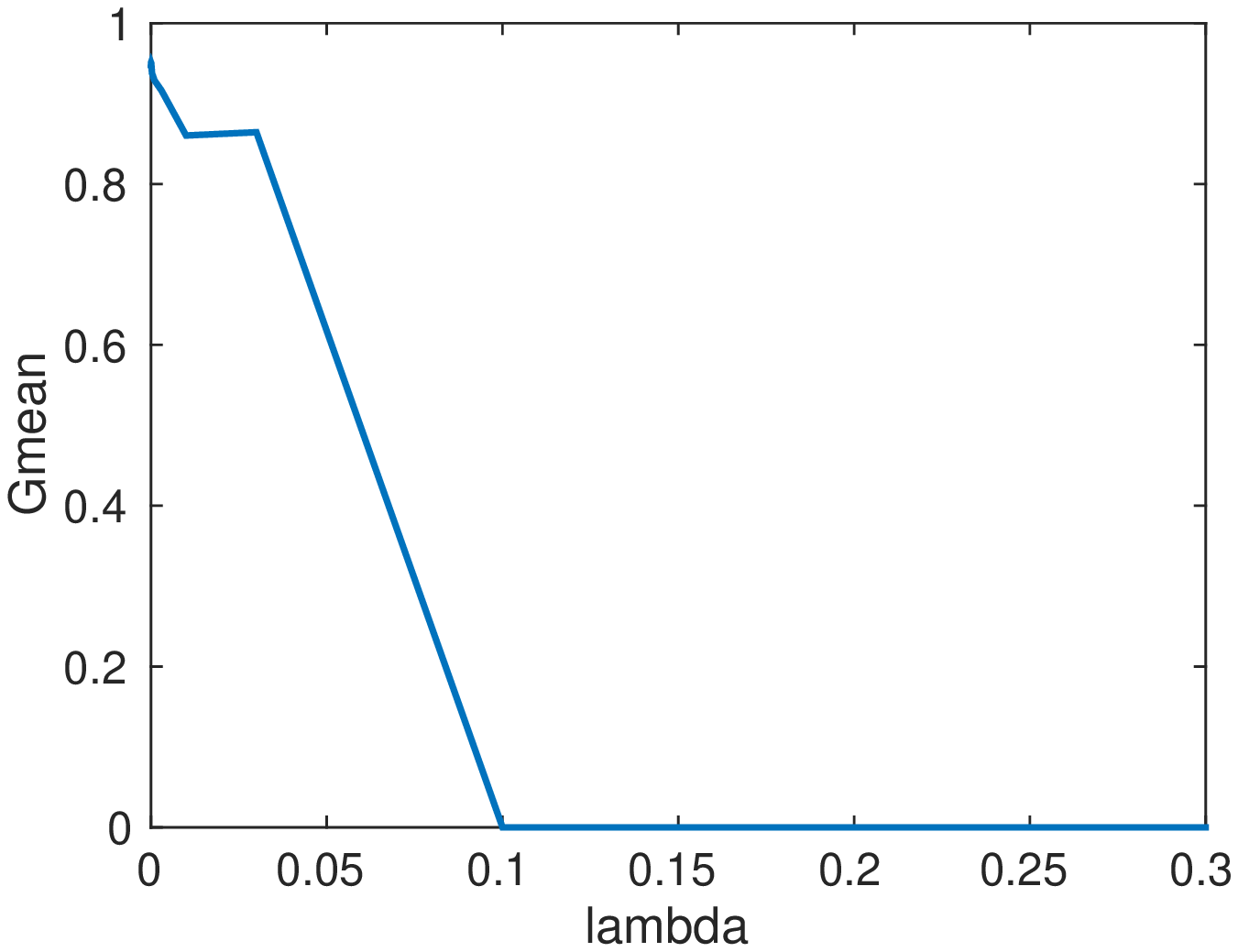}}
\caption{Gmean versus regularization parameter $\lambda$ using R-DSCIL (a) ijcnn1 (b) rcv1 (c) pageblocks (d) w8a  (e) news (f) url (g) realsim (h) webspam.}
\label{fig:gmean}
\end{figure*}

\end{enumerate}

\section{Experiments}
Below, we present empirical simulation results on various benchmark data sets as given in Table \ref{data}.
\subsection{Experimental Testbed and Setup} \label{testbed}
For our distributed implementation, we used MPICH2 library\cite{Forum:1994:MMI:898758}. We compare the performance of CILSD algorithm with that of the CSFSOL of \cite{Dayong:2015}. In the forthcoming subsections, we will show (i) The convergence of CILSD on benchmark data sets (ii) The performance comparison of CILSD, CSFSOL and CSSCD algorithms in terms of \emph{accuracy, sensitivity, specificity, Gmean} and \emph{ balanced\_accuracy} (also called $Sum$ which is =0.5*sensitivity+0.5*specificity) (iii) Speedup (iv) \emph{Gmean} versus number of cores (v) \emph{Gmean} versus regularization parameter. 

\subsection{Convergence of CILSD}
In this Section, we show the convergence of CILSD in two scenarios. In the first scenario, the convergence of CILSD is shown when we search for the best learning rate over a validation set in the range  $\{ 0.0003, 0.001,0.003, 0.01,0.03, 0.1, 0.3\}$. In the second scenario, we use the learning rate $1/L$. The convergence plot in both the scenario is shown in Figure \ref{conv}. We can clearly see that objective function converges faster with  learning rate set to $1/L$ (Obj2) than to search it over the range of possible learning rate (Obj1). These results show the correctness of our implementation as well as how to choose the learning rate.

\begin{figure*}
\centering
\subfloat[ijcnn1]{\includegraphics[width=8cm, height=6cm]{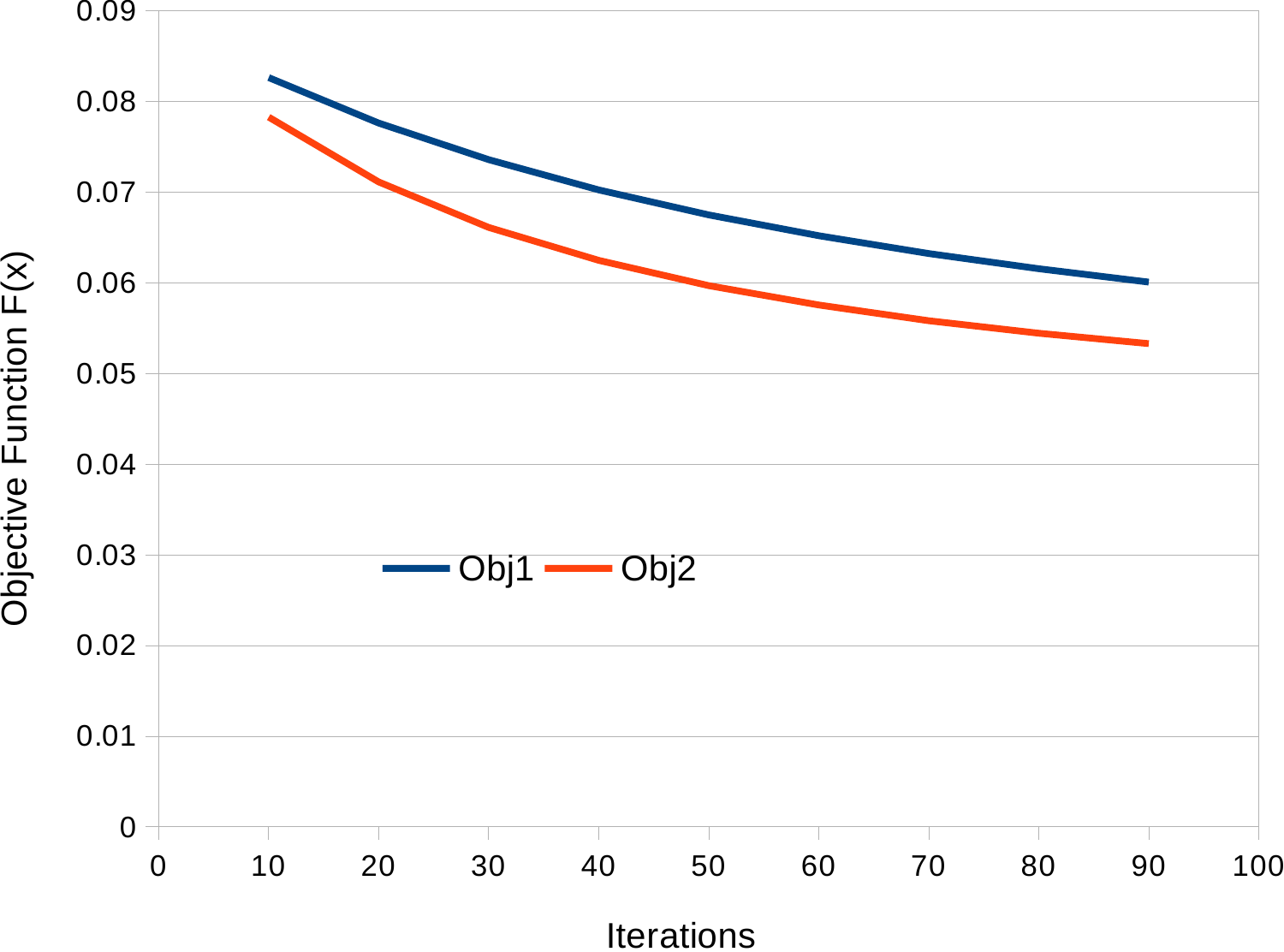}}  
\subfloat[rcv1]{\includegraphics[width=8cm, height=6cm]{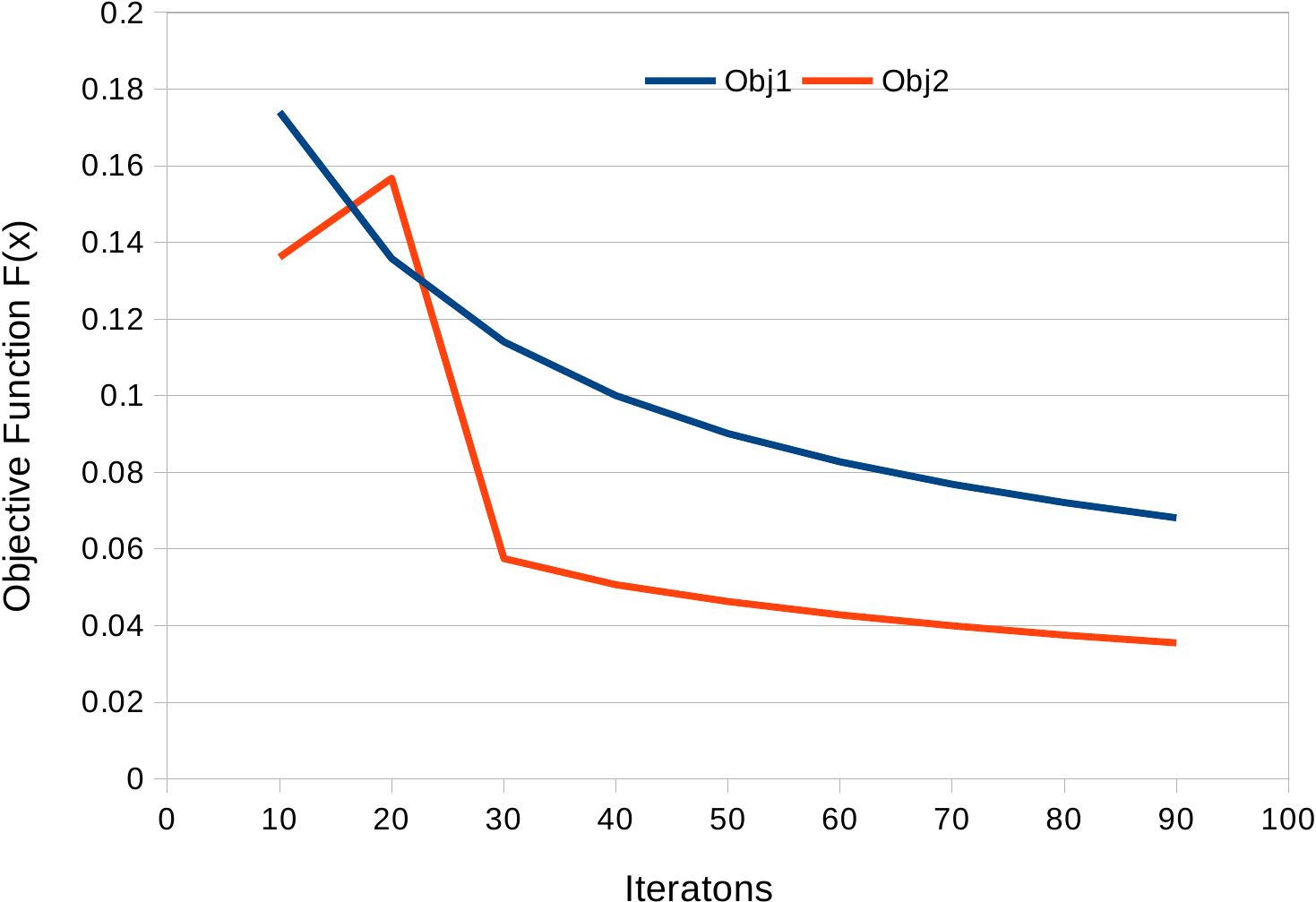}} \\
\subfloat[pageblocks]{\includegraphics[width=8cm, height=6cm]{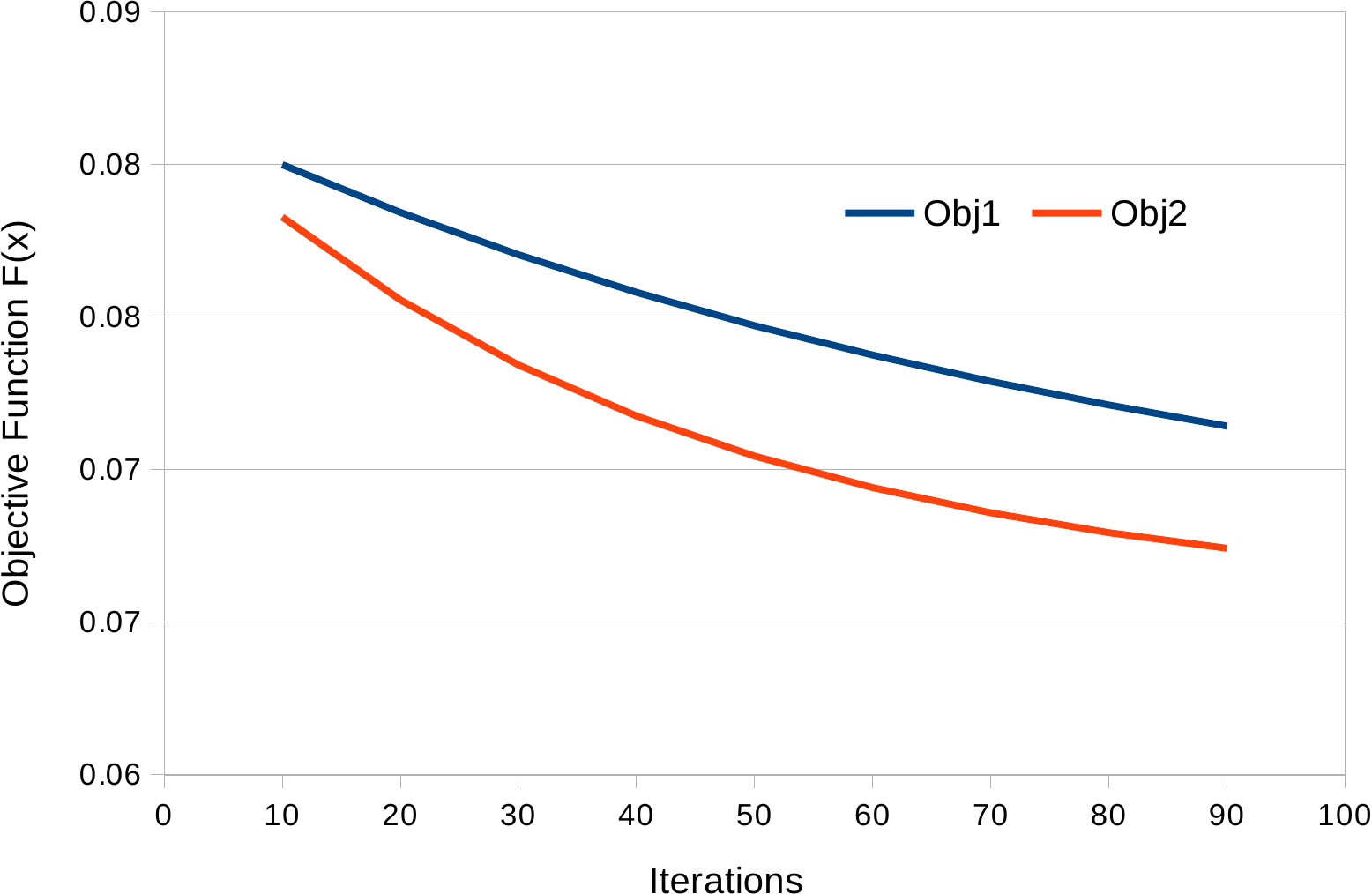}} 
\subfloat[w8a]{\includegraphics[width=8cm, height=6cm]{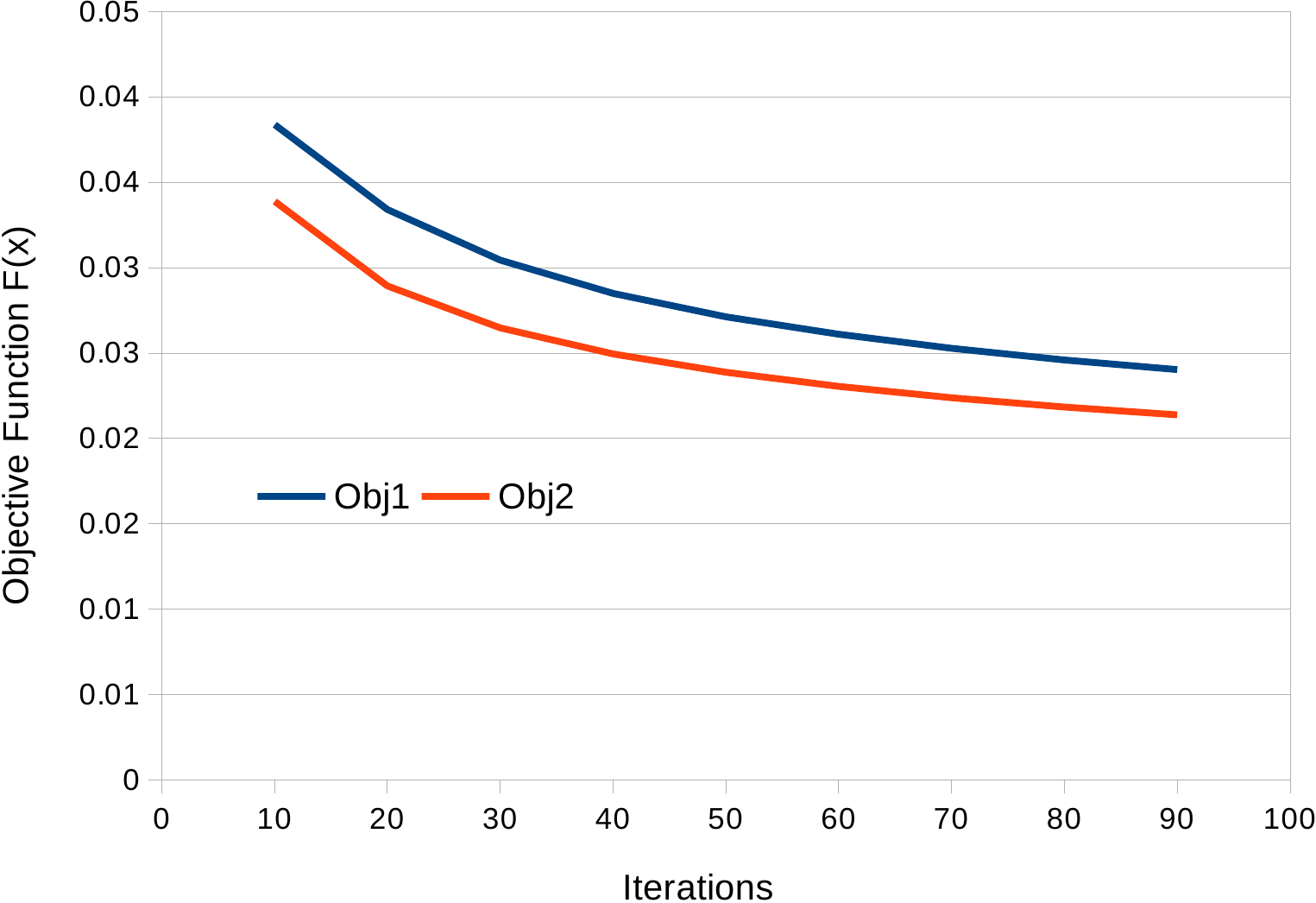}}\\
\caption{Objective function vs iterations over various benchmark data sets. (a) ijcnn1 (b) rcv1 (c) pageblocks (d) w8a. Obj1 denotes objective function when best learning rate is searched over $\{ 0.0003, 0.001, 0.003, 0.01,0.03, 0.1, 0.3\}$ while Obj2 denotes objective function value obtained with learning rate $1/L$.}
\label{conv}
\end{figure*}

\subsection{Comparative Study on Benchmark Data sets}

\begin{enumerate}
\item {\bf Performance Comparison with respect to Gmean} \\
We compare the performance of CILSD, CSFSOL and CSSCD algorithms with respect to various matrices as mentioned  in the beginning of this section. Particularly, we will focus on the \emph{Gmean} metric. The comparative results are shown in Table \ref{cilsd:tab11} and \ref{cilsd:tab12}.  From these results, we observe that CILSD achieves equal or higher \emph{Gmean} compared to the \emph{Gmean} achieved by CSSCD on 8 out of 10 data sets. On the other side, \emph{Gmean} achieved by CILSD follows closely to the \emph{Gmean} achieved by CSFSOL, an online algorithm, on most of the data sets.  In some cases, CILSD outperforms CSFSOL and CSSCD in terms of \emph{Gmean} such as on realsim, w8a etc.  If we compare the results in Tables \ref{tab11} and \ref{tab12}  with those in Tables \ref{cilsd:tab11} and \ref{cilsd:tab12}, we can see that CILSD achieves higher \emph{Gmean} compared to DSCIL (both R-DSCIL and L-DSCIL) on most of the data sets. These observations indicate the possibility of using CILSD on real data sets for class-imbalance learning in a distributed scenario.

%

\begin{table*}[]
\footnotesize
\centering
\caption{Performance comparison of CSFSOL, CSSCD, and CILSD over various benchmark data sets.}
\label{cilsd:tab11}
\begin{tabular}{|c|c|c|c|c|c|}
\hline
\multirow{2}{*}{Algorithm}  &  \multicolumn{5}{c|}{news}  \\
\cline{2-6} 
& Accuracy & Sensitivity& Specificity& Gmean& Sum  \\ \hline \hline
CSFSOL    & 0.966356&	0.982759&	0.966137&	0.974412&	0.974448 \\ \hline
CSSCD  & 0.451466&1	&0.444137&	0.666436&	0.722069
 \\ \hline
CILSD     &0.97795&	0.913793&	0.978807&	{\bf 0.945741} &	0.9463

 \\ \hline
\hline
\multirow{2}{*}{Algorithm}  &  \multicolumn{5}{c|}{rcv1}  \\
\cline{2-6} 
& Accuracy & Sensitivity& Specificity& Gmean& Sum  \\ \hline \hline
CSFSOL    & 0.957&	0.960148&	0.953312&	0.956724&	0.95673 \\ \hline
CSSCD & 0.8989&	0.871548	&0.930945&	0.900757&	0.901246
 \\ \hline
CILSD     & 0.9414&	0.9481	&0.93355&{\bf 	0.940797}&	0.940825

 \\ \hline
\hline
\multirow{2}{*}{Algorithm}  &  \multicolumn{5}{c|}{url}  \\
\cline{2-6} 
& Accuracy & Sensitivity& Specificity& Gmean& Sum  \\ \hline \hline
CSFSOL    &0.9725	&0.994505	&0.970297&	0.982327	&0.982401 \\ \hline
CSSCD & 0.947	&0.967033&	0.944994&	0.95595&	0.956014
 \\ \hline
CILSD     &0.965&	0.978022	&0.963696	&{\bf 0.970833}&	0.970859

 \\ \hline

\hline
\multirow{2}{*}{Algorithm}  &  \multicolumn{5}{c|}{webspam}  \\
\cline{2-6} 
& Accuracy & Sensitivity& Specificity& Gmean& Sum  \\ \hline \hline
CSFSOL    &0.9925	&0.952&	0.9952	&0.97336&	0.9736\\ \hline
CSSCD  & 0.988	&0.808&	1&	0.898888	&0.904
 \\ \hline
CILSD  &   0.943&	0.928	&0.944	&{\bf 0.935966}&	0.936

 \\ \hline
\hline
\multirow{2}{*}{Algorithm}  &  \multicolumn{5}{c|}{gisette}  \\
\cline{2-6} 
& Accuracy & Sensitivity& Specificity& Gmean& Sum  \\ \hline \hline
CSFSOL    &0.813&	0.896&	0.73	&0.808752	&0.8131 \\ \hline
CSSCD  & 0.451466&1	&0.444137&	0.666436&	0.722069
 \\ \hline
CILSD     &0.799	&0.6	&0.998&	{\bf 0.773822}	&0.799

 \\ \hline
\hline
\multirow{2}{*}{Algorithm}  &  \multicolumn{5}{c|}{realsim}  \\
\cline{2-6} 
& Accuracy & Sensitivity& Specificity& Gmean& Sum  \\ \hline \hline
CSFSOL    &0.8825&	0&	1	&0	&0.5 \\ \hline
CSSCD  & 0.9105&	0.33617&	0.986969&	0.576012&	0.66157
 \\ \hline
CILSD     &0.789214&	0.93617	&0.769648&	{\bf 0.848835}	&0.852909

 \\ \hline
\hline
\multirow{2}{*}{Algorithm}  &  \multicolumn{5}{c|}{ijcnn1}  \\
\cline{2-6} 
& Accuracy & Sensitivity& Specificity& Gmean& Sum  \\ \hline \hline
CSFSOL    & 0.831932	&0.607668	&0.855475&	0.721002	&0.731571 \\ \hline
CSSCD  & 0.83456&	0.827824&	0.835267&	0.831537	&0.831546
 \\ \hline
CILSD     & 0.902084	&0.507576&	0.943499	& 0.692024&0.725537

 \\ \hline

 \end{tabular}
\end{table*}

\begin{table*}[]
\footnotesize
\centering
\caption{Performance comparison of CSFSOL, CSSCD, and CILSD over various benchmark data sets.}
\label{cilsd:tab12}
\begin{tabular}{|c|c|c|c|c|c|}
\hline
\multirow{2}{*}{Algorithm}  &  \multicolumn{5}{c|}{w8a}  \\
\cline{2-6} 
& Accuracy & Sensitivity& Specificity& Gmean& Sum  \\ \hline \hline
CSFSOL    &0.970637&	0.0330396&	1	&0.181768&	0.51652 \\ \hline
CSSCD  & 0.97231	&0.563877&	0.9851&	0.745302&	0.774489
 \\ \hline
CILSD     &0.971908&	0.585903&	0.983997&	{\bf 0.759294}	&0.78495

 \\ \hline

\hline
\multirow{2}{*}{Algorithm}  &  \multicolumn{5}{c|}{covtype}  \\
\cline{2-6} 
& Accuracy & Sensitivity& Specificity& Gmean& Sum  \\ \hline \hline
CSFSOL    &  0.908817	&0.86642&	0.910199&	0.88804&	0.888309 \\ \hline
CSSCD  & 0.968433	&0	&1&	0&	0.5
 \\ \hline
CILSD   &  0.96855&	0.0227033&	0.99938&	0.150629	&0.511042

 \\ \hline
\hline
\multirow{2}{*}{Algorithm}  &  \multicolumn{5}{c|}{pageblocks}  \\
\cline{2-6} 
& Accuracy & Sensitivity& Specificity& Gmean& Sum  \\ \hline \hline
CSFSOL    &  0.793056&	0.662069	&0.81306&	0.73369&	0.737564 \\ \hline
CSSCD &  0.887163	&0.751724	&0.907846&	0.826105&	0.829785
 \\ \hline
CILSD     &0.910005&	0.727586&	0.937862&	{\bf 0.82606}	&0.832724

 \\ \hline

 \end{tabular}
\end{table*}

\item {\bf Speedup Measurements}\\
In order to see how CILSD training time varies with the number of cores, we  partition the data matrix into equal chunks across examples and distribute it to different processing nodes (cores). Each core runs a local copy of CILSD algorithm. Training time of CILSD algorithm for different partitioning of the data is shown in Figure \ref{speedup} (a) and (b). We can clearly see that training time is decreasing linearly with the number of cores for all the data sets. These results demonstrate the utility of employing multiple cores for class-imbalance learning  task.

\item {\bf Gmean versus Number of Cores}\\
We conduct the experiment to see how \emph{Gmean} varies with  different partitioning of the data. The results are shown in Figure \ref{gmeanvscore} (a) and (b). From these results, we can infer that \emph{Gmean} does not remain constant with different partitioning of the data. For some data sets such as realsim, \emph{Gmean} is decreasing with increasing number of cores. On the other hand, \emph{Gmean} is first increasing and then decreasing for rcv1 and ijcnn1. Whereas, it is continuously increasing for some data sets such as url. This chaotic behavior of CILSD may be due to different proportion of positive and negative samples in the data chunk alloted to different nodes. Above observation leads to the fact CILSD algorithm is sensitive to the different partitioning of the data and the right choice of data partitioning is important.

\item {\bf Gmean versus Regularization Parameter}\\
In this Section, the effect of sparsity promoting parameter $\lambda$ on \emph{Gmean} is demonstrated. The results are shown in Figure \ref{fig:lg}. From these results, we can clearly observe that \emph{Gmean} is quite sensitive to the setting of regularization parameter $\lambda$. For example, \emph{Gmean} on url data sets drops slowly with increasing  $lambda$ while it quickly drops to $0$ for gisette. 
\begin{figure*}
\centering
 \subfloat[]{ \includegraphics[width=8cm, height=5cm]{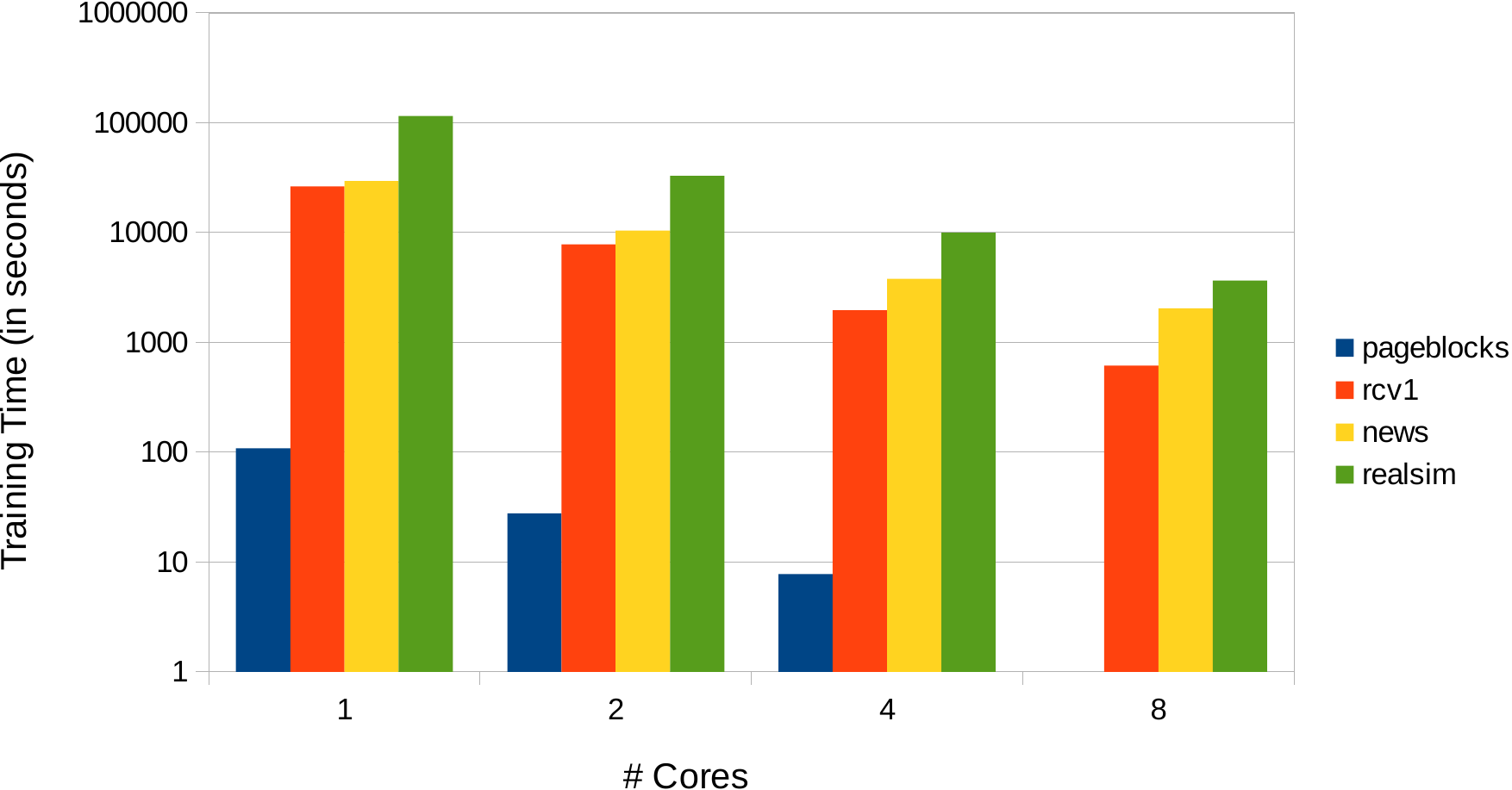}}
  \subfloat[]{ \includegraphics[width=8cm, height=5cm]{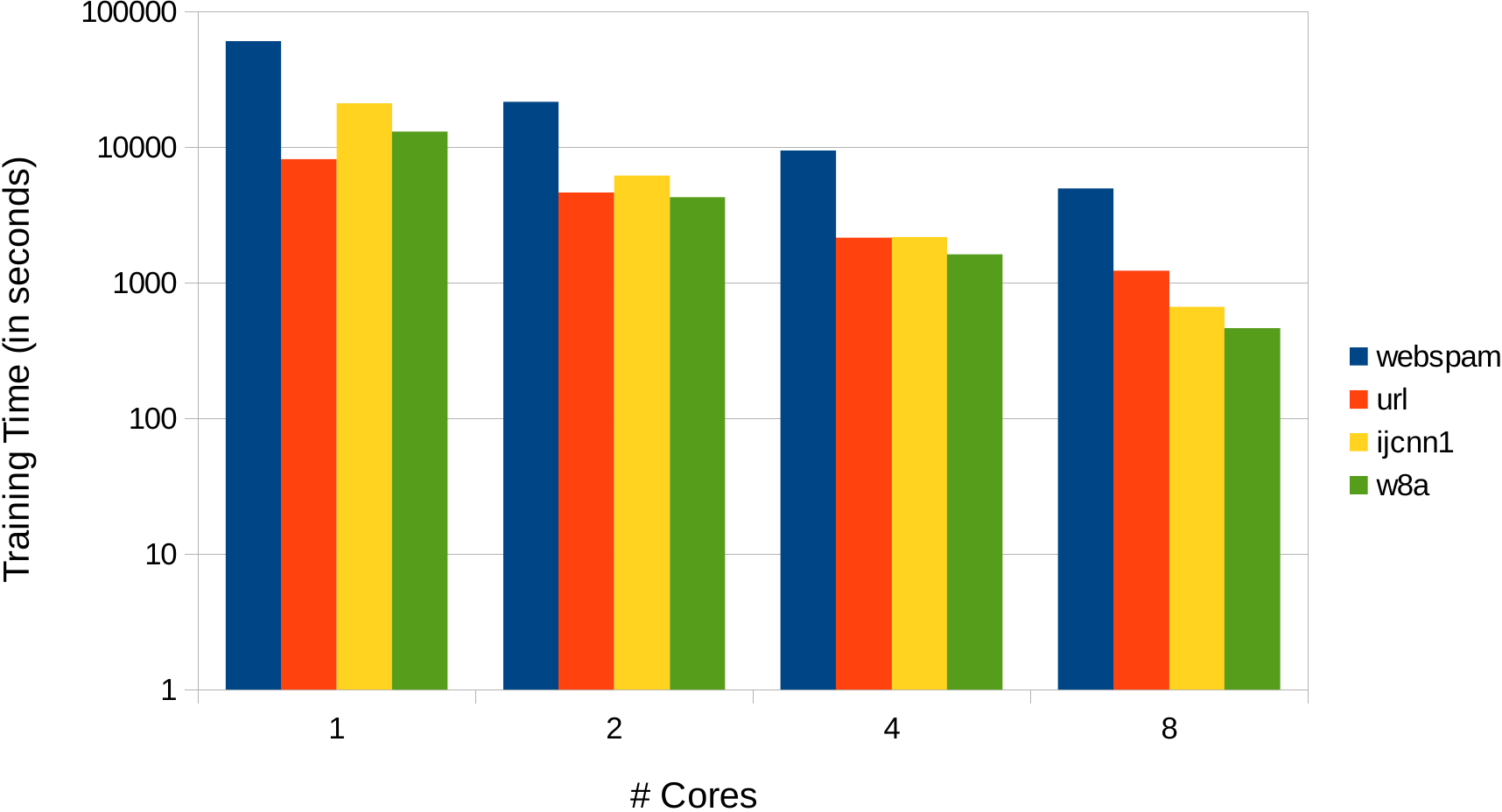}}
\caption{Training time versus number of cores to measure the speedup of CILSD algorithm. Training time in both the figures is on the log scale.}
\label{speedup}
 \subfloat[]{ \includegraphics[width=8cm, height=5cm]{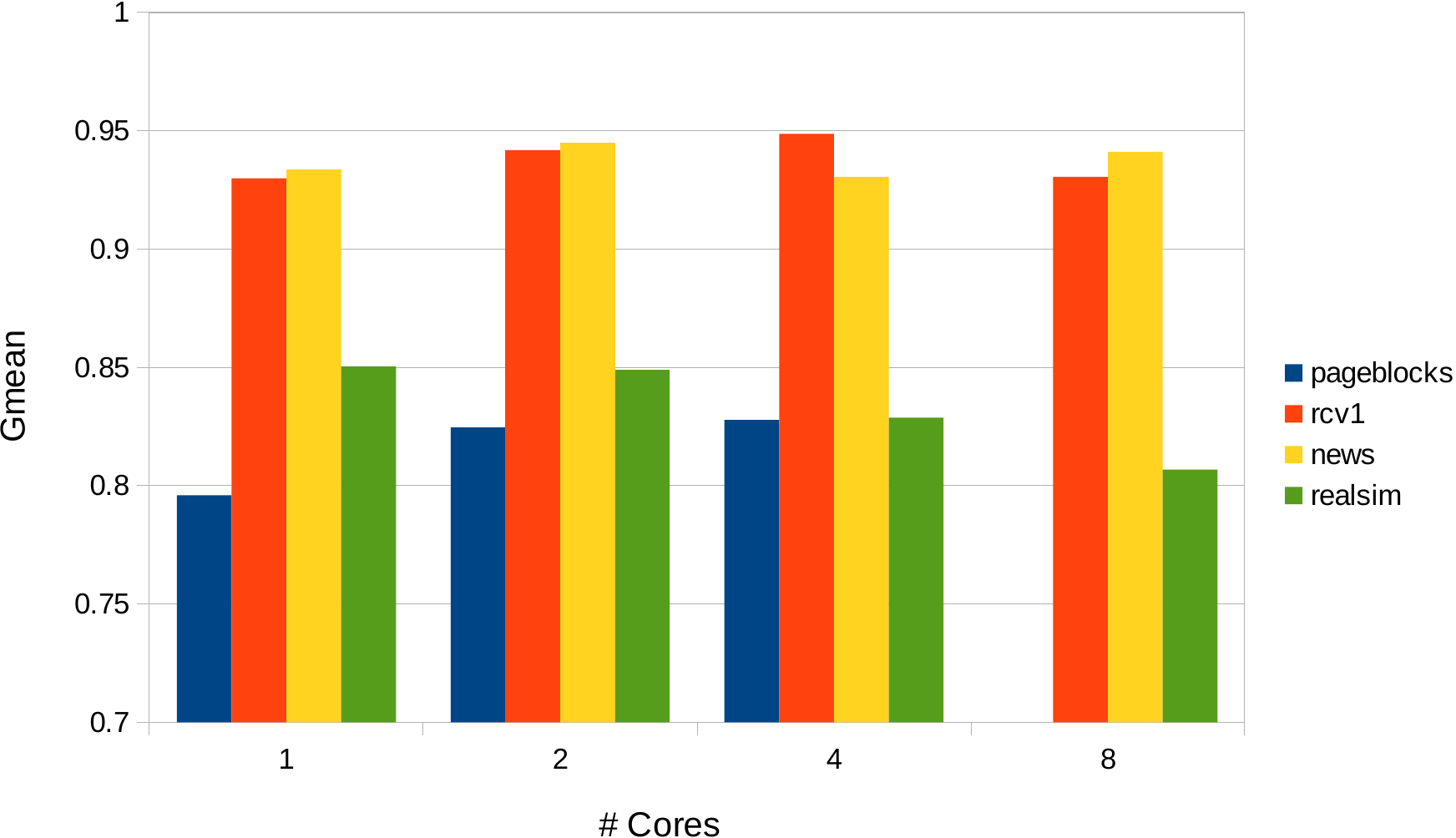}}
  \subfloat[]{ \includegraphics[width=8cm, height=5cm]{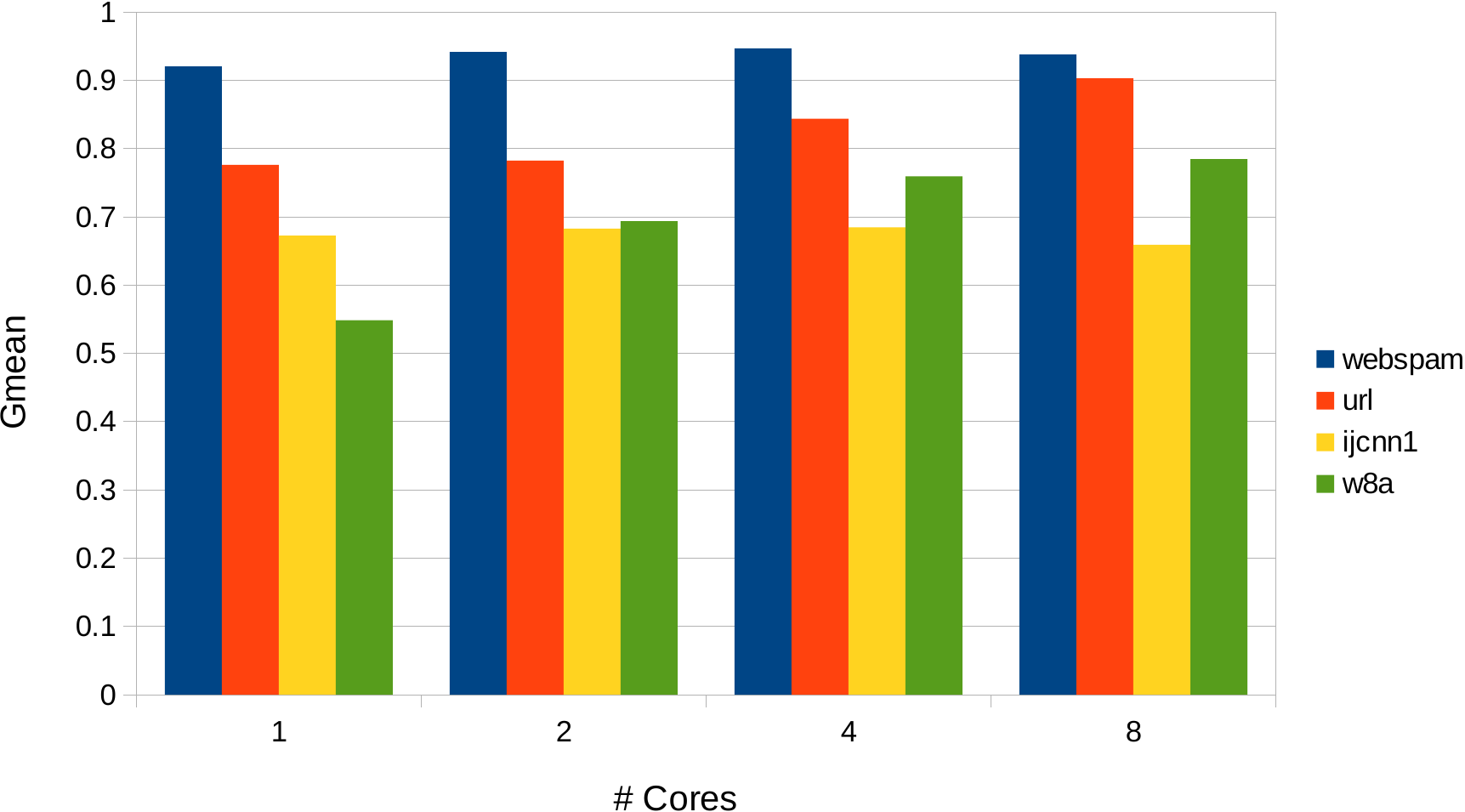}}
\caption{Gmean achieved by  CILSD algorithm versus number of cores on various benchmark data sets. }
\label{gmeanvscore}
\end{figure*}
\begin{figure*}
\centering
\subfloat[ijcnn1]{\includegraphics[width=6cm, height=4cm]{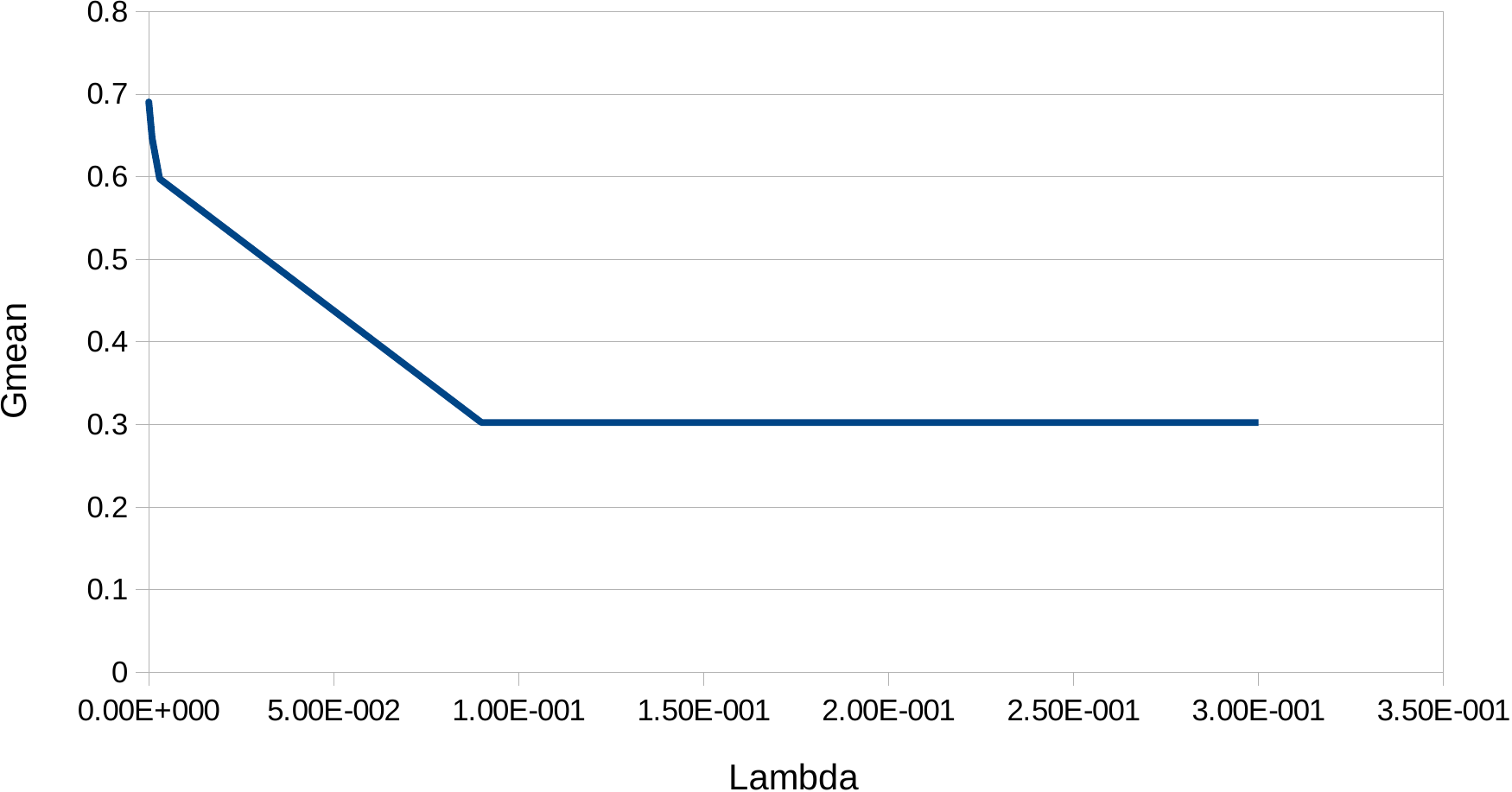}} 
\subfloat[rcv1]{\includegraphics[width=6cm, height=4cm]{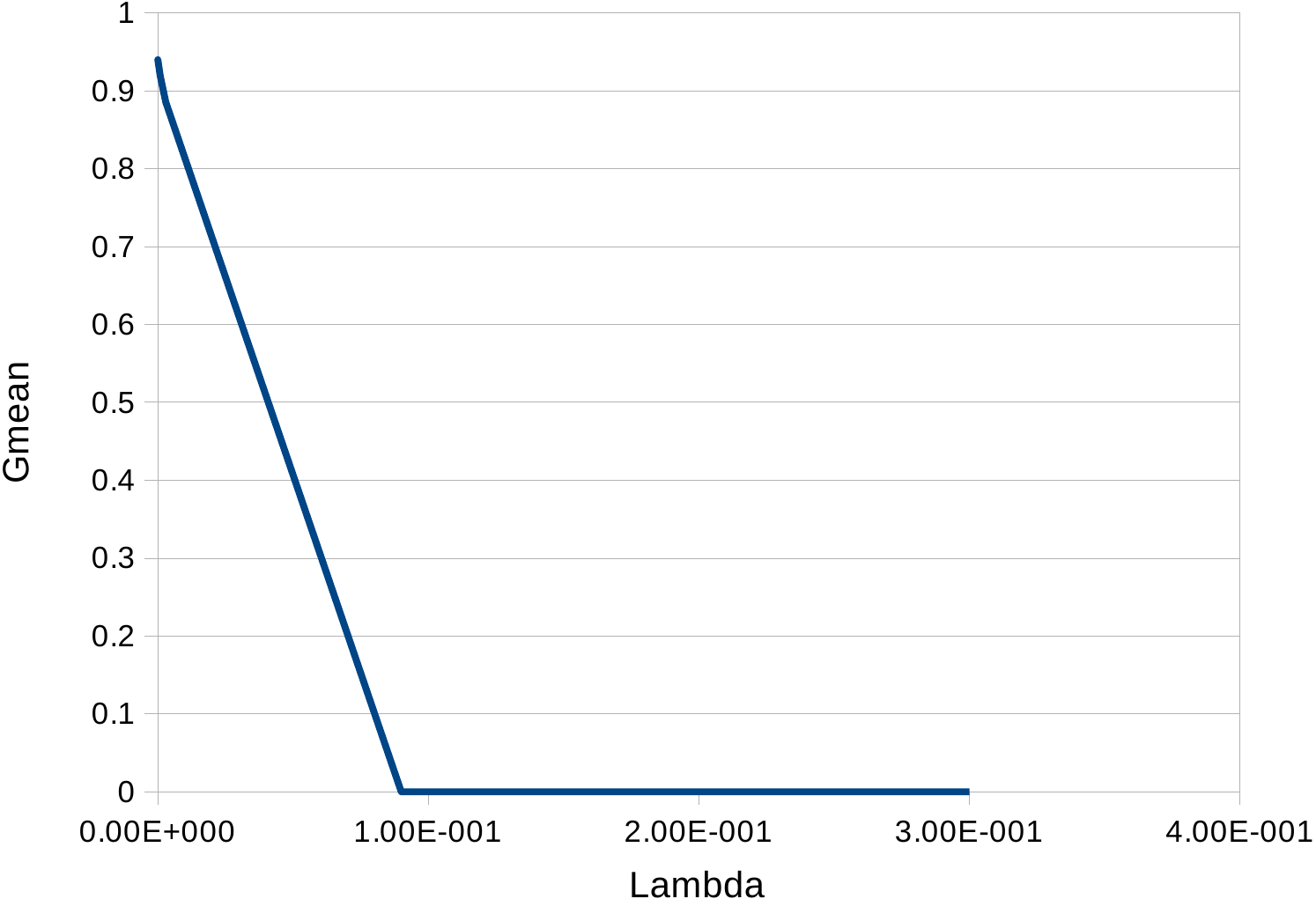}} \\
\subfloat[gisette]{\includegraphics[width=6cm, height=4cm]{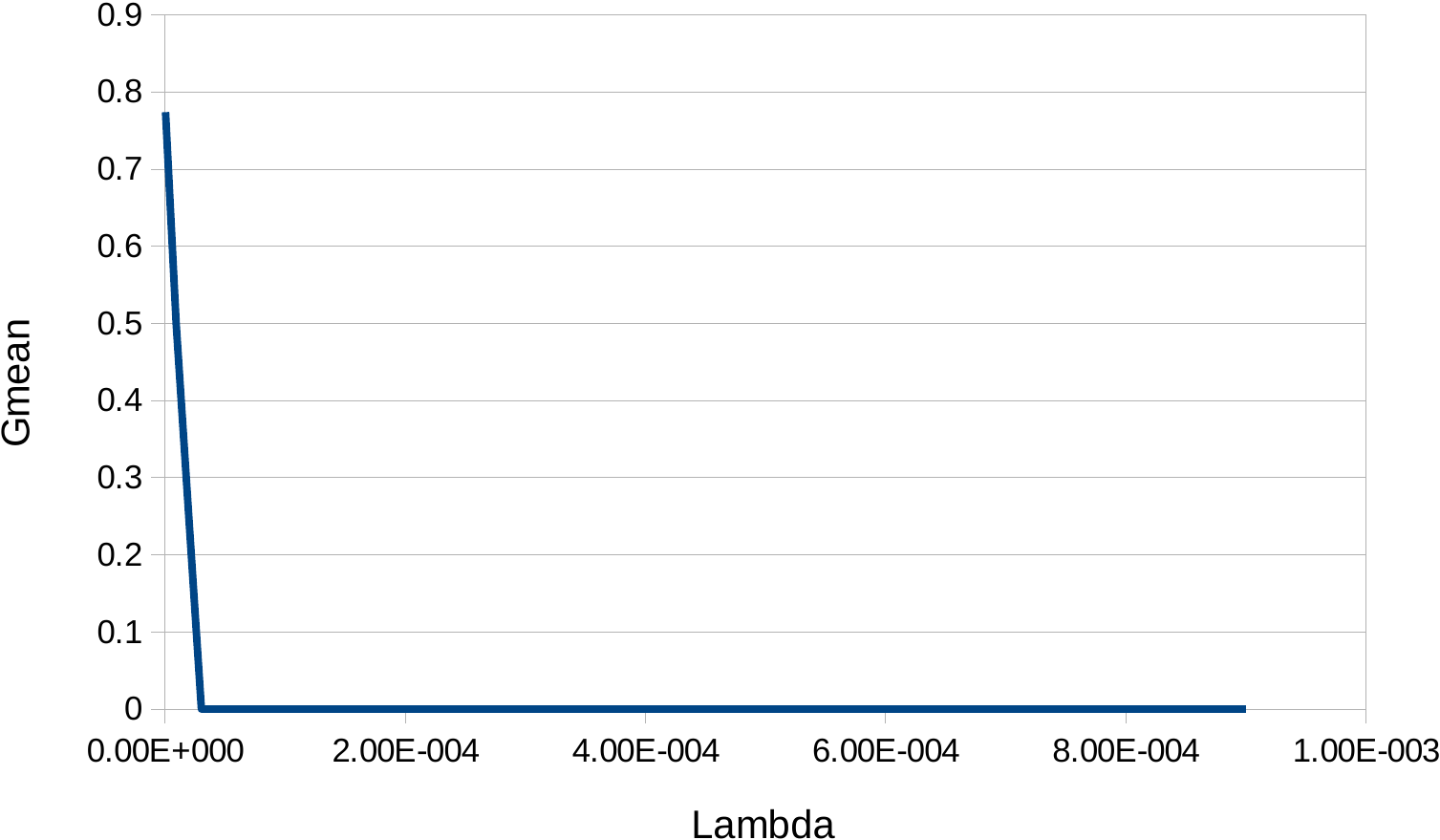}}  
\subfloat[news]{\includegraphics[width=6cm, height=4cm]{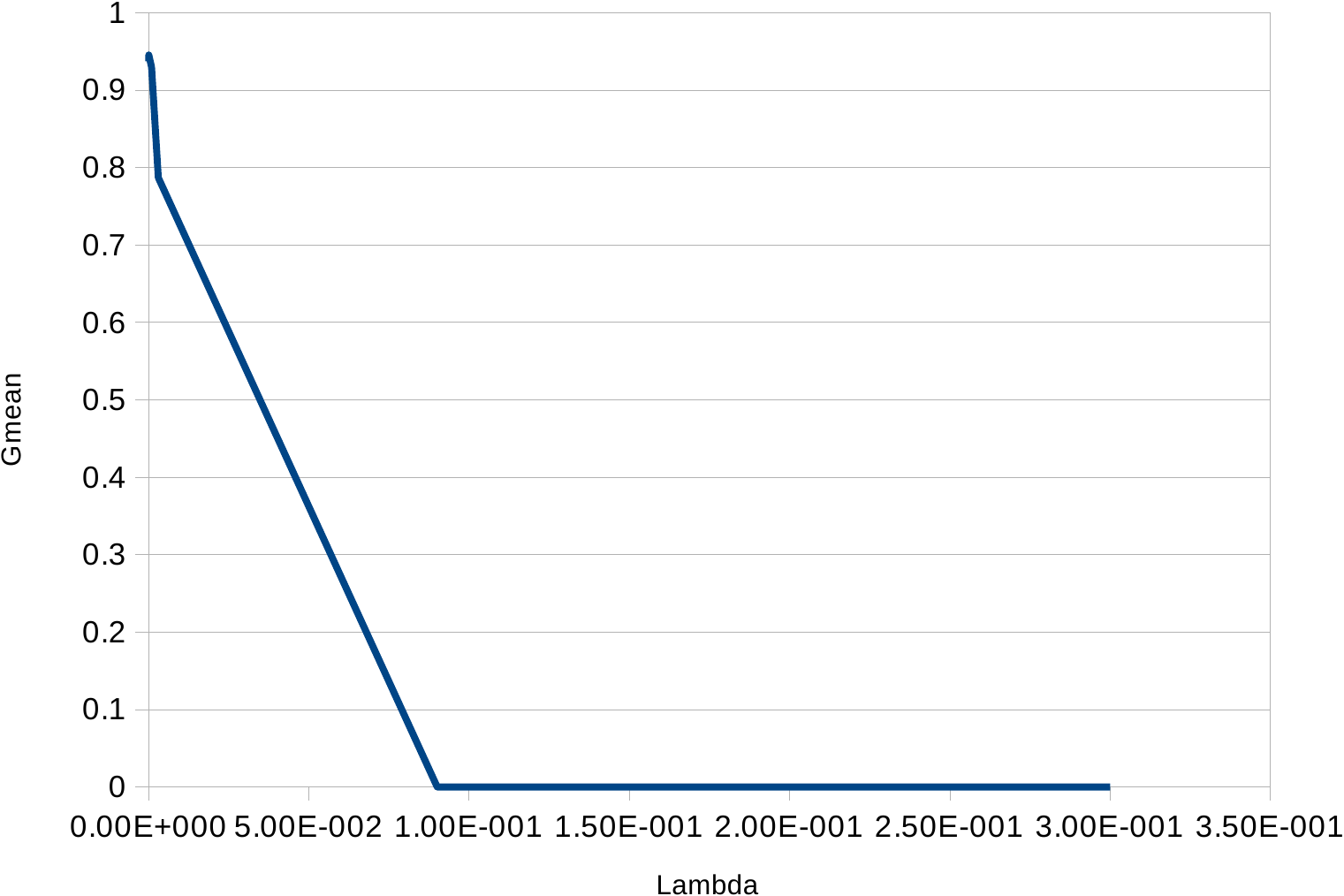}} \\
\subfloat[webspam]{\includegraphics[width=6cm, height=4cm]{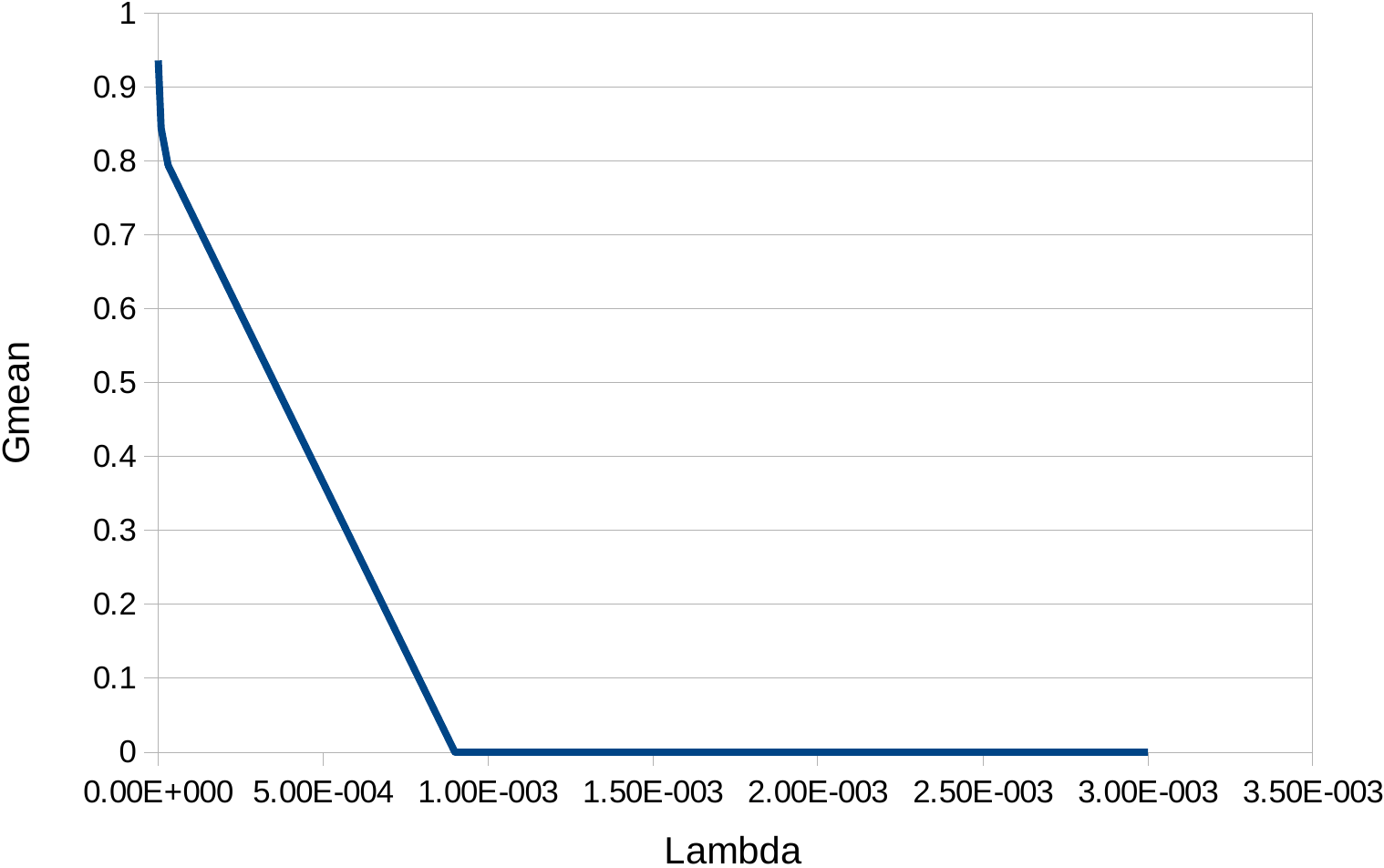}}
\subfloat[url]{\includegraphics[width=6cm, height=4cm]{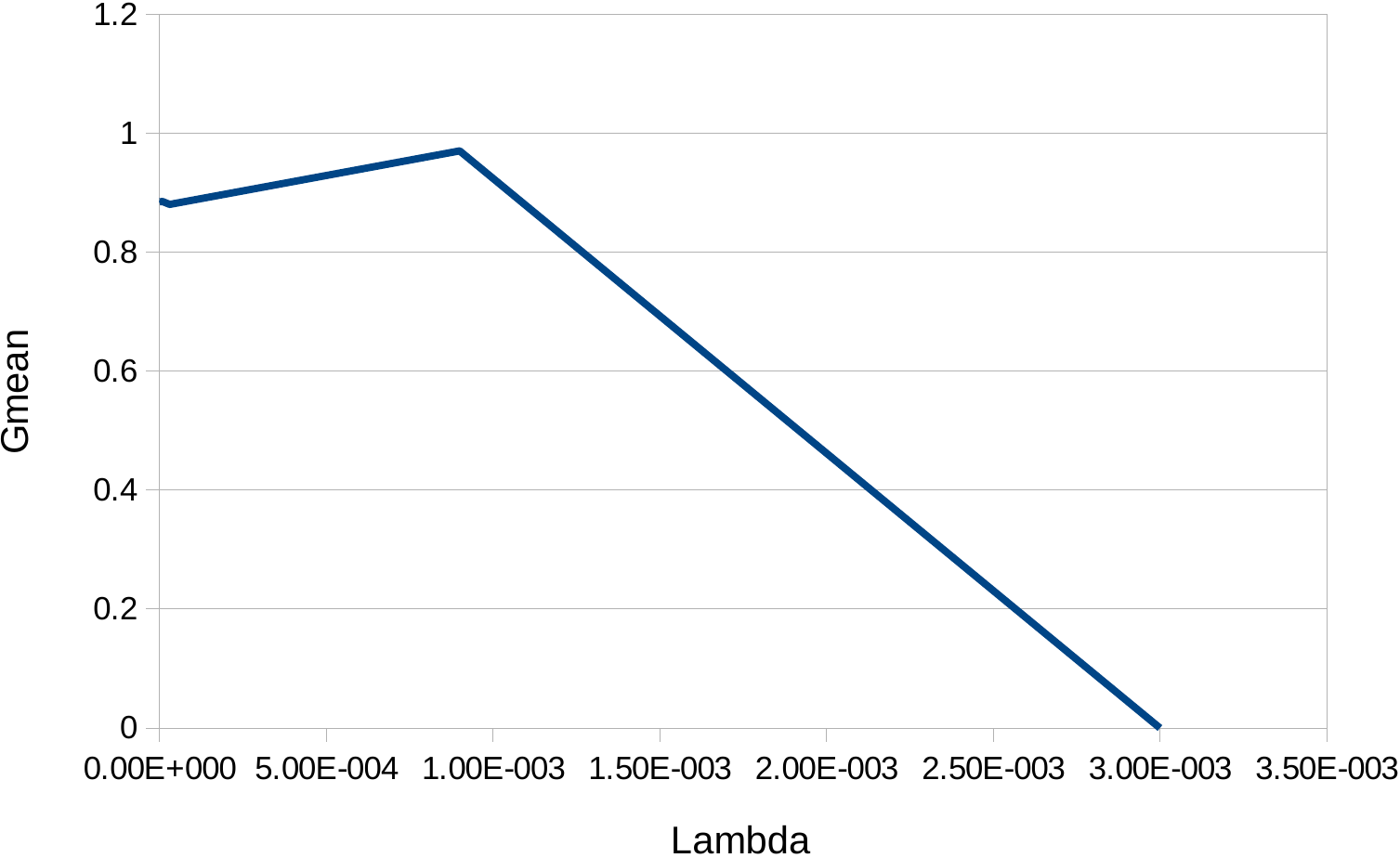}}  \\
\subfloat[url]{\includegraphics[width=6cm, height=4cm]{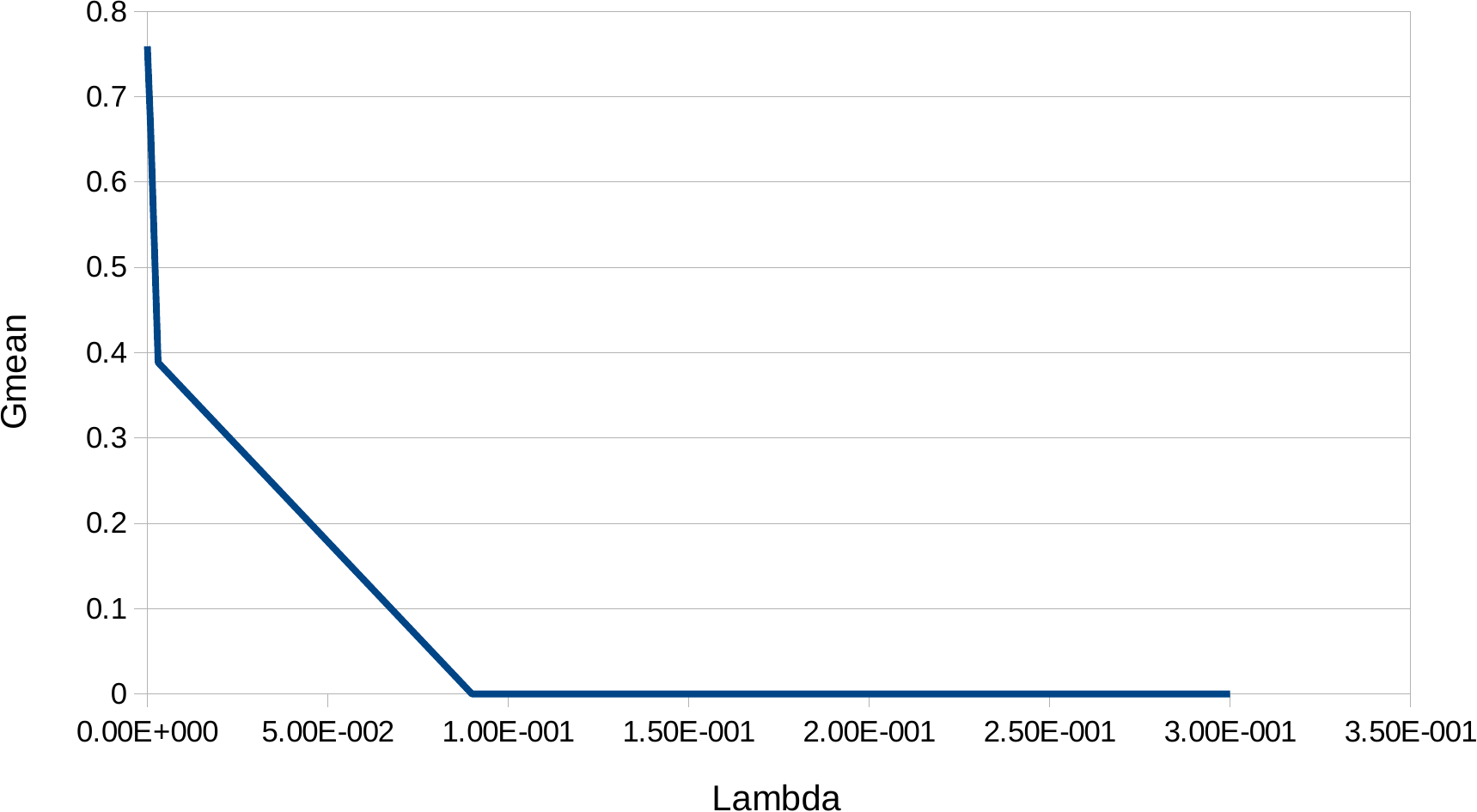}} 
\subfloat[realsim]{\includegraphics[width=6cm, height=4cm]{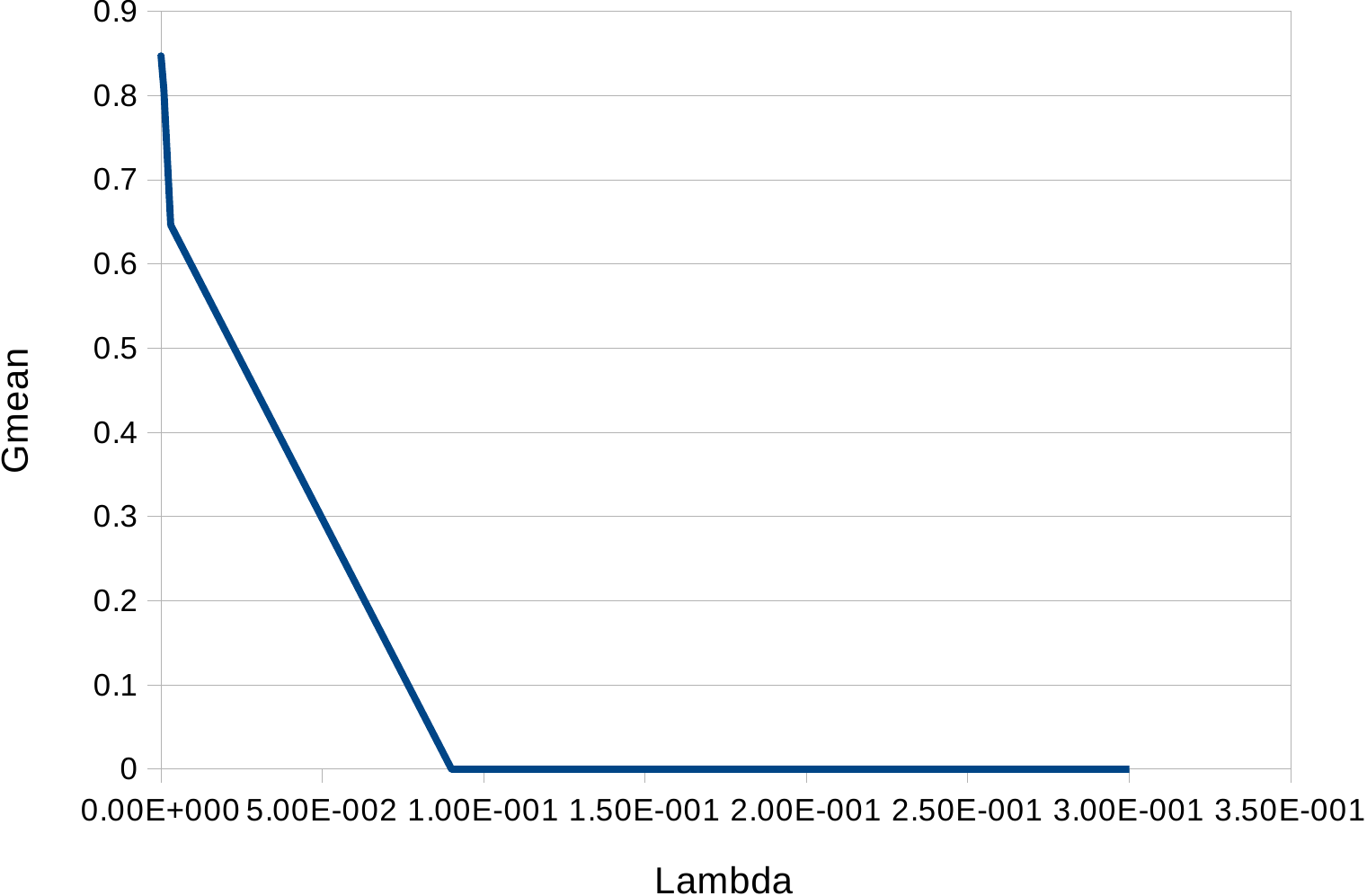}}
\caption{Effect of regularization parameter $\lambda$ on Gmean (i) ijcnn1 (ii) rcv1 (iii) gisette (iv) news (v) webspam (vi) url (vii) w8a (viii) realsim. $\lambda$ varies in \{ 3.00E-007, 0.000009, 0.00003, 0.0009
0.003
0.09
0.3
 \}} 
\label{fig:lg}
\end{figure*}

\end{enumerate}

\clearpage
\newpage

\subsection{Comparative Study on Benchmark and Real Data sets}

Breast cancer detection \cite{AT:2014,AT:2015} is a type of anomaly detection problem. One in every eight  women around the world is susceptible to breast cancer. In this section, we demonstrate the applicability of our proposed algorithm for anomaly detection in X-ray images of KDDCup 2008 data set\cite{Balaji}.  KDDCup 2008 data set contains information of 102294 suspicious regions, each region described by 117 features.  Each region is either ``benign'' or ``malignment'' and the ratio of malignment to benign regions is 1:163.19. We split the training data set into 5 chunks. The first four chunk have size 20000 candidates each and are used for training on 4 cores. The last chunk has size 22294 and used for testing.  We compare the performance of R-DCSIL and CILSD  with CSFSOL and CSOGD-I that was proposed  in \cite{Jialei2014}. It is shown in \cite{Jialei2014} that CSOGD-I outperformed many first-order algorithms such as ROMMA, PA-I, PA-II, PAUM, CPA\_{PB} etc (see Table 6 in \cite{Jialei2014}).  Hence, we only compared with CSFSOL and CSOGD-I in our experiments. We note that CSFSOL and CSOGD-I are online algorithms that do not require separate training set and test set. We set the learning rate in CSOGD-I to 0.2 as discussed in their paper and reproduced the results.  What is important here is that the ratio of malignment to benign regions in our test set is 1:123 which is not very less than the ratio of malignment to benign tumors in the original data set. Comparative performance is reported in Table \ref{kddcup} with respect to the \emph{Sum} metric as defined in the Experiment section. From the results reported in Table \ref{kddcup}, we can clearly see that the \emph{Sum} value achieved by R-DSCIL is much larger than the value obtained through CSOGD-I. The \emph{Sum} performance of CILSD remains above the \emph{Sum} performance of CSOGD-I and CSFSOL.
These observations indicate the possibility of using R-DSCIL and CILSD in real-world anomaly detection tasks.
\begin{table}
\caption{Performance evaluation of R-DSCIL, CILSD, CSFSOL and CSOGD-I on KDDCUP 2008 data set. }
\label{kddcup}
\centering

\begin{tabular}{|c|c|}
\hline
Algorithm  & Sum\\\hline
CSOGD-I &0.5741092\\ \hline
CSFSOL&0.71089\\\hline
R-DSCIL&{\bf 0.7336652}\\\hline
CILSD&{\bf 0.714153}\\
\hline
\end{tabular}
\end{table}
\newpage
\clearpage
\section{Discussion}
In the present work, we propose two algorithms for handling class-imbalance in a distributed environment on small and large-scale data sets. DSCIL algorithm is implemented in two flavors: one uses second order method (L-BFGS) and the other uses first order method (RCD) to solve the subproblem in DADMM framework. In our empirical comparison, we showed the convergence results of L-DSCIL and R-DSCIL where L-DSCIL converges faster than R-DSCIL. Secondly, \emph{Gmean} achieved by L-DSCIL is close to the \emph{Gmean} of  a centralized solution for most of the data sets. Whereas \emph{Gmean} achieved by R-DSCIL varies due to its random updates of coordinates.  Thirdly, coming to the training time comparison, we found in our experiments that R-DSCIL has cheaper per iteration cost but takes a longer time to achieve $\epsilon$ accuracy compared to L-DSCIL algorithm. Finally, the effect of varying cost, regularization parameter and the number of cores is also demonstrated. The empirical comparison showed the potential application of L-DSCIL and R-DSCIL for real-world class-imbalance and anomaly detection tasks where our algorithms outperformed some recently proposed algorithms.

Our second algorithm (CILSD)  is based on FISTA-like  update rule. We show, through extensive experiments on benchmark and real data sets, the convergence behavior of CILSD, speed up, the effect of the number of cores and regularization parameter on Gmean. In particular, CILSD algorithm does not require tuning of learning rate parameter and can be set to 1/L as in gradient descent.
Comparative evaluation with respect to a recently proposed class-imbalance learning algorithm and a centralized algorithm shows that CILSD is able to either outperform or perform equally well on many of the data sets tested. Speedup results demonstrate the advantage of employing multiple cores. We also observed, in our experiments, chaotic behavior of Gmean with respect to varying number of cores. Experiment on KDDCup data set indicates the possibility of using CILSD algorithm for real-world distributed anomaly detection task. Comparison of DCSIL and CILSD shows that CILSD convergences faster and achieves higher \emph{Gmean} than DSCIL. 
\chapter{Unsupervised Anomaly Detection  using SVDD-A Case Study}\label{Chapter6}
Anomaly detection techniques discussed in the previous chapters are \emph{supervised }, that is, they require labels of normal as well as anomalous examples to build the model. However, real world data rarely contain labels. This means the techniques discussed  in the previous chapters can not be applied. Therefore, we turn the crank to \emph{unsupervised} anomaly detection. 

In this chapter, we  study a robust algorithm, based on support vector data description (SVDD) due to \cite{Tax2004}, for anomaly detection in real data. The data set used in the experiment comes from nuclear power plant and represents the count of neutrons in reactor channels. We apply the SVDD algorithm on nuclear power plant data to detect the anomalous channels of neutron emission.  Experiments  demonstrate the effectiveness of the algorithm as well as finding anomalies in the data set. We also discuss extensions of the algorithm to find anomalies in high dimension and non linearly separable data.

\section{Introduction}
\begin{figure}{sv}
\centering
\includegraphics[width=3.5in, height=2.5in ]{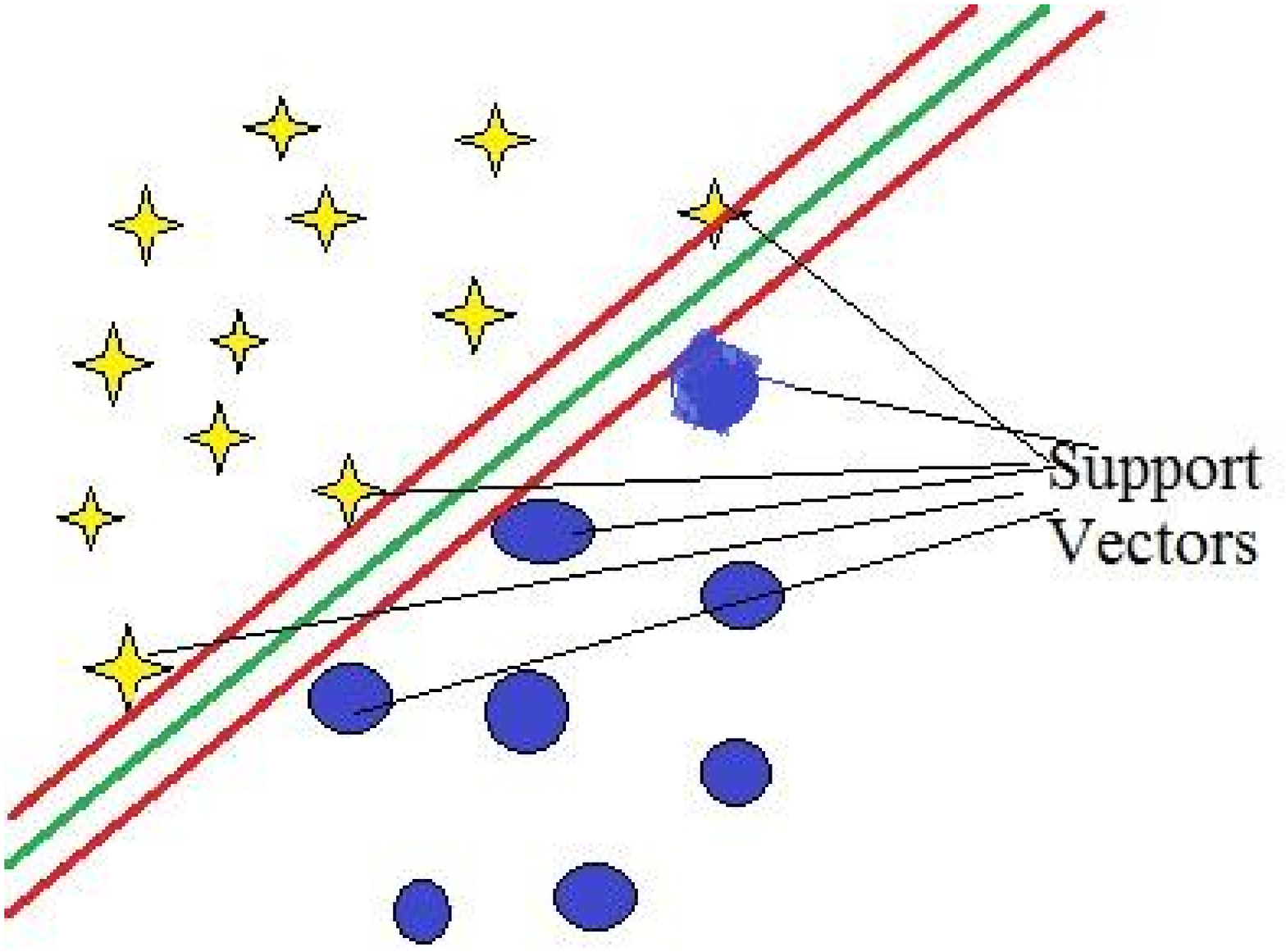}
\caption{Support vectors in two class classification problem}
\label{sv}
\end{figure}

Our case study is based on the support vector data description algorithm. Tax et al. \cite{Tax2004} first proposed the support vector data description in 2004. Their work is recently extended to uncertain data by Liu et al. \cite{Liu2013}. Original work of Tax was based on support vector classifier in an unsupervised setting. Later, this was applied in semi-supervised and supervised setting with little modification by Gornitz \cite{nico2013}.   Next, we describe some key terms related to SVDD.
\begin{definition}
Support vectors are the set of points that lie on the boundary of the region separating normal and anomalous points as shown in Fig. \ref{sv}.
\end{definition}
\begin{definition}
Support vector domain description concerns the characterization of a data set through support vectors. 
\end{definition}
\vspace{-.8cm}
\section{Support Vector Data Description Revisited }\label{svddrevisit}
\begin{figure}
\includegraphics[width=3.5in, height=2.5in ]{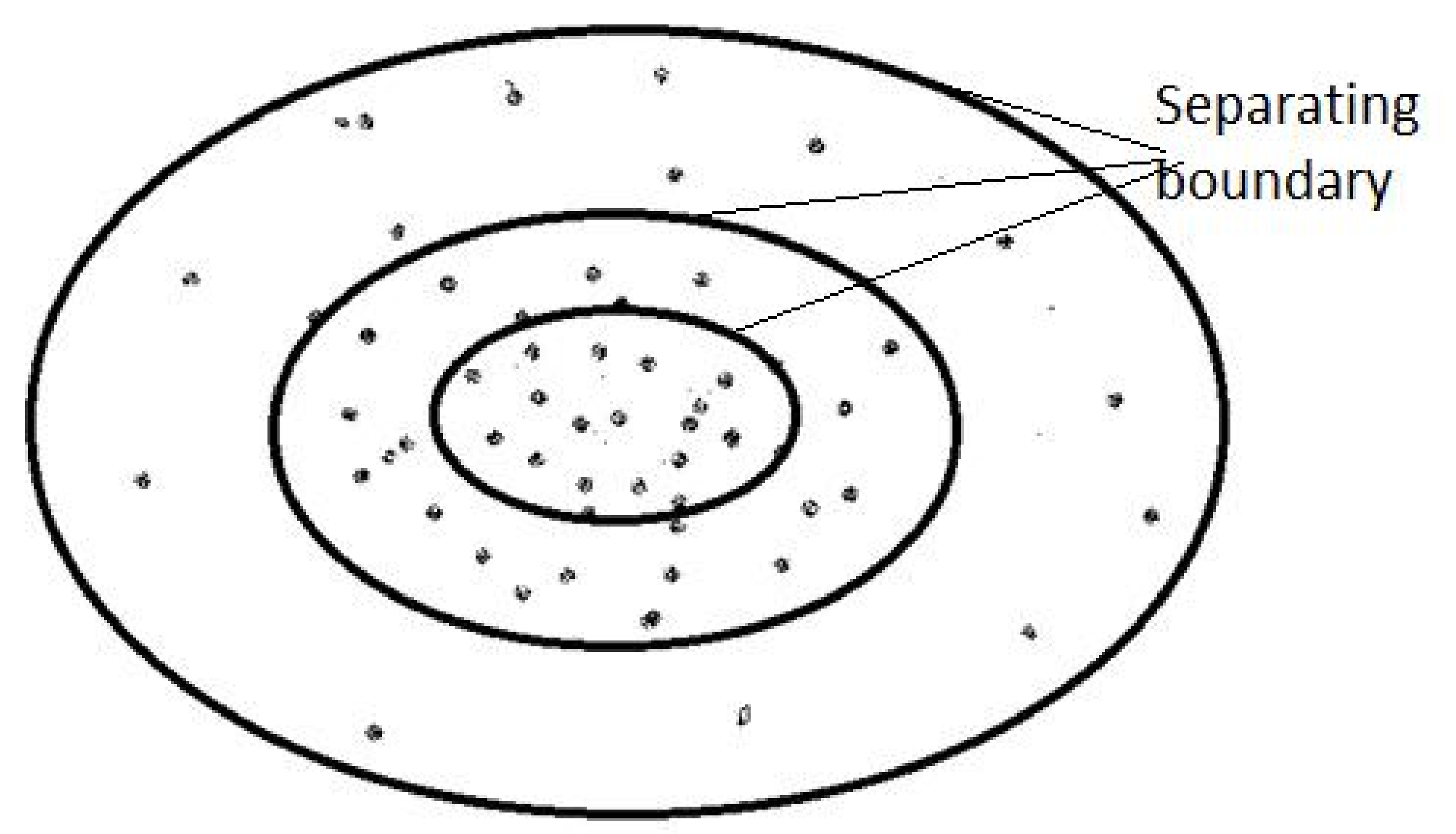}
\centering
\caption{Illustrates the spheres for data generated according to a spherical two-dimensional Gaussian distribution. The center part shows the center of mass of the training points. Anything outside the boundary can be considered as outliers.}
\label{hype}
\end{figure}

We describe support vector data description algorithm\cite{Tax2004} for completeness. 
The problem is to make a description of a training
set of data instances and to detect which (new) data instances resemble this training set. SVDD essentially is a one-class classifier \cite{Moya1993}. The basic idea of SVDD is that a good description of the data encompasses only normal instances. Outliers reside either at the boundary or outside of the hypersphere (generalization of sphere in more than 3 dimension space) containing data set as shown in Fig .\ref{hype}. The method is made robust against outliers in the training set and is capable of tightening the description by using anomalous examples. The basic SVDD algorithm is given in {\bf Algorithm 1.}

The idea of anomaly detection using minimal hypersphere (SVDD) is as follows. From the preprocessed data in the kernel matrix, we calculate the center of mass of the data. The center of mass of the data  forms the center of the hypersphere. Subsequently, we compute the distances of training points from the center of mass so as to obtain an empirical estimate of the center of mass. Empirical estimation error in center of mass is added to the max distance of training point from the center of mass to give the threshold. Now, any point in test data lying beyond threshold is classified as anomalous. 
\begin{algorithm}[t]
  \caption {Anomaly detection using SVDD}
    \label{svddalgo}
  \begin{algorithmic}[1]
    \Require Training data $X=(x_1,x_2,\dotsc,x_n)^T$ and Test data $Y=(y_1,y_2,\dotsc,y_n)^T$
    \Ensure Novel point indices
        	 \State  alculate Kernel matrix K from training data using any kernel function e.g.
RBF function   $exp(-||X-Z||^2/2\sigma ^2)$.
     	 \State   Set confidence parameter $\delta$=0.01
	\State    Compute distances of data to centre of mass
\State	Compute the estimation error of empirical centre of mass
\State	Compute resulting threshold
\State	Now compute distances of test data
\State	Indices of novel test points are calculated as:
novelindices = find (testdist2 $>$threshold) 
  \end{algorithmic}
\end{algorithm}

Below, we show how kernel matrix and center of mass (lines {\bf 1} and {\bf 3} of the algorithm) are computed because they form the heart of the algorithm.
\begin{itemize}
\item {\bf  Kernel Matrix Construction}: Kernel matrix is a matrix whose $(i,j)^{th}$ entry encodes the similarity between instances $ i$ and $j$. From implementation point of view, we have used RBF kernel (aka Gaussian kernel) given as: 
\begin{equation}
    K(i,j)=exp(-|| x-z||^2/ 2*\sigma^2 )
\end{equation}
Where $x$ and $z$ are two samples. Sigma is kernel width very similar to standard deviation. Again, we note that the final result depends upon the sigma value; careful choice of sigma is required. In our implementation, we have used $\sigma=1$. 
\item  {\bf Computing center of mass(COM)}:

Intuitively, center of mass (COM) of a set of points is same as center of gravity. More     formally, it is    defined as:-
\begin{equation}
  \phi_s= \frac{1}{m} (\sum_{i}^m \phi(X_i))
\end{equation}

       where $m$ is the size of the training set. $\phi$ is a map from input space to feature space. In fact, we  compute distances of training data from COM which is equivalent to centering the kernel matrix. For detailed explanation, see \cite{Tax2004}.

\end{itemize}

\indent Note that data can be (1) high dimension (2) Non-linearly separable. To handle case 1, we can introduce multiple kernels corresponding to each feature and learn them from the training data. To take the second case into account, we can use higher order polynomial or Gaussian kernel (a kernel is a function that maps low dimension data to higher dimension).  Since we have the model at our disposal, we now embark on evaluating it on real dataset  as described in the next Section.
\section{Experiments}
In this Section, we present the results of our numerical simulation. 

\subsection{Experimental Testbed and Setup} 
Algorithm \ref{svddalgo} is implemented in MATLAB. To perform the experiment, we constructed the hypersphere using training sample of size 600 with the assumption that the training set contains only few anomalous points. We selected  the starting 600 data points as the training sample. Although, we can also select the training samples uniformly at random. The selection of initial 600 data points as the training set is based on the assumption that reactor channel are working in the normal condition.  During testing, test sample of size 5714(=6314-600) is presented to the model. 
 In the course of our experiment, we make the following assumptions.
\begin{itemize}
\item Threshold value for the confidence parameter $\delta=0.01$. That is, the probability that test error is less than or equal to training error is 99.99\%.
\item Algorithm \ref{svddalgo} is run per feature-wise, that is, we ran the Algorithm \ref{svddalgo} for each channel to detect \emph{point} anomalies individually.
\item
Kernel used is Gaussian.
\item
All the indices in the figures are with respect to transformed data (log transformation).

\subsection{The Dataset} 
\begin{figure}
\centering
\includegraphics[width=6in, height=4in ]{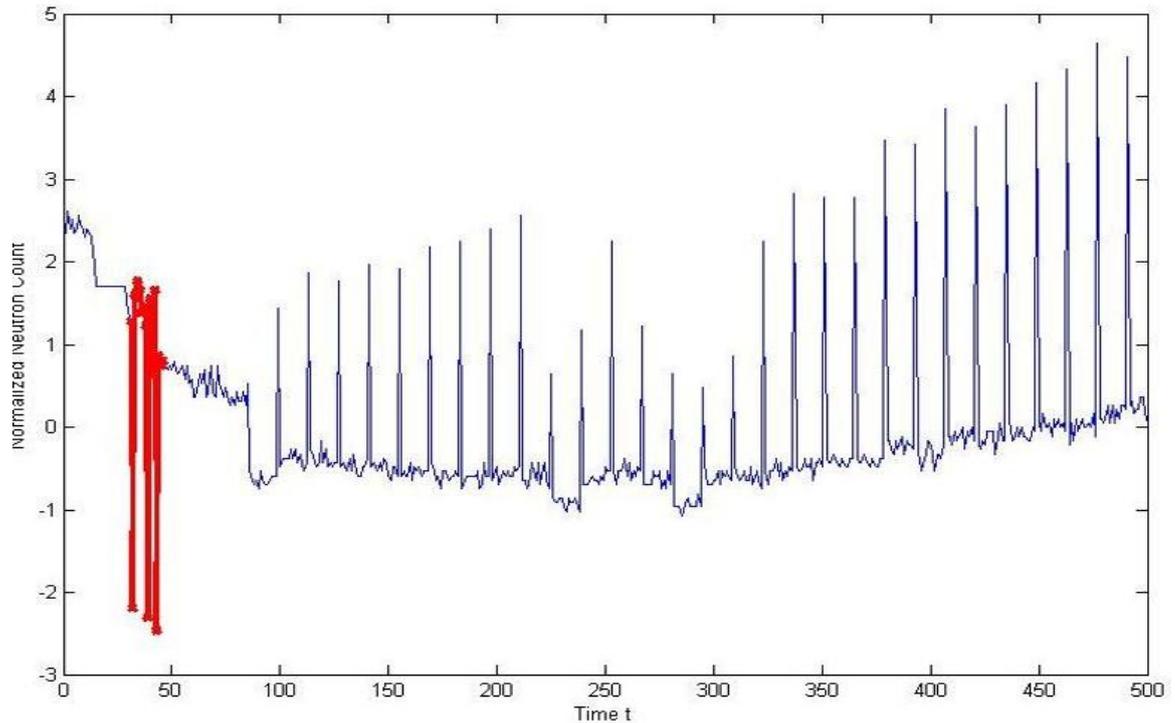}
\caption{Nuclear power plant data with marked anomalous subsequence}
\label{subano}
\end{figure}
The data set  used in the experiments come from the nuclear power plant at Bhabha Atomic Research Center, Mumbai, India.   It consists of neutron flow in nuclear reactor channel. It contains some textual data and irreverent columns. For example, detector no., Batch no. and column no. 15. Hence, we have removed them from the raw data resulting in overall 14 relevant columns(feature). A plot of the preprocessed data from single channel comprising 500 tuples is shown in Fig. \ref{subano}. 
The given data is for a duration of 6 years  (from 2005 to 2010). After preprocessing, that is, removing noise from the data, it constitutes 6314 tuples.  

\end{itemize}
\section{Results}
\begin{sidewaysfigure}
\centering
\includegraphics[width=7in, height=4.5in ]{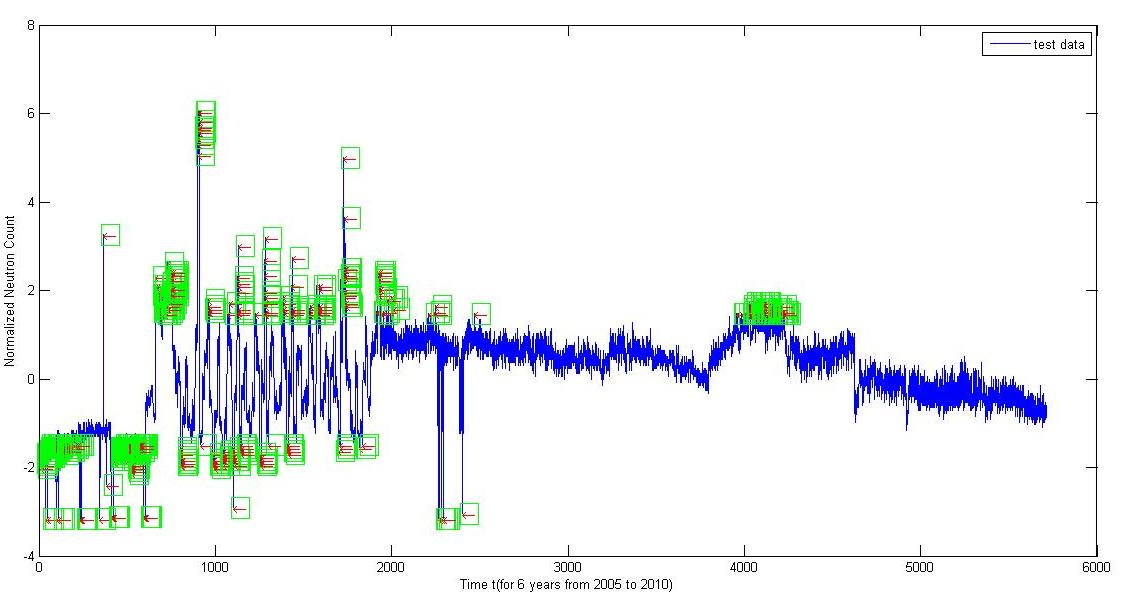}
\caption{Anomalies(marked in red) found in detector 1 of the power plant data}
\label{det1}
\end{sidewaysfigure}

\begin{sidewaysfigure}
\centering
\includegraphics[width=7in, height=4.5in ]{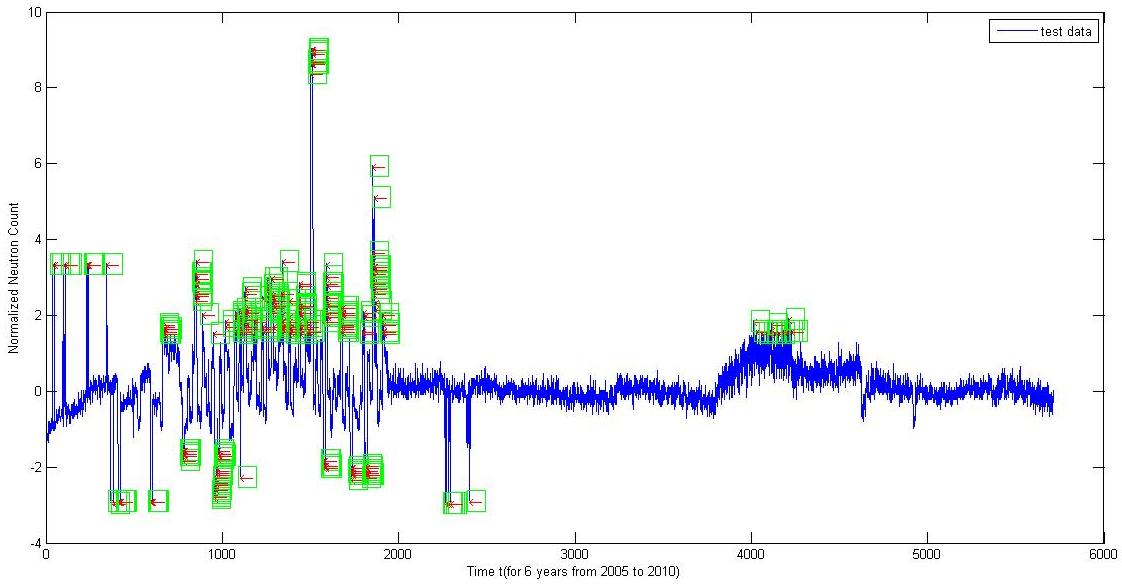}
\caption{Anomalies(marked in red) found in detector 1 of the power plant data}
\label{det2}
\end{sidewaysfigure}

\begin{sidewaysfigure}
\centering
\includegraphics[width=7in, height=4.5in ]{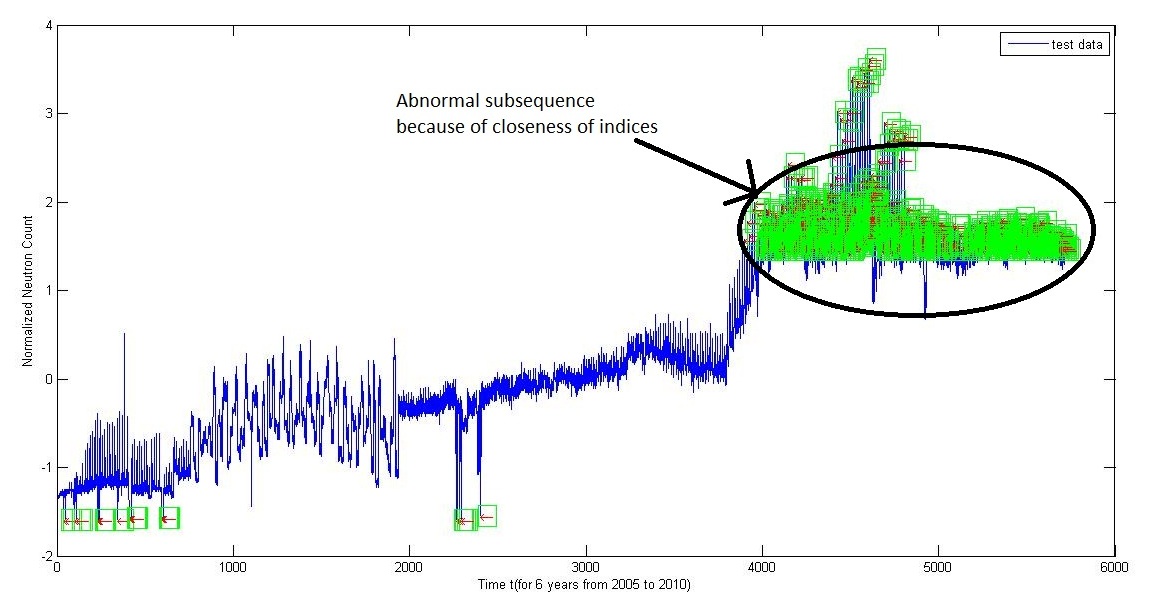}
\caption{Anomalies(marked in red) found in detector 1 of the power plant data}
\label{det3}
\end{sidewaysfigure}

 Our results of applying  Algorithm \ref{svddalgo} on the data from different channels of the reactor are shown in Fig. \ref{det1},\ref{det2},\ref{det3}. Results obtained exactly match with anomalous points (In fact neutron count was as low as 0 to 5 and as high as 170 + which is considered as abnormal flow of neutrons during some particular time period) which we verified with the expert at Bhabha Atomic Research Center). In Fig. \ref{det3}, a large number of anomalous points gets accumulated. This is verified later that this was due to technical fault in the detector for the specified duration.

\section{Discussion}
In the present work, we studied support vector data description algorithm for  anomaly detection in nuclear power plant data.  We observe that SVDD efficiently and effectively finds \emph{point} anomalies in the nuclear reactor data set. In the present work, we applied SVDD algorithm for each reactor channel individually. However, In future, we plan to use more state-of-the-art work on \emph{unsupervised} anomaly detection in the \emph{multi-variate} setting.




%

\thispagestyle{empty}
\chapter{Conclusions and Future Work}
Anomaly detection is an important task in machine learning and data mining. Due to unprecedented growth in data size and complexity, data miners and practitioners are overwhelmed with what is called big data.  Big data incapacitates the traditional  anomaly detection techniques. As such, there is an urgent need to develop efficient and scalable techniques for anomaly detection in big data.
\section{Conclusions}
 In the present research work, we proposed four novel algorithms for handling anomaly detection in big data. The PAGMEAN and ASPGD are based on online learning paradigm whereas  DSCIL and CILSD  are based on distributed learning paradigm. In order to handle the anomaly detection problem in big data, we took an  approach different from many works in the literature. Specifically, we employ the class-imbalance learning approach to  tackling \emph{point anomalies}.

PAGMEAN is an online algorithm for class-imbalance learning and anomaly detection in the \emph{streaming} setting. In chapter \ref{Chapter3}, we showed that how we can directly optimize a non-decomposable performance metric \emph{Gmean} in binary classification setting. Doing so gives rise to a non-convex loss function. We employ surrogate loss function for handling non-convexity. Subsequently, the surrogate loss is used within the PA framework to derive PAGMEAN algorithms.  We show through extensive experiments on benchmark and real data sets that PAGMEAN outperforms its parent algorithms PA and recently proposed algorithm CSOC in terms of \emph{Gmean}. However, at the same time, we observed that PAGMEAN algorithms suffer higher \emph{Mistake rate} than other  algorithms we compared with.

In chapter \ref{Chapter4}, we proposed  ASPGD algorithm for tackling anomaly detection in \emph{streaming, high dimensional} and \emph{sparse} data. We utilize  accelerated-stochastic-proximal learning framework with a cost-sensitive smooth hinge loss. Cost-sensitive smooth hinge loss applies penalty based on the number of positive and negative samples received so far. We also proposed a non-accelerated variant of ASPGD, that is, without Nesterov's acceleration called ASPGDNOACC. An empirical study on real and benchmark data sets show that \emph{acceleration} is not always helpful neither in terms of \emph{Gmean} nor \emph{Mistake rate}. In addition, we also compare with a recently proposed algorithm called CSFSOL. It is found that ASPGD outperforms CSFSOL in terms of \emph{Gmean, F-measure, Mistake-rate} on many of the data sets tested. 

In order to handle anomaly detection in \emph{sparse, high dimensional} and \emph{distributed} data, we proposed DSCIL and CILSD algorithms in chapter \ref{Chapter5}. In particular, DSCIL algorithm is based on the distributed ADMM framework that utilizes a cost-sensitive loss function. Within the DSCIL algorithm, we solve the $L_2$ regularized loss minimization problem via (1) L-BFGS method (called L-DSCIL) (2) Random Coordinate Descent method called (R-DSCIL). Firstly, Empirical convergence analysis shows that L-DSCIL converges faster than R-DSCIL. Secondly, \emph{Gmean} achieved by L-DSCIL is close to the \emph{Gmean} of  a centralized solution on most of the datasets. Whereas \emph{Gmean} achieved by R-DSCIL varies due to its random updates of coordinates.  Thirdly, coming to the training time comparison, we found in our experiments that R-DSCIL has cheaper per iteration cost but takes a longer time to achieve $\epsilon$ accuracy compared to L-DSCIL algorithm. Real world anomaly detection application  on KDDCup 2008 data set clearly shows the potential advantage of using the R-DSCIL algorithm.

CILSD, which is a cost-sensitive distributed FISTA-like algorithm, is a parameter-free algorithm for anomaly detection. We show, through extensive experiments on benchmark and real data sets, the convergence behavior of CILSD, speed up, the effect of the number of cores and regularization parameter on Gmean. In particular, CILSD algorithm does not require tuning of learning rate parameter and can be set to $1/L$ as in gradient descent algorithm.
Comparative evaluation with respect to a recently proposed class-imbalance learning algorithm and a centralized algorithm shows that CILSD is able to either outperform or perform equally well on many of the data sets tested. Speedup results demonstrate the advantage of employing multiple cores. We also observed, in our experiments, chaotic behavior of \emph{Gmean} with respect to varying number of cores. Experiment on KDDCup data set indicates the possibility of using CILSD algorithm for real-world distributed anomaly detection task. Comparison of DCSIL and CILSD shows that CILSD convergences faster and achieves higher \emph{Gmean} than DSCIL. 

We present a case study of anomaly detection on real-world data in chapter \ref{Chapter6}. Our data came from the nuclear power plant and is unlabeled. All of the algorithms discussed above can not be applied since they are based on \emph{supervised}  learning, i.e., they require labels for normal and anomalous instances. Therefore, we utilize SVDD , an unsupervised learning algorithm  for anomaly detection in real-world data. Empirical results show the effectiveness and efficiency of the SVDD algorithm. 

\section{Future Works}
In this section, we discuss some potential research directions for future. 
Our algorithms, though are scalable to high dimensions, take care of sparse  and distributed data,  have certain limitations. For example, we did not handle concept drift specifically in our setting. As a  first future work may look upon utilizing concept drift detection techniques within the framework we used. Secondly, big data is often not only distributed but also streaming in real-world. Therefore, online distributed algorithm may be developed to handle anomaly detection. As a third work, data heterogeneity may be combined with streaming, sparse and high dimensional characteristics of big data while detecting anomalies. 

Besides the above task, one may consider extending the existing framework to handle \emph{subsequence} and \emph{contextual} anomaly in big data.

\bibliographystyle{unsrt}
\bibliography{TKDE}
\end{document}